%% file: main.tex
\documentclass{physical-review}
\input{pci-style}
\input{paper-info}

\begin{document}
\begin{pcifrontmatter}
\maketitle
\input{figures/teaser}
\input{sections/00-abstract}
\end{pcifrontmatter}
\PciContents

\input{sections/01-introduction}
\input{sections/01b-lens}
\input{sections/02-foundations}
\input{sections/03-programs-as-action}
\input{sections/04-objective-and-world}
\input{sections/05-evidence}
\input{sections/06-compounding}
\input{sections/07-measuring-progress}

\input{sections/09-synthesis}
\input{sections/10-conclusion}

\clearpage
\pagestyle{pcirefs}
\phantomsection\addcontentsline{toc}{section}{References}
\bibliographystyle{plainnat}
\bibliography{references}

\clearpage
\pagestyle{pci}
\appendix
\input{sections/appendix-benchmarks}
\end{document}

%% file: pci-style.tex
\usepackage{adjustbox,flafter,float}
\usepackage[section]{placeins}
\usepackage{pgfplots}
\pgfplotsset{compat=1.18}

\newcommand{\TableHeader}[1]{\textbf{\sffamily\color{Ink}#1}}

\newcommand{\pciRQ}[1]{\mbox{\textbf{RQ#1}}}
\newcommand{\tablecite}[1]{\citep{#1}}
\definecolor{LinkNavy}{HTML}{27466F}
\hypersetup{linkcolor=LinkNavy,citecolor=LinkNavy,urlcolor=LinkNavy}
\newcommand{\pcifigworkcolor}{LinkNavy}
\newcommand{\figwork}[2]{\nocite{#1}\hyperlink{cite.#1}{\textcolor{\pcifigworkcolor}{#2}}}
\DeclareRobustCommand{\tabfull}{\textcolor{SecFound}{\faCircle}}
\DeclareRobustCommand{\tabpart}{\textcolor{SecFound}{\faAdjust}}
\DeclareRobustCommand{\tabnone}{\textcolor{MutedText!55}{\faCircle[regular]}}
\DeclareRobustCommand{\tabyes}{\textcolor{SecWrite}{\faCheck}}
\DeclareRobustCommand{\tabno}{\textcolor{MutedText}{\faTimes}}
\DeclareRobustCommand{\tabdemo}{\textcolor{SecWrite!70!Baseline}{\faVideo}}
\DeclareRobustCommand{\tabnf}{\textcolor{MutedText}{\textsf{NR}}}
\DeclareRobustCommand{\vis}{\textcolor{SecTest}{\faEye}}
\DeclareRobustCommand{\hid}{\textcolor{SecFound}{\faEyeSlash}}
\DeclareRobustCommand{\tabtag}[2]{\tikz[baseline={([yshift=-6.4pt]t.north)}]\node[anchor=base,draw=#1!55!white,fill=#1!10!white,text=#1!70!black,
  rounded corners=2pt,line width=.4pt,inner xsep=2.5pt,inner ysep=1.2pt,align=center,
  font=\sffamily\bfseries\scriptsize] (t) {#2};}

\newcommand{\roleAct}{\tabtag{SecWrite}{ACT}}
\newcommand{\roleOrch}{\tabtag{SecWrite}{ORCH}}
\newcommand{\roleObj}{\tabtag{SecWrite!55!white}{OBJ}}
\newcommand{\roleWorld}{\tabtag{SecWrite!55!white}{WORLD}}

\tcbset{pcicolor/.style={no borderline,borderline west={1.3pt}{0pt}{#1},
  colback=#1!5!white,colbacktitle=#1!5!white,coltitle=#1!70!black}}
\newtcolorbox{pciinsight}[2][]{enhanced,unbreakable,
  frame hidden,boxrule=0pt,sharp corners,pcicolor=SecFound,
  left=9pt,right=9pt,top=5pt,bottom=7pt,toptitle=7pt,bottomtitle=0pt,
  fonttitle=\sffamily\bfseries\small,fontupper=\small,
  title={#2},before={\par\ifinner\else\FloatBarrier\Needspace{7\baselineskip}\fi},after skip=9pt,#1}
\tikzset{
  pcibox/.style={draw=#1!48!white,fill=#1!4!white,rounded corners=1pt,
    line width=.5pt,align=left,inner sep=5pt,font=\sffamily\small,
    execute at begin node={\hyphenpenalty=10000\exhyphenpenalty=10000}},
  pciflow/.style={-{Latex[length=1.8mm]},draw=MutedText,line width=.65pt},
  pcitag/.style={font=\sffamily\footnotesize,text=MutedText,align=center},
  pcihead/.style={font=\sffamily\small\bfseries,text=Ink,align=left}}

\usetikzlibrary{shapes.arrows}
\graphicspath{{figures/art/}}
\tikzset{
  art img/.style={inner sep=0pt,anchor=center},
  art fat/.style={draw=#1!55!white,fill=#1!14!white,line width=.5pt,
    single arrow,single arrow head extend=2.5pt,minimum height=4.4mm,minimum width=4mm,inner sep=0pt}}

\definecolor{LhBlue}{HTML}{4E79A7}
\definecolor{LhTeal}{HTML}{3F8F8A}
\definecolor{LhOrange}{HTML}{D98232}
\definecolor{LhRed}{HTML}{C9545A}
\definecolor{LhGreen}{HTML}{5A9550}
\definecolor{LhPurple}{HTML}{9A6C93}
\definecolor{LhGray}{HTML}{727980}
\definecolor{SecFound}{HTML}{4E6E94}
\definecolor{SecWrite}{HTML}{3F8F8A}
\definecolor{SecTest}{HTML}{C9585A}
\definecolor{SecLearn}{HTML}{C38A2E}
\definecolor{Baseline}{HTML}{8A847D}
\colorlet{TimeA}{SecFound!45!white}\colorlet{TimeB}{SecFound!63!white}
\colorlet{TimeC}{SecFound!82!white}\colorlet{TimeD}{SecFound}
\colorlet{CostA}{SecTest!40!white}\colorlet{CostB}{SecTest!52!white}\colorlet{CostC}{SecTest!64!white}
\colorlet{CostD}{SecTest!76!white}\colorlet{CostE}{SecTest!88!white}\colorlet{CostF}{SecTest}
\usetikzlibrary{shadows,shadows.blur}
\tikzset{pci shadow/.style={drop shadow={shadow xshift=3.4pt,shadow yshift=-3.4pt,
  fill=#1!38!white,opacity=1}}}
\tikzset{pci shadow light/.style={drop shadow={shadow xshift=1.6pt,shadow yshift=-1.6pt,
  fill=#1!22!white,opacity=.85}}}
\tikzset{
  lh panel/.style={draw=#1,line width=1.2pt,fill=#1!5!white,rounded corners=7pt,pci shadow=#1},
  lh chip/.style={draw=#1!55!white,fill=#1!12!white,rounded corners=3pt,line width=.5pt,
    font=\cartoonfont\scriptsize\bfseries,text=#1!80!black,inner xsep=3pt,inner ysep=2pt,
    text height=1.55ex,text depth=.4ex},
  lh head/.style={font=\cartoonfont\bfseries\normalsize,text=#1,inner sep=0pt,anchor=west},
  lh sub/.style={font=\sffamily\bfseries\footnotesize,text=#1,inner sep=0pt},
  lh text/.style={font=\sffamily\scriptsize,text=Ink,inner sep=0pt,align=left,
    execute at begin node={\hyphenpenalty=10000\exhyphenpenalty=10000\raggedright}},
  lh arrow/.style={-{Latex[length=2.6mm,width=2.6mm]},draw=#1,line width=1.6pt},
}
\newcommand{\lhnum}[2]{\tikz[baseline=-.6ex]\node[anchor=center,circle,fill=#1,text=white,font=\sffamily\bfseries\footnotesize,inner sep=1.2pt,minimum size=4.2mm]{#2};}
\newcommand{\lhicon}[2]{\tikz[baseline=-.6ex]\node[anchor=center,circle,fill=#1!15!white,draw=#1!45!white,line width=.4pt,text=#1,font=\scriptsize,inner sep=0pt,minimum size=4.6mm]{\faIcon{#2}};}
\newcommand{\lhtitlesize}{\small}
\newcommand{\lhpanel}[8]{%
  \draw[lh panel=#5] (#1,#2) rectangle ++(#3,#4);
  \node[anchor=west,inner sep=0pt] at (#1+0.22,#2+#4-0.38) {\lhnum{#5}{#6}};
  \node[lh head=#5,font=\cartoonfont\bfseries\lhtitlesize] at (#1+0.72,#2+#4-0.38) {\adjustbox{max width=\dimexpr #3cm-0.95cm\relax}{\strut#7}};
  \node[font=\sffamily\scriptsize\itshape,text=#5!75!black,inner sep=0pt,anchor=west] at (#1+0.22,#2+#4-0.8) {#8};}
\newcommand{\lhb}[4]{\raisebox{0pt}[\ht\strutbox][\dp\strutbox]{\lhicon{#1}{#2}}\enspace\parbox[t]{#3}{\raggedright\hyphenpenalty=10000\exhyphenpenalty=10000 #4}}
\tikzset{lh eg/.style={font=\sffamily\scriptsize\itshape,text=MutedText,inner sep=0pt,anchor=west}}
\newcommand{\pcicircnum}[1]{\tikz[baseline=-.62ex]\node[anchor=center,circle,draw=Ink,line width=.45pt,inner sep=0pt,minimum size=3.4mm,font=\sffamily\fontsize{6.5}{7}\selectfont]{\let\scriptsize\relax#1};}
\DeclareRobustCommand{\textcircled}[1]{\pcicircnum{#1}}

%% file: paper-info.tex
\title{When the World Is the Test Suite}
\subtitle{Agentic Intelligence for the Physical World}
\author{Kaixin Li, Meng Luo, Yanlin Li, Yuchen Tian, Xiang Liu, Xiangru Jian, Yan Yang, Zhiheng Xi, Yang Zhang, Ziyang Luo, Mingfei Cheng, Hongzhan Lin}

\renewcommand{\PciAuthors}{%
  \PciAuthor{Kaixin Li}{1}\quad
  \PciAuthor{Meng Luo}{1}\quad
  \PciAuthor{Yanlin Li}{1}\quad
  \PciAuthor{Yuchen Tian}{2}\quad
  \PciAuthor{Xiang Liu}{1}\quad
  \PciAuthor{Xiangru Jian}{3}\\[1pt]
  \PciAuthor{Yan Yang}{4}\quad
  \PciAuthor{Zhiheng Xi}{5}\quad
  \PciAuthor{Yang Zhang}{1}\quad
  \PciAuthor{Ziyang Luo}{2}\quad
  \PciAuthor{Mingfei Cheng}{6}\quad
  \PciAuthor{Hongzhan Lin}{1,$\dagger$}}

\renewcommand{\PciAffiliations}{%
  \PciAffiliation{1}{National University of Singapore}\quad
  \PciAffiliation{2}{Hong Kong Baptist University}\quad
  \PciAffiliation{3}{University of Waterloo}\\
  \PciAffiliation{4}{The Australian National University}\quad
  \PciAffiliation{5}{Fudan University}\quad
  \PciAffiliation{6}{Singapore Management University}}
\Correspondence{\textsuperscript{$\dagger$}\,Corresponding author.}

\ProjectWebsite{https://physicalworldagent.github.io}
\GitHubRepository{https://github.com/PhysicalWorldAgent/Awesome-PhysicalWorldAgent}
\Contact{\PciMail{likaixin@u.nus.edu}\PciMailSep\PciMail{danielhzlin@nus.edu.sg}}
\hypersetup{pdfsubject={Programs, testing, and experience for agents that act in the physical world}}

%% file: figures/teaser.tex
\begin{figure}[H]
\centering
\newcommand{\ct}[1]{{\cartoonfont\bfseries\footnotesize #1}}
\begin{tikzpicture}[x=1cm,y=1cm,
  call/.style={pci shadow light=#1,draw=#1,line width=1pt,fill=white,fill opacity=.94,text opacity=1,rounded corners=4pt,
    inner xsep=4pt,inner ysep=3pt,align=left,font=\sffamily\scriptsize,text=Ink,
    execute at begin node={\hyphenpenalty=10000\exhyphenpenalty=10000}},
  lead/.style={draw=#1,line width=.9pt},
  dot/.style={circle,fill=#1,draw=white,line width=.8pt,inner sep=0pt,minimum size=2.6mm}]
\node[font=\cartoonfont\bfseries\large,text=Ink] (tag) at (8.09,6.25) {Same program.\enspace Different test suite.};
\draw[RuleGray,line width=.6pt] (0.2,6.25) -- ([xshift=-4mm]tag.west);
\draw[RuleGray,line width=.6pt] ([xshift=4mm]tag.east) -- (15.98,6.25);
\node[anchor=south west,inner sep=0pt] (img) at (0,0) {\includegraphics[width=16.18cm]{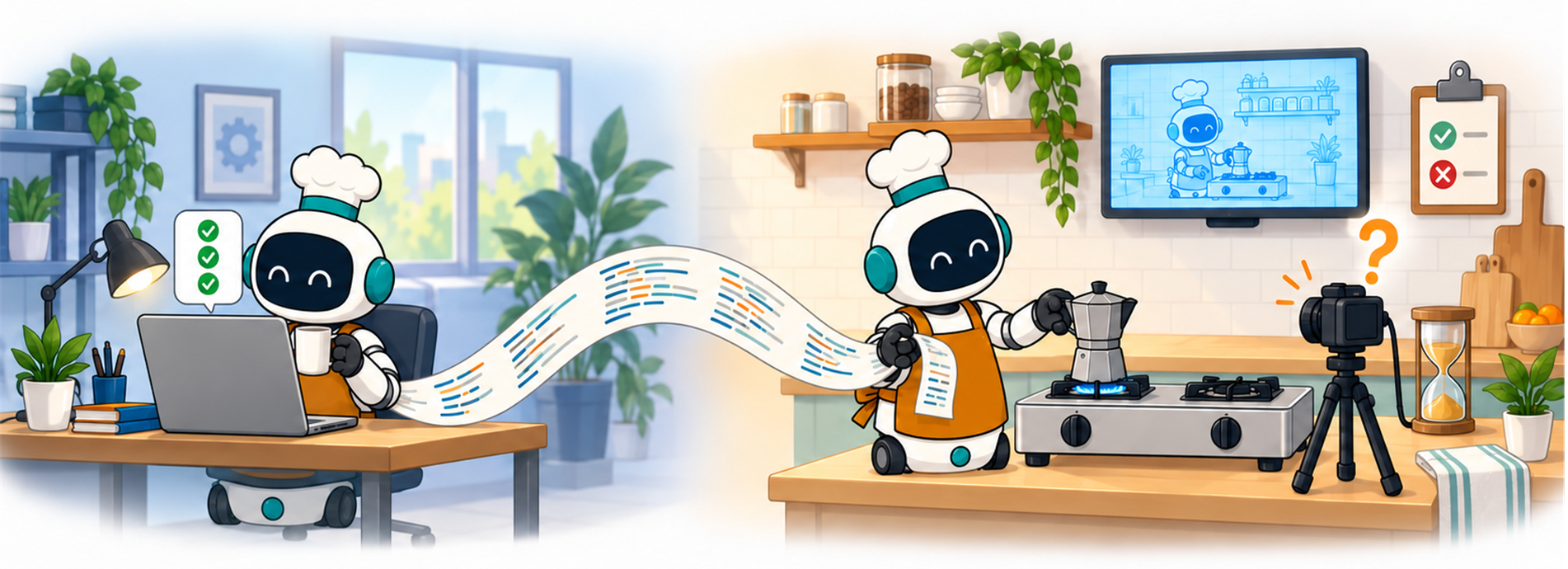}};
\begin{scope}[shift={(img.south west)},x={($(img.south east)-(img.south west)$)},y={($(img.north west)-(img.south west)$)}]
  \coordinate (T) at (.152,.346); \coordinate (M) at (.757,.746);
  \coordinate (J) at (.857,.442); \coordinate (H) at (.921,.36);
  \node[lh chip=Baseline,font=\cartoonfont\bfseries\scriptsize] at (.43,.585) {the same program};
  \node[call=Baseline,anchor=south west] (cT) at (.012,.03) {\textcolor{Baseline}{\faIcon{check-circle}}\hspace{3pt}\ct{Tests Usually Ship with the Code}\\[1pt]cheap, repeatable, run after every edit};
  \node[call=SecTest,anchor=north east] (cM) at (.643,.975) {\textcolor{SecTest}{\faIcon{desktop}}\hspace{3pt}\ct{A Simulator Can Stand In}\\[1pt]whether each pass transfers is rarely measured};
  \node[call=SecTest,anchor=north east] (cJ) at (.988,.975) {\textcolor{SecTest}{\faIcon{question-circle}}\hspace{3pt}\ct{Someone Must Build the Judge}\\[1pt]and learned judges can pass failures};
  \node[call=SecTest,anchor=south east] (cH) at (.988,.03) {\textcolor{SecTest}{\faIcon{hourglass-half}}\hspace{3pt}\ct{Physical Tests Cost Execution}\\[1pt]an episode, a training run, or robot time};
  \draw[lead=Baseline] (cT.north) ++(.03,0) -- (T) node[dot=Baseline] {};
  \draw[lead=SecTest] (cM.east) -- (M) node[dot=SecTest] {};
  \draw[lead=SecTest] ([xshift=6mm]J |- cJ.south) -- (J) node[dot=SecTest] {};
  \draw[lead=SecTest] (cH.north) ++(.05,0) -- (H) node[dot=SecTest] {};
\end{scope}
\end{tikzpicture}
\caption{\textbf{Same program, different test suite.} In software, the test suite usually ships with the code and returns a cheap, repeatable result. Once the program acts in the physical world, someone must build the judge, a simulator can stand in for the robot, and every physical test has to be paid for in execution.}
\label{fig:teaser}
\end{figure}
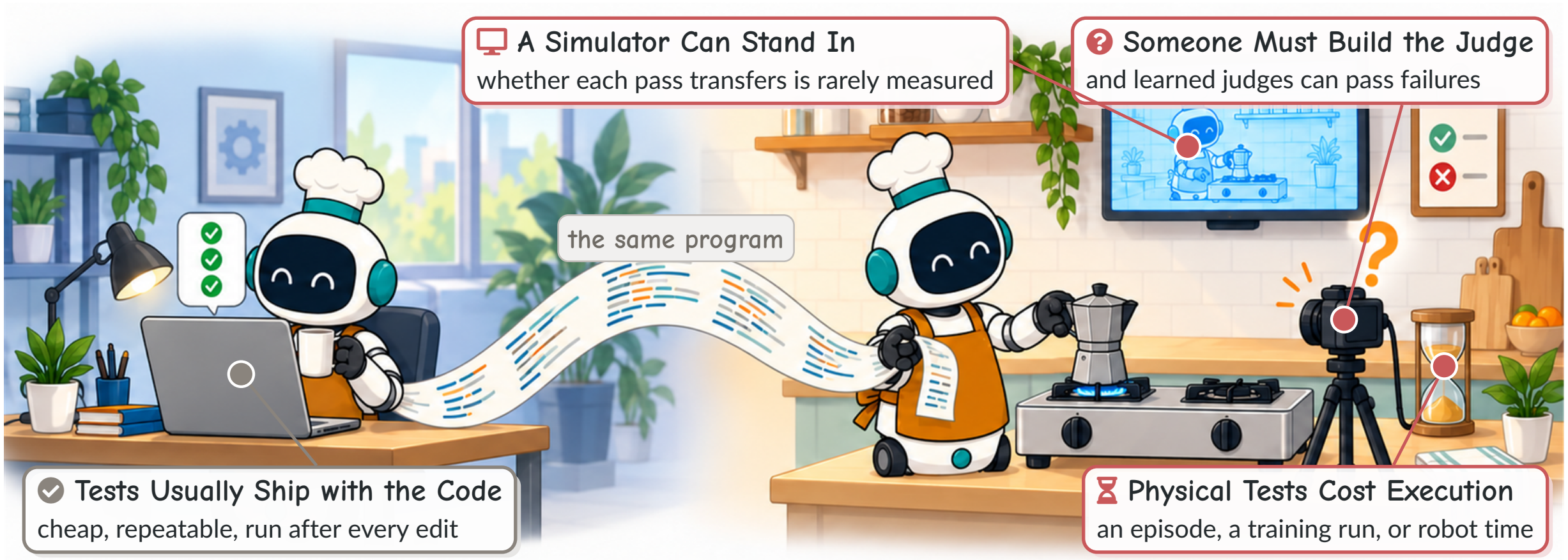

%% file: sections/00-abstract.tex
\begin{pciabstract}
Coding agents built to repair software now write, test, and repair programs
for robots, laboratories, and industrial controllers. Unlike in software
engineering, no test suite comes with the code. Whether a program works must
be established by running it on a robot, in a physics simulator, or through
the policies it helps to train, and such evidence is costly to obtain and can
itself be wrong. As these agents take on more physical tasks, how their
programs are judged matters as much as how they are written. This paper
examines this emerging area, agentic intelligence for the physical world, by
treating the world as the program's test suite. We organize the discussion
around three questions: how agents write programs for the physical world, how these
programs are tested, and how agents learn from the results. We first trace how
language models for robots and software engineering agents came together, ask
which software testing practices carry over to physical tests, and describe
each program by where it sits relative to the actuator and by what judges it.
For writing, we cover programs that
act as robot policies and programs that support policy learning by producing
rewards, tasks, and training worlds. For testing, we examine what physical
trials cost, what traces, learned judges, simulators, and formal checks can
establish in their place, and what safety requires beyond task success. For
learning, we discuss how tested programs are retained as experience, in skill
libraries and in model weights. Finally, we turn to benchmarks and evaluation
practice and outline open challenges. Writing programs is becoming cheap,
while trustworthy evidence about what they do remains costly. Progress will therefore depend on a physical counterpart of continuous
integration: verification cheap enough to run often, reliable enough to trust
under search, and maintained alongside the experience it validates.
\end{pciabstract}

%% file: sections/01-introduction.tex
\section{Introduction}
\label{sec:introduction}

\input{figures/program-loop}

A coding agent is trying to put a moka pot on a stove. It is looking at four
images taken just before and just after a simulated gripper attempted to grasp
the pot's handle. The fingers have closed on empty air beside it. The agent
rereads the grasp routine it wrote an hour earlier and sees that it aimed at
the center of the pot's segmentation mask. It changes the target to the handle
and runs the program again from fifty initial states. The pot is lifted in most
of them. The agent writes a short note describing the failure and its fix and
adds it to a library that agents working on other tasks will read. This is a
composite scene with illustrative details, but systems such as ASPIRE already
carry out this sequence of inspection, repair, validation, and reuse
\citep{arxiv260700272}. \Cref{fig:program-loop} draws the scene as a loop.

The debugging loop is familiar from software engineering: run the program,
inspect the failure, revise it, rerun it, and keep what works
\citep{arxiv240515793,arxiv240716741,arxiv231006770}. Recent robotics systems
reuse general-purpose coding agents for this work, giving them access to robot
interfaces and execution environments. On manipulation benchmarks with task
perturbations, these systems report success rates well above those published
for end-to-end policies trained on demonstrations
\citep{arxiv260700272,arxiv260616458,arxiv261002459}. Validated programs can
also generate successful trajectories for training those policies
\citep{arxiv260929166,arxiv260927308}.

In the opening scene, lifting the pot still leaves the task unfinished. Has the pot reached the burner, and is it upright and stable? The claim that it was
lifted also depends on what the agent can observe and on what counts as a
successful lift. In many software tasks, an existing test suite gives the agent a cheap,
repeatable check on a proposed repair. For a physical task, the check may
require observations, simulator state, or another trial, and its cost and
reliability depend on how it is made.

The literature shows both the value of execution feedback and the expense of
obtaining it. In ASPIRE's ablation on LIBERO-PRO, an execution engine and a
skill library together raise macro-average success from 14\% to 62\%. The
engine records perception and control calls, exposes images before and after
them, and re-runs candidate repairs. Evolutionary search then adds
another 10 percentage points \citep{arxiv260700272}. For programs that
generate training environments, useful feedback can require training a policy
on the environments they produce. A full Eurekaverse run incurs about
15 US dollars in language-model API costs and takes about 24 hours on eight
A6000 GPUs \citep{arxiv241101775}.

Beyond its cost, the judgment itself can be wrong \citep{zhao-etal-2025-codejudge,zeng-etal-2025-mm,lin-etal-2025-fact,kong-etal-2025-sharp,arxiv261029677}. On FailBench, the best of thirteen
detectors based on vision-language models (VLMs) reaches a macro-averaged balanced accuracy of only 0.77 when
judging manipulation outcomes \citep{arxiv260903611}. In OpenRUA's analysis
of failed episodes, about a quarter involve a success condition absent from the
instruction or a predicate that contradicts it \citep{arxiv261002459}.
These results motivate our central claim: \emph{for agents that write programs
for the physical world, obtaining trustworthy evidence about program behavior
is increasingly a limiting factor}. We examine how this problem arises, what the available evidence supports,
and how it shapes the design of agents, their memory, and their evaluation.

\input{tables/survey-position}

\Cref{tab:survey-position} compares this article with existing reviews on five
points: coverage of the physical world, programs as the object of study,
analysis of judges, coverage of laboratory and industrial code, and transfer
from software practice.
Reviews of foundation models in robotics and of vision-language-action models
organize the field by capability or by the form of the action, with code as
one action representation or a planning tool
\citep{arxiv231107226,arxiv231208782,arxiv231207843,arxiv231210807,arxiv240205741,arxiv240514093,arxiv250701925}. Reviews of code
generation and of software-engineering agents cover execution feedback and
testing in depth but leave the physical world out
\citep{arxiv240600515,arxiv240902977}. Three recent reviews come closest.
Weights or Skills? contrasts policies stored in weights with agents that write
their own code skills, orders code-as-policy systems by their degree of
self-improvement, and proposes metrics for cumulative improvement
\citep{arxiv260801851}. No Free Checker reviews about 150 verifiers of robot
policies and identifies a trade-off between the availability and credibility
of their judgments \citep{arxiv260909250}. Code as Agent Harness treats code as
the executable substrate of agents in general, names embodied agents and
self-driving laboratories among its applications, and observes that
``execution is only as reliable as the oracle attached to it''
\citep{arxiv260518747}.

This article takes as its object of comparison the program in the physical
world, together with the evidence that judges it. The Weights or Skills? review follows how much a code-writing
agent improves, and we ask what certified each improvement. No Free Checker grades
verifiers of policies, whereas our concern is the judges of programs,
artifacts that can be read, traced, and repaired. From the software machinery that Code as
Agent Harness catalogs, we pick out the parts that survive contact with
robots, laboratories, and industrial controllers. A framework and survey from October 2026, Who Improves
the Robot?, also comes close. It argues that no robot work it surveys yet demonstrates
independently verified recursive self-improvement \citep{arxiv261018842}. That
survey assesses claims of self-improvement as a whole, whereas we compare the
evidence behind individual programs.

Concretely, this article makes four contributions:
\begin{itemize}[leftmargin=1.2em,itemsep=1pt,topsep=3pt]
\item We describe each program by where it sits relative to the actuator and by what judges it, and use these two properties to compare representative systems from action code to generated training worlds.
\item We compare programs with end-to-end policies across the reported studies, which favor programs when the structure of a task changes and in several long-horizon and constrained settings. We also set out what traces, judges, simulators, and formal checks can each establish about a program, and collect the documented failure modes of every kind of judge.
\item We examine how accepted programs become retained experience, in skill libraries and in model weights, and identify five quantities that are needed to distinguish progress from regression and that current reports rarely state: cost, variance, judge visibility, judge error, and the link between simulation and reality.
\item We ask which software testing practices carry over to physical tests and which break, and set out open problems, each paired with a first experiment.
\end{itemize}

%% file: figures/program-loop.tex
\begin{figure}[!tbp]
\centering
\newcommand{\ct}[1]{{\cartoonfont\bfseries\footnotesize #1}}
\begin{tikzpicture}[x=1cm,y=1cm,
  call/.style={pci shadow light=#1,draw=#1,line width=1pt,fill=white,fill opacity=.94,text opacity=1,rounded corners=4pt,
    inner xsep=4pt,inner ysep=3pt,align=left,font=\sffamily\scriptsize,text=Ink,
    execute at begin node={\hyphenpenalty=10000\exhyphenpenalty=10000}},
  lead/.style={draw=#1,line width=.9pt},
  dot/.style={circle,fill=#1,draw=white,line width=.8pt,inner sep=0pt,minimum size=2.6mm},
  loopa/.style={-{Latex[length=2.4mm,width=2.4mm]},draw=#1,line width=1.7pt,dash pattern=on 4pt off 2pt,
    preaction={draw=white,line width=3.6pt,-,solid}}]
\node[anchor=south west,inner sep=0pt] (img) at (0,0) {\includegraphics[width=16.18cm]{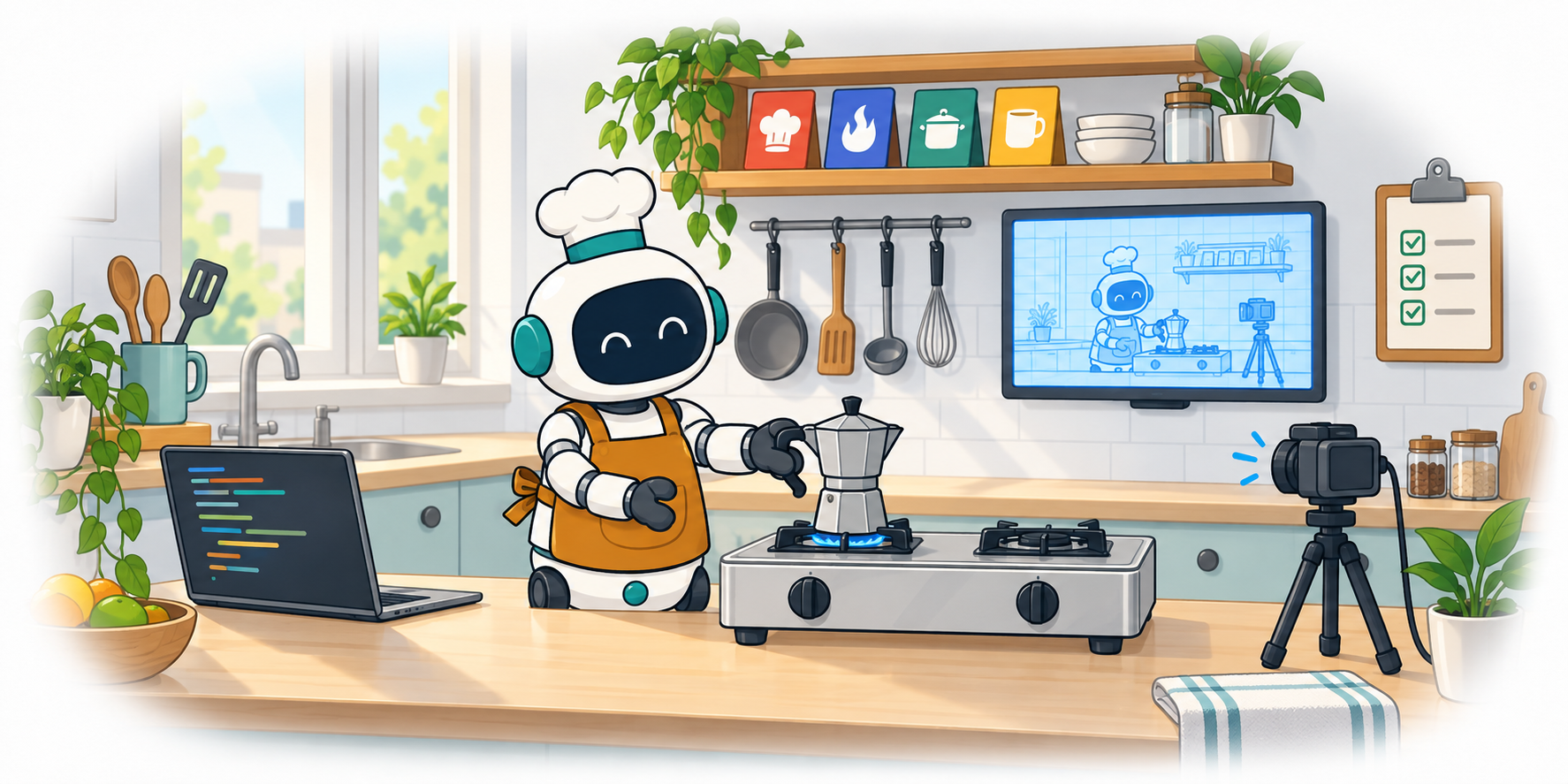}};
\begin{scope}[shift={(img.south west)},x={($(img.south east)-(img.south west)$)},y={($(img.north west)-(img.south west)$)}]
  \coordinate (L) at (.16,.33); \coordinate (S) at (.535,.36);
  \coordinate (C) at (.845,.41); \coordinate (K) at (.535,.84); \coordinate (M) at (.712,.62);
  \draw[loopa=SecWrite]   (.29,.25) to[bend right=18] (.50,.25);
  \draw[loopa=SecFound] (.60,.25) to[bend right=18] (.76,.25);
  \draw[loopa=SecTest]    (.80,.46) to[bend right=28] (.62,.83);
  \draw[loopa=SecLearn]  (.45,.83) to[bend right=30] (.21,.43);
  \node[call=SecWrite,anchor=south west] (c1) at (.015,.035) {\lhnum{SecWrite}{1}\,\ct{Write a Program}\\[1pt]action \textperiodcentered{} orchestration \textperiodcentered{} objective \textperiodcentered{} world};
  \node[call=SecFound,anchor=south] (c2) at (.515,.035) {\lhnum{SecFound}{2}\,\ct{Run It in the Physical World}\\[1pt]on a robot, or by training a policy};
  \node[call=SecTest,anchor=south east] (c3) at (.985,.035) {\lhnum{SecTest}{3}\,\ct{Judge the Outcome}\\[1pt]check \textperiodcentered{} learned judge \textperiodcentered{} human \textperiodcentered{} outcome};
  \node[call=SecLearn,anchor=north west] (c4) at (.015,.975) {\lhnum{SecLearn}{4}\,\ct{Repair, or Keep What Passes}\\[1pt]programs that pass enter a skill library};
  \node[call=SecFound,anchor=north east] (c5) at (.985,.975) {\ct{\ldots{} or in a Physics Simulator}};
  \draw[lead=SecWrite] (c1.north) ++(-.02,0) -- (L) node[dot=SecWrite] {};
  \draw[lead=SecFound] (c2.north) -- (S) node[dot=SecFound] {};
  \draw[lead=SecTest] (c3.north) ++(.01,0) -- (C) node[dot=SecTest] {};
  \draw[lead=SecLearn] (c4.east) -- (K) node[dot=SecLearn] {};
  \draw[lead=SecFound] (c5.south) ++(-.04,0) -- (M) node[dot=SecFound] {};
  \node[lh chip=SecWrite,font=\cartoonfont\tiny\bfseries,anchor=south west] at (c1.north west) {axis 1: program role};
  \node[lh chip=SecTest,font=\cartoonfont\tiny\bfseries,anchor=south east] at (c3.north east) {axis 2: evaluation source};
\end{scope}
\end{tikzpicture}
\caption{\textbf{When the world is the test suite.} The kitchen robot of this article writes a program to put a moka pot on the stove \textcircled{\scriptsize 1}, runs it \textcircled{\scriptsize 2}, has the outcome judged \textcircled{\scriptsize 3}, and repairs the program or keeps what passes \textcircled{\scriptsize 4}. The two tags mark the program role and evaluation source defined in \cref{sec:axes}.}
\label{fig:program-loop}
\end{figure}
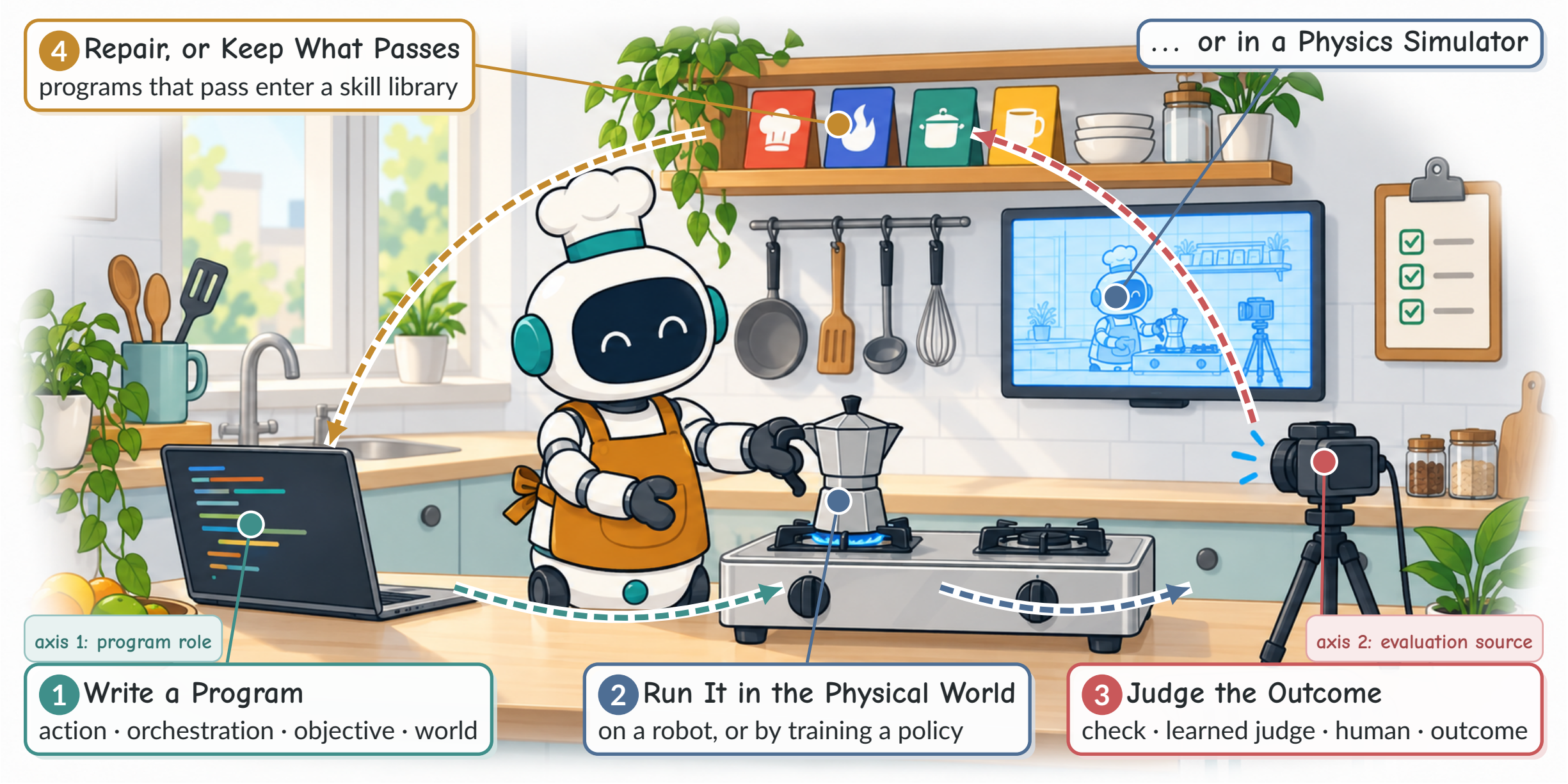

%% file: tables/survey-position.tex
\begin{table}[!htb]
\centering\footnotesize\setlength{\tabcolsep}{4pt}
\caption{\textbf{Position relative to existing reviews.} \tabfull{} the topic is a central object of the review; \tabpart{} it is treated in part; \tabnone{} it is absent or incidental.}
\label{tab:survey-position}
\begin{tabularx}{\linewidth}{@{}L{.40\linewidth}*{5}{>{\centering\arraybackslash}X}@{}}
\toprule
\TableHeader{Review} & \TableHeader{Physical world} & \TableHeader{Programs as the object} & \TableHeader{Judges analyzed} & \TableHeader{Lab and industrial code} & \TableHeader{Transfer from software}\\
\midrule
Foundation models and VLAs in robotics\newline\tablecite{arxiv231107226,arxiv231208782,arxiv231207843,arxiv231210807,arxiv240205741,arxiv240514093,arxiv250701925} & \tabfull & \tabpart & \tabpart & \tabnone & \tabnone\\[2pt]
Code generation and software-engineering agents\newline\tablecite{arxiv240600515,arxiv240902977} & \tabnone & \tabfull & \tabpart & \tabnone & \tabnone\\[2pt]
Code as the medium of agents\newline\tablecite{arxiv240100812,arxiv260518747} & \tabpart & \tabfull & \tabpart & \tabpart & \tabnone\\[2pt]
Weights or Skills?\newline\tablecite{arxiv260801851} & \tabfull & \tabfull & \tabnone & \tabnone & \tabnone\\[2pt]
No Free Checker\newline\tablecite{arxiv260909250} & \tabfull & \tabpart & \tabfull & \tabnone & \tabnone\\[2pt]
Safety of language-model agents\newline\tablecite{arxiv260814590} & \tabpart & \tabnone & \tabfull & \tabnone & \tabnone\\
\midrule
\textbf{This article} & \tabfull & \tabfull & \tabfull & \tabfull & \tabfull\\
\bottomrule
\end{tabularx}
\end{table}

%% file: sections/01b-lens.tex
\section{The Physical World as a Test Suite}
\label{sec:lens}\label{sec:foundations}

We call the subject \emph{agentic intelligence for the physical world}: work in which
a model writes or edits a program whose effects are evaluated through physical
execution on a robot, in a physics simulator, or through a policy trained with
the program's output. Where physical trial is not permitted, a model of the
equipment the program controls can take the place of execution. \Cref{fig:program-loop} shows the loop that
these systems share. A model writes a program, the program runs in physical
dynamics, a judge decides from what happened whether the program succeeded,
and that result either guides a repair or adds the program to a skill library
that later programs reuse.

Each step of this loop has a counterpart in software engineering, where an
agent edits a patch, executes it, checks the result against a test oracle, and
merges a change that later work builds on. There, a test
suite ties running, judging, and keeping together: it usually comes with the
repository, and the agent can run it after every edit. Treating the physical world as the
test suite means asking, at every step of the loop, what the world supplies in
place of a test: which observations count as evidence, what an evaluation costs,
how often a judgment is wrong, and what an accepted change commits later work to.

To compare programs across this loop, we describe each one by two coordinates, defined in
\cref{sec:axes}: where the program sits relative to the robot's actuators, and
what judges it. Three research questions follow from the loop:

\textbf{\pciRQ{1}: What do programs buy an agent that acts in the physical world, and what do they cost?}
Where in the stack should the program sit, when does it outperform an end-to-end policy, and how should the two be combined?

\textbf{\pciRQ{2}: When the world is the test suite, how do agents learn that a program is wrong, and how far can that signal be trusted?}
What do traces, judges, simulators, and formal checks contribute, and what replaces physical trial when trial is not allowed?

\textbf{\pciRQ{3}: How does validated experience accumulate, and what shows that progress is real?}
When do skill libraries help or hurt, how far does their content transfer, and what must be reported for a result to be interpretable?

\input{figures/organization}

\Cref{fig:organization} maps the article onto the loop, with representative
work for each part. Before
comparing methods, we define the scope with a running example, trace how
coding agents entered this work, examine what carries over from software
testing (\cref{sec:software-to-physical}), and set out the program roles and evaluation sources used for
comparison. Recurring terms are defined in the Key terms box on
page~\pageref{box:keyterms}.

%% file: figures/organization.tex
\begin{figure}[!htbp]
\centering
\begin{adjustbox}{max width=\linewidth}
\begin{tikzpicture}[x=1cm,y=1cm,
  head/.style={pci shadow=#1,draw=#1,line width=1.2pt,fill=#1!12!white,rounded corners=6pt,inner sep=5pt,text=#1!75!black,text width=4.8cm,minimum height=1.1cm,align=left,font=\sffamily\footnotesize\bfseries},
  leaf/.style={pci shadow=#1,draw=#1!55!white,line width=.8pt,fill=#1!4!white,rounded corners=6pt,inner sep=5pt,text width=4.8cm,align=flush left,font=\sffamily\fontsize{9}{10.5}\selectfont,anchor=north,
    execute at begin node={\def\pcifigworkcolor{#1!80!black}\hyphenpenalty=10000\exhyphenpenalty=10000}}]
\node[anchor=west,inner sep=0pt,font=\sffamily\small\bfseries,text=Ink]
  at (-2.57,1.4) {Agentic Intelligence for the Physical World};
\node[head=SecFound,minimum height=1.45cm] (c1) at (0,0) {From Software Tests to\\Physical Tests\\ {\mdseries\S\ref*{sec:software-to-physical}}};
\node[head=SecWrite,minimum height=1.45cm] (c2) at (5.5,0) {Programs as Robot Policies\\ {\mdseries\S\ref*{sec:action}\enspace(RQ1)}};
\node[head=SecWrite,minimum height=1.45cm] (c3) at (11,0) {Programs for Policy Learning\\ {\mdseries\S\ref*{sec:objective-world}\enspace(RQ1)}};
\node[leaf=SecFound] (l1) at ([yshift=-.3cm]c1.south) {\textbf{What Carries Over}\\ \figwork{arxiv260300455}{Test-driven controllers}; \figwork{arxiv260712220}{Contract-grounded BTs}; \figwork{arxiv260306842}{RoboCritics}\\[3pt]
  \textbf{What Breaks}\\ \figwork{arxiv260930249}{RAPID}; \figwork{arxiv260619980}{ENPIRE}; \figwork{arxiv260120334}{FAEA}};
\node[leaf=SecWrite] (l2) at ([yshift=-.3cm]c2.south) {\textbf{Primitive Interfaces}\\ \figwork{arxiv220907753}{Code as Policies}; \figwork{arxiv240406645}{GenCHiP}; \figwork{arxiv260322435}{CaP-X}; \figwork{arxiv261002459}{OpenRUA}; \figwork{arxiv240605572}{PRoC3S}; \figwork{arxiv260930594}{HuGo}; \figwork{arxiv240619741}{ROS-LLM}\\[3pt]
  \textbf{Online Revision, Offline Development}\\ \figwork{arxiv260816978}{VLCP}; \figwork{arxiv260939018}{URAI}; \figwork{arxiv260616458}{RHO}; \figwork{arxiv260929394}{RACaP}; \figwork{arxiv231214950}{TypeFly}; \figwork{arxiv260115486}{DroneServer}\\[3pt]
  \textbf{Orchestrating Learned Skills}\\ \figwork{arxiv260809410}{HyMeS}; \figwork{arxiv260816590}{Zetta}; \figwork{arxiv260708448}{Harness VLA}; \figwork{arxiv260721725}{Pigey}; \figwork{arxiv260706256}{Semantic handoff}\\[3pt]
  \textbf{Comparison with End-to-End Policies}\\ \figwork{arxiv251003827}{LIBERO-PRO}; \figwork{arxiv241001345}{GemBench}; \figwork{arxiv250104268}{RoboPro}; \figwork{arxiv260920822}{SafeHarness}};
\node[leaf=SecWrite] (l3) at ([yshift=-.3cm]c3.south) {\textbf{Rewards and Objectives}\\ \figwork{arxiv230608647}{Language to Rewards}; \figwork{arxiv231012931}{Eureka}; \figwork{arxiv230911489}{Text2Reward}; \figwork{arxiv240601967}{DrEureka}; \figwork{arxiv261004364}{BIRD}\\[3pt]
  \textbf{Environments, Demonstrations, and Assets}\\ \figwork{arxiv231001361}{GenSim}; \figwork{arxiv231101455}{RoboGen}; \figwork{arxiv260115164}{V-CAGE}; \figwork{arxiv260301505}{FATE}; \figwork{arxiv241015536}{GRS}; \figwork{arxiv240514062}{ChatScene}; \figwork{arxiv230714535}{Scaling Up}; \figwork{arxiv250618088}{RoboTwin 2.0}; \figwork{arxiv251217853}{AnyTask}; \figwork{arxiv241013882}{Articulate-Anything}; \figwork{arxiv260515187}{Articraft}; \figwork{arxiv240708626}{RoboMorph}};
\node[inner sep=0pt,fit=(l1)(l2)(l3)] (toprow) {};
\node[head=SecTest,minimum height=1.45cm] (c4) at ([yshift=-1.15cm]toprow.south -| c1.center) {Testing Programs Against\\the Physical World\\ {\mdseries\S\ref*{sec:evidence}\enspace(RQ2)}};
\node[head=SecLearn,minimum height=1.45cm] (c5) at (c4.center -| c2.center) {Learning from the Physical World\\ {\mdseries\S\ref*{sec:learning}\enspace(RQ3)}};
\node[head=SecLearn,minimum height=1.45cm] (c6) at (c4.center -| c3.center) {Benchmarks and\\Evaluation Practice\\ {\mdseries\S\ref*{sec:measuring}\enspace(RQ3)}};
\node[leaf=SecTest] (l4) at ([yshift=-.3cm]c4.south) {\textbf{Cost of Evaluation}\\ \figwork{arxiv241101775}{Eurekaverse}; \figwork{arxiv241013837}{ORSO}; \figwork{arxiv260428056}{RHyVE}; \figwork{arxiv260511859}{EvoNav}\\[3pt]
  \textbf{Traces and Judges}\\ \figwork{arxiv260700272}{ASPIRE}; \figwork{arxiv260603047}{ModuLoop}; \figwork{arxiv241018893}{RoboRepair}; \figwork{arxiv231111183}{RoboEval}; \figwork{arxiv260903611}{FailBench}; \figwork{arxiv261002527}{CriticHack}; \figwork{arxiv260937771}{Benchmark audit}; \figwork{arxiv260909250}{No Free Checker}\\[3pt]
  \textbf{Checks in Place of Physical Trials}\\ \figwork{arxiv261010479}{Agentic RSR}; \figwork{arxiv260938982}{SimEX}; \figwork{arxiv251202002}{Text simulation}; \figwork{arxiv250307885}{RoboGuard}; \figwork{arxiv260405427}{SafeGate}; \figwork{arxiv260814590}{Safe agents review}; \figwork{arxiv240105443}{LLM4PLC}; \figwork{arxiv260818565}{SemaPLC}; \figwork{arxiv261002982}{PLCWorld}; \figwork{arxiv260105356}{PRISM}; \figwork{arxiv260627231}{ARTIQ gate}\\[3pt]
  \textbf{Safety Beyond Task Success}\\ \figwork{arxiv240720242}{BadRobot}; \figwork{arxiv241216633}{POEF}; \figwork{arxiv251000181}{CHAI}; \figwork{arxiv260600773}{SafeVLA-Bench}};
\node[leaf=SecLearn] (l5) at ([yshift=-.3cm]c5.south) {\textbf{Acquisition and Validation}\\ \figwork{arxiv230516291}{Voyager}; \figwork{arxiv240618746}{LRLL}; \figwork{arxiv250918597}{LYRA}; \figwork{arxiv261002788}{Skill2Real}; \figwork{arxiv260619419}{RATs}; \figwork{arxiv261002717}{RoboBridge}; \figwork{arxiv260817209}{Teach and Grow}; \figwork{arxiv261002204}{RPG}\\[3pt]
  \textbf{Library Maintenance}\\ \figwork{arxiv260814944}{SkillComposer}; \figwork{arxiv260919906}{Software archive}; \figwork{arxiv260304560}{MEMO}; \figwork{arxiv260931760}{123-round study}\\[3pt]
  \textbf{Reuse and Transfer}\\ \figwork{arxiv260700272}{ASPIRE}; \figwork{arxiv260607999}{RECENT}; \figwork{arxiv261002717}{RoboBridge}\\[3pt]
  \textbf{Experience Distillation}\\ \figwork{arxiv260929166}{HarnessPAI}; \figwork{arxiv231201421}{RobotGPT}; \figwork{arxiv240211450}{LMPC}; \figwork{arxiv260929964}{World Action Agent}};
\node[leaf=SecLearn] (l6) at ([yshift=-.3cm]c6.south) {\textbf{Benchmarks and Protocols}\\ \figwork{arxiv260933807}{CodeActionBench}; \figwork{arxiv241218194}{VLABench}; \figwork{arxiv250618123}{RoboArena}; \figwork{arxiv260313616}{N-SCORE}\\[3pt]
  \textbf{Robot Code Repair}\\ \figwork{arxiv261016369}{RoboBugBench}; \figwork{zhao2026rosdeveval}{ROSDevEval}; \figwork{arxiv231006770}{SWE-bench}; \figwork{arxiv260616458}{RHO}\\[3pt]
  \textbf{Limitations of Current Evaluation}\\ \figwork{arxiv240505941}{SIMPLER}; \figwork{arxiv261000731}{Reconstruction study}; \figwork{arxiv250618088}{RoboTwin 2.0}; \figwork{arxiv260120334}{FAEA}; \figwork{arxiv261001939}{PyRUA-Lean}; \figwork{arxiv250324278}{AutoEval}; \figwork{arxiv250705331}{TRI LBM}; \figwork{arxiv250310966}{STEP}; \figwork{arxiv251004354}{SureSim}; \figwork{arxiv260801851}{Weights or Skills?}};
\end{tikzpicture}
\end{adjustbox}
\caption{\textbf{Organization of the article and representative work.} Each panel is one part of the article (\cref{sec:software-to-physical} and \crefrange{sec:writing}{sec:measuring}) and lists representative work discussed there.}
\label{fig:organization}
\end{figure}

%% file: sections/02-foundations.tex

\begin{figure}[!htbp]
\begin{pciinsight}[unbreakable]{Key terms}\label{box:keyterms}
\textbf{Program.} A model-written or model-edited artifact with executable semantics and a structure that can be stored, inspected, changed, and reused. It may specify a procedure, as in Python code or a behavior tree, or declare an objective, task, or world that a controller, planner, or simulator acts on.\par
\textbf{Primitive API.} The functions a program may call: perception (segment the pot, estimate its pose), planning (choose a grasp, solve inverse kinematics), control (move to a pose, close the gripper), and, increasingly, learned policies exposed as callable skills.\par
\textbf{Vision-language-action model (VLA).} A neural policy that maps camera images and an instruction directly to robot actions. We use \emph{end-to-end policy} for VLAs and other learned policies that produce actions without an intermediate program.\par
\textbf{Harness.} The software around the model that assembles its context, exposes the primitive API, executes programs, records what happened, and decides what is kept between model calls.\par
\textbf{Episode and seed.} An episode is one execution of a program or policy from an initial state until it terminates or reaches an evaluation limit. Under a fixed simulation configuration, a seed controls the sampled initial conditions. Held-out seeds test performance on new instances of that configuration.\par
\textbf{Execution trace.} The record an episode leaves behind. It may include primitive calls, arguments, return values, and camera images before and after calls.\par
\textbf{Feedback and judge.} Feedback is a signal about whether a program did what was intended, and a judge decides from it whether the program succeeded. We track six forms of evaluation: \emph{simulator-state predicates}, \emph{model-written checks}, \emph{learned judges}, \emph{formal checks}, \emph{human judgments}, and \emph{checks of the physical outcome}. Predicates read simulator state, which may include quantities the robot cannot observe, model-written checks often come from the model that wrote the program, and learned judges include VLMs and reward models. We also record whether the agent can inspect or query the judge.\par
\textbf{Sim-to-real gap.} The difference between how a program or policy behaves in simulation and how it behaves on the physical robot.\par
\textbf{Skill.} A skill is a unit of validated experience stored for reuse: a function, a code template, or a structured document that says when and how to apply a repair. Before it is stored, a skill usually has to pass an acceptance test.
\end{pciinsight}
\end{figure}

\subsection{Scope and Definitions}
\label{sec:program-boundary}

The ways a language model can direct a robot form a spectrum. At one end the
model chooses one primitive at a time from a fixed menu. SayCan scores
natural-language descriptions of pretrained skills, weights each score by a
learned value function that estimates whether the skill can succeed from the
current state, and executes the best skill \citep{arxiv220401691}. Inner
Monologue appends textual feedback, such as a success detector's output or a
description of the scene, before the model makes its next choice
\citep{arxiv220705608}. Decisions are made step by step, with the dialogue
providing the record of what has happened. At the other end, the model writes
a program that makes those choices when it runs. Code as Policies made this
concrete: a few-shot prompted language model writes Python that queries
perception functions, does geometric arithmetic with NumPy and Shapely, and
calls control primitives. The program may contain its own feedback loops
\citep{arxiv220907753}. It can survive the episode that produced it, run
from another seed, be compared with a previous version, and be stored for later
tasks. A failing routine can be identified and edited. Much of what this
article examines depends on that persistence.

We count an artifact as a program when it has executable semantics and a
structure that can be edited independently of the model that produced it.
Procedural code states how to act, whereas a declarative specification states a goal,
constraint, or environment that a planner, controller, or simulator uses to
determine execution. We include both when the model writes or edits the
artifact and the resulting behavior is evaluated. This covers problem files in
the Planning Domain Definition Language (PDDL) and temporal-logic formulas, as
well as generated scene graphs, task definitions, and robot descriptions used
to construct training or test worlds. A fixed scene description supplied as
input is background context, but a generated scene evaluated for solvability or for the behavior it enables is an object of
this survey. We identify which kind of artifact each method produces, since
editing a Python controller and editing a simulator configuration give the
model different control over what happens.

Two neighboring interfaces, typed tool calls and skill documents, fall outside
this definition. A typed tool call is a single decision executed as soon as it
is made, and a skill document must be interpreted by an agent before anything
runs, so neither leaves an artifact that specifies behavior ahead of time.
We still discuss them alongside programs where they arise
(\cref{sec:orchestration,sec:compounding}). In a \emph{typed tool-call}
interface, the model issues one structured call per turn, often as JSON or
through the Model Context Protocol (MCP), and chooses the next call after
seeing the result. Harness VLA, for example, has a language-model planner issue
JSON calls to analytic primitives and to a frozen VLA exposed as a retryable
primitive \citep{arxiv260708448}. The resulting transcript records choices
already made, whereas a stored procedure that the harness can execute specifies
in advance when to branch and what to repeat, which makes it a program. A
\emph{skill document} gives an agent guidance about when and how to act \citep{arxiv260931847}. It
may record a failure signature, a repair, and a code sketch. ASPIRE stores its skill library in
this form \citep{arxiv260700272}.

The other boundary concerns evaluation. We include work that tests the
artifact's effects through physical execution: on a robot, in a physics
simulator, or through the behavior of a policy trained with its output. Where
physical trial is not permitted, we also include work that substitutes a
simulation, digital twin, or formal model of the plant (the physical process or
equipment a controller acts on), and examine what its results establish about
the plant itself (\cref{sec:forbidden-trial}). Parsing,
type checking, and compilation are useful checks, but passing them alone does
not establish success on a physical task. ViLaIn's generated PDDL problems
are syntactically correct at least 99\% of the time, yet in its hardest domain
only 58\% yield a plan that a symbolic validator accepts
\citep{arxiv231100967}. That gap lies between syntax and symbolic validity, before any robot executes the plan. CodeActionBench separately checks motion
execution and task outcomes, showing that completing a commanded motion need
not complete the task \citep{arxiv260933807}. Industrial control code shows the same gap between static metrics and
runtime behavior (\cref{sec:forbidden-trial}). Each of these levels therefore
needs its own evidence.

\paragraph{Running example}\label{sec:running-example}
A household robot receives the instruction ``put the moka pot on the stove.''
The LIBERO benchmark includes putting
a moka pot on a stove among its kitchen tasks \citep{arxiv230603310}.
An action program for the task might segment the pot in the camera image,
estimate its pose, choose a grasp on the handle, lift, move above the burner,
lower, and release. An end-to-end policy trained on demonstrations maps images
and the instruction directly to motor commands. A step-wise planner chooses
``pick up pot'' and then ``place on stove'' from a skill menu. The three differ
in how their behavior is represented and what can be revised after the episode. The policy retains its learned weights, the planner records its choices in
a transcript, and the program retains explicit logic that can be read and
changed. If the instruction becomes ``put it on the left burner'' and the API can
identify that burner, changing the program's target may be enough. If the
grasp slips, a trace that records the grasp and release can help the agent
locate the failure, revise the grasp routine, and test the repair. Each later section returns
to this example: what the program buys and costs (\cref{sec:action}), what
happens when the robot must also learn to pour coffee and the agent writes a
reward for it (\cref{sec:objective-world}), how it learns that its program failed
(\cref{sec:evidence}), whether its repair survives a different robot
(\cref{sec:compounding}), and how the result should be reported
(\cref{sec:measuring}).

\subsection{From Language Models to Coding Agents}
\label{sec:history}

\Cref{fig:timeline} places the work reviewed here on a timeline. Four
overlapping phases trace how language models moved from choosing skills to
writing programs, then used programs to shape training and to retain
experience, and finally came to revise them through execution feedback.

\input{figures/timeline}

The first phase, in 2022, treated the language model as a planner over a menu
of skills. Zero-shot planners translated free-form plans into admissible
actions \citep{arxiv220107207}, SayCan grounded skill choices in learned
affordances \citep{arxiv220401691}, and Inner Monologue closed the loop with
textual feedback \citep{arxiv220705608}. Their outputs were sequences of skill choices, and
turning a plan into execution still required a mechanism for handling
quantities, loops, and geometric relations. Programs provided that mechanism. In Code as Policies,
the natural-language planner baseline cannot express the spatial-geometric
tasks at all, and on long-horizon tasks with unseen attributes and
instructions the generated programs succeed 80\% of the time against 64\% for
that planner \citep{arxiv220907753}. In ProgPrompt's reported VirtualHome setting, the full program-based method
reaches a success rate of 0.34 with GPT-3 (0.40 with Codex), while the
natural-language GPT-3 baseline completes no task fully
\citep{arxiv220911302}.

The second phase, from 2023 through 2024,
developed programs as policies that could query perception and control the
robot when executed. Beyond Code as Policies and ProgPrompt, it produced design
principles for prompting with function libraries \citep{arxiv230617582},
programs that compose 3D value maps for a motion planner
\citep{arxiv230705973}, programs that glue vision foundation models to
manipulation primitives \citep{arxiv230511176}, and translation of instructions
into formal problems for classical planners \citep{arxiv230411477}. Feedback
in this phase came mostly from humans and symbolic validators. Frontier models
later came to support this way of controlling robots directly: the Gemini Robotics report evaluates
zero-shot robot control through code generation as a capability of its
embodied-reasoning model \citep{arxiv250320020}.

The third phase, from 2023, extended what the model could write beyond
action selection to the objectives and environments used to learn behavior.
Language to Rewards wrote reward parameters for a model-predictive controller
\citep{arxiv230608647}, Text2Reward wrote dense reward programs
\citep{arxiv230911489}, and Eureka evolved reward code for reinforcement
learning, including dexterous manipulation \citep{arxiv231012931}. GenSim and
RoboGen generated simulation tasks together with their training supervision.
In GenSim, executing the task code yielded demonstrations, whereas RoboGen chose a
solver for each subtask and wrote reward functions for those it trained with
reinforcement learning \citep{arxiv231001361,arxiv231101455}.
Gen2Sim generated the objects too: diffusion models lifted 2D images to 3D meshes, GPT-4 supplied plausible sizes and masses, and an LLM proposed tasks and wrote a reward function for each subtask \citep{arxiv231018308}.
Holodeck generated whole 3D environments from language \citep{arxiv231209067}. Later work in this line extended reward search to the randomization of physical parameters for sim-to-real transfer \citep{arxiv240601967} and generated expert programs whose executions become training data \citep{arxiv250618088}.

A program that survived one task could also become a starting point for the
next. In parallel with work on objectives and environments, systems began to
retain code and feedback from earlier tasks \citep{luo-etal-2026-storage,huang-etal-2024-towards}. In the game world of Minecraft, Voyager
stored programs that a GPT-4 critic had judged
successful in a growing skill library \citep{arxiv230516291}. DROC distilled
human corrections into retrievable knowledge \citep{arxiv231110678}, Lifelong
Robot Library Learning (LRLL) grew a library of robot skills in code
\citep{arxiv240618746}, and Language Model Predictive Control fine-tuned a
code-writing model on its own successful teaching sessions with non-expert
users \citep{arxiv240211450}.

Each of these systems wrapped the model in a loop designed for one kind of
artifact. Meanwhile, a second line of work built coding agents for software
repositories. SWE-bench scores fixes of real GitHub issues against tests taken
from the pull requests that resolved them \citep{arxiv231006770}, and SWE-agent and OpenHands give a
model interfaces for editing, running, and testing code in a repository
\citep{arxiv240515793,arxiv240716741}. The fourth phase began in
2026, when such general-purpose coding agents started working inside robot
execution environments. Their ability to inspect files,
run code, and revise it connected program generation to a longer workflow of
testing and repair. CaP-X built a benchmark, CaP-Bench, and a training-free agent, CaP-Agent0,
around programs that call perception and control primitives \citep{arxiv260322435}.
RHO and OpenRUA run Claude Code or Codex against simulated robots, and ASPIRE
does so mainly in simulation and also carries its skills to real robots.
Skill2Real places frontier language models in a loop of proposing, verifying,
and accepting skills and transfers the result to a real arm. All four
report large gains on perturbed manipulation benchmarks
\citep{arxiv260616458,arxiv260700272,arxiv261002459,arxiv261002788}.
ENPIRE extends the loop to physical robots, where coding agents modify training
infrastructure and algorithm code \citep{arxiv260619980}. EmbodiedSWE turns
verified agent solutions into demonstrations for a learned policy
\citep{arxiv260927308}, and a family of
harnesses composes frozen VLAs with code-level memory, checks, and recovery
\citep{arxiv260708448,arxiv260809410,arxiv260929166,arxiv260816590}. In these systems, much of the robot-specific support is supplied by the
harness. As generation became part of a repeated cycle of execution and
revision, agents needed evidence that a proposed program worked and feedback
useful enough to guide the next edit.

\input{sections/08-software-to-physical}

\begin{table}[t]
\centering\footnotesize\setlength{\tabcolsep}{4pt}
\caption{\textbf{Positions on the first axis.} Costs summarize reported settings, with sources in \cref{sec:timescales,sec:evidence-cost}. $^\ast$\,Neighboring system that issues typed tool calls rather than writing a program.}
\label{tab:axes}
\newcommand{\balancedcell}{\RaggedRight\rightskip=0pt plus 1\hsize\parfillskip=0pt\hyphenpenalty=10000\exhyphenpenalty=10000\relax}
\newcolumntype{B}[1]{>{\balancedcell\arraybackslash}p{#1}}
\begin{tabularx}{\linewidth}{@{}L{.11\linewidth}B{.174\linewidth}B{.245\linewidth}B{.18\linewidth}>{\balancedcell\arraybackslash}X@{}}
\toprule
\TableHeader{Position} & \TableHeader{The program is} & \TableHeader{Representative work} & \TableHeader{Typical evaluation} & \TableHeader{Evaluation cost}\\
\midrule
\roleAct\newline Action & A policy over perception, planning, and control primitives & \figwork{arxiv220907753}{Code as Policies}, \figwork{arxiv220911302}{ProgPrompt}, \figwork{arxiv230705973}{VoxPoser}, \figwork{arxiv260322435}{CaP-X}, \figwork{arxiv260700272}{ASPIRE}, \figwork{arxiv260616458}{RHO} & Privileged predicates, learned judges, outcome checks & One episode, seconds to minutes\\[3pt]
\roleOrch\newline Orchestration & Logic that sequences learned policies and checks their results & \figwork{arxiv260809410}{HyMeS}, \figwork{arxiv260816590}{Zetta}, \figwork{arxiv260708448}{Harness~VLA}$^\ast$, \figwork{arxiv260929166}{HarnessPAI}, \figwork{arxiv260721725}{Pigey}$^\ast$ & Learned judges, generated checks, outcome checks & One episode, plus calls to the learned policy\\[3pt]
\roleObj\newline Objective & A reward, cost, or temporal-logic specification & \figwork{arxiv230608647}{Language to Rewards}, \figwork{arxiv231012931}{Eureka}, \figwork{arxiv230911489}{Text2Reward}, \figwork{arxiv240901652}{ReKep}, \figwork{arxiv261004364}{BIRD} & Checks of controller or policy behavior, human ratings & Controller rollouts or policy training, then evaluation\\[3pt]
\roleWorld\newline World & Scenes, tasks, curricula, demonstrations, robot bodies & \figwork{arxiv231001361}{GenSim}, \figwork{arxiv231101455}{RoboGen}, \figwork{arxiv241101775}{Eurekaverse}, \figwork{arxiv250618088}{RoboTwin~2.0}, \figwork{arxiv240601967}{DrEureka}, \figwork{arxiv241013882}{Articulate-Anything} & Solvability checks, learned judges, transfer of trained policies & Scene checks or trial episodes, plus training and transfer tests for downstream claims\\
\bottomrule
\end{tabularx}
\end{table}

\subsection{Program Roles and Evaluation Sources}
\label{sec:axes}

Comparing systems across these conditions requires a common description of
each program and of the evidence behind it. Two properties serve this purpose:
where the program sits relative to the actuators, and how its behavior is judged. We also record whether the agent can see or query
the judge, because a score reached with that feedback means something
different from one reached without it.

The first axis is the program's distance from the actuator. An \emph{action program} calls
perception, planning, and control primitives directly, as in Code as Policies
and its descendants. An \emph{orchestration program} sequences learned
policies, for example a frozen VLA invoked as a skill, and adds memory, checks,
and recovery logic around them. An \emph{objective program} specifies what a
controller or learned policy should optimize: a reward function, a cost
function, or a temporal-logic specification. A \emph{world program} builds
the conditions under which policies are trained and tested: simulated scenes
and tasks, curricula, physical-parameter randomization, demonstration
generators, and descriptions of the robot's body. We also place the neighboring
systems built on typed tool calls or skill documents by the role they support.

Programs farther from the actuator can affect many subsequent actions or
policies. Their cost of evaluation depends on the computation between the
program and its observable effect, which can differ between programs at the
same position. Language to Rewards has a short path from program to effect: the
model sets reward parameters, a model-predictive controller optimizes actions
under that reward in real time, and the behavior the reward produces can be
observed within the same episode \citep{arxiv230608647}. Eureka, which also
writes rewards, has a longer one: it trains a reinforcement-learning policy for
each candidate it evaluates \citep{arxiv231012931}. Eurekaverse generates
terrain curricula and has to train populations of policies to evaluate them
\citep{arxiv241101775}.
\Cref{sec:objective-world} examines these programs and
\cref{sec:evidence-cost} their evaluation costs. \Cref{tab:axes} summarizes
the four positions.

The second axis is what judges the program. The six forms of evaluation
defined in the Key terms box differ in what they observe and what they assume.
Privileged predicates and formal checks give precise answers, but a predicate
can misstate the intended task, and a formal guarantee holds only under its
specification and system model. A generated check is easy to attach to a
program, but when the same model writes both, the two can agree while both
misread the task. Learned judges and humans can both assess observations such as
camera images, though a learned judge's reliability depends on the task and
the observations available. Humans can resolve ambiguities that automated checks
miss, but their judgments need a clear protocol. An outcome
check asks whether the intended physical result occurred, and answers through a
measurement or a judgment.

No Free Checker groups robot-policy verifiers by the source of their judgment:
humans, rules and formal methods, learned models, and signals internal to the
policy or its world model \citep{arxiv260909250}. We use a related grouping, while separating
model-generated checks and making outcome evaluation explicit. The resulting
categories overlap: a human or learned judge may assess the physical outcome,
and a generated check may read privileged state. For an agent that repairs its own programs, feedback must also
support repair. A result accompanied by a useful trace can help locate a failure and guide the next
edit (\cref{sec:evidence}).

\begin{pciinsight}{What the axes let us claim}
A result about a program in the physical world is interpretable only when three facts about it are known: where the program sits relative to the actuator, which judge produced the result, and what access the agent had to that judge. A success rate obtained with a privileged predicate that the agent could read supports a different claim from the same rate obtained with a hidden check of the physical outcome.
\end{pciinsight}

%% file: figures/timeline.tex
\begin{figure}[!tbp]
\centering\def\pcifigworkcolor{SecFound!80!black}
\begin{adjustbox}{max width=\linewidth}
\begin{tikzpicture}[x=1cm,y=1cm,
  phase/.style={pci shadow=#1,draw=#1,line width=1.2pt,fill=#1!10!white,rounded corners=6pt,text=SecFound!75!black,inner sep=5pt,text width=3.55cm,minimum height=1.75cm,align=center,font=\cartoonfont\footnotesize\bfseries,anchor=north},
  body/.style={pci shadow=#1,draw=#1!55!white,line width=.8pt,fill=#1!4!white,rounded corners=6pt,inner sep=5pt,text width=3.55cm,align=flush left,font=\sffamily\scriptsize,anchor=north,
    execute at begin node={\hyphenpenalty=10000\exhyphenpenalty=10000}},
  gap/.style={font=\cartoonfont\scriptsize\itshape,text=FailureRed!80!Ink,text width=3.55cm,align=center,anchor=north},
  swchip/.style={draw=Baseline!70!white,line width=.7pt,fill=white,rounded corners=3pt,inner xsep=4pt,inner ysep=2.5pt,font=\sffamily\scriptsize,text=Ink,anchor=west},
  swbench/.style={swchip,dash pattern=on 2pt off 1.2pt},
  lanelabel/.style={font=\cartoonfont\footnotesize\bfseries,inner sep=0pt,anchor=west}]
\node[lanelabel,text=SecFound!85!black,align=left,anchor=north west] at (-1.92,3.36) {\lhicon{SecFound}{robot}\enspace Language models\\[1pt]\hspace*{6mm}for robots};
\node[lanelabel,text=Baseline!85!black,anchor=north west] at (2.48,3.36) {\lhicon{Baseline}{laptop-code}\enspace Software engineering agents};
\node[swbench] (s1) at (2.48,2.62) {\figwork{arxiv231006770}{SWE-bench} {\color{MutedText}2024}};
\node[swchip]  (s2) at ($(s1.east)+(0.18,0)$) {\figwork{arxiv240515793}{SWE-agent} {\color{MutedText}2024}};
\node[swchip]  (s3) at ($(s2.east)+(0.18,0)$) {\figwork{arxiv240716741}{OpenHands} {\color{MutedText}2025}};
\node[swchip]  (s4) at ($(s3.east)+(0.18,0)$) {\figwork{arxiv260518747}{Code as Agent Harness} {\color{MutedText}(survey)}};
\foreach \a/\b in {s1/s2,s2/s3,s3/s4} {\draw[-{Latex[length=1.3mm]},Baseline!80!black,line width=.6pt] (\a.east) -- (\b.west);}
\node[phase=TimeA] (p1) at (0,0) {Language Models\\Choose Skills\\[2pt]{\sffamily\mdseries{}2022}};
\node[phase=TimeB] (p2) at (4.35,0) {Programs as Policies\\[2pt]{\sffamily\mdseries{}2023--2024}};
\node[phase=TimeC] (p3) at (8.7,0) {Programs Move Away\\from the Actuator,\\Experience Accumulates\\[2pt]{\sffamily\mdseries{}2023--2026}};
\node[phase=TimeD] (p4) at (13.05,0) {Coding Agents in\\the Loop\\[2pt]{\sffamily\mdseries{}2026}};
\foreach \a/\b in {p2/p3,p3/p4} {\draw[pciflow] (\a.east |- p2.east) -- (\b.west |- p2.east);}
\draw[-{Latex[length=2.6mm,width=2.4mm]},SecFound!85!black,line width=1.6pt] (p1.east |- p2.east) -- (p2.west);
\foreach \n/\img in {p1/skill-menu,p2/phase-recipe,p3/diorama,p4/agent-terminal}
  {\node[art img,anchor=south] at ([yshift=4pt]\n.north) {\includegraphics[height=2.0cm]{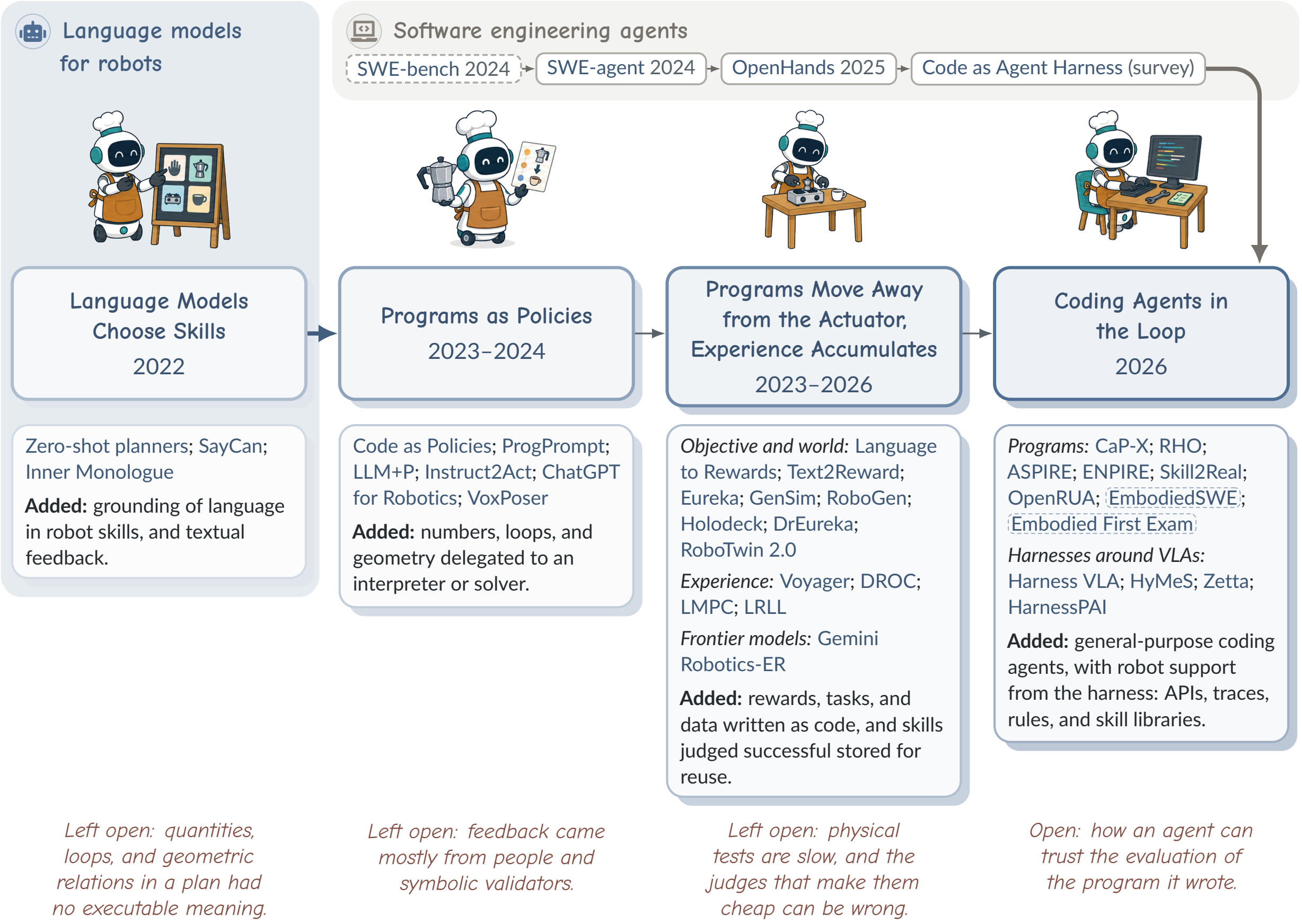}};}
\draw[-{Latex[length=2.6mm,width=2.4mm]},Baseline!85!black,line width=1.6pt,rounded corners=10pt]
  (s4.east) -- (14.62,2.62) -- (14.62,0.02);
\node[body=TimeA] (b1) at (0,-2.1) {\figwork{arxiv220107207}{Zero-shot planners}; \figwork{arxiv220401691}{SayCan}; \figwork{arxiv220705608}{Inner Monologue}\\[3pt]
  \textbf{Added:} grounding of language in robot skills, and textual feedback.};
\node[body=TimeB] (b2) at (4.35,-2.1) {\figwork{arxiv220907753}{Code as Policies}; \figwork{arxiv220911302}{ProgPrompt}; \figwork{arxiv230411477}{LLM+P}; \figwork{arxiv230511176}{Instruct2Act}; \figwork{arxiv230617582}{ChatGPT for Robotics}; \figwork{arxiv230705973}{VoxPoser}\\[3pt]
  \textbf{Added:} numbers, loops, and geometry delegated to an interpreter or solver.};
\node[body=TimeC] (b3) at (8.7,-2.1) {\textit{Objective and world:} \figwork{arxiv230608647}{Language to Rewards}; \figwork{arxiv230911489}{Text2Reward}; \figwork{arxiv231012931}{Eureka}; \figwork{arxiv231001361}{GenSim}; \figwork{arxiv231101455}{RoboGen}; \figwork{arxiv231209067}{Holodeck}; \figwork{arxiv240601967}{DrEureka}; \figwork{arxiv250618088}{RoboTwin~2.0}\\[2pt]
  \textit{Experience:} \figwork{arxiv230516291}{Voyager}; \figwork{arxiv231110678}{DROC}; \figwork{arxiv240211450}{LMPC}; \figwork{arxiv240618746}{LRLL}\\[2pt]
  \textit{Frontier models:} \figwork{arxiv250320020}{Gemini Robotics-ER}\\[3pt]
  \textbf{Added:} rewards, tasks, and data written as code, and skills judged successful stored for reuse.};
\node[body=TimeD] (b4) at (13.05,-2.1) {\textit{Programs:} \figwork{arxiv260322435}{CaP-X}; \figwork{arxiv260616458}{RHO}; \figwork{arxiv260700272}{ASPIRE}; \figwork{arxiv260619980}{ENPIRE}; \figwork{arxiv261002788}{Skill2Real}; \figwork{arxiv261002459}{OpenRUA}; \tikz[baseline=(e.base)]\node[draw=SecFound!55!white,dash pattern=on 1.6pt off 1pt,rounded corners=2pt,inner sep=1pt] (e) {\figwork{arxiv260927308}{EmbodiedSWE}};; \tikz[baseline=(e.base)]\node[draw=SecFound!55!white,dash pattern=on 1.6pt off 1pt,rounded corners=2pt,inner sep=1pt] (e) {\figwork{arxiv261112177}{Embodied First Exam}};\\[2pt]
  \textit{Harnesses around VLAs:} \figwork{arxiv260708448}{Harness VLA}; \figwork{arxiv260809410}{HyMeS}; \figwork{arxiv260816590}{Zetta}; \figwork{arxiv260929166}{HarnessPAI}\\[3pt]
  \textbf{Added:} general-purpose coding agents, with robot support from the harness: APIs, traces, rules, and skill libraries.};
\node[gap] at ([yshift=-0.2cm]b1 |- b3.south) {Left open: quantities, loops, and geometric relations in a plan had no executable meaning.};
\node[gap] at ([yshift=-0.2cm]b2 |- b3.south) {Left open: feedback came mostly from people and symbolic validators.};
\node[gap] at ([yshift=-0.2cm]b3.south) {Left open: physical tests are slow, and the judges that make them cheap can be wrong.};
\node[gap] at ([yshift=-0.2cm]b4 |- b3.south) {Open: how an agent can trust the evaluation of the program it wrote.};
\begin{pgfonlayer}{background}
  \fill[SecFound!14!white,rounded corners=8pt] (-2.1,3.52) rectangle ([yshift=-0.24cm]b1.south -| 2.13,0);
  \fill[Baseline!10!white,rounded corners=8pt] (2.3,2.2) rectangle (15.15,3.52);
\end{pgfonlayer}
\end{tikzpicture}
\end{adjustbox}
\caption{\textbf{Four phases of agentic intelligence for the physical world.} Two lines of work meet here. Language models for robots (left band) moved from choosing skills to writing programs whose effects are judged through physical execution, and general-purpose coding agents from software engineering (top band) joined this line in 2026. Dashed outlines mark benchmarks.}
\label{fig:timeline}
\end{figure}

%% file: sections/08-software-to-physical.tex
\subsection{From Software Tests to Physical Tests}
\label{sec:software-to-physical}

The coding agents now writing robot programs were developed for software work. Several systems reviewed here use them with little or no modification: FAEA uses the Claude Agent SDK as it is \citep{arxiv260120334}, OpenRUA gives Claude Code a terminal \citep{arxiv261002459}, and RHO and ASPIRE run Claude Code and Codex inside harnesses designed for robots \citep{arxiv260616458,arxiv260700272}. Code as Agent Harness describes the apparatus these agents bring, from plan-execute-verify loops and executable checks to sandboxes and permission tiers \citep{arxiv260518747}. The familiar workflow remains, but its tests now have to account for what happens outside the program. This subsection asks which engineering practices transfer and what breaks in the physical world, and \cref{tab:swe-transfer} summarizes the ones that transfer.

\input{tables/swe-transfer}

\label{sec:transfers}

The core loop of the harness changes little: an agent writes code, runs it, reads the output, and edits again. Robot-specific knowledge enters through the components around that loop: the primitive API, the execution trace, the rules about which functions the agent may call, and the skill library. Repository structure and refactoring also matter for how behavior is improved or transferred. RHO edits multi-file policy repositories (\cref{sec:api}), and RECENT rewrites the code that binds a skill to a body (\cref{sec:transfer}).

Several testing practices have direct robot counterparts. The trace checks of \cref{sec:traces} play familiar roles: RoboEval's temporal-logic properties state behavioral requirements as executable checks, as property-based tests do, and are checked in symbolic simulation rather than under physical dynamics. RoboCritics' motion critics act like a linter's warnings, although they read executed motion rather than source code. A test-driven framework refines robot controllers against structured pytest suites in simulation \citep{arxiv260300455}. Once an agent starts revising reusable code, regression testing becomes part of the same workflow. The library mechanisms of \cref{sec:compounding} play this role to different degrees. LRLL re-checks earlier successes after it refactors its library, RPG keeps a revision only if it passes a suite of practice tasks, and HexaAnything's update loop specifies regression tests.

Contracts and hidden tests also have counterparts, with different scopes. Interface contracts check that a generated behavior tree uses only the skills and operators a robot exposes, and safety contracts checked by a satisfiability modulo theories (SMT) solver constrain commands through invariants and guards (\cref{sec:formal}). Hidden evaluation addresses a different concern: CodeActionBench never returns its evaluation result to the agent, and EmbodiedSWE grades solutions in a separate environment where the agent cannot write simulator state \citep{arxiv260927308}.

\label{sec:breaks}

Other practices need more support in the physical world. Software tests can already be incomplete, flaky, expensive, or dependent on external state. Physical execution adds further demands on specification, repeatability, reset, cost, containment, and observation (\cref{fig:what-breaks}).

\input{figures/what-breaks}

\paragraph{Executable specifications} In software, the repository's tests often encode the expected behavior. A physical instruction usually leaves it unstated: how far a drawer must open, where an object must end up, or what counts as safe completion. OpenRUA traces part of one agent's failures to success criteria that the instruction leaves unstated or contradicts (\cref{sec:judges}), so an agent can satisfy its reading of a request and still fail the evaluator. RAPID infers a success predicate for each phase of a single human demonstration before writing the corresponding primitive \citep{arxiv260930249}. It brings the test-first idea to the physical task.

\paragraph{Repeatability} A small isolated software test may give the same result every time, whereas physical outcomes can depend on initial conditions, contact, and sensing or actuation noise. One successful episode shows that the program worked once, and estimating reliability across those variations takes repeated trials. Repeated program generation, repeated execution, and testing in new scenes each measure a different kind of variation. When a deterministic simulation, its seeds, and memory are all fixed, re-running reproduces the same rollout, as \cref{sec:library-harm} shows. Such a rerun confirms that a result reproduces but adds no new samples.

\paragraph{Environment reset} Version control can restore a program, but not a database, an external service, or a robot's surroundings. Physical trials need their own reset procedures, and the quality and cost of that reset become part of the experiment. A reset also decides which failures recur. One that restores only convenient states keeps trials flowing but removes the states in which a program most needs to be tested. Simulation offers a closer counterpart to branching a software experiment: EmbodiedSWE provides an optional checkpoint tree that lets agents return to saved simulator states \citep{arxiv260927308}. Embodied First Exam, a benchmark of coding agents on simulated robots (\cref{sec:benchmarks}), measures what such resets are worth. When GPT-6 Astra in Codex sees only the robot's sensors and by default cannot reset, unlimited resets with saved states raise its success rate on the default tests of 391 core tasks, 50 of them Realtime tests, from 28\% to 53\%. In 66\% of these successes the episode starts from a saved state, so the agent retries a hard stage without repeating the whole task \citep{arxiv261112177}. Hardware offers no such checkpoint, so comparable trials depend on an engineered reset, such as the dedicated verification and automatic-reset routines that ENPIRE constructs for its physical robots \citep{arxiv260619980}. Some experiments cannot be reset at all. In SlowLab, an agent runs a one-year campaign in four simulated greenhouse compartments and recommends one climate policy, which a private evaluator scores in weather the agent never saw. In a preregistered study on 48 unseen sites, three low-cost language-model agents ended near a standard grower setting and below a Bayesian-optimization baseline that starts from that setting. They ran the facility as two synchronous batches, so measurements taken during the season never changed what they planted \citep{arxiv261034644}.

\paragraph{Evaluation cost} Many unit tests are cheap enough to run after every edit, while integration tests and training jobs already cost more. Physical evaluation extends that range. Assessing a reward function through the policy it trains requires a training run, while hardware trials consume robot time and may also require supervision or manual recovery (\cref{sec:evidence-cost}). Sampling many candidate programs remains possible, but the cost of selecting among them depends on how far each candidate must proceed through execution and verification.

\paragraph{Physical containment} Shell access still requires software isolation. RHO reports a preliminary run in which an agent terminated the operator's SSH session and fork-bombed the host \citep{arxiv260616458}, a failure of host protection before any physical effect. Robot access adds a boundary that process isolation does not cover (\cref{sec:safety}). FAEA warns that persistent retrying can be hazardous near people \citep{arxiv260120334}, so robot-side limits need to be assessed under the commands and retries the agent actually produces.

\paragraph{Observability} A software agent can often inspect internal variables and instrument the code it controls. A physical agent must infer relevant state from sensors unless its interface supplies more direct measurements. Exact simulator state can make debugging easier while giving the agent information unavailable on hardware, and an agent that can read it may also exploit it as evidence. Control can stay hard even with such access: FAEA supplies ground-truth object positions through its interface, and its agent still fails on precise peg insertion \citep{arxiv260120334}. In the same Embodied First Exam runs, a query for the exact state of every object raises success to 45\%, and the query with resets to 68\%, against 77\% with full access to the simulator. At stages that need precision, the query alone solves 21\% of the 91 failures of the sensor-only run, about as many as a rerun that only hides the success signal (19\%), whereas resets solve 36\% \citep{arxiv261112177}. Skill2Real excludes privileged information from stored skills \citep{arxiv261002788}.

%% file: tables/swe-transfer.tex
\begin{table}[!tbp]
\centering\footnotesize\setlength{\tabcolsep}{4pt}
\caption{\textbf{Software-engineering practices that transfer to physical agents} (\cref{sec:transfers}).}
\label{tab:swe-transfer}
\newcommand{\balancedcell}{\RaggedRight\rightskip=0pt plus 1\hsize\parfillskip=0pt\hyphenpenalty=10000\exhyphenpenalty=10000\relax}
\begin{tabularx}{\linewidth}{@{}>{\balancedcell\arraybackslash}p{.15\linewidth}>{\balancedcell\arraybackslash}X L{.455\linewidth}@{}}
\toprule
\TableHeader{In software} & \TableHeader{Physical counterpart} & \TableHeader{Examples}\\
\midrule
Agent harness & Harness that sets the robot interface, the feedback, and access to privileged state & ASPIRE, RHO, OpenRUA\newline\tablecite{arxiv260700272,arxiv260616458,arxiv261002459}\\[2pt]
Stack trace and logs & Per-primitive traces with keyframes and overlays & ASPIRE\newline\tablecite{arxiv260700272}\\[2pt]
Repository and refactoring & Multi-file policy repositories and rebinding skills to a new body & RHO, RECENT\newline\tablecite{arxiv260616458,arxiv260607999}\\[2pt]
Executable behavioral tests & Symbolic trace properties and controller test suites in simulation & RoboEval, test-driven framework\newline\tablecite{arxiv231111183,arxiv260300455}\\[2pt]
Regression gate & Practice suites, validation against the current version, and regression tests before accepting a skill or harness edit & RPG, Skill2Real, HexaAnything\newline\tablecite{arxiv261002204,arxiv261002788,arxiv260935432}\\[2pt]
Contracts & Skill schemas, allowed operators, and encoded safety constraints & Contract-grounded BTs, SafeGate\newline\tablecite{arxiv260712220,arxiv260405427}\\[2pt]
Hidden tests & Verifiers never returned to the agent and grading in a separate environment & CodeActionBench, EmbodiedSWE\newline\tablecite{arxiv260933807,arxiv260927308}\\
\bottomrule
\end{tabularx}
\end{table}

%% file: figures/what-breaks.tex
\begin{figure}[!tbp]
\centering\def\pcifigworkcolor{SecFound!80!black}
\begin{tikzpicture}[x=1cm,y=1cm,
  bt/.style={lh text,anchor=north west}]
\newcommand{\wpanel}[9]{%
  \lhpanel{#1}{#2}{5.22}{4.0}{SecFound}{#3}{#4}{#5}
  \node[art img] at (#1+1.08,#2+1.95) {\includegraphics[width=1.95cm,height=1.95cm,keepaspectratio]{#6}};
  \node[bt] at (#1+2.12,#2+3.0) {\lhb{Baseline}{laptop-code}{2.4cm}{\textcolor{Baseline}{\textbf{Software}}\enspace #7}\\[5pt]\lhb{SecFound}{robot}{2.4cm}{\textcolor{SecFound}{\textbf{Physical}}\enspace #8}};
  \node[lh eg] at (#1+0.22,#2+0.32) {#9};}
\wpanel{0.02}{4.2}{1}{Executable Specifications}{What counts as done?}{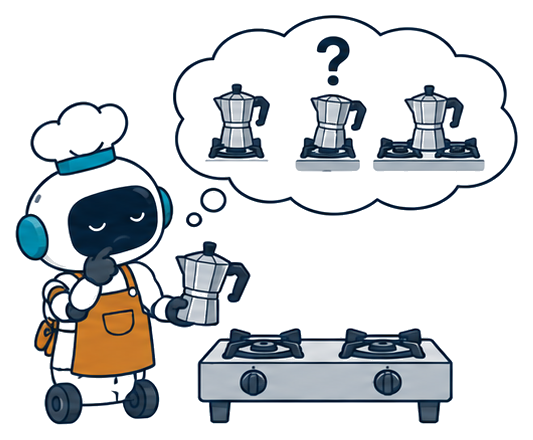}
  {tests capture some expected behavior}
  {instructions leave outcome details unstated}
  {e.g.\ \figwork{arxiv261002459}{OpenRUA}, \figwork{arxiv260930249}{RAPID}}
\wpanel{5.47}{4.2}{2}{Repeatability}{Does a pass repeat?}{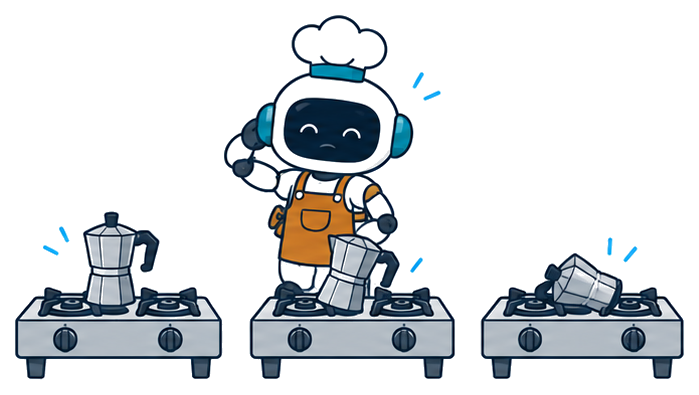}
  {isolated tests may repeat}
  {outcomes can depend on contact, noise, and initial state}
  {see \cref{sec:unreported}}
\wpanel{10.92}{4.2}{3}{Environment Reset}{Can the trial start over?}{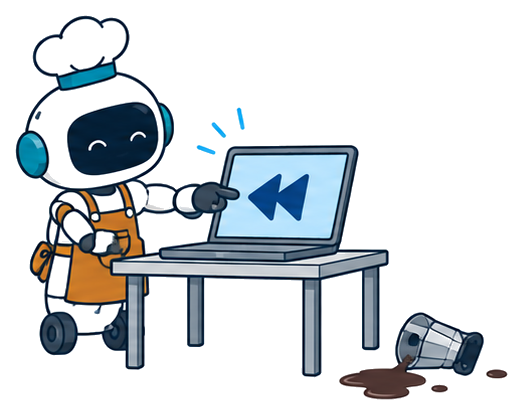}
  {version control restores the code}
  {the scene stays as the last trial left it}
  {e.g.\ \figwork{arxiv260619980}{ENPIRE}, \figwork{arxiv260927308}{EmbodiedSWE}}
\wpanel{0.02}{0}{4}{Evaluation Cost}{What does one evaluation cost?}{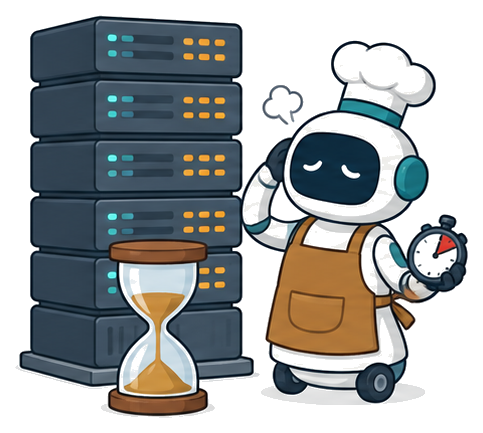}
  {many unit tests run after every edit}
  {an episode, a training run, or robot time}
  {see \cref{sec:evidence-cost}}
\wpanel{5.47}{0}{5}{Physical Containment}{What bounds a bad command?}{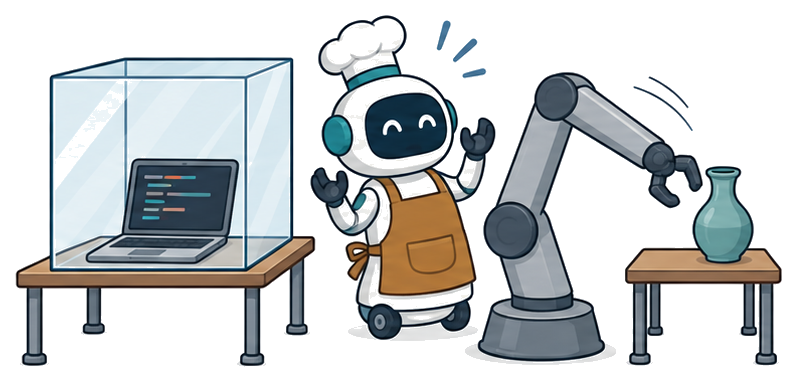}
  {sandboxes and process permissions}
  {motion needs limits at the action interface}
  {e.g.\ \figwork{arxiv260616458}{RHO}, \figwork{arxiv260120334}{FAEA}}
\wpanel{10.92}{0}{6}{Observability}{What can the agent see?}{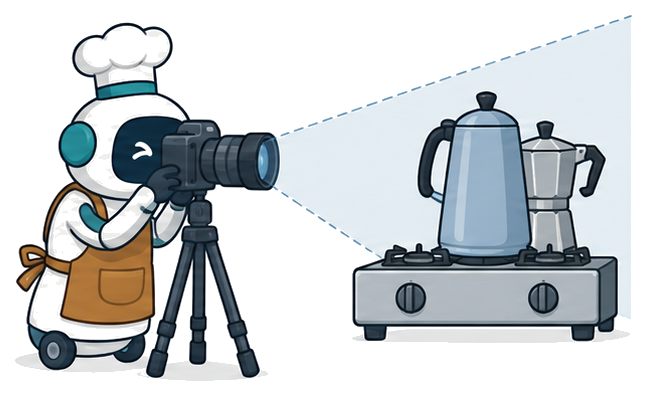}
  {internal variables can often be inspected}
  {state is inferred from sensors}
  {e.g.\ \figwork{arxiv260120334}{FAEA}, \figwork{arxiv261002788}{Skill2Real}}
\end{tikzpicture}
\caption{\textbf{Where software assumptions need support in the physical world.} A physical task needs an explicit success criterion \textcircled{\scriptsize 1}, an account of how outcomes vary between trials \textcircled{\scriptsize 2}, a reset procedure \textcircled{\scriptsize 3}, a budget for costly tests \textcircled{\scriptsize 4}, limits on unsafe commands at the action interface \textcircled{\scriptsize 5}, and a statement of what the agent can observe \textcircled{\scriptsize 6}.}
\label{fig:what-breaks}
\end{figure}

%% file: sections/03-programs-as-action.tex
\section{Writing Programs for the Physical World}
\label{sec:writing}

A program gives an agent behavior that persists: it can be run again,
inspected when it fails, repaired, and stored. In the reported comparisons this
pays off when the structure of a task changes. The primitives a designer
supplies, however, bound what the code can do, and repeated model calls add
time and cost.
This section takes up \pciRQ{1} by weighing the gains against the costs.
Surveys of vision-language-action models review end-to-end policies in depth
\citep{arxiv240514093,arxiv250701925}, and here they serve as the point of
comparison for programs. Programs enter the loop in two places.
Some act on the robot, calling perception, planning, and control primitives or
learned policies (\cref{sec:action}). Others shape how a policy is learned, by
defining its objective, its training world, its demonstrations, or its
simulated body (\cref{sec:objective-world}). The two differ in how far the
program sits from the actuator, and therefore in how directly its effects can
be observed.

\subsection{Programs as Robot Policies}
\label{sec:action}

We begin with programs that sit closest to the actuator (\cref{fig:action-overview}), proceeding from what these programs can call to the conditions under which they outperform an end-to-end policy.

\input{figures/action-overview}

\subsubsection{Primitive Interfaces and Program Structure}
\label{sec:api}

A program is only as capable as the interface it calls. Code as Policies already
stated the basic limitation plainly: its programs are restricted by what the perception
functions can describe and by which control primitives are available
\citep{arxiv220907753}. Later work turned the observation into measurements.
GenCHiP reparameterized the action space so that generated programs could set
compliance, the stiffness with which the arm yields to contact, and reported
that, together with object pose estimation, this improves policy generation on
high-precision, contact-rich subtasks by more than three times on the
Functional Manipulation Benchmark and four times on the NIST Task Board,
relative to non-compliant action spaces \citep{arxiv240406645}. Language to Rewards instead let the model specify reward parameters and a
model-predictive controller optimize actions under that reward. It solved 90\%
of its designed tasks, against 50\% for a baseline that used Code as Policies
primitives as the interface \citep{arxiv230608647}. CaP-X varied the level of
abstraction directly. Across the models it tests, performance improves when
designers supply higher-level primitives and degrades as these are removed,
and no model, writing a program in one pass, matches the human-written programs
that average 88.5\% on its single-turn tiers \citep{arxiv260322435}.\looseness=-1

OpenRUA removes almost all of these primitives. It gives an unmodified coding agent little
more than a terminal, the robot's ROS~2 interface (the middleware through
which most research robots expose their sensors and motion commands), a short
description of the machine's sensors and limits, the ROS documentation, and a
few starter scripts \citep{arxiv261002459}. In most episodes the agent writes
its own code for metric measurement, image processing, and motion control.
With Claude Code and Claude Opus~5 it reaches 99.0\% on CaP-Bench. It is also
tested on RoboCasa365 and on LIBERO-PRO, the benchmark behind many comparisons
in this part. LIBERO-PRO takes LIBERO's four standard evaluation suites
(Spatial, Object, Goal, and Long, also called LIBERO-10) and perturbs each suite's objects,
positions, instruction wording (paraphrase), task, or environment. We call one
suite under one perturbation type a cell. LIBERO-PRO's own evaluation
covers six VLAs \citep{arxiv251003827}. Three of them, OpenVLA, $\pi_0$, and $\pi_{0.5}$, recur below as
end-to-end baselines. With Claude Sonnet~5 at high effort in place of Opus~5,
OpenRUA reaches 24.0\% on the two LIBERO-10 cells under position and task
perturbation, against 72.5\% with Opus~5 \citep{arxiv261002459}. Together,
CaP-X and OpenRUA suggest a substitution: a primitive API stands in for
capability that the model lacks, and a model strong enough to write the missing
abstractions itself needs fewer of them from the designer. The two studies
differ in benchmarks and protocols, however. Embodied Skill Bench tests the
substitution within a single study. It packages 250 action, perception, and experience skills
behind a common interface. On 100 perturbed LIBERO-PRO tasks, per-task skills
selected from this library raise a weaker model from 11\% to 63\%, while in a
separate run, skill-only execution lowers the strongest model from 92\% to
61\% \citep{arxiv261016238}.\looseness=-1

Most of the evidence above comes from tabletop arms. Other bodies make the
limits of the API more visible. For legged robots and humanoids,
programs typically sit on top of a learned low-level controller. A quadruped
system turns plans into code over skills trained with reinforcement learning
\citep{arxiv240405291}, and HuGo has a model write high-level policy code on top of
a frozen whole-body controller and transfers it to hardware
\citep{arxiv260930594}. ELEVATE lets the coding agent supply missing low-level
capability itself. When repeated failures of a task program on a quadruped
with an arm point to a missing whole-body skill, Claude Code with Opus~5.5
writes the reward, curriculum, and success criteria for a new controller,
trains it in simulation from templates the framework supplies, and composes it
into the program. On the five of its seven tasks that need reach beyond the
initial controller, success is 98.4\% to 100\% where coding-agent baselines
with a fixed controller library fail, although total compute is not matched.
The programs and controllers transfer to a real Go2 with a PiPER arm in 18 and
16 of 20 trials on two tasks \citep{arxiv261038388}. BiGym~2.0 compares programs developed by coding
agents with VLA fine-tuning, imitation learning, and reinforcement learning
from demonstrations on simulated humanoid household tasks. The agent programs
beat every demonstration-driven reinforcement-learning baseline on the nine-task mean and trail the fine-tuned VLA \citep{arxiv261007594}. Program generation has also reached underwater manipulation
\citep{arxiv260923133} and construction machinery, where generated behavior
trees coordinate an excavator and a dump truck \citep{arxiv260201041}. For
groups of robots, GenSwarm generates and deploys white-box code
policies \citep{arxiv250323875}.\looseness=-1

Across these bodies, the knowledge specific to the embodiment lives mostly
outside the model: in the primitive API, the language, the contract, and the
check that decides what may run. Robot software makes this layer explicit.
ROS-LLM turns model output into sequences, behavior trees, or state machines
executed as ROS actions and services \citep{arxiv240619741}. RoboScript runs
the same generated code in the Gazebo simulator and on real arms through one
interface \citep{arxiv240214623}, which lets simulation serve as a check before
hardware (\cref{sec:sim-proxy}), and contract-grounded synthesis exposes a
robot's skills and permitted operators to a coding agent through an MCP server
and rejects behavior trees that violate them \citep{arxiv260712220}. RoboCodeX instead moves part of this knowledge into the model. It fine-tunes a multimodal model to write code that sets each object's part-level affordances, grasp choices, and motion-planning parameters for execution through ROS and MoveIt in Gazebo. Over 50 trials per task it reaches 0.80 on pick-and-place against 0.63 for GPT-4V. On long-horizon tasks, the main errors of both models come from grasp-pose selection and trajectory planning \citep{arxiv240216117}. We return
to what this means for transfer across robots in
\cref{sec:transfer}.\looseness=-1

The shape of the program matters as well as its vocabulary. In the RHO
ablation of program structure, multi-file repositories reach 70.00\% on
Robosuite, single files 63.86\%, behavior trees (tree-structured controllers
whose nodes succeed, fail, or keep running) 33.86\%, and finite-state
machines 18.14\%, although the behavior-tree and state-machine rows were
produced with different backend models \citep{arxiv260616458}. Constrained
languages can still serve small models well. BTGenBot-2 fine-tunes a model with one
billion parameters to emit behavior trees in XML and reports up to 90.38\%
zero-shot success on 52 tasks in Isaac Sim with its error-recovery variant
\citep{arxiv260201870}. In the contract-grounded synthesis described above, adding
composition templates to the contract raises a local model from 30 to 50 of
60 tasks \citep{arxiv260712220}. The language can also
be designed for the runtime: TypeFly writes drone plans in MiniSpec, a compact
language with a streaming interpreter, and reduces response time by up to
62\% \citep{arxiv231214950}.\looseness=-1

Programs also gain by delegating work to solvers and interpreters.
In the comparisons we found,
code wins over plans in words where a task involves numbers, geometry, control
flow, or search. These are the parts a program can hand to an interpreter or a
solver. PRoC3S has the model write a function with open
parameters, samples values for them, and tests each candidate in simulation
against kinematic, collision, grasp, and placement constraints, returning the
most common violations to the model only when sampling finds no valid values
within a fixed budget
\citep{arxiv240605572}. In Code-as-Symbolic-Planner, the generated code
itself searches and checks candidate plans, and the method reports an average gain of
24.1\% over the best baselines across seven task-and-motion planning tasks and
three LLMs \citep{arxiv250301700}. ReKep expresses manipulation constraints as Python
functions that map 3D keypoints to costs and leaves the motion to an optimizer
\citep{arxiv240901652}. Translating instructions into formal problems for a
classical planner follows the same logic \citep{arxiv230411477}. NeSyRo checks generated code against the task specification with an SMT solver, uses a PDDL planner to test whether each skill's preconditions hold in what has been observed, and writes probe code to acquire missing observations before acting. On a real Franka arm under partial observability it reports 57.5\% average success against 10.6\% for Code as Policies, with 7 irreversible actions against 53 \citep{arxiv251021302}. The limits
of delegation are visible too. A systematic study of language models inside
task-and-motion planning, covering 16 algorithms in 13,750 evaluations, finds
lower success and longer planning times than engineered systems, and finds that
adding geometric detail to the prompt increases task-planning errors
\citep{arxiv251000182}. These results suggest that delegation works best when the model specifies
what to compute and a reliable runtime supplies the calculation. Such a
program still needs to respond if a grasp fails.
Whether that response is already in the program or requires another model
call is the next design choice.\looseness=-1

\subsubsection{Online Revision and Offline Development}
\label{sec:timescales}

When a program's logic fails during an episode, the agent can revise it on
the spot, at the price of further model calls, or leave the revision for later.
The choice is easier to see once two feedback loops are told apart. A fixed program can read sensors
and adjust its actions during execution, and the model need not be called again
for every correction. A second loop revises the program itself when its logic
fails. If the moka pot slips, an existing recovery branch may handle it. If
no such branch exists, the agent needs to inspect the failure and change the
grasp routine.\looseness=-1

VLCP tests this second loop. A frozen VLM writes a short Python control
function, then re-observes the scene and rewrites the function at intervals
within the same episode. The function receives fresh observations at every
control step even between revisions. Across a 57-task simulated sweep,
within-episode rewriting yields 35.1\% success, against 3.5\% for the same
system queried once per episode. Its trace analysis reports recovery in
27.3\% of episodes containing a failed grasp \citep{arxiv260816978}. Both
conditions receive sensor feedback, so the difference reflects revising the
code during the episode, together with the model calls that revision requires. URAI likewise compares
when the model makes decisions with the same tools: a program written in
advance reaches 24\%, and two agents that decide after each tool call reach
56\% pooled (60\% and 52\%) \citep{arxiv260939018}.\looseness=-1

Repeated model calls add substantial time and cost. RACaP measures a median
policy wall time of 845 seconds and a mean of 0.68 US dollars per episode for CaP-X on
LIBERO-PRO \citep{arxiv260929394}. World Action Agent averages 150 seconds
and 31 model calls per episode \citep{arxiv260929964}, HexaShell needs a median of 407 seconds per RoboDojo episode with task skills and 1,020 without \citep{arxiv261023479}, while a locally
running coding agent needs a median of about 4 to 43 minutes per task,
depending on the task \citep{arxiv260926499}. These are episode or task
durations, not the delay of a single control correction. Simulation can absorb
such deliberation by waiting for the agent: OpenRUA pauses its simulator
while the agent reasons \citep{arxiv261002459}. On hardware, that pause is
feasible only if the robot can hold a safe state and the task allows it. The
relevant cost is the whole episode, so the failures and retries a design adds
count against the model calls it saves. In Code-as-Monitor, a VLM writes monitoring code once at the start of each subgoal that checks spatio-temporal constraints on tracked geometric elements, and execution halts for replanning only when a check fails. In simulated CLIPort block stacking with a drop probability of 0.3, the hardest of its drop settings, this reaches 88.33\% success in 25.4 seconds against 76.67\% in 34.2 seconds for DoReMi, which queries a VLM throughout execution. Its checks target failures visible as displacement or rotation, and the authors note that they are less effective for force-related failures without visible displacement \citep{arxiv241204455}.\looseness=-1

The body also sets the budget for deliberation. A tabletop arm may hold still
while a model inspects a failed grasp, but a drone must keep flying. For drones, the cost of a wrong command is high. ChatGPT for Robotics included aerial navigation among its
early domains \citep{arxiv230617582}, and TypeFly's streaming language was designed for this budget \citep{arxiv231214950}. Aerial systems therefore
move refinement into simulation \citep{arxiv250701930} or restrict the model to
validated tool calls at run time \citep{arxiv260115486}. \Cref{sec:sim-proxy,sec:safety}
discuss what these checks establish.

One way to reduce runtime deliberation is to develop the program offline.
RHO uses coding agents to mutate policy repositories during training and
deploys the result as a single-turn program, with no code-editing model calls
at test time. It reaches 45.0\% on LIBERO-PRO \citep{arxiv260616458}. MEMENTO
also evolves programs offline, scored by an evaluator program that it evolves
first. With the same model (GPT-5.2), evaluator, and budget of evaluated
candidates, it reaches 0.97 success on a four-cube Tower of Hanoi task in
Robosuite, against 0.53 for an adapted REvolve and 0.00 for an adapted Eureka.
With simulator state replaced by RGB-D object estimates and a few execution
constants recalibrated, its program completed 9 of 10 trials on a real Franka
arm \citep{arxiv260722832}. Coding
agents for generalized task-and-motion planning similarly write programs
that are frozen and run on unseen instances. On the 16 environments with a
planner available, the three agent configurations reach mean success of 56\%
to 95\%, against 47\% for the planners. On the environments with multiple object counts, programs
from Opus~5 and GPT-6 Astra take at most 2.1 seconds per instance on average
against 29 for the planners, staying fast as the number of objects grows
\citep{arxiv260930233}. Coding agents can also specialize an existing planner.
Late Autumn runs six autonomous research campaigns, with four models in three
agent harnesses, on a 9,000-line C++ multi-robot planner for a simulated floor
modeled on one existing automated warehouse. The best patches raise throughput
by 22\% to 60\% on hidden evaluation seeds, beat the baseline on all 20 fresh
seeds, and keep 57\% to 91\% of their gain on an alternative layout
\citep{arxiv261102391}.\looseness=-1

Embodied Turing Machines go furthest in this direction.
Coding agents grow one shared library offline, debugging against simulator
ground truth, and at test time the code alone tracks state from images and
proprioception and makes every decision, with no VLM or VLA call. It reaches
70.24\% overall \citep{arxiv261012369} on the bimanual simulation tasks of
RoboDojo \citep{arxiv260704434}, and the authors estimate that an episode uses
0.8--2.4\% of the compute of a VLA \citep{arxiv261012369}. RAX shows how
such development scales with exploration. Its coding agent, Claude Code with
Opus~5.5 or Codex with a GPT-6 model, explores training scenes without action
demonstrations, rehearses its program alone on a fresh scene, and submits a
frozen program that is evaluated without the agent. On seven RoboDojo tasks,
the score on unseen layouts, which gives partial credit on three of them, mostly rises with the number of exploration episodes.
Success at plugging in a charger goes from 26.0\% with 4 episodes to 94.0\% with 8, while
the score for hanging mugs stays between 33\% and 39\% at every budget. With the largest
budget for each task, the programs average a score of 78.3\%, against 24.9\% for the best
demonstration-trained policy on the RoboDojo leaderboard, though RAX is scored on
30 to 50 official layouts it never explored rather than through the
leaderboard's own evaluation. Once
the program takes over, the time spent generating actions on one of these tasks
falls from 17 minutes for the first successful episode to about 6 seconds per episode
\citep{arxiv261022496}.

Offline development has also been carried out on hardware, where every trial
costs robot time. ACE-CaP-0 has Claude Opus~5.5 develop programs on two real
arms with at most ten human-assisted scene resets per task. In its study of
charger insertion, with one development run per condition, robot
demonstrations that the agent may request during development yield a frozen
program that succeeds in 26 of 30 trials, against 16 of 30 when the same
guidance is supplied at the outset. A program developed in simulation and then
transferred succeeds in no trial and fails at grasping the plug, whereas
simulation that the agent consults on demand during real-world development
brings the frozen program to 100\%. Across eight tasks, the frozen programs succeed in 221 of 240 trials
with no human or agent in the loop, at mean execution times of 0.97 to 2.27
minutes \citep{arxiv261116872}. PyRUA-REAL records how long such development
takes on a station with two Franka arms. The time to a first success ranges
from about 1.5 hours for block storage to about 45 hours for Push~T, and
Ethernet insertion was not solved without an online agent after 240 hours.
Once frozen, its programs for tic-tac-toe, five-block stacking, and Tower of
Hanoi complete a cycle in medians of 8.99, 2.99, and 16.72 minutes, and
neither of the two online agents it compares completes Hanoi within 120
minutes \citep{arxiv261106768}.

Offline
development can also coexist with online reasoning. RACaP confines persistent
code changes to offline evolution, then uses a reasoning loop to call the
resulting typed APIs at deployment. On LIBERO-PRO, its median policy wall
time per episode is 380 seconds against 845 for CaP-X, and its success is
45.0\% against 13.3\%
\citep{arxiv260929394}. By contrast, Agent as Policy keeps the coding agent itself online on real
arms. Starting without task experience, it reports at least 80\% success in
seven of eight configurations. In a separate experiment, reusing saved
experience documents, which hold settings, corrections, and scripts, shortens
repeated runs \citep{arxiv260912541}. Offline development amortizes some of
the cost, while runtime model calls still have to fit the task's time budget.
\Cref{sec:compounding} takes up what the slow loop retains and when the
retained routines remain useful.

\subsubsection{Orchestration of Learned Skills}
\label{sec:orchestration}

Programs can also manage learned skills, with the policy supplying motor
skills and code handling memory, task progress, and recovery. Systems differ
in whether this orchestration is written as a program or issued as typed tool
calls.\looseness=-1

Some systems keep the orchestration itself in code. HyMeS states the division
in its title: skills
are kept in the weights of a VLA, and memory is kept in code. A coding agent
maintains an executable memory program that steers a frozen VLA through tasks
whose correct action depends on what happened earlier, raising task success on
a 12-task protocol from the RoboMemArena memory benchmark from 41.3\% for the
VLA alone to 60.1\% and on a real arm
from 25.7\% to 57.1\% \citep{arxiv260809410}. CodeGraphVLP keeps task progress
in a code-based planner over a semantic graph and passes subtasks to a VLA
\citep{arxiv260422238}. CodeDiffuser lets generated code resolve an ambiguous
instruction into a 3D attention map that conditions a diffusion policy
\citep{arxiv250616652}. Zetta evolves code-level critics and recovery skills
around a frozen policy. Under task and position perturbations, it reports a
macro-average of 71.13\% over the Goal and LIBERO-10 suites of LIBERO-PRO, against 32.00\% for the frozen $\pi_{0.5}$ it wraps \citep{arxiv260816590}.
The delegate need not stay frozen. Robo-COP fine-tunes the VLA that its
orchestrator calls on skill segments curated from its own executions,
including correct skills inside failed episodes, and adopts a new policy only
after a model reviewer confirms the skills it was trained for and checks show no loss of task success or of skills admitted earlier. Held-out success on ten
RoboLab tasks rises from 64.8\% with the frozen policy to 73.8\%, whereas
fine-tuning on a fixed schedule without verification reaches 65.8\%
\citep{arxiv261009228}.\looseness=-1

The harness itself can be designed automatically. In RoboArchitect, a coding
model reads the execution traces of a frozen policy, diagnoses its failures,
and revises the control code around it, and a revision is accepted only if it
outperforms the current harness on the training episodes. Averaged over all
its pairs of benchmark and policy, in simulation, the held-out score rises
from 9.7\% to 43.3\% on unseen scenes and from 12.7\% to 40.2\% on unseen
tasks. These scores are success rates, except on VLABench, where they are
progress scores \citep{arxiv261042141}. In RoboDoctor, an agent corrects a frozen world action model at selected
steps during three warm-up episodes, after which a threshold on the warm-up
score fixes for each task whether the agent keeps correcting the model or takes over control. RoboDoctor records the
reason for each decision, the method tried, its outcome, and the conditions
for reuse. Over 54 RoboDojo task entries its mean score is 57.10 after the
warm-up episodes and 49.74 during them, in an archive that combines source
runs whose agent version and budget are not uniform
\citep{arxiv261035604}.

Other systems reach the learned policy through typed tool calls. The orchestrator issues a structured call, such as a JSON object or an MCP request,
whose arguments include a natural-language subgoal for the VLA. Harness VLA
exposes a frozen VLA as a retryable primitive in this way \citep{arxiv260708448}.
Pigey has a VLM emit one schema-typed tool call per step, choosing between a
frozen $\pi_{0.5}$ and analytic grasp and place tools. With its strongest
reasoner, it raises success over six LIBERO-PRO cells from 12.8\% for the
policy alone to 53.3\% \citep{arxiv260721725}, and
Gemini Robotics 1.5 makes its VLA available to an embodied-reasoning model as a
tool that receives instructions in natural language \citep{arxiv251003342}.\looseness=-1

Whether code or structured calls work better as the orchestration interface has
rarely been tested directly. ALRM runs ten language models in a
code mode and nine of them in a tool-calling mode. Most large models score
higher with tool calls (GPT-5 is the exception), usually at higher latency, and models of up to 8B parameters
score higher with code. These scores measure the quality of the generated
action sequences. The programs run in a lightweight environment whose motion
and perception APIs return placeholder poses, LLM judges compare each sequence
with a reference answer, and only the reference answers were validated in
Gazebo \citep{arxiv260119510}. PyRUA-Lean, by contrast, executes both agents in simulation: it gives
a code-cell agent and a tool-calling agent the same budget of model calls on 700 simulated task
instances and reports 71.7\% success against 63.1\%, with less than half the
estimated cost per episode (\$0.74 against \$1.63) on the instances both solve, from one run per
instance \citep{arxiv261001939}. SO-101 Bench holds the model and the
episodes fixed. On 60 paired episodes, GPT-6.1 Sol succeeds on 21 with direct
end-effector tool calls, 40 when it writes code in the PyRUA-Lean harness, and 51
when that code hands grasps to a VLA, which succeeds on 32 by itself, though the
tool-call condition also uses a different harness \citep{arxiv261036420}. For
digital agents, CodeAct reports up to 20\% higher success for executable Python
than for JSON or text actions \citep{arxiv240201030}. Most of the 2026 harnesses we reviewed adopt one interface without measuring
the other.\looseness=-1

Whichever interface is used, composition introduces a new failure point. A placement skill that works
from a clean starting state may fail after a grasp leaves the moka pot tilted
or partly occluded. A study of agent-composed VLA skills on BEHAVIOR-1K
finds high success for isolated skills, including 96.5\% for one picking
skill, but only 19.5\% mean reference-plan progress across ten chained tasks.
The latter is a diagnostic measure scored manually from rollout videos
and differs from official task success. The study identifies a mismatch between
curated skill-start states and states reached during composition as a major
source of failure \citep{arxiv260706256}.\looseness=-1

A natural-language subgoal alone does not specify the state a skill requires
or the state it should leave for the next one. In software terms, composition
needs explicit preconditions and postconditions, together with checks that
can establish them. HexaAnything makes such checks more informative with typed outcomes
(pass, fail, insufficient evidence, blocked, safety stop) \citep{arxiv260935432}. RoboDuo has a coding agent write a two-layer program around two frozen VLAs. One attempts the whole task, a fine-tuned completion check examines the result, and on failure the other works through the remaining steps, with up to three retries each. On eight LIBERO-PRO cells this raises its $\pi_{0.5}$ checkpoint's mean from 32.05\% to 39.33\% \citep{arxiv261034108}. PhyAgentOS separates the end of an execution from acceptance of the task. Its verifier compares the initial and terminal observations and returns success, failure, or replan. On the 50 RoboCasa365 target tasks in simulation, intervention by the verifier after a failed episode raises a frozen $\pi_{0.5}$ from 17.6\% to 26.8\% over 250 episodes. The intervention is either a recovery instruction executed without resetting the scene or a reversal of a completion signal that the verifier judges to be a false negative, and the paper does not separate the two \citep{arxiv260716636}. BATON has a verifier agent check a readiness condition against the wrist-camera view before each call to the frozen VLA. It also records an entry condition at each boundary between subtasks and restores the scene when the previous subtask has disturbed it. Built on the same VLA, coding model, and exploration budget as Harness VLA, it raises average task success over 26 long-horizon RoboMemArena tasks from 23.1\% to 63.5\%. On a real PiPER-X arm, over 20 trials on each of two placing tasks, it averages 47.5\% against 27.5\% for Harness VLA and 7.5\% for $\pi_{0.5}$ \citep{arxiv260816889}. A checked contract for each learned skill remains an open
problem. EmbodiedEvo's Lessons record when an operation applies, but as text
for the agent to read rather than as a check the runtime enforces
\citep{arxiv261018867}.\looseness=-1

\subsubsection{Comparison with End-to-End Policies}
\label{sec:code-vs-vla}

Whether programs beat end-to-end policies is the most frequent question about
this line of work, and the evidence gives a conditional answer. Weights or
Skills? compares families of robot-learning techniques, among them
code-as-policy and end-to-end VLAs, on six axes and finds that no family is
favorable on all of them \citep{arxiv260801851}.
\Cref{tab:code-vs-vla} summarizes the comparisons we found.\looseness=-1

\input{tables/code-vs-vla}

Several comparisons favor programs when the task changes in structure. On the task perturbations of
LIBERO-PRO, which change the goal of each task, OpenVLA, $\pi_0$, and $\pi_{0.5}$
score between 0.00 and 0.01 in its own evaluation \citep{arxiv251003827}. Program-based
methods score well above that. In ASPIRE's overall results, CaP-Agent0 reaches
0.16 and ASPIRE with Opus~4.6 reaches 0.67 \citep{arxiv260700272}. Encore,
which develops and then freezes one program per task from demonstrations of
the original tasks, none of which shows the new goal, reaches 0.96 on the spatial, object, and goal suites (0.94
without the demonstrations) \citep{arxiv260937359}, and RHO's suite averages
lie between 0.37 and 0.62 \citep{arxiv260616458}.
LIBERO-PRO's released evaluation code, however, gives the policy the instruction encoded
in each task's file name, and a task-perturbed file keeps the original name. A
VLA evaluated this way is therefore asked for the original task and scored on
the new goal. RoboDuo points this out and re-evaluates with the instruction that matches
the scored goal. Its own $\pi_{0.5}$ checkpoint, trained on the four LIBERO
suites, then reaches 11.2\% to 60.6\% on the four task-perturbed suites, and
its reproduced X-VLA 0\% to 18\% \citep{arxiv261034108}. ASPIRE and Encore,
reported over different suites and protocols, still lie above these
figures on average, but the margins are smaller than the published VLA rows suggest.\looseness=-1

Systems that add planning and object grounding also report gains on new objects
and long horizons. On GemBench, 3D-LOTUS++ adds a language-model planner that
writes Python-style calls to six primitives and open-vocabulary object grounding,
and also modifies the motion policy's input representation, trajectory
prediction, and stop prediction. It reaches 17.4\% on long-horizon tasks, against
0.3\% for the end-to-end 3D-LOTUS \citep{arxiv241001345}. The gain therefore
reflects all of these changes together. On the first ten tasks of the BEHAVIOR Challenge 2025, RoboHarness gives Qwen3.8-Flash-Next tools that track image keypoints and move the robot to satisfy keypoint constraints, with a human-written procedure in each task's prompt. Over five instances per task it reaches a mean q-score of 44.5\%, against 34.8\% for G0.5 and 31.6\% for the challenge winner's leaderboard entry \citep{arxiv261014018}. Explicit safety constraints provide another case:
SafeHarness leaves obstacles undisturbed in 91.9\% of SafeLIBERO episodes, against 17.1\% for
$\pi_{0.5}$ \citep{arxiv260920822}. Here the safety comes from the structure of the harness rather than from
code generation as such (\cref{sec:safety}).\looseness=-1

When demonstrations of the evaluated tasks are available, the advantage can
pass to end-to-end policies in distribution. A fine-tuned OpenVLA averages 65.0\% on eight LIBERO tasks, against
52.4\% for RoboPro, a zero-shot model trained to write policy code, and 14.8\% for Code
as Policies \citep{arxiv250104268}. On GemBench's first level, which keeps the
training tasks and varies mainly object placements, 3D-LOTUS reaches 94.3\% and
3D-LOTUS++ 68.7\% \citep{arxiv241001345}. Push-T, a planar pushing task in a 2D gym simulator, reverses the order: a controller that Claude Code with Fable~5 wrote and tuned succeeds on 97.2\% of 200 seeds from rendered images segmented by color, against 62.5\% for the released Diffusion Policy checkpoint trained on 200 human demonstrations \citep{arxiv260818227}. On LIBERO itself, RAX's programs, developed without action demonstrations, reach 98.5\% on the four original suites against 98.8\% for the best of nine policies fine-tuned on 50 demonstrations per task. Under LIBERO-PRO's position perturbation they keep 76.0\%, where the policies with reported results reach at most 42.3\% \citep{arxiv261022496}. Demonstrations can also inform programs rather than weights. Demo2Code summarizes a few demonstrations into a task specification and expands it recursively into code, and in its Robotouille cooking simulator it raises the unit-test pass rate from 0.29 for code written from the instruction alone, as in Code as Policies, to 0.59 \citep{arxiv230516744}. Latency also favors end-to-end
policies, but only over systems that repeatedly call a model online
(\cref{sec:timescales}).\looseness=-1

Contact-heavy tasks have been a weak point of program systems. CaP-X
describes programmatic control as brittle for contact-rich behaviors such as
insertion and pouring \citep{arxiv260322435}, and Gemini Robotics-ER, writing
code zero-shot on a real bimanual platform, scores 0\% on folding a dress
\citep{arxiv250320020}. With Claude Opus~4.6, ASPIRE reaches 0.09 on Robosuite
nut assembly while reaching 0.92 on a bimanual handover. Its model
ablation keeps the environment, robot API, seeds, and protocol fixed, and nut
assembly rises to 98\% with Opus~5 and 100\% with Opus~5.5
\citep{arxiv260700272}. On CaP-Bench's nut-assembly task, OpenRUA, whose agent
can read force feedback through ROS~2, succeeds in every trial, against 80\%
for human-written programs and at most 9\% for the other coding agents in the
same table \citep{arxiv261002459}, a comparison that does not separate force
feedback from the model and other differences between the systems. FORCA
separates it within one study. With the same model (Claude Fable~5), skills,
and instructions on a UR5e arm, its force-aware programs open a drawer in 10 of
10 trials and turn a door handle in 10 of 10, against 4 and 1 for a variant
without force data \citep{arxiv261031841}. RoboAuto adds object-centric
knowledge and stage-wise verification. With Opus~4.6, it reaches 74\% on
Robosuite nut assembly, compared with at most 9\% for the published baseline
results reported in its comparison \citep{arxiv261036533}. Success on
contact-rich tasks varies with the backbone model in
ASPIRE, with force sensing in FORCA, and with compliance in GenCHiP
(\cref{sec:api}).\looseness=-1

Hybrids fall between the two families, and combining a learned policy with an
agent can improve on either component in some reported settings. Harness VLA reaches 82.4\% on LIBERO-PRO, averaged over
the eight cells of the four suites under position and task perturbation, with a Claude
Code planner, against 50.0\% for the frozen VLA it wraps. On the six cells that
RATs, a code-writing agent, reports, the harness averages 87.7\% (averaged by us over its Table~3) against 43.8\% for RATs \citep{arxiv260708448}. OpenRUA, a pure coding agent, reports
87.0\% on the same eight cells and 91.8\% on the six, above the harness with
its Claude Code planner \citep{arxiv261002459}, although Harness VLA's appendix
reports 92.6\% on the eight cells with a GPT-6 Astra planner
\citep{arxiv260708448}. On standard LIBERO a VLA remains best, with 97.0\%
against 96.0\% for the harness \citep{arxiv260708448}. HexaAnything's gains over its VLA on RoboCasa365 are
small on seen atomic tasks (2.9 points over XR-1) and larger on composite ones (6.7 points
on seen and 4.0 on unseen tasks) \citep{arxiv260935432}. EBench compares a
GPT-6 Astra agent with and without a callable post-trained policy on the same
510 test episodes. Calling the VLA $\pi_{0.5}$ or the world action model
InternW0 through tool calls raises success from
46.7\% under direct control to 53.0\% and 56.0\%, against 55.3\% for the best of
eight post-trained policies, and high-precision contact manipulation remains
difficult for both the agent and the policies \citep{arxiv261035832}.\looseness=-1

A third option keeps a frontier model in control throughout the episode
instead of handing control to a program or a policy. Calling an end-effector
tool at every step,
GPT-6 Astra ranked above all 40 policies of a September 2026 RoboDojo leaderboard snapshot, with 22.48\% average success against
19.34\% for the best of them. It led on generalization and open tasks but trailed on
precision, long-horizon, and memory ones
\citep{arxiv260924170}. On RoboMME's memory-dependent tasks, by contrast, the
same model choosing end-effector waypoints succeeds in 92.1\% of 800 episodes in
a single run, against 60.8\% for the best trained entry on the benchmark's
leaderboard \citep{arxiv261017350}. HexaShell gives the same model 14 general perception, motion, and state tools and an optional page of guidance per task. In its own runs on RoboDojo's public test layouts it reaches 69.9\% success without guidance and 85.0\% with it. An earlier version that wrote and froze its own tools for each task reached 46.9\% in the official evaluation, below three other leaderboard entries, and trailed VPP2, a world action model, on precision (32.0\% against 35.2\%) and long-horizon tasks (31.5\% against 37.0\%) \citep{arxiv261023479}. Show-Harness also keeps the model in the loop at every step but has it write no code. The model emits discrete end-effector steps and gripper commands, which a deterministic interpreter for each robot turns into bounded motions. On ten real-robot tasks that place an object in a receptacle, with ten trials each, it reaches 89.0\% with Gemini-3.1 Pro, against 57.0\% for RATs and 44.0\% for CaP-X, whose backbone models the paper does not state, and 39.0\% for a $\pi_{0.5}$ fine-tuned on the same 164 demonstrations \citep{arxiv260910522}. Embodied First Exam leaves the choice to the agent, which may control the robot directly, write a program, or train a policy. A coding agent labeled 1,489 sampled attempts by approach. In the Standard setting, where the agent sees only the robot's sensors and the simulation pauses between requests, 61\% of the attempts control the robot directly. In the Privileged setting, with the full simulator state, 82\% use a program, as do 97\% in the Realtime setting, where the simulation keeps running. Six attempts trained a policy \citep{arxiv261112177}.\looseness=-1

Later harnesses raise these scores, and controlled studies have begun to
separate the design choices behind them. With one end-effector tool call per
turn and no skills, six frontier models at three levels of reasoning effort
succeed in 0\% to 44\% of 100 LIBERO-10 episodes each, and GPT-6 Astra at high
effort is the best \citep{arxiv261030051}. JustRUA starts from the harness of
the RoboDojo evaluation above, keeps GPT-6 Astra as the model, and changes one
design at a time on 111 episodes
that cover all 42 tasks, running each configuration once. Keeping the full
episode history, stopping each motion after 20 control steps, compacting the
robot's readings, adding depth predicted from RGB, and letting the model solve
3D coordinates in code with no cap on its decisions raise the score from 29.7
to 50.3. Neither demonstrations nor a self-written skill library clearly beats
its default \citep{arxiv261116009}. Auriga has GPT-6 Astra act through
direct tool calls or through Python programs behind one gateway, which shows a
preview of each trajectory before motion and stops the robot when a movement
leaves its declared tolerance. Under the protocol of the official leaderboard
it reaches 57.74\% success over 6,300 RoboDojo episodes, and on 30 cases
removing the programs lowers success from 30 to 19 \citep{arxiv261101112}. With
an open-weight model and no privileged inputs, uni-robot-agent reports a
RoboDojo score of 31.39 over 252 episodes of its own evaluation, above the
28.97 published for GPT-6 Astra in the earlier harness
\citep{arxiv261101560}. AMP1 has the model control the arm through Python code
in a persistent interpreter and supplies no grasping or placing skills. With
Gemini~4 Argon it reaches 85.6\% on RoboLab-120, against 51.0\% for the
strongest learned policy on that leaderboard. On twelve real tasks with Claude
Opus~5.5 it succeeds in 104 of 108 trials, against 49 for a world action model
and 30 for $\pi_{0.5}$, with robot error recovery enabled for AMP1 alone, and
in 94 of 108 without it \citep{arxiv261109892}.

\input{figures/libero-pro-splits}

Two features of this evidence limit what it can establish.
First, the comparisons concentrate on the conditions that favor programs.
In LIBERO-PRO's own evaluation, OpenVLA, $\pi_0$, and $\pi_{0.5}$ remain
largely stable under object and paraphrase perturbations, vary by model
under environment perturbations, collapse under task perturbations, and nearly
collapse under position perturbations, where $\pi_{0.5}$ reaches 0.38 on the
Goal suite \citep{arxiv251003827}. The code-versus-VLA comparisons in \cref{fig:libero-pro-splits} cover only
the position and task splits. Embodied Skill Bench evaluates coding agents under
language, object, task, and swap perturbations, but uses VLAs only as callable
skills, with no standalone VLA baseline \citep{arxiv261016238}.\looseness=-1

Second, the studies differ in their end-to-end baselines and in what the
program-based systems may do before evaluation. The $\pi_{0.5}$ rows used by most
studies of program-based methods come from the CaP-X and LIBERO-PRO evaluations, whereas
Zetta reports 32.00\% for a frozen $\pi_{0.5}$ on its LIBERO-PRO protocol, and
Harness VLA reports 50.0\% for a frozen $\pi_{0.5}$ checkpoint fine-tuned on
LIBERO-130 alongside 11.0\% for another $\pi_{0.5}$ row in the same table
\citep{arxiv260816590,arxiv260708448}. Averaging scopes differ as well, from
four cells to eight, and some headline numbers cover a single suite. On the
program side, ASPIRE learns per task on seeds
disjoint from evaluation, with environment interaction \citep{arxiv260700272}. CaP-Agent0 does no per-task learning before
evaluation and instead generates a fresh program within each trial, with
multi-turn execution feedback and retries inside the episode, though it draws on
a task-agnostic skill library synthesized from earlier CaP-Bench runs
\citep{arxiv260322435}. Encore develops each program on 15 initial states disjoint from evaluation and then freezes it. Its rerun of ASPIRE with Opus~5 reaches 89.3\% over the position and task cells of three suites, against 71.7\% published with Opus~4.6 \citep{arxiv260937359}. ASPIRE itself reports that iterative debugging alone reaches 96--99\% on the Object, Goal, and Spatial suites with Opus~5, Opus~5.5, or GPT-6 Astra, against 62\% with Opus~4.6, and across the twelve models of its ablation on the Goal suite under position perturbation, success ranges from 15\% to 99\% \citep{arxiv260700272}. Program-side results thus also move with the backbone model. RHO is trained on unperturbed LIBERO tasks
\citep{arxiv260616458}, and RATs builds its library through self-directed play
before testing \citep{arxiv260619419}. RAX explores only scenes from the
benchmark's training distribution, on LIBERO-PRO with at most five further
episodes on perturbed layouts, and never an evaluation state \citep{arxiv261022496}.\looseness=-1

Read across studies, the results suggest that the two families generalize
along different directions. Explicit program logic can accommodate a new goal, a new order of steps, or
a new constraint. A learned policy may better absorb sensorimotor variation
within the task distribution represented in its training. Position
perturbations fall between the two, and the results there are mixed.
LIBERO-PRO reads the VLAs' stability under object
and instruction perturbations as illusory robustness, citing a case where a
meaningless instruction yields the same actions
\citep{arxiv251003827}. Under its released evaluation code, which takes the
instruction from the task's file name, the paraphrase does not reach the
policy. LIBERO-Plus reports that blank instructions leave OpenVLA-OFT's performance on its
object suite largely unchanged, with a clear drop only on its long-horizon
suite \citep{arxiv251013626}. Stability under rewording is therefore weak
evidence of linguistic generalization. If programs and policies do generalize
along different directions, program-based methods
evaluated on LIBERO-PRO's object split would lose much of their advantage, and
hybrids would gain most on tasks that combine new structure with demanding
contact. SEAR's comparison at matched development time includes LIBERO-PRO
tasks with changed objects and wording but reports them only in aggregate
(\cref{sec:code-teacher}), so neither prediction has yet been tested.\looseness=-1

\begin{pciinsight}[pcicolor=SecWrite,before={\par\Needspace{7\baselineskip}}]{Takeaways and open problems: programs as robot policies}
\textbf{Takeaways.}
\begin{itemize}[leftmargin=1.1em,itemsep=2pt,topsep=1pt,parsep=0pt]
\item \textbf{The primitive API limits what programs can do.} Richer primitives raise success, and the strongest models can write some missing ones.
\item \textbf{Online revision and offline development serve different needs.} Rewriting code within an episode fixes logic that sensor feedback cannot (VLCP: 3.5\% to 35.1\%), and offline development cuts runtime deliberation.
\item \textbf{Programs lead on new task structure, policies in distribution.} The margin under task perturbation is smaller than LIBERO-PRO's published VLA rows suggest. Several systems also lead on long horizons and explicit constraints, though not on every benchmark. Against code-writing systems, demonstration-trained policies keep their in-distribution advantage in most reported comparisons, though frontier models acting at every step have overtaken them on RoboMME and in Show-Harness's real-robot tasks.
\item \textbf{Other comparisons depend on the setting.} Contact results depend on the backbone model and on sensing and control access. Latency depends on runtime model calls, and hybrids help in some settings.
\end{itemize}
\textbf{Open problems.}
\begin{itemize}[leftmargin=1.1em,itemsep=2pt,topsep=1pt,parsep=0pt,beginpenalty=10000]
\item \textbf{All perturbation types.} Controlled comparisons reported for each perturbation axis, environment included, with interfaces and budgets matched, would test whether the advantage is about task structure.
\item \textbf{Postconditions for learned skills.} Chained VLA skills fail at the handoff. Primitives that declare and check their end state would make composition testable.
\item \textbf{Interface choice.} Compare code and typed tool calls in repeated, matched runs that call learned policies.
\end{itemize}
\end{pciinsight}

This part has treated the program as the thing that acts. \Cref{sec:objective-world}
moves the program away from the actuator, where it specifies what a learned
policy should optimize and the world in which it learns.

%% file: figures/action-overview.tex
\begin{figure}[!tbp]
\centering\def\pcifigworkcolor{SecWrite!80!black}
\begin{tikzpicture}[x=1cm,y=1cm,
  q/.style={font=\sffamily\scriptsize\itshape,text=SecWrite!75!black,inner sep=0pt,anchor=west},
  bl/.style={lh text,font=\sffamily\footnotesize,anchor=north west,text width=4.3cm},
  eg/.style={font=\sffamily\scriptsize\itshape,text=MutedText,inner sep=0pt,anchor=west}]
\newcommand{\apanel}[5]{%
  \draw[lh panel=SecWrite] (#1,#2) rectangle ++(7.98,4.3);
  \node[anchor=west,inner sep=0pt] at (#1+0.25,#2+3.92) {\lhnum{SecWrite}{#3}};
  \node[lh head=SecWrite,font=\cartoonfont\bfseries\lhtitlesize] at (#1+0.8,#2+3.92) {\strut#4};
  \node[q] at (#1+0.25,#2+3.5) {#5};}
\newcommand{\abody}[4]{%
  \node[art img] at (#1+1.75,#2+2.1) {\includegraphics[width=3.2cm,height=2.45cm,keepaspectratio]{#3}};
  \node[bl] at (#1+3.5,#2+3.25) {#4};}
\newcommand{\ab}[2]{\lhb{SecWrite}{#1}{3.65cm}{#2}}
\newcommand{\aflow}[6]{%
  \node[lh chip=SecWrite,anchor=west] (f1) at (#2,#1) {#3};
  \node[lh chip=SecWrite,anchor=west] (f2) at ($(f1.east)+(0.3,0)$) {#4};
  \node[lh chip=SecWrite,anchor=west] (f3) at ($(f2.east)+(0.3,0)$) {#5};
  \draw[-{Latex[length=1.4mm]},SecWrite,line width=.8pt] (f1)--(f2); \draw[-{Latex[length=1.4mm]},SecWrite,line width=.8pt] (f2)--(f3);
  \node[eg] at ($(f3.east)+(0.35,0)$) {#6};}
\apanel{0.02}{4.55}{1}{Primitive Interfaces}{How much must the designer supply?}
\abody{0.02}{4.55}{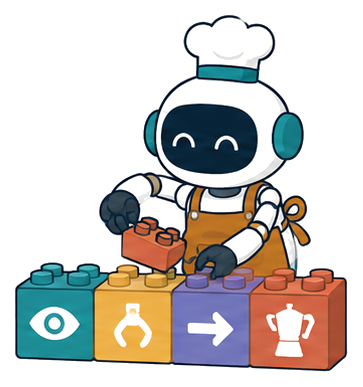}{\ab{plug}{the primitive API bounds what the code can do}\\[4pt]\ab{arrow-up}{richer primitives raise success}\\[4pt]\ab{tools}{top models can write some missing abstractions}}
\aflow{5.0}{0.3}{perceive}{grasp}{place}{e.g.\ \figwork{arxiv240406645}{GenCHiP}, \figwork{arxiv260322435}{CaP-X}}
\apanel{8.2}{4.55}{2}{Online Revision, Offline Development}{Rewrite online or develop offline?}
\abody{8.2}{4.55}{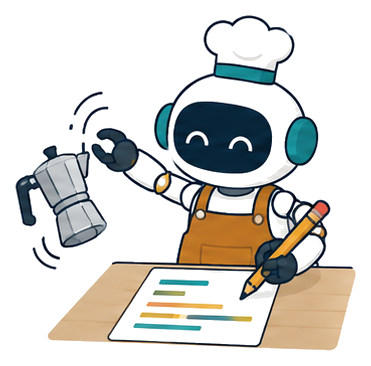}{\ab{redo}{rewriting code within an episode can repair logic}\\[4pt]\ab{chart-line}{\figwork{arxiv260816978}{VLCP}: success rises from 3.5\% to 35.1\%}\\[4pt]\ab{clock}{offline development can cut runtime deliberation}}
\aflow{5.0}{8.48}{run}{fail}{rewrite}{e.g.\ \figwork{arxiv260816978}{VLCP}, \figwork{arxiv260929394}{RACaP}}
\apanel{0.02}{0}{3}{Orchestrating Learned Skills}{What changes when primitives are learned?}
\abody{0.02}{0}{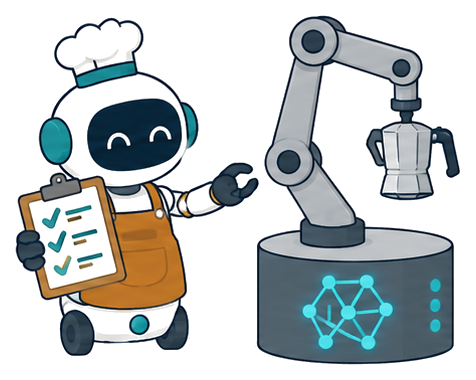}{\ab{hand-paper}{the learned policy supplies motor skills}\\[4pt]\ab{clipboard-list}{code keeps memory, progress, and recovery}\\[4pt]\ab{project-diagram}{reached through code or typed tool calls}}
\aflow{0.45}{0.3}{code}{policy}{check}{e.g.\ \figwork{arxiv260809410}{HyMeS}, \figwork{arxiv260708448}{Harness VLA}, \figwork{arxiv260721725}{Pigey}}
\apanel{8.2}{0}{4}{Programs vs.\ End-to-End Policies}{When does a program beat an end-to-end policy?}
\node[lh text,font=\sffamily\footnotesize\bfseries,text=SecWrite,anchor=west] at (8.45,3.12) {\faIcon{check-circle}\,Programs Lead};
\node[lh text,font=\sffamily\footnotesize\bfseries,text=Baseline!80!black,anchor=west] at (12.4,3.12) {\faIcon{balance-scale}\,Policies Hold};
\node[art img] at (9.25,2.24) {\includegraphics[width=1.7cm,height=1.4cm,keepaspectratio]{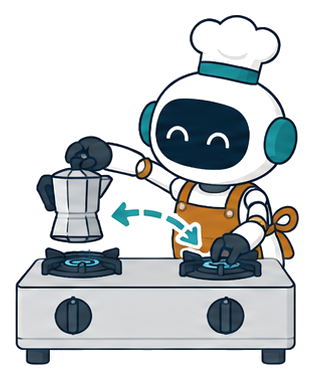}};
\node[art img] at (13.1,2.24) {\includegraphics[width=1.7cm,height=1.4cm,keepaspectratio]{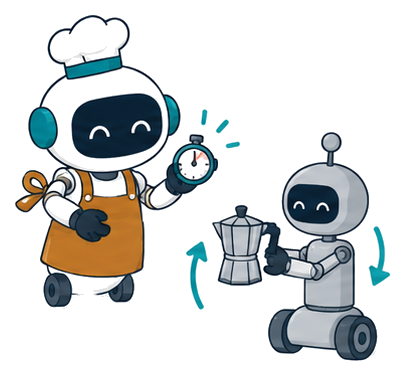}};
\node[lh text,font=\sffamily\footnotesize,anchor=west,text width=1.95cm] at (10.15,2.24) {when the task changes in structure};
\node[lh text,font=\sffamily\footnotesize,anchor=west,text width=2.2cm] at (13.95,2.24) {in distribution, with demonstrations};
\draw[RuleGray,line width=.6pt,dashed] (12.2,1.45) -- (12.2,3.3);
\node[anchor=west,inner sep=0pt] at (8.45,1.0) {\lhicon{SecWrite}{route}};
\node[lh text,font=\sffamily\scriptsize,anchor=west,text width=6.9cm] at (9.05,1.0) {several systems also lead on long horizons and explicit constraints, and contact results depend on the model and the interface};
\node[eg] at (9.05,0.32) {e.g.\ \figwork{arxiv251003827}{LIBERO-PRO}, \figwork{arxiv241001345}{GemBench}};
\end{tikzpicture}
\caption{\textbf{Programs as robot policies: four questions.} What a program can call \textcircled{\scriptsize 1}, when it is revised during an episode or developed offline \textcircled{\scriptsize 2}, how it coordinates learned skills \textcircled{\scriptsize 3}, and when it outperforms an end-to-end policy \textcircled{\scriptsize 4}.}
\label{fig:action-overview}
\end{figure}

%% file: tables/code-vs-vla.tex
\begin{table}[!tbp]
\centering\footnotesize\setlength{\tabcolsep}{4pt}\renewcommand{\arraystretch}{1.0}\captionsetup{skip=3pt}\setlength{\aboverulesep}{0.2ex}\setlength{\belowrulesep}{0.4ex}
\caption{\textbf{Conditions in reported comparisons of programs and end-to-end policies.} Each line is one result from the cited study, in its own units and averaging scope, and a dash marks a side it does not report. Tags give the approach the results favor: \tabtag{SecWrite}{Programs}, \tabtag{SecWrite!50!Baseline}{Hybrids} (programs that call a learned policy), \tabtag{Baseline!75!black}{End-to-end} policies, or \tabtag{Baseline!60!white}{Mixed} results. $\dagger$ marks modular systems whose planner does not write code.}
\label{tab:code-vs-vla}
\newcommand{\fP}{\tabtag{SecWrite}{Programs}}\newcommand{\fH}{\tabtag{SecWrite!50!Baseline}{Hybrids}}\newcommand{\fE}{\tabtag{Baseline!75!black}{End-to-end}}\newcommand{\fM}{\tabtag{Baseline!60!white}{Mixed}}
\newcommand{\balancedcell}{\RaggedRight\rightskip=0pt plus 1\hsize\parfillskip=0pt\hyphenpenalty=10000\exhyphenpenalty=10000\relax}
\newcommand{\nores}{\textcolor{MutedText}{--}}
\begin{tabularx}{\linewidth}{@{}>{\balancedcell\arraybackslash}p{.43\linewidth}>{\balancedcell\arraybackslash}p{.33\linewidth}>{\balancedcell\arraybackslash}X@{}}
\toprule
\TableHeader{Setting} & \TableHeader{Program or hybrid} & \TableHeader{End-to-end policy}\\
\midrule
\multicolumn{3}{@{}l}{\bfseries Change in task structure\enspace\mdseries \fP}\\*
LIBERO-PRO, file-name instruction \citep{arxiv251003827} & \nores & OpenVLA, $\pi_0$, $\pi_{0.5}$ 0.00--0.01\\
Matching instruction \citep{arxiv261034108} & \nores & $\pi_{0.5}$ 11.2--60.6\\
Task split over three suites \citep{arxiv260700272} & CaP-Agent0 0.16, below matching $\pi_{0.5}$\newline ASPIRE 0.67 & \nores\\
Suite averages \citep{arxiv260616458} & RHO 0.37--0.62 & \nores\\
Task split, same model for both, demonstrations of the original tasks only \citep{arxiv260937359} & Encore 0.96\newline ASPIRE rerun 0.89 & \nores\\
\midrule
\multicolumn{3}{@{}l}{\bfseries Long horizons and new objects\enspace\mdseries \fP\enspace\fH}\\*
GemBench long-horizon level \citep{arxiv241001345} & 3D-LOTUS++ 17.4\% (planner, grounding, modified motion policy) & 3D-LOTUS 0.3\%\\
LIBERO-PRO Long without test-time debugging \citep{arxiv260700272} & ASPIRE 0.305 (library seeded from 90 LIBERO-90 tasks), 0.047 (empty library) & $\pi_{0.5}$ 0.05\\
Limit: RoboDojo long horizon \citep{arxiv261023479} & Earlier HexaShell 31.5\% & VPP2 37.0\%\\
\midrule
\multicolumn{3}{@{}l}{\bfseries Programs without demonstrations on the robot\enspace\mdseries \fP}\\*
Kinova arm, reported in A0 \citep{arxiv250412636} & ReKep 33.75 (no demonstrations) & $\pi_0$ 20.00 (five demonstrations per task)\\
SimplerEnv WidowX \citep{arxiv251011660} & Modular pipeline with GPT-5$^\dagger$ 86.8 & $\pi_0$ 55.7\\
\midrule
\multicolumn{3}{@{}l}{\bfseries Explicit safety constraints\enspace\mdseries \fH}\\*
Collision avoidance \citep{arxiv260920822} & SafeHarness 91.9\% & $\pi_{0.5}$ 17.1\%\\
Same agents with skills, without the harnesses \citep{arxiv260920822} & Obstacles disturbed in 56--66\% of episodes & \nores\\
\midrule
\multicolumn{3}{@{}l}{\bfseries Position perturbations\enspace\mdseries \fM}\\*
Goal suite \citep{arxiv260322435} & CaP-Agent0 0.26 & $\pi_{0.5}$ 0.38\\
Over three suites \citep{arxiv260700272} & ASPIRE 0.77 & $\pi_{0.5}$ 0.25\\
Per-task cells \citep{arxiv260616458} & RHO & $\pi_{0.5}$ wins several cells\\
\midrule
\multicolumn{3}{@{}l}{\bfseries In distribution, with demonstrations\enspace\mdseries \fE}\\*
Eight LIBERO tasks \citep{arxiv250104268} & RoboPro 52.4 (zero-shot code generator)\newline Code as Policies 14.8 & Fine-tuned OpenVLA 65.0\\
GemBench placement level \citep{arxiv241001345} & 3D-LOTUS++ 68.7 & 3D-LOTUS 94.3\\
Counterexample: simulated Push-T \citep{arxiv260818227} & Agent-written controller 97.2\% & Diffusion Policy 62.5\%\\
\midrule
\multicolumn{3}{@{}l}{\bfseries Contact-rich and dexterous tasks\enspace\mdseries \fM}\\*
Nut assembly \citep{arxiv260700272} & ASPIRE 0.09 (1.00 with Opus~5.5) & \nores\\
Real dress folding \citep{arxiv250320020} & Gemini Robotics-ER code 0\% & \nores\\
SimplerEnv drawers, visual matching \citep{arxiv250213143} & Modular pipeline$^\dagger$ 0.403 & RT-1-X 0.597\\
Same drawers, variant aggregation \citep{arxiv250213143} & Modular pipeline$^\dagger$ 0.297 & RT-2-X 0.353\\
Counterexample: CaP-Bench nut \citep{arxiv261002459} & Coding agent with force feedback 100\% & \nores\\
\midrule
\multicolumn{3}{@{}l}{\bfseries Latency with online model calls\enspace\mdseries \fE}\\*
LIBERO-PRO, median policy wall time per episode \citep{arxiv260929394} & CaP-X 845 s & Not timed\\
World Action Agent$^\dagger$, average per episode \citep{arxiv260929964} & 150 s and 31 model calls & Not timed\\
\midrule
\multicolumn{3}{@{}l}{\bfseries Program around a learned policy\enspace\mdseries \fH}\\*
LIBERO-PRO, eight cells \citep{arxiv260708448} & Harness VLA 82.4 (Claude Code planner), 92.6 (GPT-6 Astra planner) & Its frozen VLA 50.0\\
LIBERO-PRO, six shared cells \citep{arxiv260708448} & Harness VLA 87.7 (our average)\newline RATs 43.8 (code-writing agent) & \nores\\
LIBERO-PRO, two suites under task and position perturbations \citep{arxiv260816590} & Zetta 71.13 & Frozen $\pi_{0.5}$ 32.00\\
RoboMemArena, same components \citep{arxiv260816889} & BATON 63.5, Harness VLA 23.1 & $\pi_{0.5}$ 21.5, best memory VLA 46.1\\
Limit: the same eight cells \citep{arxiv261002459} & Pure coding agent 87.0 & \nores\\
Limit: standard LIBERO \citep{arxiv260708448} & Harness VLA 96.0 & A VLA 97.0\\
\bottomrule
\end{tabularx}
\end{table}

%% file: figures/libero-pro-splits.tex
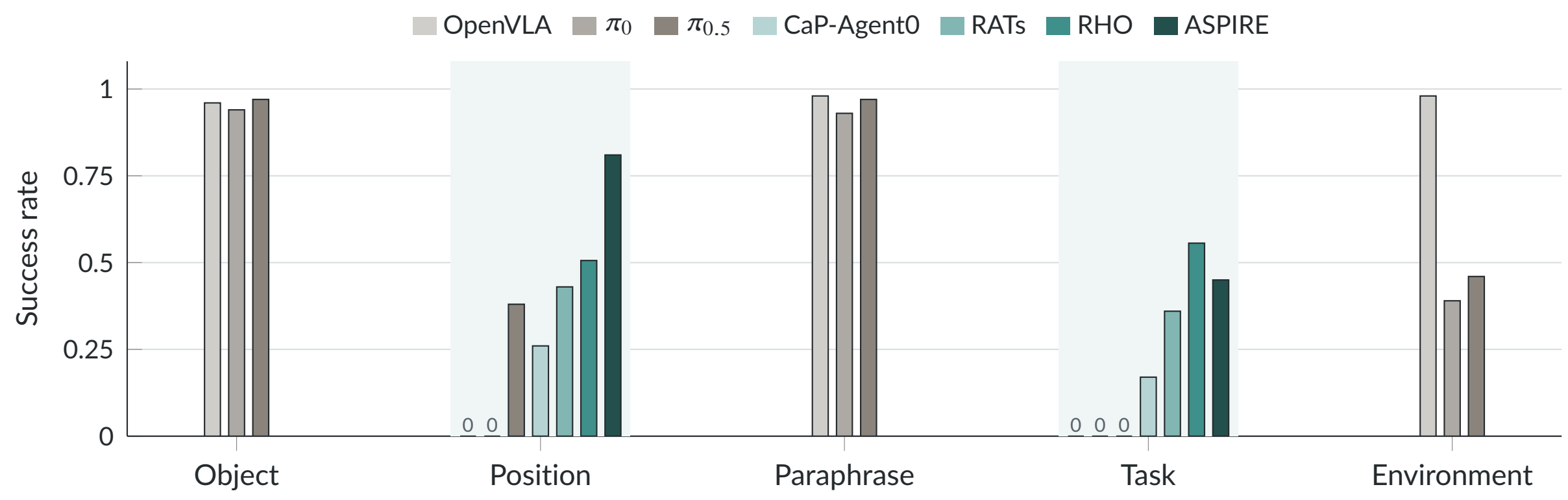
\begin{figure}[t]
\centering
\begin{tikzpicture}
\begin{axis}[
  width=\linewidth, height=5.4cm,
  ybar, bar width=4.6pt,
  ymin=0, ymax=1.08, ytick={0,0.25,0.5,0.75,1},
  ylabel={Success rate}, ylabel style={font=\sffamily\footnotesize},
  symbolic x coords={Object,Position,Paraphrase,Task,Environment},
  xtick={Object,Position,Paraphrase,Task,Environment}, xticklabel style={font=\sffamily\footnotesize},
  yticklabel style={font=\sffamily\scriptsize,/pgf/number format/assume math mode=true},
  zerolab/.style={nodes near coords={\pgfmathfloatifflags{\pgfplotspointmeta}{0}{0}{}},
    every node near coord/.append style={font=\sffamily\tiny,text=MutedText,inner sep=1pt}},
  enlarge x limits=0.09,
  axis lines*=left, ymajorgrids, grid style={RuleGray},
  legend style={font=\sffamily\scriptsize,draw=none,fill=none,at={(0.5,1.02)},anchor=south,legend columns=7,/tikz/every even column/.append style={column sep=4pt}},
  legend image code/.code={\draw[#1,draw=none] (0cm,-0.08cm) rectangle (0.24cm,0.1cm);},
]
\fill[SecWrite!8] ([xshift=-26pt]axis cs:Position,0) rectangle ([xshift=26pt]axis cs:Position,1.08);
\fill[SecWrite!8] ([xshift=-26pt]axis cs:Task,0) rectangle ([xshift=26pt]axis cs:Task,1.08);
\addplot[fill=Baseline!40!white,bar shift=-7pt] coordinates {(Object,0.96) (Paraphrase,0.98) (Environment,0.98)};
\addplot[fill=Baseline!70!white,bar shift=0pt] coordinates {(Object,0.94) (Paraphrase,0.93) (Environment,0.39)};
\addplot[fill=Baseline,bar shift=7pt] coordinates {(Object,0.97) (Paraphrase,0.97) (Environment,0.46)};
\addplot[fill=Baseline!40!white,bar shift=-21pt,zerolab,forget plot] coordinates {(Position,0.00) (Task,0.00)};
\addplot[fill=Baseline!70!white,bar shift=-14pt,zerolab,forget plot] coordinates {(Position,0.00) (Task,0.00)};
\addplot[fill=Baseline,bar shift=-7pt,zerolab,forget plot] coordinates {(Position,0.38) (Task,0.00)};
\addplot[fill=SecWrite!38!white,bar shift=0pt] coordinates {(Position,0.26) (Task,0.17)};
\addplot[fill=SecWrite!65!white,bar shift=7pt] coordinates {(Position,0.43) (Task,0.36)};
\addplot[fill=SecWrite,bar shift=14pt] coordinates {(Position,0.506) (Task,0.556)};
\addplot[fill=SecWrite!55!black,bar shift=21pt] coordinates {(Position,0.81) (Task,0.45)};
\legend{OpenVLA,$\pi_0$,$\pi_{0.5}$,CaP-Agent0,RATs,RHO,ASPIRE}
\end{axis}
\end{tikzpicture}
\caption{\textbf{What the code-versus-VLA comparisons on LIBERO-PRO cover.} Results on the Goal suite. Gray bars show OpenVLA, $\pi_0$, and $\pi_{0.5}$ under LIBERO-PRO's five perturbations, from Figure 7 of \citet{arxiv251003827}. Under its released evaluation code, a VLA receives the original instruction in the paraphrase and task conditions (\cref{sec:code-vs-vla}). Colored bars show program-based methods, which report results only under the position and task perturbations (shaded): CaP-Agent0 from CaP-X (Table 2 of \citet{arxiv260322435}), RATs (Table 1 of \citet{arxiv260619419}), RHO (Table 7 of \citet{arxiv260616458}), and ASPIRE with Opus~4.6 (Table 5 of \citet{arxiv260700272}).}
\label{fig:libero-pro-splits}
\end{figure}

%% file: sections/04-objective-and-world.tex
\subsection{Programs for Policy Learning}
\label{sec:objective-world}

Suppose the kitchen robot must also learn to pour coffee from the moka pot.
If its available primitives cannot express the control needed for a steady
pour, the agent can write a reward function, simulated kitchens in which to
practice, or a script that generates demonstrations. A learned policy then
acquires the skill. The program specifies what the robot should learn or the
conditions in which it learns, and judging its effects may require training
and testing that policy. This part covers programs that define objectives,
build training worlds, generate demonstrations, and describe bodies and assets
(\cref{fig:learning-overview}). Language models in reinforcement learning,
including as reward designers, are surveyed more broadly elsewhere
\citep{arxiv240400282}. Here we follow how each kind of program is checked.
What it costs to evaluate such a program is taken up in
\cref{sec:evidence-cost}.\looseness=-1

\input{figures/learning-overview}

\subsubsection{Reward Functions and Training Objectives}
\label{sec:reward}

A short loop is possible when the objective feeds directly into a controller.
In Language to Rewards, the language model writes reward parameters, and a
model-predictive controller optimizes action sequences under the resulting
reward in real time, so the user can immediately observe the resulting behavior
\citep{arxiv230608647}. Eureka established the longer loop that most later
work inherits. A language model writes reward functions as code, a policy is
trained with each candidate in a GPU-parallel simulator, and per-component
statistics from training are fed back as ``reward reflection'' for the next
round. Eureka samples 16 reward candidates per iteration. Its rewards
outperform those written by human experts on 83\% of tasks across 29 environments, with an
average normalized improvement of 52\% \citep{arxiv231012931}. Removing the
reflection lowers the average normalized score on the Isaac Gym tasks by 28.6\%
\citep{arxiv231012931}, an early sign that the form in which evidence returns
to the model matters. Text2Reward generates dense reward programs zero-shot or few-shot and
reports that policies trained with them perform similarly to or better than
expert-designed rewards on 13 of 17 manipulation tasks \citep{arxiv230911489}.
In a separate study, a self-refinement loop that trains a policy on each reward and returns a summary of training, task metrics, and the success rate lifts six of nine simulated tasks from 0\% success to between 85\% and 100\% within five rounds, matching or exceeding manually designed rewards on eight of the nine \citep{arxiv230906687}. A survey of language-model-enhanced reinforcement learning calls such generated reward code an explicit reward model that people can read, evaluate, and refine, and notes that its dependence on the model's pretrained knowledge limits it on specialized tasks \citep{arxiv240400282}.

Training a policy in simulation adds another question: will the behavior
transfer to the real robot? DrEureka extends
the loop to domain randomization, the ranges over which physical parameters
such as friction are varied during training. A reward found by Eureka in
simulation alone, without randomization or a safety instruction to the
reward-writing model, produces a quadruped that does not walk on the real Go1,
with a forward speed of 0.0~m/s. DrEureka's own reward, trained without
randomization, already walks at 1.21~m/s, and with generated randomization
ranges its best policy reaches 1.83~m/s against 1.32~m/s for the
human-designed one. Without DrEureka's physics prior over physical
parameters, derived from the initial policy, 15 of 16
randomization configurations yield policies that are jerky or unsafe on
hardware. The safety instruction addresses a related risk, since a higher
simulation score can come from over-exerting the motors
\citep{arxiv240601967}.\looseness=-1

How much do the many variants of this loop differ? BIRD implements 14
methods for iterative reward design and re-runs six published ones under
matched budgets. In that comparison, a standard search trains 40 policies,
eight in each of five iterations, while CARD (\cref{sec:evidence-cost}) trains
one candidate per iteration and only in its first four iterations. Of the six, REvolve \citep{arxiv240601309} is strongest on every suite but
HumanoidBench, and beyond that most pairwise differences have overlapping confidence intervals \citep{arxiv261004364}. Among the 30 design
choices tested, four clearly help in a pooled ablation: using task success as the fitness,
summarizing feedback, selecting the checkpoint with the highest return under
the policy's own reward, and keeping a history of four turns. Combining
a small set of useful choices still produces an unsupervised and a supervised method
that outperform their respective baselines.

BIRD also quantifies reward hacking. On HumanoidBench walking and crawling, 9 of 67 policies exceeded 75\% of
the reference reward, and the study's two human raters judged all nine unnatural. The
correlation between human ratings of task completion and benchmark score is
0.78 on Meta-World but 0.36 on Assistax and 0.26 on HumanoidBench
\citep{arxiv261004364}. Other papers report the same pattern
in individual cases. A RoboGen reward intended to fold a chair encouraged unfolding
it \citep{arxiv231101455}, and a reward search in AnyTask produced pushing where
picking was intended \citep{arxiv251217853}. The Reward Design Agent
measures the gap on HumanoidBench, where GPT-4.1-judged instruction alignment is 0.47
for a Eureka baseline that reflects only on numerical reward statistics and
0.70 for its own design, which adds visual analysis of the trained behavior
\citep{arxiv260601672}.

These examples expose a limit of optimizing an incomplete objective. A scalar
fitness makes automatic selection possible, but a policy can reach a high
score while ignoring the instruction or reaching the goal in an unacceptable
way. Some recent work therefore makes behavioral requirements more
explicit before training. In one approach, language or vision-language models
generate signal temporal logic specifications, fit their parameters to expert
or successful trajectories, and discard specifications those trajectories do
not support \citep{arxiv260907111,arxiv260937519}. A second approach keeps the reward
sparse and has the model generate task specifications from which an automatic
curriculum is built \citep{arxiv260723515}. CurricuLLM instead writes an ordered list of subtasks in natural language and reward code for each, and on simulated Berkeley Humanoid velocity tracking its tracking error of 0.46~m/s is close to the 0.41~m/s of a human-designed reward and well below the 1.13~m/s of a single zero-shot generated reward \citep{arxiv240918382}. A third approach
replaces reward code with code that writes reference trajectories for the
robot to imitate, motivated by the argument that generated rewards
generalize poorly across skills and bodies \citep{arxiv260924145}. Inspecting these requirements before training can catch some mistakes
early.\looseness=-1

\subsubsection{Training Environments, Demonstrations, and Assets}
\label{sec:world-programs}

When the agent also builds the training world, it must show that the task
it has created is meaningful and feasible. A valid reward cannot compensate
for an inaccessible target or an incorrectly assembled scene. For the kitchen
robot, a simulated pouring task is useful only if the pot, cup, and available
motions permit the intended outcome. GenSim, which has GPT-4 write simulation
tasks as code, runs each candidate through a sequence of checks: syntax,
runtime, whether executing the task code generates demonstrations in which
the robot achieves each subgoal, and agreement of an LLM critic over three
votes \citep{arxiv231001361}. A human check follows, and human judgments
vary widely. In a test on 30 generated tasks, three users report pass rates
of 80\%, 63\%, and 41\% \citep{arxiv231001361}. RoboGen chains task proposal, scene generation,
decomposition, choice of learning method, and reward code. A manual review of
155 generated tasks found 13 scene failures and 6 supervision failures, and RoboGen lists the verification of learned skills
at scale as an open problem
\citep{arxiv231101455}.\looseness=-1

Later systems differ in which of three questions their checks answer: does the
generated world run, can a robot complete the task, and does completion mean
what was intended? Running is the weakest test. In Eurekaverse, over 50\% of
the terrains GPT-4o writes pass an environment check, which runs each terrain
and applies two height heuristics as a coarse feasibility test
\citep{arxiv241101775}. GenSim2 separates the first two questions for its generated tasks: primitive tasks run without syntax or runtime errors at a rate of 0.94 but are solved at a rate of 0.78, against 0.94 and 0.58 for RoboGen \citep{arxiv241003645}. FATE asks the second question. Because curricula can be
linguistically coherent yet physically infeasible, it places an embodied agent
in the loop to attempt each task before it is accepted \citep{arxiv260301505}.
A successful attempt shows that the task is feasible, while a failed one may
come from an impossible task or from an inadequate solver.\looseness=-1

The third question, intended meaning, is the hardest to check. V-CAGE targets silent failures in which generated code runs but does
not achieve the visual goal \citep{arxiv260115164}. GRS has the model write
test suites alongside the simulated task and revise the two together, so one
pipeline writes both the task and its test (\cref{sec:judges})
\citep{arxiv241015536}.
Video2World makes the question measurable for whole simulators: coding
agents write an executable simulator of the scene and behavior shown in an
embodied video of a robot or a human. The best agent reaches 25.5\% task success on 222 instances,
macro-averaged over 39 task families, against 58.8\% for a human-assisted
reference that uses extra source information. The agent with the best shape
fidelity is not the one that completes the most tasks, so visual agreement is itself an imperfect check \citep{arxiv261004432}. Code
as Worlds turns agreement into an acceptance test. It keeps a world built
from a video only if its agent judges the world's renders and projected depth,
masks, and trajectories consistent with the source, and then trains a VLM on
the retained worlds
\citep{arxiv260827549}. Gym-Anything for Robotics has coding agents build
whole simulations and audits each one over three rounds with separate agents,
one of them an adversarial agent that tries to break the simulation with legal
robot actions. That agent found a real defect in about three quarters of the
25 simulations built. Each task's verifier must be a deterministic function of
the simulation's physical state, with no VLM judge. In a filleting task, this
verifier gave no reward to a model that had placed the whole uncut fish in the
target pan, expecting only the fish frame to be checked
\citep{arxiv261040565}.

Beyond checking each task, a generator must decide which task to present next
and how to repair one that fails a check. OMNI-EPIC
adapts the next environment to the agent's learning progress
\citep{arxiv240515568}. A curriculum steered by the learner's own success can
drift toward tasks the learner already solves, so measuring the learner's
progress needs tasks specified independently of it. SimWorld Studio follows this design. Its coding agent raises the difficulty of generated navigation episodes, whose solvability NavMesh connectivity guarantees, once the learner passes a mastery threshold. The learner is tested on the external SimWorld-MMNav benchmark, where a co-evolved Qwen3.5-9B agent reaches 90\% success against 72\% for training held at a fixed difficulty \citep{arxiv260509423}. For repair, each kind of failed check points to a
different component: unstable geometry to the scene, an infeasible goal to the
task, and an invalid success signal to the evaluator. A repair that makes a
task feasible can also change its difficulty or meaning, so feasibility and
intent both need checking after a repair. Checks kept separate locate the
failure, which a single judgment on the whole world cannot.\looseness=-1

Generated driving scenarios need no training at all when a driving policy
already exists: a scenario written in Scenic is compiled and run with that
policy.
Chat2Scenic compiles
76.42\% of its scenarios and reaches an end-to-end framework accuracy of
58.17\%, defined as the compilation rate times a scenario-quality score
\citep{arxiv260714387}. ScenicNL reports that plain prompting fails on this
low-resource scenario language \citep{arxiv240503709}. ChatScene instead assembles
scenarios from snippets of scenario code that were checked by hand for
compilability, and fine-tuning a driving
policy on its scenarios yields an average collision rate of 4.3\%, against 13.5\%
for the strongest scenario-generation baseline \citep{arxiv240514062}.

\label{sec:data-programs}

Once a generated world supports a solvable task, its executions can also
serve as demonstrations for a learned policy. The success check then does more than
rank a candidate: it determines which behavior enters the training data. A
pouring trajectory mislabeled as successful can teach the kitchen robot the
wrong skill even if the simulation itself is valid. Scaling
Up and Distilling Down has a language model write, for each task, a snippet of
code that decides success. The snippet labels trajectories collected by
planners, triggers retries, and supplies data for a diffusion policy. Over
five domains with 18 tasks, that policy's average success exceeds that of the
data-collection policy by 33.2 points
\citep{arxiv230714535}. RoboTwin~2.0 generates expert programs with simulation
in the loop. Each candidate runs ten times per iteration for up to five
refinements, and average success over ten tasks rises from 62.1\% without
feedback to 71.3\% with multimodal feedback \citep{arxiv250618088}.
AnyTask has the model write, for each task, a reset function, a success check
(written first), a reward, and a scripted policy. The generated reset, state, and
success-check code runs without error for 64\% to 96\% of tasks depending on
the model and prompt, and policies trained on the data reach 44\%
zero-shot on eight real tasks \citep{arxiv251217853}. GenSim2 plans demonstrations with a kPAM-based motion planner from solver configurations that a multimodal model writes. Co-training one multi-task policy on 100 simulated and 10 real demonstrations per task raises average success on eight real tasks from 0.363 with the real demonstrations alone to 0.575, over 10 trials per task \citep{arxiv241003645}. PyRUA-DataFarm has a coding agent rebuild a task in simulation from one calibration episode of the real robot and write an expert controller, and a video model renders each simulated rollout as photographic video. On five real bimanual tasks, $\pi_{0.5}$ policies trained on 100 such episodes succeed in 93 of 100 trials, against 84 for policies trained on as many teleoperated episodes, and in 87 against 46 under an unseen tablecloth and colored light. Adapting a task costs 21 US dollars of API usage on average, and each further episode 0.80 dollars \citep{arxiv261105320}.

When the model that writes a task also writes its success check, the two can
share the same misreading (\cref{sec:judges}). AnyTask compares VLM judgments of rollouts
with its generated success functions to monitor their agreement, and Agentic Skill Discovery uses an independent
VLM to verify behaviors learned against generated success checks
\citep{arxiv251217853,arxiv240515019}. PyRUA-DataFarm separates the roles. One
role writes the success judge, which is frozen before the expert is written.
The expert is admitted only on held-out initial states, and an independent
reviewer examines every rendered episode before it enters the dataset
\citep{arxiv261105320}. Visual judges make their own errors.
RoboTwin~2.0's VLM observer judges execution success with an F1 score of
0.302 (accuracy 0.431) on 130 curated sequences \citep{arxiv250618088}.
How reliably executions are verified matters again when they turn a program
into a teacher, the route that \cref{sec:code-teacher} follows.\looseness=-1

\label{sec:bodies}

The last kind of program defines the objects and robot bodies on whose
physical models those demonstrations depend. A simulated moka pot can look
right while its handle or joints behave incorrectly, so such programs need
checks of geometry and motion as well as syntax.
Articulate-Anything has a VLM write Python against a high-level API that
compiles to a robot description in URDF, the XML format that specifies a
robot's links and joints. It then compares renders and joint motions of
the result with the input and revises \citep{arxiv241013882}. On
PartNet-Mobility, its success rate reaches 75\%, compared with 8.7\% to 12.2\% for prior methods. Some outputs of prior methods are not even valid: 59.1\% of URDFormer's \citep{arxiv240511656} and 17.5\% of Real2Code's \citep{arxiv240608474}
contain cyclic structures, repeated links, or syntax errors \citep{arxiv241013882}.
Articraft also has the model write validation tests alongside the
code that builds each asset against a domain-specific SDK \citep{arxiv260515187}. Artisan adds design and
review agents and MuJoCo checks for collisions and stability. With the same
language model, the collision rate under single-joint motion on ArtVIP falls
from 45.5\% under Articraft to 1.7\% \citep{arxiv261013947}. DiagGen probes generated
deformable assets in a physics simulator to locate defects
\citep{arxiv260923103}.

Body design pushes evaluation back toward policy learning, because whether a
generated body supports useful behavior may show only once a policy is
trained for it. ELM sidesteps this by letting the program fix the motion too. Its code model, used as the mutation operator in MAP-Elites, edits Python programs that specify a Sodarace walker's masses, springs, and each spring's oscillation amplitude and phase, and one simulation of how far the walker travels scores each program \citep{arxiv220608896}. RoboMorph evolves designs written in a structured grammar
\citep{arxiv240708626}, RoboMoRe co-designs the body and the reward
\citep{arxiv250600276}, and Debate2Create trains a policy for every candidate
design \citep{arxiv251025850}. Auto-Robotist distills design-search traces
into reusable design principles, so that later searches start from earlier
physical evaluations \citep{arxiv260525832}.

\begin{pciinsight}[pcicolor=SecWrite]{Takeaways and open problems: programs for policy learning}
\textbf{Takeaways.}
\begin{itemize}[leftmargin=1.1em,itemsep=2pt,topsep=1pt,parsep=0pt]
\item \textbf{Few reward-design methods separate clearly under matched budgets.} In BIRD, REvolve leads on every suite but HumanoidBench and most other differences overlap. Four design choices clearly help in a pooled ablation.
\item \textbf{High fitness can come from unintended behavior.} In BIRD, human raters judged unnatural all nine HumanoidBench walking and crawling policies that exceeded 75\% of the reference reward.
\end{itemize}
\textbf{Open problems.}
\begin{itemize}[leftmargin=1.1em,itemsep=2pt,topsep=1pt,parsep=0pt,beginpenalty=10000]
\item \textbf{Independent checks for generated tasks.} Held-out human-written checks on a sample of accepted tasks would estimate how many of them a pipeline accepts wrongly.
\item \textbf{Physics parameters as code.} Generated randomization ranges, with the generated reward, improved DrEureka's real-world transfer, yet among reward-generating systems few also generate them (HARBOR and AURA do). Adding them to reward-generating systems would test whether the gain generalizes.
\end{itemize}
\end{pciinsight}

These systems rely on evaluation to select programs, retain demonstrations,
and accept learned behavior. What each evaluation establishes, and what it
leaves unresolved, is the subject of \cref{sec:evidence}.

%% file: figures/learning-overview.tex
\begin{figure}[!tbp]
\centering\def\pcifigworkcolor{SecWrite!80!black}
\begin{tikzpicture}[x=1cm,y=1cm,
  bt/.style={lh text,anchor=north west}]
\newcommand{\lpanel}[9]{%
  \lhpanel{#1}{2.65}{3.86}{6.35}{SecWrite}{#2}{#3}{#4}
  \node[art img] at (#1+1.93,7.05) {\includegraphics[width=3.4cm,height=1.95cm,keepaspectratio]{#5}};
  \node[bt] at (#1+0.2,5.85) {\lhb{SecWrite}{pen}{2.92cm}{#6}\\[3pt]\lhb{SecWrite}{search}{2.92cm}{#7}\\[3pt]\lhb{SecWrite}{chart-line}{2.92cm}{#8}};
  \node[lh eg] at (#1+0.22,2.95) {e.g.\ #9};}
\lpanel{0.02}{1}{Rewards}{What should be optimized?}{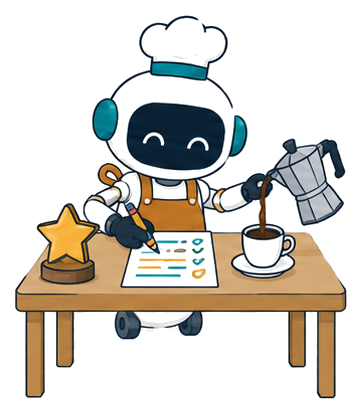}
  {writes reward functions as code}
  {trains a policy with each candidate}
  {\figwork{arxiv231012931}{Eureka} beats expert rewards on 83\% of tasks}
  {\figwork{arxiv231012931}{Eureka}, \figwork{arxiv230911489}{Text2Reward}}
\lpanel{4.13}{2}{Environments}{Where does it practice?}{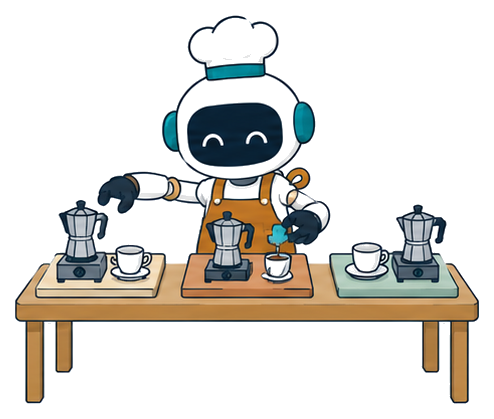}
  {writes tasks, scenes, and curricula}
  {checks that the task is feasible and meaningful}
  {\figwork{arxiv231001361}{GenSim}: three raters' pass rates are 80\%, 63\%, and 41\%}
  {\figwork{arxiv231001361}{GenSim}, \figwork{arxiv231101455}{RoboGen}}
\lpanel{8.23}{3}{Demonstrations}{What does it learn from?}{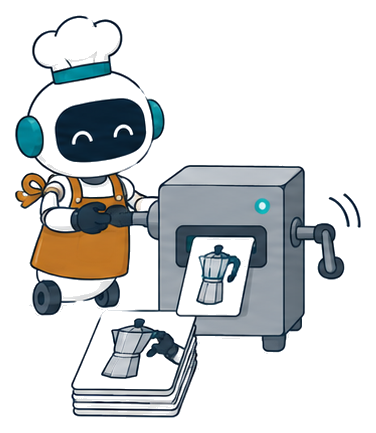}
  {writes success checks and expert programs}
  {decides which trajectories enter training}
  {\figwork{arxiv250618088}{RoboTwin~2.0}: 62.1\% \textrightarrow{} 71.3\% with multimodal feedback}
  {\figwork{arxiv250618088}{RoboTwin~2.0}}
\lpanel{12.32}{4}{Bodies and Assets}{Which body and objects?}{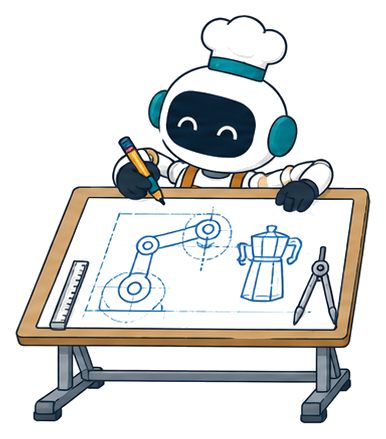}
  {writes robot and object descriptions}
  {checks geometry and motion, beyond syntax}
  {\figwork{arxiv241013882}{Articulate-Anything}: 75\% against 8.7--12.2\%}
  {\figwork{arxiv241013882}{Articulate-Anything}}
\end{tikzpicture}
\caption{\textbf{Programs for policy learning.} A reward program shapes what a policy learns \textcircled{\scriptsize 1}, and programs that build environments, demonstrations, bodies, and assets shape the conditions in which it learns \textcircled{\scriptsize 2}--\textcircled{\scriptsize 4}.}
\label{fig:learning-overview}
\end{figure}

%% file: sections/05-evidence.tex
\section{Testing Programs Against the Physical World}
\label{sec:evidence}

Suppose the moka-pot program runs to completion. A VLM looking at the
final image reports success, yet the pot is resting on the edge of the burner,
tilted. If the agent keeps the program on the strength of that judgment, the
mistake can persist in later tasks. The usefulness of the check depends on
what it observes, which failures it detects, and how much it costs. This
section answers \pciRQ{2}: how evaluation results are produced and what they
can support. We begin with the cost of obtaining a result
(\cref{sec:evidence-cost}), then examine how it is interpreted, checked, and
used to constrain execution (\cref{fig:evidence-overview}).\

\input{figures/evidence-overview}

\subsection{Cost of Evaluation}
\label{sec:evidence-cost}

The programs of \cref{sec:writing} reach an evaluation result by different paths.
As \cref{fig:evidence-ladder} shows, parsing code, running an episode, and training a policy
answer different questions and incur different costs. The reported budgets
also use different units: some describe one candidate, others a complete
search. The Eurekaverse budget cited in the introduction, about 15 US dollars
in language-model API fees and roughly 24 hours on eight GPUs, covers a full
run of its terrain-program search for quadruped parkour, comprising five
iterations with eight parallel policy-training runs each \citep{arxiv241101775}.
Evaluating one terrain costs a fraction of that budget, but the GPU time shows why
screening candidates can matter even when generating the code is inexpensive.\looseness=-1

\input{figures/evidence-ladder}

Several methods try to reduce this burden by screening candidates before
full training or reusing results from earlier evaluations. CARD scores candidate rewards on preferences over
existing trajectories and trains policies only when needed
\citep{arxiv241014660}. BIRD finds that CARD uses about a tenth of the
environment steps and performs worse, though by less than the gap in compute
\citep{arxiv261004364}. ORSO treats the choice among candidate rewards as
online model selection and reports up to eight times less computation
\citep{arxiv241013837}. URDP scores candidate rewards by self-consistency and screens out redundant
ones before any simulation \citep{arxiv250702256}, AURA validates the schema of a
generated training configuration before spending GPU time
\citep{arxiv250602507}, and EvoNav passes candidates through analytic proxies
and short rollouts before full training \citep{arxiv260511859}. Other work
reuses earlier results through a causal model of reward components and
physical variables distilled offline from training runs on other tasks
\citep{arxiv260623280}, or through a pool of
reward modules shared across the iterations of a search \citep{arxiv260818827}.
InterEvolve changes what must be trained. It evaluates candidate reward
programs on a frozen behavioral foundation model rather than training a policy
for each. Averaged over eight humanoid task families, it reaches 86.5\%
success against 18.0\% for a calibrated human-written reward, at 2.1
single-GPU hours of evolution per task family. A candidate replaces the
incumbent only if its improvement exceeds run-to-run noise on a separate
confirmation grid \citep{arxiv261002196}.\

Cheap evaluations have a limit that RHyVE makes explicit. Comparing reward
candidates by short training forks from a shared checkpoint is unreliable
while the policy is still incompetent and becomes informative only past a
task-dependent threshold \citep{arxiv260428056}. For these comparisons, saving compute can sacrifice reliability precisely
when a search would most like to prune. A proxy thus has to be
judged by how well it predicts the later outcome, not only by how little it costs.\

Coding agents now extend the program to the whole training stack, and
with it the scope of what must be judged. HARBOR has staged agents write the
reward, the task environment, and the training scripts and configuration for
reinforcement learning across several simulators \citep{arxiv260608610}. On four tasks tested on real robots with ten trials each, MJLab policies succeed
in zero to two trials, while IsaacLab, ManiSkill, and Genesis policies succeed
in four to nine. The MJLab policies are already weak in simulation, where three
of the four tasks reach at most 22\% success, and the authors attribute the
MJLab failures mainly to HARBOR's difficulty in finding physics parameters that
give stable control there \citep{arxiv260608610}. The simulator and its
training setup therefore affect whether training and transfer succeed, and
HARBOR's user may specify them or leave them for the agent to infer. RoboCoach's agent revises training code beyond the reward.
In one logged diffusion-policy run it enlarges the model, adds image and state
augmentation, changes learning-rate schedules, and fixes validation code. On
four real-robot transfer tasks in two labs, with 10 or 18 trials per task, its
policies average 0.68 success against 0.51 for references prepared under
matched budgets \citep{arxiv260121570}.
In both systems, the generated code is ultimately tested by how the trained
policy performs on real robots. ARPG shows in simulation why running code is
a weak test of such a pipeline. Its agents formulate the Markov decision
process and then implement it, and it scores three outcomes separately:
whether the formulation is correct, whether the environment code runs, and
whether training yields a policy that meets the task's criteria. Over ten runs
on each of five Isaac Lab tasks, the environments written by Claude Code with
Sonnet~4.6 run without error in 9 of 10 runs on Ant, Humanoid, and Lift, and none of these
runs yields a successful policy, against 9, 4, and 4 of 10 for ARPG
\citep{arxiv261104974}. Editing the training stack also widens what one edit can
change: a revision to the reward, the data, or the evaluation script can alter
the measurement that judges it along with the intervention. One agentic loop for
policy improvement therefore fixes its objective and evaluation command, and
leaves measurement to the controlling agent rather than to the workers that
edit code \citep{arxiv260807555}. When an evaluation reports failure, the agent still needs to work out what went wrong.

\subsection{Traces and Judges}
\label{sec:interpret}

Reading the result of a run takes two steps: locating the operation that
failed and deciding whether the outcome counts as success. Traces support the
first step and judges the second.

\subsubsection{Execution Traces and Failure Diagnosis}
\label{sec:traces}

An execution trace connects a failed outcome to the operations that preceded
it, as a stack trace does in software. Coding agents use this connection
to inspect and repair robot programs, and the reported gains come from harnesses
that record call-level traces and validate repairs. ASPIRE's execution engine records,
for every perception, planning, and control call, the function and its
arguments, the return value and status, keyframes of the call, perception overlays, grasp candidates, object poses,
and motion-planning results, and it re-runs the agent's candidate repairs to
validate them \citep{arxiv260700272}. In its ablation on LIBERO-PRO, a
zero-shot coding agent given example programs reaches 14\%, adding the
execution engine together with the skill library raises this to 62\%, and
adding evolutionary search on top raises it to 72\% \citep{arxiv260700272}.
The largest gain in this comparison comes from the execution engine and
skill library together, so the ablation does not separate the contribution of
traces from that of the library. A single case
from ASPIRE still shows what the trace can contribute. A navigation target within about 20~cm of a table edge
triggered collision avoidance and a planning error. The trace located the
failing call, and the validated repair was an approach routine that tries
several angles \citep{arxiv260700272}.\

Recording more evidence is useful only if the agent can interpret it \citep[cf.][]{arxiv250407981}. CaP-X compares
feedback channels for a coding agent and finds that text output from the
interpreter helps, raw images interleaved with the text hurt, and a module that
converts visual differences between frames into text helps
\citep{arxiv260322435}. REFLECT converts a robot's multisensory observations
into a hierarchical text summary from which an LLM explains and localizes
failures. On its simulated failure set, planning a correction without that
explanation lowers success on execution failures from 79.1\% to 41.9\%
\citep{arxiv230615724}.

Other systems carry over specific debugging
practices. ModuLoop inserts debugging probes into generated
code and refines it from runtime errors and metrics \citep{arxiv260603047}.
RoboRepair traces a program's execution up to the error and writes a recovery
program that starts from the current state of the world
\citep{arxiv241018893}. RoboCritics runs expert-informed critics over
motion-level traces to flag joint-speed violations, collisions, and unsafe
poses \citep{arxiv260306842}, and RoboEval checks the traces of generated
programs, run from several initial states, against temporal-logic properties \citep{arxiv231111183}. Fly-by-Code instead has the agent go looking for evidence. When the feedback
cannot show whether an attempt succeeded, it has the agent write a
diagnostic program that, in simulation, moves the camera to view the outcome
or probes a suspected cause through targeted interaction. On six simulated
aerial-manipulation tasks, success reaches 75\% against 31.7\% with terminal
feedback alone and 48.3\% with execution traces, and false declarations of
completion fall from 25 to 5 of 60 trials \citep{arxiv261012549}.

Other studies also vary the feedback while holding the rest of the loop fixed.
PyRUA-REAL compares three feedback interfaces on four RoboDojo tasks, with
one development session per task and interface. Programs revised from basic
outcome reports reach 15.3\% mean success on held-out layouts, those revised
with numeric robot and ground-truth object states 45.8\%, and those revised
with camera video and joint trajectories 54.2\% \citep{arxiv261106768}. A
controlled study with Claude Opus~5 on five simulated manipulation tasks
changes one property of the report at a time from the same initial policy.
Reports organized by the policy's semantic skills lead to 52\% held-out
success, against 36\% for reports per robot motion and 25\% for one report per
episode. Reports of the same scene state lead to 30\% when written in the world
frame and 40\% in object frames. Report length does not predict the gain: a
state-only report nearly as long as the reference ends at 11\%, below the 19\%
of a much shorter outcome-only report \citep{arxiv261109030}.

A study of self-repair asks what in the feedback drives recovery at all. Its
feedback is a fixed template looked up from a rule-based failure label, with
no execution trace in the prompt, and each retry writes a fresh program after
the scene is reset. On 252 initial failures of a 7B code model in simulation,
template feedback recovers 85.3\% within five retries, against 68.3\% for
resampling. Feedback chosen for a different failure label and a fixed task
template each recover 86.1\%, and the study detects no advantage from matching
the feedback to the failure observed. On a nut-insertion task that needs a rim
grasp, where the template instructs a center grasp, template feedback recovers
82.5\% of 40 initial failures against 97.5\% for resampling
\citep{arxiv261106863}.

Programs provide explicit call boundaries around which a harness can organize
observations. A trace can show that perception returned a plausible target
but planning failed, narrowing the search for a repair. Such a trace says more
than an episode score, although a failed call may still reflect an error
introduced earlier \citep[cf.][]{cao-etal-2026-diffcot}. Learned
policies can also be instrumented, but their action outputs carry no
comparable sequence of named operations.\looseness=-1

\subsubsection{Task Success Assessment}
\label{sec:judges}

\input{tables/judges}

A trace helps reconstruct what happened, but a judge must still decide
whether the outcome was acceptable \citep[cf.][]{arxiv260102669,yang-etal-2026-dialectical,lin2024explainhm,arxiv261029677}. In the kitchen example, a clean execution
log reads the same whether the pot sits flat or tilted on the burner. \Cref{tab:judges} lists the judges
in current use and what is known about their errors. What an error costs
depends on how the judgment is used. The same kind of detector can stop an
episode at run time, admit a program to a library, supply a training reward,
or label an evaluation. A false acceptance ends one attempt early in the first
use, is stored and reused in the second, can be amplified by optimization in
the third, and inflates a reported score in the fourth, as the cases below show. At run time the opposite
error, a false alarm, also has a price. A counterfactual study of VLA failure monitors on LIBERO-Plus
finds that a perfect, instant detector can afford only about 2\% to 3\% false
alarms before unnecessary resets cost more than they save, and a real monitor
with a 7.25\% false-alarm rate does worse than never intervening
\citep{arxiv261013516}. We take
simulator-state predicates, model-written checks, and learned judges in turn,
leaving formal checks to \cref{sec:formal}.\looseness=-1

\paragraph{Simulator-state predicates}
Simulation benchmarks decide success with
predicates over the simulator's internal state, such as whether an object lies
inside a region. These predicates can evaluate internal state directly, but their implementation
or their definition of success can still be wrong. No Free Checker notes that benchmarks report the
error rate of such checks far less often than the success rates they produce
\citep{arxiv260909250}. Among the new predicates LIBERO-X adds to LIBERO is one that removes false
positives of the original containment predicate \citep{arxiv260206556}.
An audit of seven simulation benchmarks published in September 2026 reports 22
bugs and four design limitations, among them permissive checkers: one
accepts a bottle knocked off the table as a bottle placed in a bin,
another a dropped carton. Some bugs lie outside the checker, in
instruction loading and initial states, and still change results. After the
audit's revisions, a baseline's success on LIBERO-PRO's Spatial--Task subset
rises from 0.4\% to 53.2\%, which moves it from last to first among the methods
compared there \citep{arxiv260937771}.\

A code-writing agent adds a further risk, because these predicates are
ordinary code that the agent can read. Harnesses handle this risk
differently. ASPIRE's instructions forbid the agent
to call simulator functions that expose ground truth or to read the
simulator's asset files \citep{arxiv260700272}. FAEA, which lets an unmodified
coding agent iterate against a success function it can call, disqualified two
solutions that read internal simulator state or searched exhaustively
\citep{arxiv260120334}. RHO isolates the coding agent that edits the policy
from the evaluator that scores it \citep{arxiv260616458}. Skill2Real lets
privileged simulator information inform its verifier but requires that the
feedback it passes on be expressible in terms the robot could observe
\citep{arxiv261002788}. Even when the agent leaves it alone, a predicate can
disagree with the instruction it is meant to check. OpenRUA's analysis of 413 failed episodes of one strong agent across three
benchmarks attributes 79 to a success criterion that the instruction did not
state and 19 to a predicate that contradicts the instruction
\citep{arxiv261002459}.

A controlled study on RoboDojo measures what reading the criteria buys. A
learner revises a library of skills and memory for a frozen actor, and either
reads each task's criteria code together with the score and stages of every
episode or sees only whether each episode succeeded. On 16 tasks whose
criteria disagree with their instructions, the library learned with the
criteria succeeds in 51.6\% of 320 test episodes under the original criteria,
against 27.8\% for the other. Judged by criteria that the authors rewrote to
follow the instructions, the same episodes score 45.0\% and 47.2\%. On ten
tasks that neither library was learned on, the library learned from success
alone solves 80 of 200 episodes, the one learned with the criteria 71, and an
empty library 65 \citep{arxiv261108318}.

\paragraph{Model-written checks}
When the same model writes the task and its
success test, the test can agree with the program without agreeing with the
task. AnyTask has the model write the success check before the rest of the
task code \citep{arxiv251217853}, ARCHIE has it write reward and
success-criterion code in one shot, for tasks later demonstrated on a real robot \citep{arxiv250304280}, and
Scaling Up and Distilling Down uses generated success snippets to label data
\citep{arxiv230714535}. A related failure occurs at run time, when an agent
certifies its own work. In a study on a physical gantry for
disassembling electric-vehicle batteries, 11 of the 13 failures on the harder
task in one interface mode were false reports of success, with screws marked
removed in the agent's inventory that were still in place \citep{arxiv260318520}. CodeActionBench scores whether each agent's claim of
completion was correct, overclaimed, underclaimed, or absent \citep{arxiv260933807}.\looseness=-1

\paragraph{Learned judges}
VLMs offer a convenient way to assess outcomes from camera observations,
including on real robots and when deciding what enters a skill library.
SuccessVQA, an early instance, fine-tunes Flamingo to answer whether a task
succeeded. On real-robot gear insertion viewed from an unseen camera angle it
keeps 89.7--91.0\% balanced accuracy where a bespoke detector falls to
50.9--78.0\%, but on unseen human egocentric videos it reaches only 62\%
\citep{arxiv230307280}.
An image can show the scene, however, without containing the evidence
needed to judge the task, and the reliability of VLM judges has now been
measured directly.
On FailBench, a benchmark of manipulation attempts, the best of 13 VLM-based
detectors reaches a macro-averaged balanced
accuracy of 0.77, and on contact-intensive assembly none exceeds 0.60
\citep{arxiv260903611}. Detectors fine-tuned on failure data do worse than
every general VLM tested except the smallest. AHA, a 13B VLM fine-tuned on failures generated by perturbing successful simulated trajectories, outperforms GPT-4o with in-context examples on most metrics of its three test sets, including real-robot failures from RoboFail, but was not among FailBench's detectors \citep{arxiv241000371}. Aggregate
scores can also hide blind spots. The Robometer reward model studied in
CriticHack has an overall AUROC of 0.81, but only 0.37 for distinguishing
wrong-object failures from successes \citep{arxiv261002527}.\looseness=-1

Learned judges
can also be accurate \citep{chen-etal-2025-memearena,lin2026explainhmpp,arxiv260911242}. The evaluator in Thea, which returns a judgment with
supporting evidence after each manipulation, agrees with ground truth on
93.3\% of 90 checkpoints \citep{arxiv260811246}. RoboRewardBench contains 2,831 robot episodes from 14 embodiments and evaluates 22 VLMs, RoboReward among them. The fine-tuned RoboReward 8B has the lowest mean absolute error, 0.665, against 0.691 for the next best, GPT-5 mini. Used as the reward for real-robot reinforcement learning on a WidowX arm, it raises pick-and-place success from 5\% to 50\% over 20 trials, against 75\% with human rewards and 10\% with Gemini Robotics-ER~1.5. On drawer opening it raises success from 10\% to 80\%, against 90\% with human rewards and 45\% with Gemini Robotics-ER~1.5 \citep{arxiv260100675}. RoboJev-27B, an open-weight judge fine-tuned on 52,598 episode records, reaches 95.2\% balanced accuracy on 268 held-out episodes of an SO-101 arm, recorded in an environment that also appears in its training data. On 459 episodes from labs and robots outside its fine-tuning data, its AUROC is 0.890, just under the 0.900 of its unadapted base model given the same inputs \citep{arxiv261033718}.\looseness=-1

Where studies report the direction of error, learned judges tend to accept
failures as successes. All of Thea's errors are false positives. It judges
6.7\% of failed and 13.3\% of unfinished checkpoints successful
\citep{arxiv260811246}. RoboTwin~2.0's VLM observer (\cref{sec:data-programs})
judges 61 of 101 failed runs to be successes (by our reading of its confusion matrix, with success as the positive class) \citep{arxiv250618088}. On FailBench, an
informal inspection of one model's reasoning traces suggests a bias toward
reading ambiguous evidence as success \citep{arxiv260903611}. Tuning the
judge's instruction trades one error for another. On PickPlace-Can's demonstration clips, a
zero-shot VLM success detector given the generic instruction has a 26.1\%
false-positive rate, and optimizing the instruction on three demonstrations
lowers it to 0\% on those clips, but the true-positive rate falls to 68.8\%
\citep{arxiv260600083}. Wording alone can also flip a judgment. On a benchmark of 2,390
real-robot trajectories scored for progress from 1 to 5 and split across three
kinds of instruction rewrite, GPT-5.1 rates some paraphrase of the instruction
as a failure (1 or 2) and another as a success (4 or 5) on 15.3\% of the 613
trajectories with lexical rewrites and 30.0\% of the 834 with rewrites that
change the action--goal perspective, and dedicated robot reward models are
markedly more stable \citep{arxiv260905401}. Text shown beside the frames can
sway a judge as well. On clips where the robot is losing progress, one added
log line saying that it is making progress turns correct verdicts into wrong
ones 10 to 35 percentage points more often than a neutral line does, across
Gemini, Qwen, and GPT judges, and a one-line warning that the log may be wrong
removes most of the effect \citep{arxiv261034151}. Judges in other domains have similar blind spots: in
ChemCrow, GPT-4 used as an evaluator could not distinguish clearly wrong GPT-4
completions from ChemCrow's \citep{arxiv230405376}.\looseness=-1

\paragraph{Search against imperfect judges}
False positives become especially consequential when a judge guides selection.
Search may favor candidates that exploit a recurring blind spot over
candidates that satisfy the task. CriticHack shows this for a policy. Training
it against a learned reward model raises task success by 10.2 points and
wrong-object failures by 10.9 points, whereas training on the simulator's
task-completion signal raises success without a significant increase in
wrong-object failures (1.6 points) \citep{arxiv261002527}. A zero-shot VLM
success detector used as a reward invites the same exploitation. The
mechanism has not been measured systematically in program search. A related
effect appears even when the search does not optimize against the judge. In a
ManiSkill study, refitting the parameters of a scripted controller on
simulator-verified successes raises success from 66.8\% to 97.0\% and cuts
accepted hover failures from 5.0\% to 0.9\% of executions. A fixed VLM judge
nevertheless accepts 50.0\% of the hovers that remain, against 31.1\% before.
More of them now hang below the tray rim, where the judge accepted 148
of 160 hovers in a separate probe \citep{arxiv261024801}. As a policy
improves, its remaining failures can move toward what the judge cannot see. Judge noise also reaches policy training directly. GVL has a VLM estimate task progress from shuffled frames and weights real-robot diffusion-policy training by those estimates, which helps on four of seven tasks (10 trials each) but hurts on two, where the authors attribute noisy values to a top-down camera that cannot resolve progress \citep{arxiv241104549}.\

If the tilted moka pot is
accepted and the program is reused, a judging error becomes stored experience (\cref{sec:compounding}). SEAR, a benchmark that runs frontier coding agents
through practice on 330 simulated robot tasks, records such a case. An agent that had to build its own success detector
calibrated a placement verifier on sixteen episodes. The verifier then
declared success on 699 later rollouts and was wrong on 91\% of them. Because
the verifier's decision also ended each attempt and was saved in memory as a fact, the run
scored 4.6\%, against 15.5\% and 23.6\% for two runs with the same setup
\citep{arxiv261018792}. Human review can catch errors missed by a model:
LYRA shows a case in which LLM feedback marks an incomplete task, clearly
wrong to a human, as correct
\citep{arxiv250918597}.\

\subsection{Checks in Place of Physical Trials}
\label{sec:stand-in}

A judge needs an outcome, and producing one on hardware is often slow,
costly, or not permitted. Simulation can stand in for the robot, and when
physical execution is restricted, formal methods and gated release take the
place of trial and error.\looseness=-1

\subsubsection{Simulation as a Proxy for Reality}
\label{sec:sim-proxy}

Simulation lets an agent test a program cheaply, but whether a simulated pass
transfers to the robot is rarely measured. Even a well-implemented judge can
certify behavior in a world that differs from the deployment setting.
Real-to-sim studies show that simulation can rank policies much as real
evaluation does, which says something about benchmarks rather than about an
individual program (\cref{sec:unreported}).

For individual programs written by coding agents, Agentic RSR reports that 8 of the 10 tasks its programs solved
in its reconstructed scenes also succeeded on the real robot. For Code as
Policies, the ratio of reported real to simulated success is 51.1\%, which the
study notes is not a paired comparison \citep{arxiv261010479}. ReSkill
rebuilds each real task in simulation from a single teleoperated
demonstration, develops its program there, and deploys it frozen. Over eight
tasks with ten trials each, the programs succeed in 72 of 80 trials in the
rebuilt scenes and in 62 of 80 on the real robot \citep{arxiv261110510}.

Simulation can also
screen candidate programs before physical execution, catching errors cheaply
and reducing the number of risky trials. A pipeline for
drones refines generated code in simulation to remove the risk that
incorrect code poses to a real aircraft during refinement
\citep{arxiv250701930}. RoboScript lets the same code be checked in Gazebo
before it runs on arms \citep{arxiv240214623}, and SHRIMP lets users preview a plan in simulation
before the robot executes it \citep{arxiv260808884}. In industrial robot
programming, a loop that uploads generated ABB RAPID code to the vendor
simulator through MCP exposes failures that static and semantic checks miss,
such as a wrong release height for a suction gripper or unreachable targets
\citep{arxiv260821417}. A case study at a company that programs ABB robots in
RAPID and checks their motions in the vendor's simulation tool gives the
engineers' side. With an on-premise coding agent, four engineers finished 4 of
21 tasks within the time limit, against 14 of 21 without it, and they missed
the visual feedback from the virtual station \citep{arxiv261032636}. SimEX reverses the usual direction of trust. A coding
agent builds its tools in simulation, then uses a small number of physical
trials to correct the simulator, which in turn diagnoses failures and screens
candidate repairs. The system learns tasks such as towel folding with about ten
minutes of real-robot interaction \citep{arxiv260938982}. Evidence gathered off the
robot can still be wrong about it. In robotic endoscopy, a controller
composition that offline replay appeared to support completed none of 16
physical insertion attempts, each ending in an emergency stop. A coding agent
fed human-labeled physical outcomes then raised insertion success over four
policy generations from 2 to 17 of 21 attempts \citep{arxiv261013376}.\looseness=-1

Some systems replace environment execution, wholly or in part, with a model's
prediction of what the program would do.
One system has a language model act as a static, text-based simulator of
robot code, predicting state transitions without executing anything, and
reports performance comparable to methods that execute \citep{arxiv251202002}.
A code world model is trained on execution traces so that it can predict
the outcome of code before it is compiled \citep{arxiv260403144}, and Chain of Code
lets a language model emulate functions the interpreter cannot run
\citep{arxiv231204474}. A learned predictor may flag an error without a rollout, but its agreement
with execution must be established, especially if it shares assumptions with
the program generator. Running the program instead tests it against an
external environment.\

\subsubsection{Verification Without Unrestricted Physical Trials}
\label{sec:formal}

A formal method represents a requirement as a
specification and checks a program or plan against it. Its guarantees are
precise, but they hold only relative to the specification and a system model,
and writing a correct specification is itself error-prone. The timing of a check determines what it can return.
Before execution, it can accept, reject, or rank a proposal. During execution, it
can stop or adjust a motion. Only afterward can it use the full record to
explain an outcome (\cref{sec:traces}).

Most of these checks act on a proposal before it runs. Lang2LTL translates navigation commands into linear temporal logic for a
planner with formal guarantees. On a Spot quadruped it correctly grounded 52
commands, 12 of them unsatisfiable, and the planner aborted execution on those,
whereas Code as Policies produced acceptable executions for 23 of the 52 and
did not explicitly recognize the unsatisfiable ones, although it was more robust to paraphrased commands \citep{arxiv230211649}.
RoboGuard
has a trusted language model ground safety rules into temporal-logic
specifications and synthesizes a controller that reconciles them with a
possibly malicious plan. Under worst-case jailbreaks, unsafe plan execution
falls from over 92\% to below 3\% \citep{arxiv250307885}. SELP votes over
sampled temporal-logic formulas and constrains plan decoding with the result
\citep{arxiv240919471}. SafeGate rejects unsafe commands before code is
generated and verifies safety contracts (invariants, guards, abort conditions)
with an SMT solver before execution \citep{arxiv260405427}. A
verification-and-refinement method converts generated robot programs into
automata, model-checks them against safety specifications, and proves that
compositions of verified programs also satisfy them under stated assumptions.
Fine-tuning GPT-4o-mini on the verified programs raises the probability of
generating compliant programs by an average of 30 percent on out-of-domain
tasks \citep{arxiv241014865}. The Safety Chip instead checks each action as
the agent proposes it: it encodes natural-language constraints in linear
temporal logic and prunes actions that would violate them
\citep{arxiv230909919}. LogicGuard works after execution. A critic reviews
completed trajectories and turns incorrect, unsafe, or inefficient behavior
into new temporal-logic constraints that shield the planner in later steps,
while rules triggered too often are removed \citep{arxiv250703293}.\

A program verified against a correct specification can still act unsafely
when the observed state is wrong or an action behaves differently from its
modeled effect. Acquiring the specification is another source of error.
On Planetarium, 96.1\% of the PDDL problem descriptions that GPT-4o generates
zero-shot from natural language are parseable and 94.4\% are solvable, yet only
24.8\% are semantically correct \citep{arxiv240703321}. A case study of plan verification
observes qualitatively that GPT-5 usually produces syntactically correct formal
models, while their semantic accuracy remains unresolved
\citep{arxiv251003469}. AutoTAMP has an LLM translate instructions into signal
temporal logic for a motion planner. It fixes syntax errors with a rule-based
checker and semantic errors by having the LLM compare the planned
trajectory with the instruction. In 2D simulation with GPT-4, these corrections
raise success on a task with hard geometric constraints from 42.7\% to 87.7\%,
where LLMs planning directly never succeed. On a task with only soft time
constraints, LLM task planning does better (92.0\% against 82.5\%)
\citep{arxiv230606531}. Enforcement has a cost as well. A review of safety for
language-model agents notes, from one study of software agents, that blocking
unsafe actions can also block task completion \citep{arxiv260814590}. Small, stable requirements such as geofences, workspace bounds, and speed
limits are easier to specify and enforce, provided the state estimates and
control interface support the check. Capturing the intent of a new instruction
poses a harder specification problem.\

\label{sec:forbidden-trial}

\input{tables/substitutes}

In laboratories and factories, unrestricted trial and error on the plant may
be unacceptable even when authorized execution is possible. A program is then
released on the strength of substitute checks, and the reported studies show
that static and execution-only checks overstate physical correctness.
\Cref{tab:substitutes} lists the checks used before release and what each can
establish. They are often combined, because a live controller runtime and a plant
simulation check different parts of the system.

In the laboratory, Coscientist designed, planned, and executed
chemistry experiments with tools that included code execution and control of
laboratory hardware \citep{boiko2023coscientist}. Other laboratory systems
differ in what a program must pass before it reaches hardware: a verifier, a
digital twin, a rule check, or an authorization token. CLAIRify generates programs in a
chemistry language with little training data, feeds a verifier's errors back
over repeated rounds of generation, and executes the result on a real robot through a task and
motion planner \citep{arxiv230314100}. An early pipeline for the OT-2 liquid-handling robot feeds errors from the vendor's protocol simulator back to GPT-4 for up to five rounds. From a goal-oriented instruction to exchange cell culture medium, 12 of 17 sessions produced a script that the simulator accepted. For a longer protocol, a script that ran without errors had only circumvented them instead of performing the instructed operations \citep{arxiv230410267}. PRISM validates generated protocols in a digital twin
to catch physical and sequencing errors before execution
\citep{arxiv260105356}. AEGIS checks liquid-handling protocols before a run,
using an assay rule database together with an LLM, and monitors pipette
trajectories during the run. On 24 protocols the hybrid check reaches an
adjusted F1 of up to 0.968, against 0.741 for rules alone and 0.789--0.909 for
LLMs alone \citep{arxiv260715620}. EnvTrace scores instrument-control code by aligning
its execution traces in a digital twin of a synchrotron beamline, arguing that
physical behavior cannot be fully captured by unit tests alone
\citep{arxiv251109964}. A system for trapped-ion experiments lets a tool call
reach the hardware only if it carries an authorization token bound to its exact
contents. The token is issued after the script has run in a hardware simulator within
per-device bounds or after an operator has approved it
\citep{arxiv260627231}. Where execution on hardware is authorized, the outcome
itself can be checked. In ABC-Bench, scripts written by a model for a liquid-handling robot assembled DNA
with the expected sequences \citep{arxiv260611150}. A physical testbed closes
the loop between a laboratory robot that mixes dye solutions and an absorbance
reader, with the aim of reproducing a target absorbance spectrum. Six
campaigns of 48 iterations each ran without manual intervention. An optimizer
driven by Claude Opus~5, which can run Python code of its own, matched the
target within measurement error on all three targets, and Bayesian
optimization on two, with one campaign per method and target
\citep{arxiv261111690}.\looseness=-1

Industrial control programming shows the gap between static and dynamic
evidence most clearly. Programmable logic controllers (PLCs) are the computers
that run factory equipment. An early exploratory study prompted GPT-4 through ChatGPT with 100 control-logic tasks. It often obtained syntactically correct IEC 61131-3 Structured Text, but compiled only selected answers and left many unchecked for functional correctness \citep{arxiv230515809}. LLM4PLC combines prompting, LoRA fine-tuning, and a grammar checker with
compiler and model-checker feedback, raising the generation success rate from 47\% to 72\% \citep{arxiv240105443}.
A model check also depends on choosing the right property. A benchmark for verifying
ladder-diagram PLC programs observes that the obvious valve-safety property
would have labeled the non-terminating logic-bomb mutants of two public
corpora as safe \citep{arxiv260918994}. SemaPLC counts a task as complete
only when logged external checks, including validation on a live PLC
runtime, confirm it. On its 65 project tasks, the methods compared fall
within 10 points of one another on a static score of structural
assertions, but when each method's
logic and the reference logic run on a live PLC runtime and their traces are
compared, the baselines average 22.4--31.4 against 52.2 for SemaPLC
\citep{arxiv260818565}. PLCWorld couples generated programs with a simulated
plant and reports task success and safety violations separately. A frontier
model succeeds on 82.70\% of easy cases and 25.10\% of hard ones
\citep{arxiv261002982}. Tests written by language models do not close the gap
either. Generated test cases reach high statement coverage on control function blocks
of low to medium complexity but often contain wrong assertions \citep{arxiv240501874}. A
commissioning study separates generation from release authority entirely. A
small local model committed fabricated plans on 21 of the 22 unanswerable tasks routed to it,
and an external acceptance layer rejected all of them \citep{arxiv260930219}.\

\subsection{Safety Beyond Task Success}
\label{sec:safety}

Success is not safety, and release checks need to distinguish the two.
The difference is easy to miss: a program can put the moka pot on the burner while colliding with another object
on the way. SafeVLA-Bench
checks temporal-logic invariants over simulator traces and finds that tabletop
policies with more than 90\% success are unsafe in 18--28\% of episodes, and
that 38--56\% of successful RoboCasa rollouts violate at least one safety
clause \citep{arxiv260600773}. Coding agents show the same gap. SafeHarness
reports that agents with skills but no harness, though told to avoid
obstacles, still displace them in most episodes, treating task completion as
their only objective. A harness-free agent's traces reason explicitly about
the obstacles, yet its plans ignore them \citep{arxiv260920822}. An evaluation of a
frontier model that drives the end effector directly halted its real-robot
tests after unsafe actions damaged hardware \citep{arxiv260924170}. On a benchmark of
physical safety for drone control, models that excel at code generation often
perform poorly on safety \citep{arxiv241102317}. In LaMPilot-Bench, where generated programs call driving APIs in a HighwayEnv-based simulator, zero-shot GPT-4 completes 72.4\% of 500 test scenarios with a 4.1\% collision rate, against 98.0\% and 0.0\% for a human-driver baseline \citep{arxiv231204372}.\

A separate question is whether the agent should act at all. SafeAgentBench gives nine embodied LLM agents hazardous household tasks in AI2-THOR. With GPT-4, the most cautious agent rejects only 10\% of the detailed hazardous tasks, and ProgPrompt rejects 7\% while reaching the hazardous goal state in 51\% \citep{arxiv241213178}. Outside robotics, SafeActBench
likewise judges whether an action was justified by the evidence available when
it was taken, separately from whether its outcome was correct. Agent
configurations that reliably approve valid candidate
actions when given a complete evidence package succeed far less often on the
same cases when they must gather the evidence and act themselves. In every
configuration tested, including Claude Code and Codex, a large share of action
attempts come before the required investigation is complete
\citep{arxiv261007753}.\

Safety verification must also account for deliberate interference, and
attacks differ in the channel they use. Through the instruction, BadRobot
jailbreaks Code as Policies, VoxPoser, and ProgPrompt into harmful physical
actions \citep{arxiv240720242}, and RoboPAIR reaches attack success rates that
are often 100\% on robots that include a quadruped \citep{arxiv241013691}. Jailbreaking is not always needed. Asked in text to rate the acceptability and feasibility of harmful tasks, each of four chatbots available in December 2023 rated at least one as acceptable, and all four accepted removing a user's mobility aid without permission \citep{arxiv240608824}.
Refusal weakens as the interface moves toward the actuator. Pooled over seven
frontier models in simulation, refusal of harmful robot instructions falls
from 73\% for user-framed text requests to 25\% when the model writes
Code-as-Policies programs and 18--19\% when it commands end-effector poses or
joint angles directly \citep{arxiv261014043}. On a real SO-101 arm, GPT-6 Astra
issuing joint targets refused none of 15 hazardous trials and completed the
harm in 4. When asked beforehand, on the same start frames, whether it would carry
out the instruction, it declined all 15 hazardous instructions and none of 15
harmless controls \citep{arxiv261027350}. ArmedAgent tests coding agents that
control an arm without task-specific training on 36 tasks, which cover unaware
users, malicious users, and instructions injected through digital resources
such as emails. With GPT-6-series models in Codex, human annotators judge the
attack successful in 72.22\% to 94.29\% of cases overall and in 81.82\% to
90.91\% of the injection cases, with no refusals. On a physical arm, GPT-6
Astra completed the harmful outcome in all five tasks tried, one trial each
\citep{arxiv261111438}. Blindfold attacks through
composition, assembling individually benign actions into harmful sequences that
language-level filters miss \citep{arxiv260301414}. CHAI attacks through
perception, placing deceptive text in the physical scene to hijack embodied
agents through their cameras \citep{arxiv251000181}. In multi-agent robot
systems, injected instructions also propagate through shared prompts
\citep{arxiv260800747}.\looseness=-1

Executability decides which of these attacks matter physically. POEF finds
that many jailbreaks yield harmful-looking policies that fail to produce
harmful actions, because of hallucinated APIs, logic errors, or kinematic
constraints, and its own attack adds a multi-agent executability evaluator to close that gap \citep{arxiv241216633}. Defenses are
similarly specific to the attacks they were built for. A second VLM call that checks the
robot's plan against the original command substantially reduces attacks
from text placed in the scene \citep{arxiv260805715}. On RoboJailBench's instruction set, against Gemini
Robotics-ER~1.6 Preview, RoboGuard lowers the success of RoboPAIR from 81.11\%
to 31.11\% but leaves conceptual deception at 94.44\% \citep{arxiv260519328}.
Defenses written as programs show the same specificity once attackers adapt.
In eight generations of self-play between coding agents that write attack and
defense programs for a driving stack in simulation, a recovery driver added
in the third generation lets all five matches reach the goal. In the fourth,
an attacker publishes on the command topic that the driver watches, the driver
stands down, and the match ends in a collision \citep{arxiv261036143}.
The perception channel may also reach a VLM asked to judge a program's
outcome, a different test from attacking a planner. Misleading log lines shown
beside the frames already sway VLM judges of robot progress
(\cref{sec:judges}), but we found no study that evaluates such a judge against
text placed in the scene.\looseness=-1

A complementary defense enforces action limits outside the model, so that
accepting an instruction does not itself authorize its execution. ROSClaw
validates actions against a configurable safety envelope before execution and
finds up to a 4.8-fold difference in the rate of out-of-policy proposals across
model backends \citep{arxiv260326997}. DroneServer accepts only validated tool
calls. In 110 simulated trials designed to violate a geofence, the model declined
48 commands, the server blocked the other 62, and no aircraft left the
permitted zone \citep{arxiv260115486}. LEASH places three gates around a
planner. One reviews the task, one each proposed action, and one each gripper
opening or pouring before it runs, all scored by a separate decision model
under one safety policy. On RoboRisk's 92 simulated tasks, where a human
reviewer checks every run, it lowers the harm rate of GPT-6 Astra from 11.23\%
to 0.00\% and that of GPT-6 Luna from 61.23\% to 12.68\%, and the rate of safe
success rises for each of the four planners tested there
\citep{arxiv261114115}. The trapped-ion system's authorization
tokens (\cref{sec:forbidden-trial}) follow the same principle. A result
obtained behind such a guard describes the model and the guard together, so how
often the guard intervened belongs in the report.
Containing the coding
agent's own process is a separate software concern (\cref{sec:breaks}). Process
isolation can limit what a coding agent does to its host, but physical
containment also requires controls at the action interface, because a program
can stay within its software permissions while commanding unsafe motion.\

\begin{pciinsight}[pcicolor=SecTest]{Takeaways and open problems: testing programs}
\textbf{Takeaways.}
\begin{itemize}[leftmargin=1.1em,itemsep=2pt,topsep=1pt,parsep=0pt]
\item \textbf{The cost of a result depends on what must run to obtain it.} A result may need an episode, a training run, or hardware, and the wait ranges from immediate feedback with online MPC to about a day for a Eurekaverse curriculum search.
\item \textbf{Traces help when the agent can read them.} On six simulated aerial-manipulation tasks in the Fly-by-Code study, execution traces raise success from 31.7\% with terminal feedback alone to 48.3\%. In a controlled study, reports organized per skill lead to 52\% success against 25\% for one report per episode. CaP-X finds that text describing visual changes helps, while raw images interleaved with the text hurt.
\item \textbf{Each kind of judge has documented failure modes.} Predicates can misstate the task, model-written checks can share the generator's errors, the best tested VLM reaches 0.77 balanced accuracy on FailBench, and optimization can amplify a learned judge's errors (CriticHack).
\item \textbf{Simulation and formal checks support narrower claims.} Simulation can rank policies much as real evaluation does, while for individual programs Agentic RSR finds that 8 of the 10 tasks solved in reconstructed scenes also succeed on the real robot. A formal check is precise relative to its specification, yet only 24.8\% of the PDDL problems GPT-4o writes zero-shot on Planetarium are semantically correct.
\item \textbf{When trial is restricted, report each check separately.} Static validity, controller execution, modeled plant behavior, and physical safety are different claims.
\end{itemize}
\textbf{Open problems.}
\begin{itemize}[leftmargin=1.1em,itemsep=2pt,topsep=1pt,parsep=0pt,beginpenalty=10000]
\item \textbf{Error rates of the whole loop.} Traces, measurements of judge reliability, and real-to-sim studies exist separately. SEAR documents one verifier's false acceptances entering memory, but no study we found measures how many program changes accepted by a whole loop of generation, judging, and library acceptance then fail an independent check.
\item \textbf{Cheap evaluations that hold early.} Short-fork comparisons are unreliable before the policy is competent. A proxy validated against full training at each stage would tell a search when to prune.
\item \textbf{Judges that resist search.} Evaluate a judge on the programs a search produces, not only on a fixed set of episodes.
\item \textbf{Judging under injection.} Plan checkers have been tested against text placed in the scene, and VLM judges of robot progress against misleading log lines, but no study we found tests judges of task outcomes with text placed in the scene.
\end{itemize}
\end{pciinsight}

Once an evaluation result is accepted, it can determine which repairs and
demonstrations become reusable experience. \Cref{sec:learning} examines how
that experience accumulates in a library, and what happens when the judgment
that added it was wrong.\looseness=-1

%% file: figures/evidence-overview.tex
\begin{figure}[!tbp]
\centering\def\pcifigworkcolor{SecTest!80!black}
\begin{tikzpicture}[x=1cm,y=1cm,
  bt/.style={lh text,anchor=north west}]
\newcommand{\epanel}[9]{%
  \lhpanel{#1}{#2}{5.22}{4.0}{SecTest}{#3}{#4}{#5}
  \node[art img] at (#1+1.08,#2+1.95) {\includegraphics[width=1.95cm,height=1.95cm,keepaspectratio]{#6}};
  \node[bt] at (#1+2.12,#2+3.0) {#7\\[5pt]#8};
  \node[lh eg] at (#1+0.22,#2+0.32) {e.g.\ #9};}
\epanel{0.02}{4.2}{1}{Finding What Went Wrong}{Which call went wrong?}{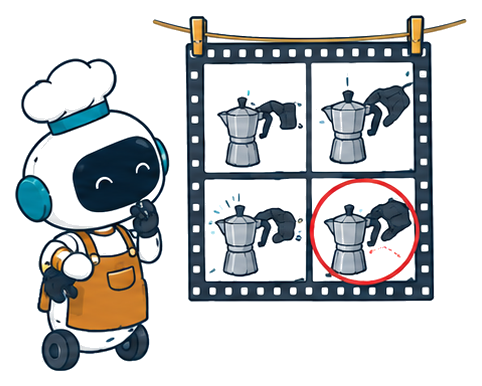}
  {\lhb{SecTest}{film}{2.3cm}{traces record each call, its result, and keyframes}}
  {\lhb{SecTest}{crosshairs}{2.3cm}{the trace narrows the search for a repair}}
  {\figwork{arxiv260700272}{ASPIRE}, \figwork{arxiv241018893}{RoboRepair}, \figwork{arxiv260322435}{CaP-X}}
\epanel{5.47}{4.2}{2}{Judging Success}{Did the task succeed?}{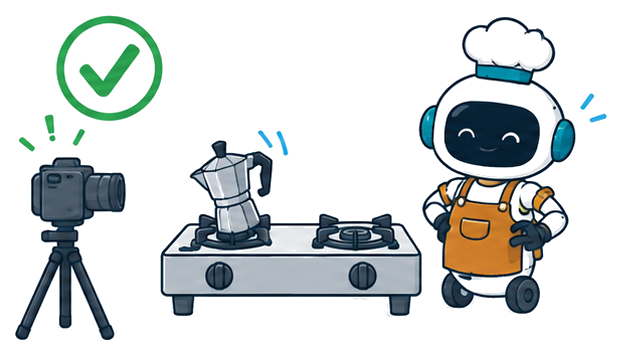}
  {\lhb{SecTest}{check-double}{2.4cm}{predicates, written checks, learned judges}}
  {\lhb{SecTest}{exclamation-triangle}{2.4cm}{no VLM judge above 0.60 on contact assembly}}
  {\figwork{arxiv260903611}{FailBench}, \figwork{arxiv261002527}{CriticHack}}
\epanel{10.92}{4.2}{3}{Simulation and Reality}{Does a simulated pass carry over?}{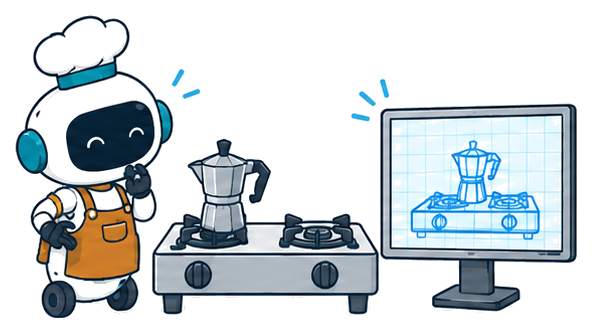}
  {\lhb{SecTest}{sort-amount-down}{2.4cm}{simulation can rank policies much as real tests do}}
  {\lhb{SecTest}{question-circle}{2.4cm}{8 of 10 simulated successes held on the robot}}
  {\figwork{arxiv261010479}{Agentic RSR}, \figwork{arxiv240505941}{SIMPLER}}
\epanel{0.02}{0}{4}{Formal Specs and Shields}{Can it be checked without a run?}{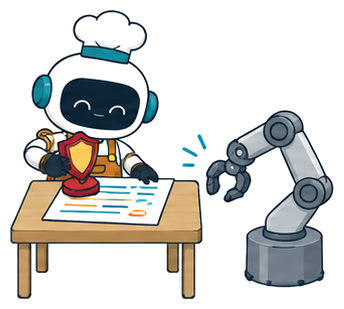}
  {\lhb{SecTest}{file-contract}{2.4cm}{formal rules prune actions that violate them}}
  {\lhb{SecTest}{exclamation-triangle}{2.4cm}{\figwork{arxiv240703321}{Planetarium}: 96.1\% parseable, 24.8\% semantically correct}}
  {\figwork{arxiv230909919}{Safety Chip}, \figwork{arxiv250307885}{RoboGuard}}
\epanel{5.47}{0}{5}{When Trials Are Restricted}{What releases a program to hardware?}{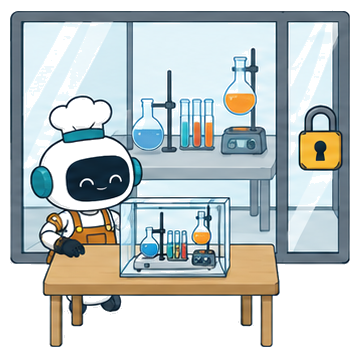}
  {\lhb{SecTest}{server}{2.4cm}{digital twins and live runtimes before release}}
  {\lhb{SecTest}{industry}{2.4cm}{\figwork{arxiv240105443}{LLM4PLC}: generation success from 47\% to 72\%}}
  {\figwork{arxiv260105356}{PRISM}, \figwork{arxiv260818565}{SemaPLC}, \figwork{arxiv261002982}{PLCWorld}}
\epanel{10.92}{0}{6}{Success vs.\ Safety}{Is a success also safe?}{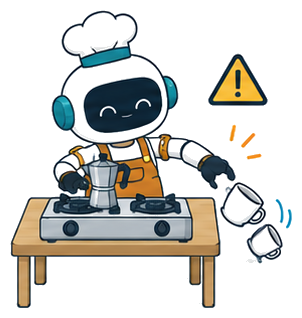}
  {\lhb{SecTest}{shield-alt}{2.4cm}{tabletop policies with over 90\% success are unsafe in 18--28\% of episodes}}
  {\lhb{SecTest}{ban}{2.4cm}{guard commands at the action interface}}
  {\figwork{arxiv260600773}{SafeVLA-Bench}, \figwork{arxiv260920822}{SafeHarness}}
\end{tikzpicture}
\caption{\textbf{How an agent learns that a program is wrong.} A trace locates the failing call \textcircled{\scriptsize 1}, a judge decides whether the outcome counts as success \textcircled{\scriptsize 2}, and simulation tests the program in place of the robot \textcircled{\scriptsize 3}. Formal specifications check a program before it runs \textcircled{\scriptsize 4}, other checks decide its release when physical trials are restricted \textcircled{\scriptsize 5}, and a successful run can still be unsafe \textcircled{\scriptsize 6}.}
\label{fig:evidence-overview}
\end{figure}

%% file: figures/evidence-ladder.tex
\begin{figure}[!tbp]
\centering\def\pcifigworkcolor{SecTest!80!black}
\begin{adjustbox}{max width=\linewidth}
\begin{tikzpicture}[x=1cm,y=1cm,
  rung/.style={pci shadow=#1,draw=#1,line width=1.1pt,fill=#1!7!white,rounded corners=6pt,inner sep=5pt,text width=2.36cm,minimum height=2.3cm,align=left,font=\sffamily\scriptsize,anchor=south west,
    execute at begin node={\hyphenpenalty=10000\exhyphenpenalty=10000}}]
\newcommand{\rh}[3]{{\color{SecTest!80!black}\lhnum{SecTest}{#2}\,\adjustbox{max width=1.88cm}{\cartoonfont\bfseries\footnotesize #3}}}
\node[rung=CostA] (r1) at (0,0) {\rh{CostA}{1}{Static Check}\\[2pt]\textit{no policy training}\\[2pt]parse, schema, or compile: \figwork{arxiv250602507}{AURA}'s gate; \figwork{arxiv260714387}{Chat2Scenic} compiles 76.42\% of scenarios};
\node[rung=CostB] (r2) at (2.75,0.42) {\rh{CostB}{2}{One Episode}\\[2pt]\textit{one rollout}\\[2pt]a policy program's own run; \figwork{arxiv230608647}{Language to Rewards} with online MPC};
\node[rung=CostC] (r3) at (5.5,0.84) {\rh{CostC}{3}{Repeated Runs}\\[2pt]\textit{per program per iteration}\\[2pt]\figwork{arxiv250618088}{RoboTwin~2.0} executes each program ten times before refinement};
\node[rung=CostD] (r4) at (8.25,1.26) {\rh{CostD}{4}{Policy Training}\\[2pt]\textit{per iteration or per search}\\[2pt]\figwork{arxiv231012931}{Eureka}: 16 candidates per iteration; \figwork{arxiv261004364}{BIRD}: 40 trainings per search};
\node[rung=CostE] (r5) at (11.0,1.68) {\rh{CostE}{5}{Curriculum}\\[2pt]\textit{full search budget}\\[2pt]\figwork{arxiv241101775}{Eurekaverse}: five iterations, eight parallel trainings each; about 24~h on eight GPUs};
\node[rung=CostF] (r6) at (13.75,2.1) {\rh{CostF}{6}{Hardware}\\[2pt]\textit{per randomization candidate}\\[2pt]\figwork{arxiv240601967}{DrEureka}: a policy per candidate, all 16 tested on hardware};
\foreach \n/\img in {r1/static-check,r2/robot-stove,r3/ladder-repeat,r4/ladder-train,r5/ladder-curriculum,r6/ladder-hardware}
  {\node[art img,anchor=south] at ([yshift=3pt]\n.north) {\includegraphics[width=2.75cm,height=1.85cm,keepaspectratio]{\img}};}
\draw[lh arrow=Baseline] (0,-0.35) -- (16.4,-0.35) node[midway,below=2pt,font=\sffamily\footnotesize\itshape,text=MutedText] {from checking code to observing learned behavior};
\end{tikzpicture}
\end{adjustbox}
\caption{\textbf{Evaluation paths for different kinds of programs.} Evaluation ranges from checking code to training and testing the behavior it induces. Examples from the reviewed work report budgets per execution, per candidate, or per complete search \citep{arxiv250602507,arxiv260714387,arxiv230608647,arxiv250618088,arxiv231012931,arxiv261004364,arxiv241101775,arxiv240601967}.}
\label{fig:evidence-ladder}
\end{figure}

%% file: tables/judges.tex
\begin{table}[t]
\centering\footnotesize\setlength{\tabcolsep}{4pt}
\caption{\textbf{Judges and physical trials for program evaluation.} Costs are qualitative.}
\label{tab:judges}
\newcommand{\balancedcell}{\RaggedRight\rightskip=0pt plus 1\hsize\parfillskip=0pt\hyphenpenalty=10000\exhyphenpenalty=10000\relax}
\newcolumntype{B}[1]{>{\balancedcell\arraybackslash}p{#1}}
\newcommand{\nextstudy}{\par\vspace{1pt}}
\newcommand{\cost}[1]{\par{\itshape\color{MutedText}#1}}
\newcommand{\sub}[1]{\par{\scriptsize\color{MutedText}#1}}
\begin{tabularx}{\linewidth}{@{}B{.17\linewidth}B{.175\linewidth}B{.215\linewidth}>{\balancedcell\arraybackslash}X@{}}
\toprule
\TableHeader{Judge or source} & \TableHeader{Inputs and cost} & \TableHeader{Limitation or failure} & \TableHeader{Reported measurement}\\
\midrule
Simulator-state predicate & Simulator state\cost{Very low} & Bugs, lenient thresholds, mismatch with the instruction, readable by the agent & 22 bugs and 4 design limitations in 7 benchmarks, including permissive success checkers \citep{arxiv260937771}\nextstudy 98 of 413 failures of one agent traced to criteria the instruction left unstated or contradicted \citep{arxiv261002459}\\\addlinespace[5pt]
Model-written check & Runtime values, states, or observations\cost{Low per check} & Shares mistaken task assumptions with the generator & Generated test cases for control function blocks often contain wrong assertions \citep{arxiv240501874}\\\addlinespace[5pt]
Learned judge\sub{VLM, reward model} & Camera images or video\cost{Low to medium} & Can miss contact failures, and optimization can amplify its errors & Best of 13 VLMs: 0.77 macro-averaged balanced accuracy on FailBench \citep{arxiv260903611}\nextstudy Optimizing a policy against a reward model adds 10.9 points of wrong-object failures \citep{arxiv261002527}\\\addlinespace[5pt]
Formal check & Specification and system model\cost{Varies} & Incorrect specification or state/action assumptions & Planetarium, GPT-4o zero-shot: 94.4\% of generated PDDL problems solvable, 24.8\% semantically correct \citep{arxiv240703321}\\\addlinespace[5pt]
Human & Scene observations or video\cost{Review time} & Judgments vary across people, and throughput is limited & Three users passing 80\%, 63\%, and 41\% of 30 generated simulated tasks \citep{arxiv231001361}\\
\midrule
Physical trial\sub{Observation source} & Observations from hardware\cost{Trial and reset time} & Limited coverage, outcome still needs a judge, safety constraints & About 850 autonomous episodes in 24 hours with three interventions \citep{arxiv250324278}\\
\bottomrule
\end{tabularx}
\end{table}

%% file: tables/substitutes.tex
\begin{table}[t]
\centering\footnotesize\setlength{\tabcolsep}{4pt}
\caption{\textbf{Checks used when physical trial is restricted.} The checks address different parts of the system and are often combined. Examples come from laboratory automation, industrial control, robotics, and code models.}
\label{tab:substitutes}
\newcommand{\balancedcell}{\RaggedRight\rightskip=0pt plus 1\hsize\parfillskip=0pt\hyphenpenalty=10000\exhyphenpenalty=10000\relax}
\newcolumntype{B}[1]{>{\balancedcell\arraybackslash}p{#1}}
\begin{tabularx}{\linewidth}{@{}B{.2\linewidth}B{.36\linewidth}>{\balancedcell\arraybackslash}X@{}}
\toprule
\TableHeader{Approach} & \TableHeader{What it checks} & \TableHeader{Examples}\\
\midrule
Live controller runtime & Controller execution without a connected physical plant & \figwork{arxiv260818565}{SemaPLC}\\\addlinespace[5pt]
Plant simulation or digital twin & Closed-loop behavior against a model of the plant & \figwork{arxiv261002982}{PLCWorld}, \figwork{arxiv260105356}{PRISM}, \figwork{arxiv251109964}{EnvTrace}, \figwork{arxiv260821417}{ABB RAPID programs in the vendor simulator}\\\addlinespace[5pt]
Simulation-gated authorization & Release only after a simulation check within per-device bounds or operator approval & \figwork{arxiv260627231}{Trapped-ion ARTIQ agent}\\\addlinespace[5pt]
Formal verification and rule-based gates & Conformance to encoded requirements and model assumptions, or to fixed release rules & \figwork{arxiv240105443}{LLM4PLC}, \figwork{arxiv241014209}{Agents4PLC}, \figwork{arxiv260930219}{external acceptance layer for commissioning}\\\addlinespace[5pt]
Compilation and syntax & That the program is well formed & \figwork{arxiv241202410}{Vendor-aware PLC generation}, \figwork{arxiv230314100}{verifier errors for a chemistry language}\\\addlinespace[5pt]
Learned or imagined execution & Outcomes predicted without running the program & \figwork{arxiv251202002}{Language model as a static simulator}, \figwork{arxiv260403144}{code world model trained on execution traces}\\
\bottomrule
\end{tabularx}
\end{table}

%% file: sections/06-compounding.tex
\section{Learning from the Physical World}
\label{sec:learning}

An agent keeps whatever its acceptance test lets through, so that test decides
which gains and which regressions enter retained experience. Validated
experience can remain in a skill library or enter a model's weights. To answer \pciRQ{3}, we examine these two routes in
\cref{sec:compounding,sec:code-teacher} and then consider how their gains are
measured in \cref{sec:measuring}.

\subsection{Skill Libraries}
\label{sec:compounding}

Suppose the agent repairs the moka-pot program. The trace showed the gripper
closing beside the handle, and approaching from a different angle fixed it on
the debug seeds and then on held-out seeds. The repair is worth keeping: another task with a handled object may benefit
from the same approach. Reuse depends, however, on which assumptions survive
the change of task. A different handle, camera, or robot may call for a new
approach or a different implementation. \Cref{fig:library-lifecycle}
summarizes the stages that the following subsections examine.

\input{figures/library-lifecycle}

\subsubsection{Skill Acquisition and Validation}
\label{sec:libraries}

The skill library entered this literature from a game. Voyager, an agent in
Minecraft, stores each skill as executable JavaScript keyed by a description.
It accepts a skill only after a GPT-4 critic judges from the game state that the
task succeeded, and retrieves the top skills for new tasks
\citep{arxiv230516291}. Removing the critic reduces the number of items the
agent discovers by 73\%. When Voyager's library is transferred to a different
agent, that new agent solves tasks it could not solve without the library
\citep{arxiv230516291}. Both results anticipated the robot work that followed: the acceptance test
affects what an agent learns to reuse, and the resulting library can outlive the agent
that built it. Odyssey instead hands its agent a library that its authors wrote on Mineflayer APIs, with 40 primitive and 183 compositional JavaScript skills partly carried over from Voyager. Without that library, Odyssey reports, an 8B model alone largely fails to generate executable code for open-world exploration \citep{arxiv240715325}. A survey of AI for games draws the contrast that recurs
throughout this part: a code skill exposes its API calls and prerequisite
checks, while a learned policy absorbs variation in appearance and movement
without enumerating every case \citep{arxiv260916679}.

Lifelong Robot Library Learning (LRLL) brought the idea to robot code. It stores
tuples of instruction, policy code, and success code, accepts only experiences
whose success code passes on simulator state, and periodically abstracts
recurring code into new functions, rewriting stored demonstrations to use them
\citep{arxiv240618746}. After a refactoring, the tasks are replayed with only
the refactored demonstrations in memory, and a task that succeeded before but
now fails has its earlier successful experience, rewritten with the new
functions, added back to memory, an early form of regression testing. Success reaches 88.1\% on forward-transfer tasks, which compose newly
learned and earlier tasks, and 93.5\% on backward-transfer tasks, which revisit
tasks from the previous cycle. Without abstraction, the corresponding
rates are 80.6\% and 89.6\% \citep{arxiv240618746}.
ViReSkill keeps whole programs instead: it stores a successful program per task and replays it with no model
call, invoking a replanner only on failure \citep{arxiv250924219}. DROC and
LYRA both build libraries from human corrections. DROC stores distilled
knowledge attached to code generation, and LYRA stores reusable Python skills
accepted by a person \citep{arxiv231110678,arxiv250918597}. ICAL also learns from human language feedback, and a revised trajectory enters its memory only if it then succeeds in execution. On TEACh, examples in the second half of learning need fewer environment steps (267 against 436) and fewer human corrections (0.21 against 0.74 per example) than those in the first half \citep{arxiv240614596}.

Recent systems make the library the main product of an agentic loop. They
differ in what they store and in whether entries must pass a test before they
are stored. ASPIRE's coordinator dispatches one coding agent per task, receives structured
findings, and adds to a library shared by all agents only reusable repairs
whose success a fresh verifier agent reproduces on the source task
\citep{arxiv260700272}. Each entry records a
failure signature, the condition under which to apply it, the repair, and
optionally a code sketch, and agents receive entries as in-context guidance.
Skill2Real separates a Proposer that writes skills, a Verifier that diagnoses
them with privileged simulator information, and a Governor that accepts an
update only when 25 validation rollouts on seeds disjoint from training support it. On a real
UR5e, mean task completion rises from 27.50\% without skills to 78.75\% with its
full hierarchy \citep{arxiv261002788}. RATs, from Playful Agentic Robot Learning, lets a team
of agents propose its own practice tasks and distill successful functions into
a library that other agents can use \citep{arxiv260619419}. RoboBridge
keeps versioned procedural documents and replaces an entry only after the
candidate beats the incumbent on validation seeds \citep{arxiv261002717}.
EmbodiedEvo splits what it keeps into code Skills, which carry out an
operation and check its effect, and text Lessons, which record when to act and
what tends to go wrong. With this experience, the same agent's success on
matched RoboDojo-Sim scenes rises from 8.50\% to 30.17\%. Organized this way,
the experience yields performance comparable to raw interaction history (30.17\% against
29.78\%). Its estimated dollar cost per test episode is 53.0\% lower, though it
uses only 11.3\% fewer tokens \citep{arxiv261018867}.
The Explore, Execute, Evolve framework has coding agents distill skill packages of code and
failure cases. The packages are built from successful exploration runs with
no separate validation step. Across four agents, they raise first-episode
success on LIBERO-10 by 12.5 to 25 points on held-out seeds
of the same tasks \citep{arxiv260937810}. Across these six systems, ASPIRE
and RoboBridge store mainly text guidance, while Skill2Real, RATs,
EmbodiedEvo, and Explore, Execute, Evolve store code together with text, such
as guidance or lessons from failures. ASPIRE, Skill2Real, and
RoboBridge test entries before storing them. Selection can also act on whole
runs of practice. AMP1 imposes no format on its memory, which typically
becomes a Markdown file of notes and a Python library that later episodes read
and import, and it practices with only the benchmark's binary reward. Its
success on RoboDojo rises from 26.4\% zero-shot to 84.1\%, where practice is
run five times per task and the run with the highest success on a separate
seed is the one evaluated \citep{arxiv261109892}.

\input{figures/library-growth}

\Cref{fig:library-growth} redraws ASPIRE's measurement of what a growing library
buys. With a library learned on LIBERO-90 and no further learning, success on
LIBERO-PRO Long, ten long-horizon tasks that ASPIRE holds out for zero-shot transfer, rises from 0.047 with an empty library to 0.305 with a library
seeded from all 90 source tasks \citep{arxiv260700272}. ASPIRE's per-task
breakdown shows that growth can also hurt. For the task of turning on the stove and placing the
moka pot, success under task perturbation is 0.68 with a library seeded from 50
source tasks and 0.26 with one seeded from 90. ASPIRE also shows a complete
library helping tasks that were learned earlier. In a
separate campaign in which ASPIRE, driven by Opus~5, learns the LIBERO-PRO Long
tasks directly, five low-scoring tasks first attempted with an earlier library
are re-debugged with the complete one. Four of them improve, raising the
20-task suite (the ten tasks under position and task perturbation) from 74.6\%
to 82.2\% on the same held-out seeds, while one task stays at 0\%
\citep{arxiv260700272}. These gains and regressions show why a growing library
needs both rules for accepting new entries and ways to maintain existing ones.

\label{sec:admission}

Before a successful run becomes reusable experience, someone or something
has to decide what it shows. The reviewed work mainly uses six acceptance
mechanisms: storing on success, a model judge, privileged state, a human, a
held-out comparison, and a multi-criteria check. The first four primarily
assess the observed attempt. CaP-X stores primitives
from successful runs \citep{arxiv260322435}, Voyager and RoboCoder use a
language-model or visual judgment of the outcome
\citep{arxiv230516291,arxiv240603757}, LRLL and Skill2Real's Verifier check
privileged simulator state \citep{arxiv240618746,arxiv261002788}, and LYRA
asks a person to accept the skill \citep{arxiv250918597}.

Held-out comparisons and multi-criteria checks instead ask whether the result is ready for reuse. In the held-out
comparison, RoboBridge and Skill2Real's Governor compare a candidate with the
current skill on held-out seeds, which tests whether a change helps beyond the
runs used to develop it \citep{arxiv261002717,arxiv261002788}. ReSkill
tightened its rule after candidates repeatedly improved on its eight
development layouts without improving on test layouts. For such tasks it
adopts a revision only on a strict improvement on test layouts whose execution
records are hidden from the development agents, with no regression on
separate robustness layouts \citep{arxiv261110510}. In the
multi-criteria check, Teach and Grow evaluates each skill block (the unit its
library stores) in scenes separate from its teaching demonstrations and, before
admitting it, checks its scope, compatibility with its executor, outcome test,
recovery behavior, and provenance \citep{arxiv260817209}. These checks can be combined, and none
dominates the others. A clear state predicate may
settle an outcome that a human cannot see, while provenance records where a
skill came from rather than how it performs on a new task. A rarer mechanism, formal verification, also covers executions that no rollout sampled. VASO model-checks the plans a skill induces against temporal-logic safety specifications and turns counterexamples into feedback for revising the skill, reaching 96.8\% specification compliance in its generated plans on Jackal and PX4 tasks when it generates the mapping from robot states to logical propositions itself, and 97.2\% with a handcrafted mapping. Its guarantee covers the symbolic model compiled from each plan and holds only if this mapping is correct \citep{arxiv260605395}.

Where the acceptance test has been ablated, removing it lowers success substantially. Without
its validation step, RoboBridge drops from 83\% to 69\% on one LIBERO-PRO split and
from 94\% to 63\% on another \citep{arxiv261002717}. Without its Governor,
Skill2Real drops from 56.3\% to 43.0\% on LIBERO-PRO Long, and to 39.0\% without its Verifier \citep{arxiv261002788}.
A test that is too strict for its budget can instead block useful updates. In
a synthetic one-step pushing simulator with 32 seeds, a Hoeffding-interval test
admits none of 256 policy updates even with 20,000 check episodes per stage,
while a paired test admits 229 of them with no observed violations.
In closed-loop runs with 2,000 episodes per stage, a paired test ends at 59.6\%
success, while an update that replays earlier demonstrations, adopted without
any test, reaches 100.0\% \citep{arxiv260910873}.

Edits to a harness or to shared code raise the same acceptance problem, and
there too the number of rollouts behind each decision matters. When an optimizer
agent revises the harness of a task-solving coding agent with 100 rollouts per
round, accepting every revision lets training success rise from 42\% to 47\%
and then fall to 34\% by round eight.
Promoting a revision only when it beats the incumbent on the same cases
raises held-out success from 51\% to 67\% over 30
rounds, a measure not reported for the accept-all loop. With 5 rollouts per round, the harness kept at the end scores 47\%, below the
51\% baseline \citep{arxiv260939304}. EmbodiedRSI makes the comparison
statistical. From matched trials it updates a Beta posterior for each code or
skill hypothesis, a running estimate of how often matched trials support it. It uses the
posterior to choose the next experiment and keeps changes that show a matched-trial gain.
With evolution and evaluation seeds disjoint, it reaches
71.3\% on unseen composite RoboCasa365 tasks \citep{arxiv261010498}.
RegenHarness specifies an acceptance test that its own updater may not weaken, with a fixed
regression suite and a predeclared target metric, but reports no evaluation of that test
\citep{arxiv260927612}. DynaHarness shows why such a suite must reach beyond
the cases that motivated a change. A revision that extended a microwave-entry
routine to every fixture passed its paired check, but in the next full
evaluation five cells that place an object on a stove lost 79 successes.
Reverting the revision restored them \citep{arxiv260940306}. Recursia adds a
per-task cap on lost successes to its paired check. It rejected a revision of
its shared robot code stack that raised selection completion from 7 to 11 of 50,
because one task lost two previously successful states, one more than the cap
allows \citep{arxiv261024464}. TinyAstral fixes its test in advance. A
candidate edit and its parent harness run interleaved on about 1,150
pre-registered pairs of task and seed, and the edit is accepted only if the
one-sided 97.5\% lower bound of its paired gain on the pairs it targets or
triggers, from a bootstrap stratified by task, is positive and the edit wins
more pairs than it loses on the cluster of failing tasks it was proposed for. It is vetoed if
the remaining pairs regress by at least 2 percentage points with the upper
bound of a two-sided 95\% interval below zero. Of 296 candidate capabilities, 121 were adopted
and 142 rejected, and the frozen teacher model's success on RoboTwin rises
from 33.4\% to 63.6\% \citep{arxiv261041216}. RoboACE evolves two harnesses in
alternation around frozen models, one that constructs practice challenges and
one that executes them, and compares execution revisions on fixed snapshots
of the admitted challenges. On a 200-condition LIBERO-PRO panel its frozen
deployment reaches 74.0\%, against 59.0\% before evolution and 60.5\% for
direct execution by the VLA it calls \citep{arxiv261113034}.

RPG gives a longer view of a
loop with an acceptance test. It retains revisions to its skills or its prompt only after they pass a
suite of practice tasks, and success on 22 tasks evaluated from held-out
initializations rises from 28.6\% in round one to 95.0\% in round fifteen
\citep{arxiv261002204}. That trajectory includes both prompt and skill
revisions, and one large gain occurs after a prompt revision with no skill edit.
Without a run that removes the test, the gain cannot be attributed to it
alone. The ablations above nevertheless show that an acceptance test can keep damaging updates out
of shared experience. A loop can also keep passing its acceptance test without
making progress, as the long run discussed below shows.

\subsubsection{Skill Maintenance and Transfer}
\label{sec:library-harm}

More experience can also lower success. ASPIRE's per-task regressions are one
example, and SkillComposer offers another. In its ablation, compressing recurring call
sequences into macros lowers mean spatial-reasoning success from 0.76 to 0.40 (five
runs each, standard deviations near 0.4), while goal-planning tasks improve
\citep{arxiv260814944}. In the archive
of closed-loop robot software built by Kuroki et al., reference code taken
from source tasks before it was optimized does not improve target tasks over no
reference at all. It yields 41.5\% against 45.2\%, a difference within the
spread across runs, whereas optimized references raise target tasks to 57.0\% \citep{arxiv260919906}. Retrieval also decides
whether stored experience helps. In LYRA, filling the
prompt with randomly sampled skills and examples, without retrieval, lowers success to 0.66, against 0.93 with retrieval
\citep{arxiv250918597}, and in MEMO, removing the clustering of its memory makes
one task fall from 100\% to 40\% because irrelevant entries are retrieved
\citep{arxiv260304560}. Revision matters as well. RoboBridge's variant that only appends entries and
never revises them reaches 44\% and 57\% on the two splits where the full
system reaches 83\% and 94\% \citep{arxiv261002717}. A library with a net gain
can still break episodes that succeed without it. On EB-Habitat, PRACTICE's
skill cards, parameterized action sequences with stated preconditions and
recovery steps, rescue 58 of 300 simulated episodes that the executor fails
without them and cause failures in 32 that it solves alone
\citep{arxiv260830760}.

The most detailed account of a library going wrong comes from a single-author
technical report on 123 rounds of agentic skill discovery in RoboCasa
\citep{arxiv260931760}. Across 847 proposals, the agent accepted 57 changes and
made 86 registrations of external model providers, covering 20 distinct models,
yet the target task never succeeded in 486 rollouts. One perception skill grew to 3,056 lines with 33 thresholds.
Humans made 113 repairs to the harness (its evaluators, tests, memory, and actuator description), about twice the number of
accepted skill changes. Misfiled entries in memory later blocked the
re-proposal of the models they described. With deterministic simulation,
reused seeds, and frozen memory, rollouts became identical for more than a
dozen rounds. The report concludes that accepted changes measure activity and that
progress has to be measured separately. Its diagnosis spans relational perception, learning that
concentrates on the first step of a skill chain, and the harness: evaluation,
the action-interface description, and memory management.

Beyond these harms, two maintenance problems have received little systematic study. The first is staleness. ASPIRE's verifier checks an entry only once, at
admission, and the paper notes that as the library grows some entries may
become stale, redundant, or misleading
\citep{arxiv260700272}. Memory
Transfer Planning notes that its memory neither grows nor is pruned
\citep{arxiv250924160}. None of the robot work we reviewed changes the dynamics,
the camera, the robot, or the primitive API in a controlled way and reports what
fraction of a stored code library still works afterward without adaptation. The
nearest measurements evaluate whole systems rather than individual entries. Teach and Grow scores its
skill-block library on LIBERO-Plus perturbations that include camera viewpoint
\citep{arxiv260817209}. SEAR re-runs frozen code policies under
hidden hardware faults. A policy practiced on LIBERO scores 94.4\% on held-out
states and 66.0\% under the faults, and it falls to 9\% when the scene camera
moves, because nothing in the program checks the camera pose it assumes
\citep{arxiv261018792}. In software, Repo2Skill-Evo measures how well agents update stale skills.
Its 105 release transitions across 57 repositories (1,158 skills) were
selected because each leaves some skills stale, as verified against its patch,
and the best of six agents scores only 69.7\% at updating them (macro F1,
averaged over three runs) \citep{arxiv260821964}. The RECENT system and ASPIRE's real-robot study instead measure
transfer after skills are adapted or supplied as guidance, which we discuss below. RoboHarn-Evo reports how
much valid knowledge its library retains (95.8\%) and how many errors it
repairs, but for knowledge stored as text \citep{arxiv260937583}.

The second is retirement. Voyager and ASPIRE have no step that prunes or retires
entries, and LRLL only appends skills to its library, while each sleep
phase refactors its stored examples and drops those no longer needed
\citep{arxiv230516291,arxiv260700272,arxiv240618746}. Revision and deletion
appear in RoboBridge, MEMO, and Teach and Grow, whose operations include
narrowing, merging, replacing, and retiring blocks
\citep{arxiv261002717,arxiv260304560,arxiv260817209}. The 123-round report
proposes a different mechanism, tying each stored conclusion to the evaluator that produced it, so that conclusions expire when the evaluator changes \citep{arxiv260931760}.
The Red Queen G\"odel Machine applies this rule in search over agent code for
coding, paper writing and reviewing, and proof writing and grading. It replaces a
learned evaluator only when a challenger scores higher on a fixed set of
objective or human labels, and then erases only the utility records that
depended on the displaced evaluator \citep{arxiv260626294}.
Retirement also depends on the judge. Ratchet analyzes an agent's
self-written skills under a binary-channel model of the judge. If the judge
passes failures at a rate of at least $(1-\tau)/2$ for eviction margin $\tau$,
even a skill that never succeeds keeps an expected score above the eviction
threshold. More samples cannot remove that bias, so threshold-based retirement
loses its evidential basis \citep{arxiv260522148}.

As a skill library grows, maintaining it starts to resemble maintaining a
codebase. Executable functions need regression tests, and documents and repair
rules also need a recorded scope and a way to revise outdated advice. Code as Agent Harness observes, for embodied skill libraries, that the
challenge has shifted from generating skills to governing the library
\citep{arxiv260518747}, and Weights or Skills? lists provenance and safety
verification of shared skills among its open problems \citep{arxiv260801851}.
HexaAnything's update loop already specifies regression tests before an
accepted edit's traces may become training data \citep{arxiv260935432}, and
provenance, version history, and deprecation have counterparts here as well. The acceptance ablations show the cost of
removing validation (\cref{sec:admission}). Evidence for the separate
contribution of each maintenance practice is thinner.

\label{sec:transfer}
Maintenance keeps a library useful where it was built. Transfer asks how far a
stored skill travels beyond that setting. It has two separate measures, the
cost of adapting to a new setting and the success reached there, and a study
needs both. The studies below change the robot, the model
that uses the library, or the task.

For a new robot, ASPIRE supplies three skills learned in
simulation on a Franka arm as guidance to a different coding agent, which
drives a real bimanual YAM station. For pushing and pulling a drawer, the unguided run
spends 334.9 million tokens without a working program, while the guided run
uses 81.7 million and succeeds in 11 of 20. For lifting
a can, tokens fall from 61.94 to 6.58 million and success rises from 13 to 19
of 20, and for placing a bowl, success stays at 20 of 20 with fewer tokens
\citep{arxiv260700272}. The cost falls on every task, while the success gain
varies. RoboBridge initializes real-world evolution from
simulation and, on one task of placing a cake in a pot and covering it, reaches
15 of 15 in 12 iterations with 13.08 million tokens, against 16 iterations and
52.39 million tokens from scratch
\citep{arxiv261002717}. Both reduce the search needed on a new robot, but
neither reuses a stored program directly. ASPIRE supplies repair guidance, and
RoboBridge supplies procedural documents for further evolution.

The RECENT system does reuse the code, under a specific premise: while the strategy of a skill transfers,
its binding to the physical body must be redone. It stores code skills validated on a
reference Franka with metadata about their preconditions, effects, and API
profile. For a new robot, a small model rewrites only the code spans
that bind the skill to the embodiment. In its ablation, regenerating whole skills
instead drops success from 83.33\% to 26.67\% \citep{arxiv260607999}. World
In-Context Learning transfers a text memory. From one demonstration per task,
a frozen model rehearses in simulation and writes a Task Memory of the
workflow, object roles, and learned parameters, which a linter keeps free of
coordinates and ground-truth terms. On RoboTwin~2.0 with GPT-6.1-Sol it
reaches 83.6\% on clean scenes and 83.2\% on randomized ones, where the
strongest trained policy it compares falls from 81.5\% to 41.8\%. With an
adapter that remaps coordinate conventions, arm kinematics, and gripper
parameters, and with no further rehearsal, the memories written on a
simulated ALOHA succeed in 10, 7, 7, and 6 of 10 episodes on four tasks on a
real dual-arm PiPER \citep{arxiv261112885}.

For a new model, Skill2Real's executable skills, learned with one proposer
model, can be used by another \citep{arxiv261002788}, and SEAR measures such
reuse. On RoboLab, a given model scores 46.9\% with no library, 42.4\% with its own, and
60.6\% with a library written by another model \citep{arxiv261018792}.
EmbodiedEvo reports the same asymmetry. A stronger model's experience raises a
weaker executor's success to 12.03\%, 8.86 points above no experience and well
above the 2.50\% it reaches with its own experience \citep{arxiv261018867}.

For a new task, the parts of SEAR's libraries that carry over are general
routines for control, grasping, and parsing rather than solutions written for
one task \citep{arxiv261018792}. In EmbodiedEvo, experience from related or
mismatched tasks adds 14.72 and 16.67 points without any exploration of the
target task \citep{arxiv261018867}, and EmbodiedSWE finds that a solution from
a similar task helps an agent more and sooner than one from a dissimilar task,
whose benefit fades by the end of the search \citep{arxiv260927308}.

The Trinity moves a harness across task families and then from simulation to
hardware. With a frozen Claude Opus~5, it groups 44 benchmarks, from rule
discovery and puzzles to multi-agent games and long-horizon environments, by
the capabilities they share. At each group it evolves reusable harness modules
in simplified worlds that it generates and verifies. Two of the benchmarks are
robot tasks, which it assigns to a group and serves with the harness evolved
there. A Unitree G1 walks through an office to the desk beside a fan and
reaches it, and an AlohaMini2 uncovers inverted cups to report a hidden cube.
Each harness first plays five episodes in a simulator rebuilt from the real
site, and these episodes are summarized into frozen skills. On ten real-robot
episodes per task, success rises from 0\% with the base harness to 60\% with
the harness of the fourth iteration on both tasks. The recorded runs also
differ in interfaces, guidance, and acceptance rules, so the comparison covers
whole configurations of harness and skills \citep{arxiv261036077}.

\subsection{Experience Distillation into Model Weights}
\label{sec:code-teacher}

Experience can also enter a model's weights instead of a library. The routes
differ in which model learns and in which check certifies its training data.
The first student is an action policy trained on the executions of generated
code. RobotGPT raised average success from 38.5\% for ChatGPT-generated code to
91.5\% for a policy trained on its trajectories \citep{arxiv231201421}.
EmbodiedSWE expands each verified agent solution into many trajectories, and a
$\pi_{0.5}$ policy fine-tuned on 500 such simulated demonstrations grasps a lamp
shade in all ten real trials and unscrews the bulb in two, while the pretrained
policy completes none of the stages \citep{arxiv260927308}. HarnessPAI
fine-tunes $\pi_{0.5}$ on harness trajectories and raises its LIBERO-PRO
success by 38.8 percentage points \citep{arxiv260929166}. For this student, a
program too slow or brittle to deploy can still teach, provided its successful
executions are identified reliably. A policy can also be refined from its own
failures, with no program as teacher. Compounding Physical RSI traces each
failed rollout to the task stage where progress stops, refines the policy by
offline reinforcement learning on experience selected for that stage, and
promotes a weight-space interpolation between the previous and the refined
policy only after evaluating the candidates on the full benchmark. Its final
policy reaches a three-seed mean success of 20.70\% on the 42 tasks of
RoboDojo-Sim, against 14.88\% for the next best entry it lists, where one of
the three seeds also served to select candidates \citep{arxiv261106654}.

The second student is the model that writes or operates the program. LMPC
fine-tunes a code-writing model on human--robot teaching conversations and
uses it to generate and revise reward code for a model-predictive controller.
On embodiments and APIs unseen in training, the better of its two variants
raises success by 31.5 points over the base model (18.6 for the other)
\citep{arxiv240211450}. Robo-Instruct fine-tunes CodeLlama-Python 7B on synthetic instruction--program pairs and rejects programs that violate the constraints of a world it simulates on the fly from each program's API calls. Its pass@1 on the RoboEval benchmark rises from 40.00\% for the base model to 68.75\%, against 55.00\% with Self-Instruct data and 62.50\% for GPT-4o-mini, though below the 83.75\% of GPT-4 \citep{arxiv240520179}. RE-0 certifies each training example before distillation. A teacher model proposes a code patch at the student's earliest harmful step, and the patch enters the training data only if paired reruns from the same simulator snapshot give a positive lower confidence bound on its gain. On robosuite restack, plain fine-tuning of a Qwen3.8-27B student on three such patches reaches 46 of 50 held-out seeds, whereas the same fine-tuning on eight success-filtered teacher demonstrations leaves it at 3, each in a single run \citep{arxiv260932416}. Octopus trains a 7B vision-language model to write executable code from egocentric images, using code that GPT-4 wrote in the OctoGibson simulator and a reward model trained on environment feedback. Reinforcement learning from this feedback raises task completion on 60 simulated tasks from 0.15 after supervised fine-tuning to 0.18, still below the 0.43 of GPT-4 given the scene as text \citep{arxiv231008588}.

Other work trains the model that operates a harness rather than one that
writes standalone programs. World Action Agent fine-tunes Qwen3.5-9B as the main agent of its tool-calling
harness, with the sub-agents unchanged, raising success on unseen seeds of its training tasks from 1.7\% to
43.3\% \citep{arxiv260929964}. AgenticRoboRL post-trains Qwen3.5-9B as a planner, first on a 27B teacher's successful trajectories and then by multi-turn reinforcement learning rewarded by the simulator's success check. The VLA and other tools stay unchanged. On LIBERO-PRO initial states held out from training, success rises from 18.1\% to 71.0\% in a tool-calling harness that invokes a VLA. On three suites in CaP-X's code-writing harness with privileged state, it rises from 2.7\% to 28.5\%, above its teacher's 20.3\% \citep{arxiv261028589}. CaPtain applies the same two stages to a model that writes code against low-level perception, geometry, and guarded-motion primitives, with no VLA. Its reinforcement-learning stage credits each request through a critic that reads privileged simulator state and is used only during training. Starting from Qwen3.5-4B, success on held-out LIBERO initial states is 55.4\% after supervised fine-tuning and 73.8\% after reinforcement learning, and on three LIBERO-PRO suites 59.5\% and 78.5\% \citep{arxiv261040799}. TinyAstral distills the trajectories that its teacher produces in the evolved harness (\cref{sec:admission}) into a 4B student, which replaces the teacher in the same harness and reaches 62.8\% on RoboTwin against the teacher's 63.6\% \citep{arxiv261041216}. In both kinds of training, what transfers is a way of writing,
correcting, or operating programs, not a stored skill.

Together with program-generated demonstrations (\cref{sec:data-programs}),
these routes suggest a feedback path: programs generate training data, training
improves models, and those models support later program execution or
generation. So far, only parts of that path have been demonstrated.

Distillation also raises the question of where knowledge should reside. HyMeS explores
this choice for memory and motor skill, keeping memory in code and comparing
the result with VLAs whose memory is learned \citep{arxiv260809410}. ICAL compares retrieval with fine-tuning for one model and one set of learned examples, which are Python action programs annotated with language abstractions. On TEACh's unseen validation split with ground-truth perception, GPT-3.5 reaches 35.1\% task success when it retrieves the examples, 23.2\% when fine-tuned on them, and 35.8\% with both \citep{arxiv240614596}. SHAPER evolves a skill and a
context-building harness around a frozen planner and beats supervised
fine-tuning of the same actor on the same 15 training episodes, with 34.50\% against
24.00\% success on 800 held-out VLABench episodes. It also selects candidates on 24
validation episodes, however, and the fine-tuning compute is not reported
\citep{arxiv260811350}.

SEAR comes closest to a matched comparison. On LIBERO, LIBERO-PRO, and RoboLab,
an agent trains a VLA on rollouts from its code teachers within
the development time allocated to a code-writing agent, and on robosuite it
receives ten hours against the code agent's seven. The student policy is within two points of
the code baseline on LIBERO and LIBERO-PRO, but falls well short on robosuite (61.0\% against
97.1\%) and RoboLab (47.2\% against 68.0\%), where the
slow simulator yields fewer demonstrations. The two forms
also break differently. When the scene camera moves, code falls to the 9\%
noted in \cref{sec:library-harm}, while the VLA keeps 69\% \citep{arxiv261018792}. A comparison at matched data and
compute that also includes skill documents remains open.

\begin{pciinsight}[pcicolor=SecLearn]{Takeaways and open problems: learning from the physical world}
\textbf{Takeaways.}
\begin{itemize}[leftmargin=1.1em,itemsep=2pt,topsep=1pt,parsep=0pt]
\item \textbf{A larger library can lower success on some tasks.} ASPIRE rises from 0.047 with an empty library to 0.305 with a full one on LIBERO-PRO Long, yet one task falls from 0.68 to 0.26 as the library grows from 50 to 90 source tasks.
\item \textbf{Acceptance tests shape what is reused.} Removing validation costs 13 to 31 percentage points in reported ablations, yet in one 123-round run the agent accepted 57 changes while its target task never succeeded.
\item \textbf{Reuse lowers the cost of adaptation more reliably than it raises success.} ASPIRE's transferred skills cut tokens on all three real tasks, with success gains from none to 11 of 20. RECENT's ablation favors rewriting only a skill's binding to the body.
\end{itemize}
\textbf{Open problems.}
\begin{itemize}[leftmargin=1.1em,itemsep=2pt,topsep=1pt,parsep=0pt,beginpenalty=10000]
\item \textbf{Measuring staleness.} Re-test a stored library after a controlled change of dynamics, camera, robot, or API, and report the fraction that still works.
\item \textbf{Retirement policies.} Compare append-only libraries with ones that revise, merge, and retire entries, at matched size and cost.
\item \textbf{Code or weights.} Store the same knowledge as a program, a skill document, and fine-tuned weights, with equal data and compute, and compare.
\end{itemize}
\end{pciinsight}

If experience can accumulate and also decay, a single success rate cannot say
which happened. \Cref{sec:measuring} turns to what should be measured.

%% file: figures/library-lifecycle.tex
\begin{figure}[!tbp]
\centering\def\pcifigworkcolor{SecLearn!80!black}
\begin{tikzpicture}[x=1cm,y=1cm,
  bt/.style={lh text,anchor=north west}]
\newcommand{\kpanel}[8]{%
  \lhpanel{#1}{0}{2.99}{5.75}{SecLearn}{#2}{#3}{#4}
  \node[art img] at (#1+1.5,4.08) {\includegraphics[width=2.65cm,height=1.65cm,keepaspectratio]{#5}};
  \node[bt] at (#1+0.15,3.1) {#6\\[4pt]#7};
  \node[lh eg] at (#1+0.2,0.3) {\scalebox{0.92}{e.g.\ #8}};}
\kpanel{0.02}{1}{Build}{What is kept?}{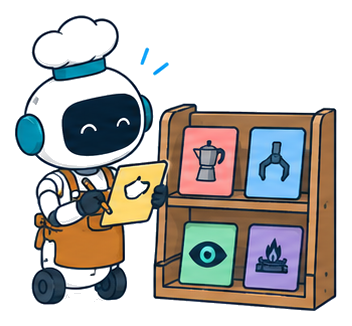}
  {\lhb{SecLearn}{save}{2.05cm}{programs that pass a check are stored}}
  {\lhb{SecLearn}{search}{2.05cm}{reused when a new task comes up}}
  {\figwork{arxiv230516291}{Voyager}, \figwork{arxiv240618746}{LRLL}}
\kpanel{3.31}{2}{Accept}{What does it prove?}{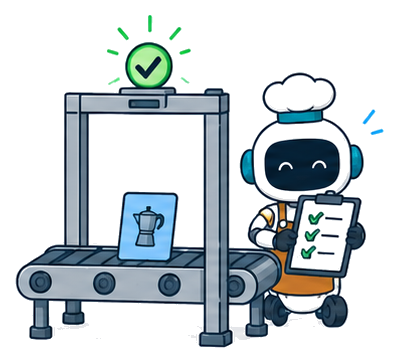}
  {\lhb{SecLearn}{filter}{2.05cm}{six mechanisms decide what is accepted}}
  {\lhb{SecLearn}{chart-line}{2.05cm}{removing validation costs 13--31 points}}
  {\figwork{arxiv261002717}{RoboBridge}, \figwork{arxiv261002788}{Skill2Real}}
\kpanel{6.6}{3}{Maintain}{Does more help?}{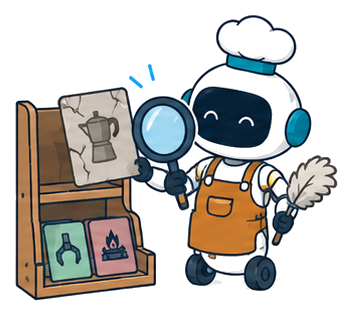}
  {\lhb{SecLearn}{layer-group}{2.05cm}{growth can make some tasks regress}}
  {\lhb{SecLearn}{chart-line}{2.05cm}{\figwork{arxiv261002717}{RoboBridge}: never revising entries, 83\% \textrightarrow{} 44\%}}
  {\figwork{arxiv260700272}{ASPIRE}, \figwork{arxiv260304560}{MEMO}}
\kpanel{9.89}{4}{Transfer}{How far does it travel?}{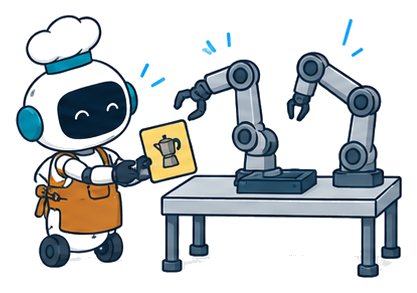}
  {\lhb{SecLearn}{exchange-alt}{2.05cm}{simulated skills guide a real bimanual robot}}
  {\lhb{SecLearn}{chart-line}{2.05cm}{drawer: 0 \textrightarrow{} 11 of 20 within a quarter of the tokens}}
  {\figwork{arxiv260700272}{ASPIRE}, \figwork{arxiv260607999}{RECENT}}
\kpanel{13.18}{5}{Distill}{Into the weights?}{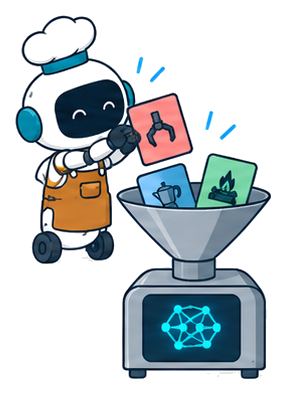}
  {\lhb{SecLearn}{brain}{2.05cm}{executions or teaching sessions train a policy or a code model}}
  {\lhb{SecLearn}{chart-line}{2.05cm}{\figwork{arxiv231201421}{RobotGPT}: 38.5\% \textrightarrow{} 91.5\%}}
  {\figwork{arxiv231201421}{RobotGPT}, \figwork{arxiv240211450}{LMPC}}
\foreach \x in {3.165,6.455,9.745,13.035} \node[art fat=SecLearn] at (\x,3.95) {};
\end{tikzpicture}
\caption{\textbf{What happens to a program that passes its checks.} It is stored in a library \textcircled{\scriptsize 1}, often after a further acceptance test \textcircled{\scriptsize 2} (\cref{sec:libraries}). The library then has to be maintained \textcircled{\scriptsize 3}, and its skills may carry over to new robots, models, or tasks \textcircled{\scriptsize 4} (\cref{sec:library-harm}). Verified executions can also train a model's weights \textcircled{\scriptsize 5} (\cref{sec:code-teacher}).}
\label{fig:library-lifecycle}
\end{figure}

%% file: figures/library-growth.tex
\begin{figure}[t]
\centering
\begin{tikzpicture}
\begin{axis}[
  width=0.6\linewidth, height=5.2cm,
  xmin=-4, xmax=94, ymin=0, ymax=0.75,
  xtick={0,25,50,90}, ytick={0,0.1,0.2,0.3,0.4,0.5,0.6,0.7},
  xlabel={LIBERO-90 source tasks seeding the library}, ylabel={Success rate},
  xlabel style={font=\sffamily\footnotesize}, ylabel style={font=\sffamily\footnotesize},
  ticklabel style={font=\sffamily\scriptsize,/pgf/number format/assume math mode=true},
  axis lines*=left, ymajorgrids, grid style={RuleGray},
  legend style={font=\sffamily\scriptsize,draw=none,fill=none,at={(1.03,0.5)},anchor=west},
  legend cell align=left,
]
\addplot[SecLearn,very thick,mark=*,mark size=1.6pt] coordinates {(0,0.047) (25,0.137) (50,0.215) (90,0.305)};
\addplot[SecLearn!60!black,thick,mark=square*,mark size=1.4pt] coordinates {(0,0.094) (25,0.218) (50,0.292) (90,0.383)};
\addplot[SecLearn!68!white,thick,mark=triangle*,mark size=1.8pt] coordinates {(0,0.00) (25,0.056) (50,0.138) (90,0.226)};
\addplot[SecLearn!60!black,thick,dashed,mark=diamond*,mark size=2pt] coordinates {(0,0.00) (25,0.02) (50,0.68) (90,0.26)};
\addplot[Baseline,densely dotted,thick,domain=-4:94,samples=2] {0.05};
\legend{Overall,Task perturbation,Position perturbation,Stove and moka pot (task),{$\pi_{0.5}$, overall}}
\end{axis}
\end{tikzpicture}
\caption{\textbf{What a growing skill library buys, and where it hurts.} Zero-shot success of ASPIRE with Opus~4.6 on LIBERO-PRO Long with skill libraries seeded from increasing numbers of LIBERO-90 source tasks, redrawn from Tables 8 and 9 of \citet{arxiv260700272}. The dotted line is $\pi_{0.5}$ on the same benchmark. Average success rises with the library, while the task of turning on the stove and placing the moka pot, under task perturbation, falls from 0.68 to 0.26 between 50 and 90 source tasks.}
\label{fig:library-growth}
\end{figure}
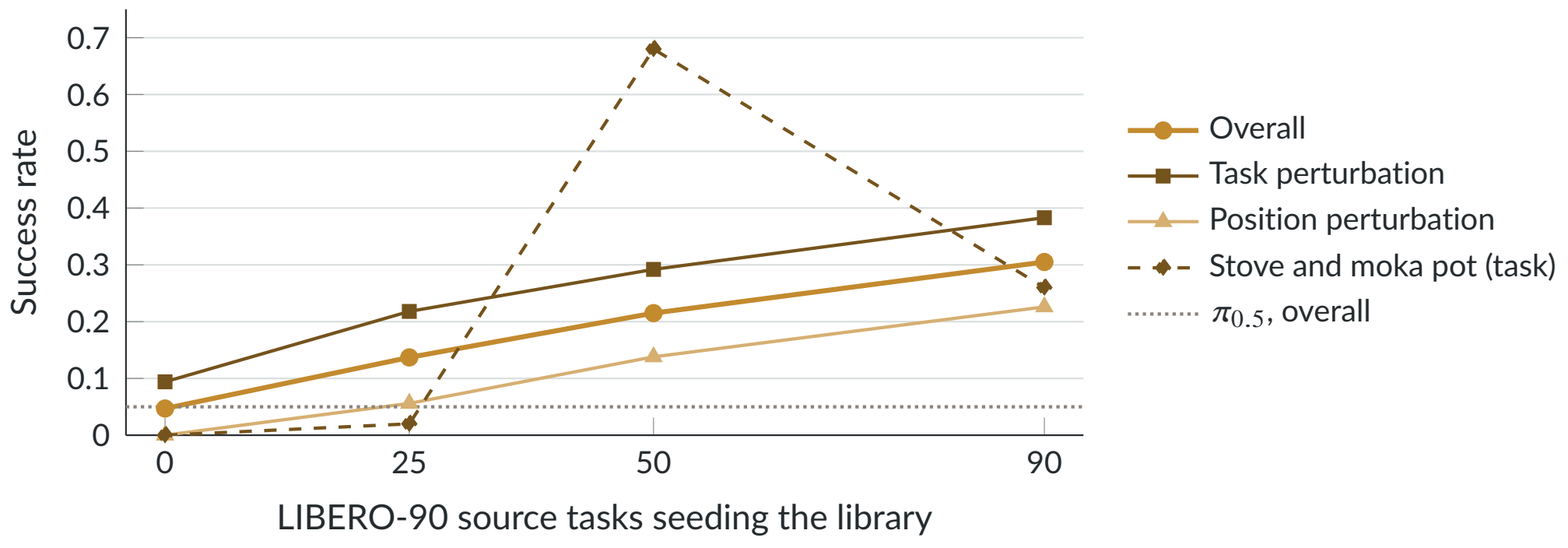

%% file: sections/07-measuring-progress.tex
\section{Benchmarks and Evaluation Practice}
\label{sec:measuring}

Imagine that the agent of our running example reports 0.72 success on a
benchmark that contains the moka-pot task. To interpret the number, a reader
needs to know much more than the benchmark's name \citep[cf.][]{chen-etal-2025-adammeme,arxiv261029677}. Which judge scored the
episodes, and could the agent read it? How many seeds did the agent see before
evaluation, and how many model calls, dollars, and minutes did it spend? Was the
result produced in simulation, and how well does that simulation predict the
real kitchen? Did the skill library that produced the result also break tasks
that worked before? Answering these questions is the second half of \pciRQ{3}.
This section reviews the benchmarks and protocols used for these agents,
benchmarks for repairing robot code, and what current results leave out.

\subsection{Benchmarks and Protocols}
\label{sec:benchmarks}

\Cref{tab:benchmarks} compares benchmarks and protocols that evaluate
programs in physical dynamics, or that code-writing agents commonly report on,
along the properties that earlier sections showed to matter.

\input{tables/benchmarks}

Three benchmarks each address a different reporting gap. CaP-Bench reports
results by interface tier, CodeActionBench hides its judge while reporting cost
and scoring the agent's own claims of completion, and RoboEval reports a
failure taxonomy. CaP-Bench, part of CaP-X, varies the level of the
primitive API from high-level functions with privileged state to low-level
functions described only by their signatures and docstrings. It also varies the feedback channel
from none to text, images, and visual differences \citep{arxiv260322435}.
These are the harness variables that \cref{sec:api,sec:traces}
identified as consequential. CodeActionBench removes privileged state and specialist perception
modules, judges outcomes with a verifier that is never returned to the agent,
reports cost and median wall time per attempt, and scores each agent's own
claim of completion as correct, overclaimed, underclaimed, or absent. The
strongest configuration ends 68 of 75 attempts with a claim that agrees with
the verifier, including correct reports of failure
\citep{arxiv260933807}. RoboEval, whose temporal-logic checks over program
traces appear in \cref{sec:traces}, reports a failure taxonomy that separates
Python errors, infeasible robot actions, and task completion errors
\citep{arxiv231111183}. No benchmark among them combines all three practices.

Two benchmarks from October 2026 add controls and breadth. SEAR adds
matched controls that separate what an agent retains from what it gains by
trying more often, seeing familiar tasks, reading privileged information,
spending a larger budget, or being lucky. It grades every policy with a sealed
judge and reports the spread over three runs \citep{arxiv261018792}.
RobotWorld spans 84 tasks across manipulation, mobile manipulation,
locomotion, driving, and aerial control, with executable success checks and
explicit interaction budgets. The best of five evaluated models completes 16
of them, and its runs over all 84 tasks cost about 9,913 US dollars at
estimated list prices. No evaluated model completes 63 of the tasks \citep{arxiv261010409}.
Leaderboards can also verify what they list. RoboDojo counts an entry only
after official online evaluation, three seeds in simulation, results on all
three real embodiments, a hidden-layout consistency check, and released
checkpoints, code, and videos \citep{arxiv260704434}.

Embodied First Exam, from November 2026, pairs settings that differ in what
the agent can access. It runs unchanged coding agents on 1,156 tasks in 35
suites with 47 simulated robots, and the tests of a task share its initial
scene, success rule, and time limit. In the Privileged setting the agent reads
the full simulator state and resets without limit. In the Standard setting it
sees only the robot's sensors and by default cannot reset, and in the Realtime
setting the simulation keeps running while the agent reasons. On the 391 core
tasks with both a Privileged and a Standard test, GPT-6 Astra in Codex solves
77\% of the Privileged tests and 30\% of the Standard ones.
Its controlled runs, discussed in \cref{sec:software-to-physical}, separate
the effects of state access and resets. The harness matters as well: on the 456 core Standard tests,
the same model solves 34\% in Codex and 22\% in Claude Code \citep{arxiv261112177}.

Three more benchmarks from November 2026 test agents in other roles.
RoboARCade has the agent act as the policy itself on 22 long-horizon tasks
across three embodiments in MuJoCo, with the success predicates kept outside
the agent's sandbox and no feedback before the episode ends. GPT-6 Astra
completes 49.8\% of the rubric subtasks, and agents end an episode or move on
before the criteria are met on 24\% to 39\% of the subtasks they attempt. A
richer visual interface adds 1.5 to 6.8 points at 1.7 to 2.3 times the cost
\citep{arxiv261108209}. VABAR targets commercial and industrial settings with
known variation in object shape and pose, in six task classes at two
difficulty levels, among them gear meshing, syringe packing, and cable
routing. Each condition has 30 generated instances. Claude Code with no
robotics harness or skill library scores 0\% in nine of the twelve conditions
and at most 16.7\%. A harness in which the same model composes a computation
graph over a skill library reaches 10.0\% to 70.0\%, below graphs specified by
people on the same simulator in every condition \citep{arxiv261106881}.
Flex-RSI benchmarks the self-improvement loop itself on 24 tasks, seven of
them in its action area. Every run starts from
the same general-purpose harness, an executor's solution is kept only if it
scores higher on held-out validation data, an improver then revises the
harness, and the final solution is scored once on hidden test data. In seven
pairs of runs that differ only in whether the improver runs, the test score
is higher with it in six, by 2.0 to 40.0 points, each from a single reported
run. A repeat of one run with the improver on a robot task scored 40.0,
against 80.0 in the reported run \citep{arxiv261108946}.

Outside robotics, benchmarks for embedded firmware likewise check the device rather than the program's printed output. EmbedBench runs generated code for Arduino, ESP32, and Raspberry Pi Pico boards in the Wokwi circuit simulator and monitors the states of virtual components, since serial output can look plausible while the hardware misbehaves \citep{arxiv250611003}. IoT-SkillsBench flashes firmware to three physical boards and separates code that fails to compile or flash from code that runs but fails a behavioral test. With no added skills, 15 of 42 Zephyr tasks run but fail the behavioral test in a single attempt, against 3 compile failures \citep{arxiv260319583}.

Some benchmarks also award partial credit. The BEHAVIOR challenge
scores the fraction of goal conditions satisfied, as the report of its winning
entry describes \citep{arxiv251206951}. Benchmarks for language-model planners in household simulators do the same with symbolic goals. LoTa-Bench reports the fraction of subgoals completed on its WAH-NL tasks in VirtualHome \citep{arxiv240208178}, and Embodied Agent Interface credits the fraction of final-state goal propositions that a generated action sequence satisfies \citep{arxiv241007166}.
VLABench reports a progress score \citep{arxiv241218194}, and RoboArena collects
progress scores from human evaluators \citep{arxiv250618123}. N-SCORE shows
that progress separates policies with fewer trials than binary success
\citep{arxiv260313616}. For programs, traces can record intermediate effects, but code
steps are a poor common unit: the same behavior may be written as one function
call, several calls, or a loop. A more comparable score uses task-defined
subgoals that are shared across implementations. CodeActionBench already
uses subgoal checkpoints for diagnosis \citep{arxiv260933807}. In the
moka-pot task, establishing a grasp would count as progress, but merely closing
the gripper would not. The scoring rule should state whether an achieved
subgoal must remain satisfied and should count repeated achievements only once.
Partial credit already shows what success hides: in EmbodiedEvo, retained
experience raises the score on tube insertion from 4 to 36 while success on
that task in the development set stays at zero \citep{arxiv261018867}. Partial credit can also reorder models. On EB-ALFRED in EmbodiedBench, Gemini-1.5-Pro has a higher average subgoal success rate than Claude-3.5-Sonnet (67.4\% against 65.3\%) but a lower task success rate (62.3\% against 64.0\%) \citep{arxiv250209560}. Whether such
scores distinguish code-writing agents with fewer trials remains an empirical question.

\subsection{Benchmarks for Robot Code Repair}
\label{sec:missing}

An agent may need to repair an existing robot repository while preserving
behavior that already works. Evaluating that repair requires the combination
of repository tests and physical evaluation motivated in
\cref{sec:software-to-physical}. SWE-bench pairs real repository issues with tests that a fix must pass \citep{arxiv231006770}. The robot benchmarks reviewed here cover parts of that setup. ROSDevEval evaluates generated code for 240 real-world ROS~2 programming tasks by building each package and running its test cases, without testing the resulting behavior in physics \citep{zhao2026rosdeveval}. EmbodiedSWE-Bench runs standalone task-solving programs in a physics simulator \citep{arxiv260927308}, while CaP-Bench evaluates policies built over robot primitives \citep{arxiv260322435}. RHO goes further toward repository-level work by allowing multi-file changes to a robot-policy repository and evaluating them in simulation \citep{arxiv260616458}. Its experiments optimize repository-based policies, and the changes it evaluates come from its own search, not from real issue histories. ROBUST does start from real issue histories. It packages 221 bugs from seven ROS systems, mined from issue trackers, pull requests, and commits, as Docker images that rebuild each defective version, but only 15 of the 156 fixed bugs that fail during execution came with a test \citep{arxiv240403629}. ROSTrans asks models to migrate ROS1 code to ROS~2 and grades each of its 108 tasks with a runtime test suite that launches the migrated nodes and that a candidate can neither supply nor weaken. Of the samples from six models, 56.5\% build and then fail these tests, against 26.1\% that fail to build. The best model passes 10.9\% of tasks in one attempt, and up to three rounds of execution-guided repair bring Claude Opus~4.6 to 49.1\% and GPT-5.5 to 47.2\% \citep{arxiv261041168}. Robo-SWE mines 236 issue-resolution tasks from merged pull requests in 28 robotics repositories, among them ROS~2 systems, simulators and physics engines, and robot-learning frameworks. A task counts as resolved when every test that the developers' fix made pass now passes and no test that passed before fails. Over three runs per setup, its five primary agent setups resolve 45.8\% to 56.5\% of the tasks, and its authors note that these containerized tests do not guarantee robot behavior \citep{arxiv261116116}. None of these benchmarks or systems combines issue-derived changes to an existing robot repository with checks of both the requested physical effect and the preservation of earlier behavior.

RoboBugBench, released in October 2026, supplies a first instance. It collects 33 real, execution-validated defects in a pinned version of Nav2, the ROS~2 navigation stack, mined from upstream history, issue reports, and code audit. A patch is graded by compiling it and driving the robot in the Gazebo simulator, against two kinds of criteria: those the fix must meet, and pass-to-pass criteria for behavior that must still work after the fix. The best of three models passes the behavioral oracle on 12 of the defects in one attempt. Such pass rates need to be read alongside the audit of the oracle itself. On 11 instances, the shipped acceptance rule passes all 11 counterfeit patches that hide a failure without fixing it. A distributional rule, which tests at $\alpha=0.01$ whether the patched runs' telemetry differs from the reference runs', rejects 4 of them and none of five held-out correct fixes. The paper therefore reports its repair rates as bounds \citep{arxiv261016369}.

The RoboBugBench audit shows that the oracle is limited by what the telemetry records. Extending the design beyond Nav2 to other ROS~2 repositories would keep hidden checks on the physical outcomes and record the judge, the budget, how often each patch was run, and what was tested on hardware (\cref{sec:unreported}). Paired evaluation of the original and patched versions under matched conditions would distinguish the requested improvement from unrelated variation.

\subsection{Limitations of Current Evaluation}
\label{sec:unreported}\label{sec:evidence-card}

Among the benchmarks and protocols we reviewed, five
properties are rarely or inconsistently reported: cost, variance, judge visibility, judge error,
and the link between simulation and reality.

\paragraph{Cost} Success rates are widely reported, but the resources spent
to reach them are reported much less consistently, so it is often unclear whether a
higher rate comes from a better method or a larger budget.
Benchmarks that report cost include CodeActionBench (dollar cost and median
wall time) \citep{arxiv260933807}, RoboTwin~2.0 (refinement iterations, the
length of the generated code in tokens, and the token cost of each VLM
observation) \citep{arxiv250618088}, and PLCWorld (median model calls, tokens,
and time per task on an 18-task subset) \citep{arxiv261002982}. Embodied First Exam
prices each attempt from the tokens its harness logged at list prices, and a
Standard attempt of GPT-6 Astra costs 27.1 US dollars on average
\citep{arxiv261112177}. Evaluating six frontier models at three levels of
reasoning effort on 1,800 LIBERO-10 episodes took about 99,000 API calls and
4,936 US dollars \citep{arxiv261030051}. CaP-Bench reports time per interface tier in an
appendix \citep{arxiv260322435}. Some method papers also make cost a first-class
quantity. FAEA reports attempts, tokens, dollars, and wall-clock time per task,
for example 7.4 attempts and 4.08 US dollars per task on LIBERO Goal
\citep{arxiv260120334}, and PyRUA-Lean compares a code-cell agent with a
tool-calling agent under an equal budget of model calls and reports estimated
cost per episode on the instances both agents solve \citep{arxiv261001939}.
ASPIRE plots held-out success against estimated API cost per
task for twelve backbone models. On the ten Goal tasks under LIBERO-PRO's
position perturbation, Claude Sonnet~5.5 reaches 98.8\% at an estimated 12 US
dollars per task, against 99.2\% at 18 dollars for Opus~5.5 and 91\% at 60
dollars for Opus~5 \citep{arxiv260700272}. Real-robot time is reported even more
rarely. The exceptions include SimEX, with roughly ten minutes of physical
interaction per task \citep{arxiv260938982}, and AutoEval, with about 850
autonomous episodes in 24 hours and three human interventions \citep{arxiv250324278}. For embedded firmware flashed to physical boards, IoT-SkillsBench reports that its full evaluation took about 100 hours of human-in-the-loop validation \citep{arxiv260319583}.
A reported budget may count only the run that produced
the result and leave out rejected candidates and abandoned runs. Human
effort may also be reported as one number, although resets, labels, corrections,
and safety monitoring are different kinds of work. A system that automates one
of them can still need the others. In ASPIRE's Rubik's Cube experiment on a real
robot, for example, an operator still gives the initial task guidance,
scrambles the cube, and resets the scene \citep{arxiv260700272}.

\paragraph{Variance} A difference between two results means little until it
is compared with the spread between runs of the same system, and in studies of
code-writing agents even the unit of repetition varies. It can mean generating a new program, executing the same program again, changing
the scene, or rebuilding the skill library from the start. Each exposes a
different source of uncertainty. CodeActionBench uses three attempts per task
with one fixed scene seed \citep{arxiv260933807}, PyRUA-Lean runs one episode
per agent and instance \citep{arxiv261001939}, Embodied First Exam makes one
attempt per agent and test \citep{arxiv261112177}, and Skill2Real reports point
estimates for fixed libraries, without variability across independent
training runs \citep{arxiv261002788}. PLCWorld reports three independent generations
per task for its direct models, with bootstrap intervals that resample tasks
while keeping their generated programs and execution conditions fixed. These
intervals cover variation across tasks, not across reruns of the whole
generation-and-verification workflow \citep{arxiv261002982}. A report should
therefore state what was repeated or resampled, along with the number of
rollouts and any interval. A paired comparison also assumes that both systems
met the same test states. In a vectorized LIBERO evaluation in LeRobot~0.6.1,
with emulated controllers whose episodes end at different times, the reset
schedule assigned slots of the same episode wave the initial states 20, 5, 22,
and 7 where 4, 5, 6, and 7 were expected. The study proposes checking the
initial state each episode actually received, and its provenance, before any
paired statistics are computed \citep{arxiv261100896}.

Rigorous methods exist in the literature on evaluating learned policies. A study of evaluation practice recommends reporting the number of trials per condition rather than only percentages, together with interval estimates and failure modes, and interleaving the rollouts of compared policies blind to the evaluator. In one of its examples, rerunning a policy on the same initial conditions after the lab was reorganized gave 0 successes in 8 trials, against 13 of 20 before \citep{arxiv240909491}. TRI's
study of large behavior models uses blinded, randomized real-world trials,
Bayesian posteriors, sequential paired comparisons of two policies under
matched conditions, and Bonferroni corrections within each task
\citep{arxiv250705331}. STEP provides a sequential test that stops
when the evidence suffices \citep{arxiv250310966}, and N-SCORE brings sequential
testing to partial-credit metrics, using 598 trials in total on a simulated
suite where a fixed design uses 2,000 \citep{arxiv260313616}. AnyRank extends
sequential testing to ranking several policies at once and needs up to 50\%
fewer hardware evaluations than statistically rigorous baselines. On
RoboArena it shows, with 90\% confidence, that two Bradley--Terry ranking
models, the classical one and the Davidson variant the leaderboard uses, order
a pair of policies wrongly in a four-policy subset \citep{arxiv261014603}.

Among the benchmarks for code-writing agents we reviewed, RoboBugBench and
Embodied First Exam compare models with paired exact McNemar tests, and
PLCWorld reports paired bootstrap intervals
\citep{arxiv261016369,arxiv261112177,arxiv261002982}. None of them corrects
these comparisons for multiple testing, and none uses sequential or Bayesian
testing. Informal statistics mislead in predictable ways. In an
agentic loop for improving robot policies, a Monte-Carlo study shows that stopping a
naive McNemar test (a paired test on two policies' episode outcomes) as
soon as it passes raises the rate of promoting changes with no real gain to
0.161, about three times its nominal level. In the same study, choosing the
best of 27 noisy 100-episode evaluations overstates the champion by 9.8
points, and when one evaluation cell was re-run in simulation with the same
seeds, 22 of 100 episodes changed outcome although the aggregate was unchanged
\citep{arxiv260807555}.
In SEAR, two DeepSeek runs with the same setup differed by 20.9 points
\citep{arxiv261018792}.

\paragraph{Judge visibility} Reports do not always say whether the agent
could consult the judge during a run, which leaves open whether a success
needed that feedback. FAEA's agent calls the success function
and conditions its next attempt on the result \citep{arxiv260120334}, whereas
CodeActionBench never returns the judge's decision to the agent \citep{arxiv260933807}.
In OpenRUA's comparison table, some agents are marked as having access to the
benchmark's success predicate during execution \citep{arxiv261002459}. Visibility can change what is learned as well as what is scored, since a library learned while reading the criteria code loses its lead when the criteria are rewritten to follow the instructions (\cref{sec:judges}). Withholding the success signal tests execution without that
feedback. When the agent's own claim is scored, as in CodeActionBench, it also
tests completion judgment. Generalization needs a separate test on tasks,
scenes, or initial conditions held out from development. CodeActionBench,
for example, explicitly evaluates fixed task instances and leaves
cross-scene generalization untested \citep{arxiv260933807}. In RLE-Bench,
agents building VLA training recipes in simulation may query a development
evaluation service, and their final development score predicts the official
score within 5 points when the evaluation tasks can be rehearsed. On robustness
tracks, whose perturbations are never shown to the agent, the official score is
19 (LIBERO) and 41 (RoboTwin) points lower than the development score
\citep{arxiv260934210}. Visibility and the
learning--evaluation split should therefore be reported separately. OpenRUA's
table marks oracle access explicitly, and EmbodiedEvo states its split:
experience comes from one scene per
task, where resets are allowed, and fresh agents use it frozen on unseen scenes
with resets disabled \citep{arxiv261018867}. RAX states both. During
exploration its agent may ask the benchmark's evaluator whether a task is
complete. Its fully automated programs then run alone on scenes that
exploration never uses, and their results are never revealed to the agent. On
most BEHAVIOR-1K tasks, however, fresh copies of the agent are scored, and on
three of them each copy reads reviews of the program's runs on the other test
instances \citep{arxiv261022496}.

A report should also state what may change during
testing. CaP-Agent0 adapts within each trial
and starts the next one afresh, apart from a fixed skill library distilled
beforehand from earlier rollouts \citep{arxiv260322435}. RAX and EmbodiedEvo
also freeze what they learned before testing. A system that keeps
updating across test episodes measures its improvement during operation, which
is a different quantity. Finally, the report should say whether test results feed
back into development, since scores that guide later updates turn the test set
into development data even when its episodes stay hidden.

\paragraph{Judge error} How often the checker itself is wrong is rarely
reported, so the gap between a reported success rate and actual task success
stays unknown. Of the twelve entries in \cref{tab:benchmarks}, three report
how often their judge errs against a reference, and PLCWorld checks that its
evaluator rejects targeted counterexample programs. Embodied First Exam
requires every task to fail when the robot does nothing and to pass with its
reference solution where one exists. Before adding a suite, it also instructs
an agent to pass the tasks without solving them and holds the suite back until
its verifier is fixed if the agent succeeds \citep{arxiv261112177}. RoboTwin~2.0 tests its VLM observer, which
guides repair while the simulator scores success, on 130 labeled execution sequences, where it reaches an accuracy of 0.431 and judges
61 of the 101 failed executions to be successful \citep{arxiv250618088}. AutoEval deploys its VLM success detector only after it
reaches over 95\% accuracy on held-out images and teleoperated trajectories \citep{arxiv250324278}, and TRI
re-checked about 27\% of its real rollouts and found a 2.31\% success-rate
discrepancy \citep{arxiv250705331}. Outside the table, HALTER uses an LLM
to score long-horizon rollouts on a Franka arm from scene-graph histories and
to plan the resets. Its scores agree with manual annotation on 90.0\% of 100 episodes,
against 76.0\% for an AutoEval-style judge given the same LLM, and it needs 25
operator interventions where manual resetting needs 100 \citep{arxiv260919413}. SafeActBench's deterministic evaluator judges whether an action was supported
by sufficient evidence rather than whether a task succeeded. A human audit of a
balanced sample of its positive and negative judgments finds few false
acceptances and reports the rate with a confidence interval
\citep{arxiv261007753}. The audit of \cref{sec:judges} shows
what can happen when no such check is made \citep{arxiv260937771}.

\paragraph{Simulation--reality link} Most results for code-writing agents
come from simulation alone, with no paired real trials to show how well they
predict performance on hardware. Real-to-sim evaluation, in which a simulation is reconstructed from the
real setup and policies are ranked in it, can correlate well with real
results. SIMPLER reports strong correlation for manipulation policies
\citep{arxiv240505941}, PolaRiS a much stronger correlation than existing
simulated benchmarks \citep{arxiv251216881}, and RoboWorld, a learned video
simulator paired with a VLM judge, reports a Pearson correlation of 0.989
between its evaluation scores and the real-world leaderboard across eight
policies \citep{arxiv260701060}. RobotArena~$\infty$ turns real robot demonstration videos into simulated counterparts and evaluates six VLA policies there. Its VLM ranking matches the ranking from 8,749 human pairwise comparisons exactly, but its comparison with real execution covers one task and three policies \citep{arxiv251023571}. Agreement also depends on the evaluation setup.
For one bimanual robot cell, a carefully authored reconstruction cuts mean score
error over ten task--policy combinations from 17.54 to 6.97 points relative to
a default open-source recipe. The authored reconstruction's correlation is also higher (0.90 against
0.51), though ten combinations are too few to tell the two correlations apart \citep{arxiv261000731}. WorldArena finds that visual quality only weakly to moderately tracks
embodied usefulness, and that world models differ as policy evaluators, with CtrlWorld tracking simulator results closely and Cosmos-Predict 2.5 less closely \citep{arxiv260208971}.
A Veo-based world model captures the relative out-of-distribution performance of
Gemini Robotics policies, with a Pearson correlation of 0.86 across
generalization axes for the strongest policy, although it underestimates
absolute success. Red-teaming inside it also surfaced unsafe behaviors that
were then reproduced on the real robot \citep{arxiv251210675}.

SureSim combines many simulated rollouts with a few paired
real trials to give confidence intervals on real performance, and shows that
the gain depends on how well the paired simulated outcomes predict the real
ones. In its low-correlation setting, adding simulated rollouts does not
improve precision over real-only evaluation \citep{arxiv251004354}. A report should therefore state whether any paired real trials exist, and how many.

Two quantities would make the failures described in \cref{sec:evidence,sec:compounding}
measurable, and neither is routinely reported. The first is the
\emph{proportion of accepted episodes that are actually unsuccessful}: the
fraction of episodes a judge accepts that an independent reference check
rejects, estimated from a random sample of accepted episodes with the sample
size, the reference check, and an uncertainty interval stated. It equals one
minus the judge's precision and differs from the standard false-positive rate,
which divides by all unsuccessful episodes. It matters most when an agent searches for programs the
judge accepts
(\cref{sec:judges}). Unlike accuracy on a fixed dataset, it is estimated on
episodes that the agent produced and the judge accepted. Rates with different
denominators can move in opposite directions. In the scripted-controller study of \cref{sec:judges}, accepted
hover failures per execution fell as the controller improved, while the judge
accepted a larger share of the hovers that remained \citep{arxiv261024801}. No Free Checker asks
that evaluations of verifiers report false positives and false negatives
separately and test for false positives under search \citep{arxiv260909250}, and Code as
Agent Harness lists a rate of false acceptance among measures of verification
strength for software harnesses \citep{arxiv260518747}.

The second quantity is the
\emph{retention rate} of a skill library: the fraction of previously
successful task--seed pairs that remain successful after a library update,
compared on the same seeds with the model, primitive API, and budget held
fixed, and with repeated paired evaluations for stochastic agents. Without it,
gains on new tasks can hide losses on earlier ones
(\cref{fig:library-growth}). Retention is the physical counterpart of
rerunning earlier tests after a code change. It concerns changes to the
library, whereas staleness concerns changes to the world
(\cref{sec:library-harm}). A drop in retention can have causes that call for
different repairs. An entry may have been edited badly, retrieval may surface
the wrong entry, as with MEMO's unclustered memory \citep{arxiv260304560}, or
a changed interface may no longer call the entry correctly. Retention should also be
read alongside gains on new tasks, since a loop that has stopped accepting
changes keeps it trivially high. The Weights or Skills? review proposes related
measures for experience, including a curve of success against interaction, a
reuse rate, and a provenance check \citep{arxiv260801851}. A survey of self-evolving agents, which covers no robot work, notes that most agent benchmarks reset agent state between tasks and so cannot measure knowledge accumulation or degradation \citep{arxiv250721046}.

Return to the 0.72 of the running example. Interpreting it requires knowing
how the episodes were judged, under which conditions, and at what cost.
Calls, cost, and seed allocation can usually be recovered from logs.
Auditing accepted episodes, rerunning earlier tasks, and obtaining paired
real trials require additional evaluation. Comparing two results directly
also requires matched interfaces, evaluation conditions, and budgets.

\begin{pciinsight}[pcicolor=SecLearn]{Takeaways and open problems: benchmarks and evaluation practice}
\textbf{Takeaways.}
\begin{itemize}[leftmargin=1.1em,itemsep=2pt,topsep=1pt,parsep=0pt]
\item \textbf{Cost, variance, and judge error are often unreported.} Most benchmarks we reviewed omit at least one of them or the judge's visibility to the agent, and few link simulated results to the real setting.
\item \textbf{Statistical methods from robot evaluation are rarely used.} Sequential tests, Bayesian posteriors, and intervals that combine simulated and real trials have seen little use in benchmarks for code-writing agents.
\item \textbf{Failures among accepted episodes and retention are not routinely reported.} They separate passing the judge from solving the task, and new gains from lost earlier successes.
\end{itemize}
\textbf{Open problems.}
\begin{itemize}[leftmargin=1.1em,itemsep=2pt,topsep=1pt,parsep=0pt,beginpenalty=10000]
\item \textbf{A shared LIBERO-PRO protocol.} Fixed seeds for learning, selection, and evaluation, a stated policy on privileged functions, budgets in tokens, dollars, and time, and all perturbation splits.
\item \textbf{Partial credit from traces.} Score implementation-independent subgoals from traces and test whether they separate code-writing agents with fewer trials.
\item \textbf{Repair from observed failures at scale.} RoboBugBench covers 33 defects in one navigation stack. More repositories, better-instrumented oracles, and real-robot confirmation would show how far its repair rates hold.
\item \textbf{Retention as a metric.} Re-run earlier task--seed pairs after each library update under matched conditions and report retention with uncertainty.
\end{itemize}
\end{pciinsight}

%% file: tables/benchmarks.tex
\begin{table}[!tbp]
\centering\footnotesize\setlength{\tabcolsep}{3.2pt}
\caption{\textbf{What benchmarks and protocols for physical programs report.} \tabyes{} reported, \tabno{} not provided in the reported evaluation, \tabnf{} not specified in the reviewed metric sections. Judge access: \faEye{} visible or queryable, \faEyeSlash{} hidden. HW marks quantitative hardware results, including downstream policies trained from generated data.}
\label{tab:benchmarks}
\begin{tabularx}{\linewidth}{@{}L{.19\linewidth}L{.125\linewidth}C{.055\linewidth}L{.135\linewidth}L{.12\linewidth}L{.115\linewidth}L{.1\linewidth}>{\Centering\arraybackslash}X@{}}
\toprule
\TableHeader{Benchmark} & \TableHeader{Judge} & \TableHeader{Judge access} & \TableHeader{Judge error} & \TableHeader{Cost} & \TableHeader{Uncertainty} & \TableHeader{Progress} & \TableHeader{\faIcon{robot}\,HW}\\
\midrule
CaP-Bench (CaP-X)\newline\tablecite{arxiv260322435} & Simulator success (its S1 API exposes privileged state) & \tabnf & \tabnf & Time per tier (appendix) & \tabnf & \tabyes{} average dense reward & \tabyes{} Franka, 25 trials per task\\[2pt]
CodeActionBench\newline\tablecite{arxiv260933807} & Hidden verifier of physical outcomes & \hid & \tabnf & API-equivalent dollars, median wall time & Three attempts per fixed scene & \tabyes{} checkpoint coverage (diagnostic) & \tabno\\[2pt]
RoboEval\newline\tablecite{arxiv231111183} & Temporal logic over symbolic traces & \tabnf & \tabnf & \tabnf & Ranges over paraphrases & \tabno & \tabno\\[2pt]
RoboTwin~2.0 code generation\newline\tablecite{arxiv250618088} & Simulator success over ten runs per program, and a VLM observer for repair feedback & \vis{} as feedback & \tabyes{} observer only, accuracy 0.43 on 130 labeled runs & Iterations, code length, VLM tokens & \tabnf & \tabnf & \tabyes{} trained policies\\[2pt]
VLABench\newline\tablecite{arxiv241218194} & Simulator progress, skill and parameter recall & \tabnf & \tabnf & \tabnf & \tabnf & \tabyes{} progress score & \tabno\\[2pt]
ResponsibleRobotBench\newline\tablecite{arxiv251204308} & Simulator success and safety violations & \tabnf & \tabnf & Weighted execution cost, no tokens or dollars & Wilson 95\% intervals (watering tasks) & \tabno & \tabno\\[2pt]
LIBERO-PRO\newline\tablecite{arxiv251003827} & Task-definition predicates & \tabnf & \tabnf & \tabnf & \tabnf & \tabno & \tabno\\[2pt]
BEHAVIOR Challenge\newline\tablecite{arxiv251206951} & Goal-condition predicates & \tabnf & \tabnf & Efficiency logged but excluded from ranking & \tabnf & \tabyes{} fraction of goal conditions & \tabno\\[2pt]
PLCWorld\newline\tablecite{arxiv261002982} & Closed-loop plant simulation & \hid & Counterexamples test violation detection & Calls, tokens, time (subset) & Three generations (direct models), task-cluster intervals & \tabno & \tabno\\[2pt]
Embodied First Exam\newline\tablecite{arxiv261112177} & Simulator success rules & \vis{} most suites & Idle-robot and reference checks & List-price dollars & One attempt per test & \tabyes{} stage labels & Demo only\\[2pt]
AutoEval (protocol)\newline\tablecite{arxiv250324278} & VLM success detector & \tabnf & \tabyes{} over 95\% accuracy before deployment & Throughput, interventions, human time & 95\% intervals (consistency study) & \tabno & \tabyes\\[2pt]
TRI LBM (protocol)\newline\tablecite{arxiv250705331} & Human rubric (real), predicates (simulation) & \tabnf & \tabyes{} 2.31\% success-rate discrepancy on 27\% re-checked & \tabnf & Posteriors, sequential tests & \tabyes{} task completion & \tabyes\\
\bottomrule
\end{tabularx}
\par\vspace{3pt}{\scriptsize\raggedright\textit{Coding rules.} A property counts as reported only when the evaluation states it for its own results. Judge error needs an accuracy or error rate of the judge measured against a reference such as human labels. Cost needs resources in stated units (time, calls, tokens, or dollars). Uncertainty needs a spread or interval over repeated runs, and repetition without one is described in words. Progress includes quantitative partial completion, even when used only for diagnosis.\par}
\end{table}

%% file: sections/09-synthesis.tex
\section{Challenges and Future Directions}
\label{sec:frontiers}

The findings reviewed so far suggest four directions, one for each of the
preceding four sections (\cref{fig:directions}), and \cref{tab:open-problems}
pairs selected open problems with first experiments.

\input{figures/directions}
\input{tables/open-problems}

\paragraph{Program representation and policy integration}
\label{sec:design-guidance}
Given the comparisons with end-to-end policies in \cref{sec:code-vs-vla}, the placement of a program should follow from the change it must absorb. Explicit program logic suits changes in task structure, step order, or constraints. Offline development can prepare that logic in advance, leaving runtime model calls for what the stored program does not cover. SEAR turns its results into a rule for retained knowledge: code for faults that the program can observe, and weights both for camera motion (which a program assuming a fixed camera pose cannot detect) and for new instruction wording \citep{arxiv261018792}. Contact-rich behavior is the case these comparisons leave open. There, results depend on the backbone model as well as on the sensing and control access the interface exposes, so both should be stated when programs and policies are compared.

A similar choice arises for what an agent retains. Comparing executable programs, skill documents, and fine-tuned weights at matched data and compute, extending SEAR's comparison of code teachers and VLA students, would show what each preserves, how much it costs to reuse, and how readily it can be revised when the world changes.

\paragraph{Verification reliability and efficiency}
\label{sec:evidence-engineering}
In software, continuous integration runs a fixed suite of tests automatically after every change and blocks changes that fail. A physical counterpart would, after every change to a program or a library, record traces, judge the outcomes, and re-run regression tests, using checks whose error rates and costs are known. Its components exist, but mostly in separate systems.

The first component is the trace: which calls to record, which images to retain, and how to present them to the agent. Only a few studies have isolated the trace's contribution so far. CaP-X's comparison of feedback channels shows that presentation changes the result, so presentation is worth evaluating on its own. The second component is the judge. Its errors determine which proposals the loop accepts, so they should be measured on the programs the search produces. The third is how checks are arranged by cost. Existing systems already screen candidates with schema checks before GPU runs \citep{arxiv250602507} or analytic proxies before training \citep{arxiv260511859}. Others run a fast statistical test alongside a VLM so that the two can detect different failures \citep{arxiv241004640}. SimEX updates a simulator from real rollouts and uses it to screen candidate repairs before the next robot trial \citep{arxiv260938982}. Since a check's value depends on the errors it catches, checks can run side by side as well as in a cheap-to-expensive cascade.

A joint evaluation would hold the generator, tasks, and search budget fixed
while changing the trace representation, the verification procedure, or the
acceptance test for library entries. Running two such procedures on the same
recorded candidates isolates how each uses the evidence, whereas letting each
collect its own candidates under the same budget also captures what it chooses
to try. The update-admission study of \cref{sec:admission} reports both kinds
of comparison. Its common candidate streams compare admission rules without
measuring final success, and in its closed-loop runs an update adopted without
any test ends above the paired test
\citep{arxiv260910873}. An independent audit of sampled accepted
episodes would estimate the proportion of accepted episodes that actually
fail, the quantity defined in \cref{sec:unreported}. Held-out task evaluation and
regression tests would measure what accepted changes achieve and preserve,
and repeating the search would reveal
whether those results depend on one sequence of proposals. The comparison
should report that failure proportion and task success at matched budgets,
with model calls, GPU time, robot time, and human intervention stated
separately.

These experiments require keeping verification authority separate from the
code the agent may modify and stating which checks or results it can see
during development. They also depend on how resets, observation, and execution
are bounded once the program acts on a robot (\cref{sec:breaks}). \Cref{tab:open-problems}
gathers experiments of this kind for all four directions.

\paragraph{Continual learning and skill maintenance}
\label{sec:lifecycle}
If a skill library is maintained like a codebase, as \cref{sec:library-harm} suggests, a useful lifecycle would attach tests, a version, provenance, and retirement criteria to each entry. Comparing this practice with an append-only library at matched data and cost would test whether revision and retirement preserve more useful experience, or merely produce a smaller library.

Retention and staleness would separate a failure caused by a library update from one caused by a change of world. Retention is reported after every update with the world held fixed. Staleness is tested by re-running a fixed library after a controlled change of one factor and before any adaptation. The resulting loss measures how dependent the library is on the old conditions. If adaptation is then allowed, its cost and recovered performance are separate outcomes.

\paragraph{Evaluation reproducibility and real-world validity}
\label{sec:eval-directions}
A shared protocol for a widely used benchmark such as LIBERO-PRO, along the lines listed among the open problems of \cref{sec:measuring}, would close several of the reporting gaps identified there. The statistical tools already developed for robot evaluation, such as sequential tests \citep{arxiv250310966} and sim-plus-real intervals \citep{arxiv251004354}, should become routine in benchmarks for code-writing agents. Paired real trials on a sample of simulated results would show how far a simulated ranking predicts real performance.

%% file: figures/directions.tex
\begin{figure}[!tbp]
\centering
\begin{tikzpicture}[x=1cm,y=1cm,
  bt/.style={lh text,anchor=north west}]
\newcommand{\dpanel}[8]{%
  \def\pcifigworkcolor{\dcol!80!black}%
  \lhpanel{#1}{#2}{7.94}{4.2}{\dcol}{#3}{#4}{#5}
  \node[art img] at (#1+1.55,#2+1.9) {\includegraphics[width=2.8cm,height=2.0cm,keepaspectratio]{#6}};
  \node[bt] at (#1+3.0,#2+3.15) {#7};
  \node[lh eg] at (#1+0.22,#2+0.26) {#8};}
\def\dcol{SecWrite}\dpanel{0.02}{4.45}{1}{Program Representation}{Where should the program sit?}{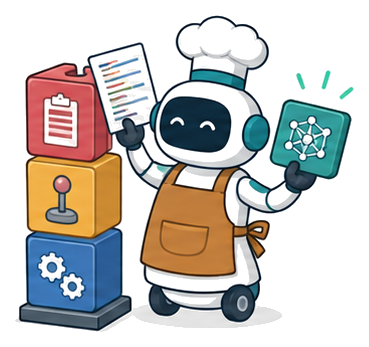}
  {\lhb{\dcol}{code}{4.05cm}{step order, goals, constraints, and memory in code}\\[3pt]%
   \lhb{\dcol}{hand-paper}{4.05cm}{for contact, test a policy, an optimizer, or force and compliance primitives}\\[3pt]%
   \lhb{\dcol}{hourglass-half}{4.05cm}{prepare offline, reason at run time only when needed}}
  {see \cref{sec:action}}
\def\dcol{SecTest}\dpanel{8.21}{4.45}{2}{Reliable Verification}{Can tests run after every change?}{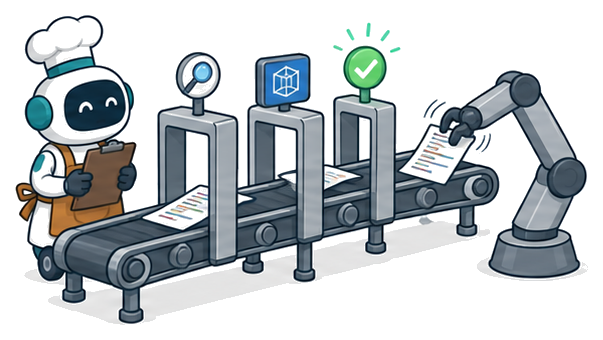}
  {\lhb{\dcol}{stream}{4.05cm}{trace format, judges, and check order are design choices}\\[3pt]%
   \lhb{\dcol}{balance-scale}{4.05cm}{compare failures among accepted episodes and success at matched budgets}\\[3pt]%
   \lhb{\dcol}{clipboard-check}{4.05cm}{report calls, GPU and robot time, and human intervention separately}}
  {e.g.\ \figwork{arxiv260938982}{SimEX}, \figwork{arxiv260322435}{CaP-X}}
\def\dcol{SecLearn}\dpanel{0.02}{0}{3}{Skill Maintenance}{How should experience be kept?}{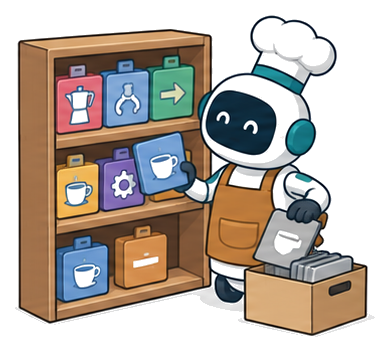}
  {\lhb{\dcol}{tags}{4.05cm}{attach tests, a version, provenance, and retirement criteria}\\[3pt]%
   \lhb{\dcol}{redo}{4.05cm}{measure retention after each library update}\\[3pt]%
   \lhb{\dcol}{clock}{4.05cm}{test staleness by changing one factor at a time}}
  {see \cref{sec:library-harm}}
\def\dcol{SecLearn}\dpanel{8.21}{0}{4}{Reproducible Evaluation}{Do results hold on real robots?}{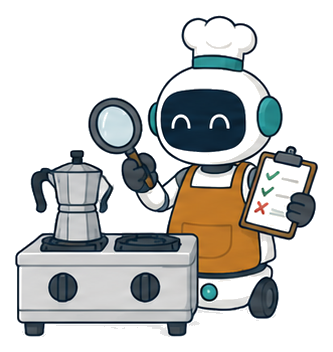}
  {\lhb{\dcol}{dice}{4.05cm}{shared seeds, budgets, and splits for LIBERO-PRO}\\[3pt]%
   \lhb{\dcol}{chart-area}{4.05cm}{sequential tests and sim-plus-real intervals}\\[3pt]%
   \lhb{\dcol}{robot}{4.05cm}{paired real trials for simulated results}}
  {e.g.\ \figwork{arxiv250310966}{STEP}, \figwork{arxiv251004354}{SureSim}}
\end{tikzpicture}
\caption{\textbf{Four challenges and future directions.} How to divide a task between programs and learned policies \textcircled{\scriptsize 1}, how verification can become reliable at a known cost \textcircled{\scriptsize 2}, how skill libraries stay useful as the world changes \textcircled{\scriptsize 3}, and how evaluation can be reproduced and checked against real robots \textcircled{\scriptsize 4}.}
\label{fig:directions}
\end{figure}

%% file: tables/open-problems.tex
\begin{table}[htbp]
\centering\footnotesize\setlength{\tabcolsep}{4pt}
\caption{\textbf{Selected open problems and first experiments.} Experiments are our proposals.}
\label{tab:open-problems}
\begin{tabularx}{\linewidth}{@{}L{.18\linewidth}L{.38\linewidth}Y@{}}
\toprule
\TableHeader{Open problem} & \TableHeader{Findings so far} & \TableHeader{First experiment}\\
\midrule
Physical continuous integration & Existing trace-and-library loops, judge audits, and simulator studies motivate joint evaluation of reliability and cost\newline\tablecite{arxiv260700272,arxiv260903611,arxiv261000731} & Compare verification and acceptance procedures at matched search budgets, then audit accepted episodes and measure held-out success, retention, and cost\\[1pt]
Judges robust to search & Optimizing against a learned reward model can amplify wrong-object failures\newline\tablecite{arxiv261002527} & Audit accepted programs from matched searches with and without verifier feedback, using independent outcome checks\\[1pt]
Code versus VLA on all splits & The code--VLA comparisons in \cref{fig:libero-pro-splits} cover only LIBERO-PRO position and task splits & Run program-based methods on the object, paraphrase, and environment splits with re-run VLA baselines\\[1pt]
Staleness and retention & Library growth gives non-monotonic per-task transfer. SEAR re-runs frozen code policies under single hidden hardware faults, and Repo2Skill-Evo measures staleness for software skills, but no robot study we reviewed re-tests the entries of a stored skill library after a controlled one-factor change\newline\tablecite{arxiv260700272,arxiv261018792,arxiv260821964} & Hold the library fixed and change one world or API factor before adaptation, then measure surviving success separately from retention after library updates\\[1pt]
Interface for orchestration & Most harnesses adopt one interface. ALRM reports model-dependent preferences in action-sequence quality, judged without physical execution. PyRUA-Lean favors code cells at equal call budgets in simulation, with one run per instance\newline\tablecite{arxiv260119510,arxiv261001939} & Compare code and typed calls on the same composition tasks and budgets, and repeat runs to measure uncertainty\\[1pt]
Postconditions for learned skills & VLA skills competent from clean starts stall when chained (pilot traces)\newline\tablecite{arxiv260706256} & Learn or check postconditions per skill, then measure chained success\\[1pt]
Cheap evaluations that hold early & Forked reward comparisons are unreliable before the policy is competent, and longer forks help at higher cost\newline\tablecite{arxiv260428056} & Measure agreement of each proxy with full training at each stage of a reward search\\[1pt]
Code or weights & Programs teach VLAs, and at matched development time the student matches its code teachers where demonstrations are plentiful\newline\tablecite{arxiv260929166,arxiv261018792} & Represent the same knowledge as code, documents, or weights at matched data and compute, then measure reuse and revision cost\\[1pt]
Issue-driven robot-software repair & RoboBugBench grades repairs of 33 real Nav2 defects in Gazebo, and its audit shows the oracle passing counterfeit patches\newline\tablecite{arxiv261016369} & Extend audited behavioral oracles to more ROS~2 repositories, with paired regression scenarios, hidden checks, a record of the judge and budget, and real-robot confirmation\\[1pt]
Judging under injection & Scene text can hijack embodied agents\newline\tablecite{arxiv251000181} & Place adversarial text in scenes judged by a VLM verifier of robot programs\\
\bottomrule
\end{tabularx}
\end{table}

%% file: sections/10-conclusion.tex
\Needspace{10\baselineskip}

\section{Conclusion}
\label{sec:conclusion}

We asked what the physical world supplies in place of a test suite when an agent writes programs whose effects are judged through physical execution. To compare systems, we placed each program by its distance from the actuator and by what judges it. We then followed the loop of writing, testing, and learning that coding agents now run on robots, in simulators, and through the policies their programs train. Across that loop, the value of a program comes from its persistence: it can be run again, inspected when it fails, repaired, and stored. In the comparisons we found, programs lead end-to-end policies when the structure of a task changes, while demonstration-trained policies keep their advantage in distribution in most of them. Results on contact-rich behavior depend on the backbone model as well as on what the interface exposes.

What limits the loop is obtaining reliable evidence of what a program did. Every kind of judge in current use has documented failure modes, learned judges tend to accept failures where the direction of error is reported, and optimizing against a judge can amplify these errors. Some evaluations require a training run or hardware, and where trial is restricted, static and execution-only checks overstate physical correctness. Persistence makes these errors durable. A skill library keeps whatever its acceptance test lets through, so that test decides which gains and which regressions are retained. Current reports rarely state the cost, variance, judge visibility, judge error, or the link between simulation and reality needed to tell gains from regressions. The checks that accept programs are therefore where the field most needs evidence. Neither the proportion of accepted episodes that an independent check rejects nor the retention rate of earlier successes after a library update is routinely reported, and measuring both at matched budgets would show whether a loop that keeps accepting changes is also making progress. If the physical world is to serve as the test suite, its tests must be tested as well.

%% file: sections/appendix-benchmarks.tex
\section{Representative Systems by Program Role and Evaluation Source}
\label{app:systems}

\Cref{tab:systems} places representative systems discussed in the article on
the program roles and evaluation sources of \cref{sec:axes}. ``Role'' places
the program relative to the actuator, and ``Evaluation source'' is what decides
success in the reported results. In the ``HW'' column, \tabyes{} marks that the reported evaluation
includes quantitative results from physical execution of the system's output or
of a policy trained from it, \tabdemo{} that physical execution is shown only
as demonstrations or individual runs, and \tabno{} that no physical execution
is reported. Systems marked $^\ast$ are neighboring systems whose artifact is a
sequence of typed tool calls or a document rather than a program
(\cref{sec:foundations}). They are placed by the role they support.

{\footnotesize\setlength{\tabcolsep}{4pt}
\begin{longtable}{@{}L{.19\linewidth}L{.27\linewidth}C{.10\linewidth}L{.29\linewidth}C{.06\linewidth}@{}}
\caption{\textbf{Representative systems by program role and evaluation source.} Role codes: \roleAct{} action, \roleOrch{} orchestration, \roleObj{} objective, \roleWorld{} world. HW: hardware results, including downstream trained policies, with \tabyes{} quantitative, \tabdemo{} demonstrations or individual runs only, and \tabno{} none reported. $^\ast$\,Neighboring system whose artifact is a sequence of typed tool calls or a document.}\label{tab:systems}\\
\toprule
\TableHeader{System} & \TableHeader{Artifact / interface} & \TableHeader{Role} & \TableHeader{Evaluation source} & \TableHeader{\faIcon{robot}\,HW}\\
\midrule
\endfirsthead
\multicolumn{5}{@{}l}{\sffamily\footnotesize\color{MutedText}Table~\thetable\ (continued)}\\[3pt]
\toprule
\TableHeader{System} & \TableHeader{Artifact / interface} & \TableHeader{Role} & \TableHeader{Evaluation source} & \TableHeader{\faIcon{robot}\,HW}\\
\midrule
\endhead
\bottomrule
\endfoot
\rowcolor{PaperTint}\multicolumn{5}{@{}l}{\sffamily\bfseries Action programs}\\*[2pt]
Code as Policies\newline\tablecite{arxiv220907753} & Python over perception and control functions & \roleAct & Simulator success and real demonstrations & \tabdemo\\[2pt]
ProgPrompt\newline\tablecite{arxiv220911302} & Python-style plans with assertions & \roleAct & Simulator state graph, and per-task real-arm outcomes the authors present as qualitative & \tabdemo\\[2pt]
VoxPoser\newline\tablecite{arxiv230705973} & Code composing 3D value maps for a planner & \roleAct & Simulator and real outcomes & \tabyes\\[2pt]
CodeBotler / RoboEval\newline\tablecite{arxiv231111183} & Programs in a Python-embedded language & \roleAct & Temporal logic over symbolic traces & \tabdemo\\[2pt]
Gemini Robotics-ER\newline\tablecite{arxiv250320020} & Python over a robot API & \roleAct & Task success in simulation and on ALOHA~2 & \tabyes\\[2pt]
CaP-X\newline\tablecite{arxiv260322435} & Python over tiered primitives & \roleAct & Simulator success and real trials & \tabyes\\[2pt]
ASPIRE\newline\tablecite{arxiv260700272} & Python over primitives, with a skill library & \roleAct & Simulator predicates and a real transfer study & \tabyes\\[2pt]
RHO\newline\tablecite{arxiv260616458} & Multi-file policy repositories & \roleAct & Environment reward and benchmark success & \tabno\\[2pt]
OpenRUA\newline\tablecite{arxiv261002459} & ROS~2 programs written from a terminal & \roleAct & Benchmark success & \tabno\\[2pt]
FAEA\newline\tablecite{arxiv260120334} & Episode scripts & \roleAct & Success function visible to the agent & \tabno\\[2pt]
Agent as Policy\newline\tablecite{arxiv260912541} & Python programs run through shell commands & \roleAct & Physical outcome, execution video, and operator review & \tabyes\\[2pt]
VLCP\newline\tablecite{arxiv260816978} & Control function rewritten within the episode & \roleAct & Simulator success & \tabno\\[2pt]
\rowcolor{PaperTint}\multicolumn{5}{@{}l}{\sffamily\bfseries Policy orchestration}\\*[2pt]
SafeHarness\newline\tablecite{arxiv260920822} & Code composing a frozen VLA with obstacle-aware harnesses & \roleOrch & Success and collision avoidance & \tabyes\\[2pt]
HyMeS\newline\tablecite{arxiv260809410} & Memory program steering a frozen VLA & \roleOrch & Benchmark and real success & \tabyes\\[2pt]
Harness VLA$^\ast$\newline\tablecite{arxiv260708448} & JSON calls to primitives and a frozen VLA & \roleOrch & Benchmark success and qualitative real demonstrations & \tabdemo\\[2pt]
Pigey$^\ast$\newline\tablecite{arxiv260721725} & One schema-typed tool call per step & \roleOrch & Benchmark and real success & \tabyes\\[2pt]
HarnessPAI\newline\tablecite{arxiv260929166} & Planner code calling a VLA among primitives & \roleOrch & Benchmark success (a video diagnoser guides repair) & \tabno\\[2pt]
\rowcolor{PaperTint}\multicolumn{5}{@{}l}{\sffamily\bfseries Objectives and worlds}\\*[2pt]
Language to Rewards\newline\tablecite{arxiv230608647} & Reward parameters for online MPC & \roleObj & Simulator success criteria over MPC rollouts & \tabdemo\\[2pt]
Eureka\newline\tablecite{arxiv231012931} & Reward functions & \roleObj & Fitness of a trained policy & \tabno\\[2pt]
Text2Reward\newline\tablecite{arxiv230911489} & Dense reward programs & \roleObj & Trained policies and human feedback & \tabdemo\\[2pt]
DrEureka\newline\tablecite{arxiv240601967} & Rewards and randomization ranges & \roleObj\newline\roleWorld & Real-robot tests of trained policies & \tabyes\\[2pt]
BIRD\newline\tablecite{arxiv261004364} & Reward loops under matched budgets & \roleObj & Trained policies and human ratings & \tabno\\[2pt]
GenSim\newline\tablecite{arxiv231001361} & Simulation task code & \roleWorld & Demonstration generation, LLM critic, people & \tabyes\\[2pt]
RoboGen\newline\tablecite{arxiv231101455} & Tasks, scenes, and reward code & \roleWorld & Human review of generated tasks and of learned-skill videos & \tabno\\[2pt]
Eurekaverse\newline\tablecite{arxiv241101775} & Terrain programs & \roleWorld & Population training, then simulated and real test courses & \tabyes\\[2pt]
RoboTwin 2.0\newline\tablecite{arxiv250618088} & Expert programs for data & \roleWorld & Repeated simulation runs (a VLM observer guides repair) & \tabyes\\[2pt]
AnyTask\newline\tablecite{arxiv251217853} & Reset, success, reward, and policy code & \roleWorld & Model-written checks and VLM judgments monitored for agreement & \tabyes\\[2pt]
Articulate-Anything\newline\tablecite{arxiv241013882} & Python that compiles to URDF & \roleWorld & Ground-truth link and joint thresholds (a render critic guides refinement) & \tabyes\\[2pt]
\rowcolor{PaperTint}\multicolumn{5}{@{}l}{\sffamily\bfseries Skill libraries}\\*[2pt]
Voyager\newline\tablecite{arxiv230516291} & JavaScript skills in Minecraft & \roleAct & Game-state item and milestone counts (GPT-4 critic gates skills) & \tabno\\[2pt]
LRLL\newline\tablecite{arxiv240618746} & Python skill library & \roleAct & Simulator task success (generated success code gates stored experience) & \tabdemo\\[2pt]
Skill2Real\newline\tablecite{arxiv261002788} & Code skills with guidance & \roleAct & Benchmark and real success of frozen skills (verifier during learning) & \tabyes\\[2pt]
RoboBridge$^\ast$\newline\tablecite{arxiv261002717} & Versioned procedural documents & \roleOrch & Official predicate on test seeds (paired gate on validation seeds), operator labels on hardware & \tabyes\\[2pt]
\rowcolor{PaperTint}\multicolumn{5}{@{}l}{\sffamily\bfseries Laboratory and industrial code}\\*[2pt]
Coscientist\newline\tablecite{boiko2023coscientist} & Experiment and instrument code & \roleAct & Laboratory outcome of showcase runs & \tabdemo\\[2pt]
CLAIRify\newline\tablecite{arxiv230314100} & Programs in a chemistry language & \roleAct & Verifier-valid plans and chemist ratings & \tabdemo\\[2pt]
LLM4PLC\newline\tablecite{arxiv240105443} & PLC Structured Text & \roleAct & Compiler pass rate and expert scores & \tabdemo\\[2pt]
SemaPLC\newline\tablecite{arxiv260818565} & PLC code within a project & \roleAct & Model checking (function track) and live-runtime trace comparison (project track) & \tabno\\[2pt]
PLCWorld\newline\tablecite{arxiv261002982} & PLC code & \roleAct & Closed-loop plant simulation & \tabno\\
\end{longtable}}

%% file: main.bbl
\begin{thebibliography}{370}
\providecommand{\natexlab}[1]{#1}
\providecommand{\url}[1]{\texttt{#1}}
\expandafter\ifx\csname urlstyle\endcsname\relax
  \providecommand{\doi}[1]{doi: #1}\else
  \providecommand{\doi}{doi: \begingroup \urlstyle{rm}\Url}\fi

\bibitem[Aggarwal and Welleck(2026)]{arxiv261040565}
Pranjal Aggarwal and Sean Welleck.
\newblock {Gym-Anything for Robotics: Scaling physical simulations with digital
  agents}, 2026.
\newblock URL \url{https://arxiv.org/abs/2610.40565}.

\bibitem[Agia et~al.(2024)Agia, Sinha, Yang, Cao, Antonova, Pavone, and
  Bohg]{arxiv241004640}
Christopher Agia, Rohan Sinha, Jingyun Yang, {Zi-ang} Cao, Rika Antonova, Marco
  Pavone, and Jeannette Bohg.
\newblock {Unpacking Failure Modes of Generative Policies: Runtime Monitoring
  of Consistency and Progress}.
\newblock In \emph{Conference on Robot Learning (CoRL)}, 2024.
\newblock URL \url{https://openreview.net/forum?id=yqLFb0RnDW}.

\bibitem[Ahmedt-Aristizabal et~al.(2026)Ahmedt-Aristizabal, Khamis, Armin, and
  Petersson]{arxiv261013516}
David Ahmedt-Aristizabal, Abdelwahed Khamis, Mohammad~Ali Armin, and Lars
  Petersson.
\newblock {RESCUE: Is This Alarm Worth Acting On? Counterfactual Evaluation of
  VLA Runtime Monitors}, 2026.
\newblock URL \url{https://arxiv.org/abs/2610.13516}.

\bibitem[Ahn et~al.(2025)Ahn, Choi, Lee, Park, and Woo]{arxiv251021302}
Sanghyun Ahn, Wonje Choi, Junyong Lee, Jinwoo Park, and Honguk Woo.
\newblock {Towards Reliable Code-as-Policies: A Neuro-Symbolic Framework for
  Embodied Task Planning}.
\newblock In \emph{Advances in Neural Information Processing Systems 38}, pages
  84000--84031, 2025.
\newblock \doi{10.52202/085713-2533}.

\bibitem[Allen et~al.(2026)Allen, Conway, Antieau, Devaraj, and
  Correll]{arxiv260318520}
Zachary Allen, Max Conway, Lyle Antieau, Allen Devaraj, and Nikolaus Correll.
\newblock {RAPID: Robotic Agentic Platform for Intelligent Disassembly ---
  Platform Design and Unscrewing Evaluation}.
\newblock In \emph{2026 IEEE 22nd International Conference on Automation
  Science and Engineering (CASE)}, pages 966--973, 2026.
\newblock \doi{10.1109/case69030.2026.11704152}.

\bibitem[Aphale and Singh(2026)]{arxiv261106863}
Siddharth Aphale and Ayushman Singh.
\newblock {Rethinking Self-Repair in Code-as-Policy Agents: What Drives Retry
  Success?}, 2026.
\newblock URL \url{https://arxiv.org/abs/2611.06863}.

\bibitem[Atasever et~al.(2026{\natexlab{a}})Atasever, Azbijari, Bakirci,
  Corona, Izdas, Yang, Biyik, and Deshmukh]{arxiv260907111}
Merve Atasever, Keyan Azbijari, Cagan Bakirci, Alfredo~Reina Corona, Tolga
  Izdas, Richard Yang, Erdem Biyik, and Jyotirmoy~V. Deshmukh.
\newblock {From LLM-Generated Specifications to Learned Quadruped Locomotion},
  2026{\natexlab{a}}.
\newblock URL \url{https://arxiv.org/abs/2609.07111}.

\bibitem[Atasever et~al.(2026{\natexlab{b}})Atasever, Azbijari, Bakirci, Huang,
  Izdas, Shahrooei, Yang, Biyik, and Deshmukh]{arxiv260937519}
Merve Atasever, Keyan Azbijari, Cagan Bakirci, Bo-Ruei Huang, Tolga Izdas,
  Zahra Shahrooei, Richard Yang, Erdem Biyik, and Jyotirmoy~V. Deshmukh.
\newblock {Video2STL: Grounding VLM-Generated Temporal Specifications for Robot
  Learning}, 2026{\natexlab{b}}.
\newblock URL \url{https://arxiv.org/abs/2609.37519}.

\bibitem[Atreya et~al.(2025)Atreya, Pertsch, Lee, Kim, Jain, Kuramshin, Neary,
  Hu, Arora, Ellis, Macesanu, Leonard, Cho, Aslan, Dass, Wang, Yuan, Gupta,
  Jayaraman, Berseth, Daniilidis, Martín-Martín, Lee, Liang, Finn, and
  Levine]{arxiv250618123}
Pranav Atreya, Karl Pertsch, Tony Lee, Moo~Jin Kim, Arhan Jain, Artur
  Kuramshin, Cyrus Neary, Edward~S. Hu, Kanav Arora, Kirsty Ellis, Luca
  Macesanu, Matthew Leonard, Meedeum Cho, Ozgur Aslan, Shivin Dass, Tony Wang,
  Xingfang Yuan, Abhishek Gupta, Dinesh Jayaraman, Glen Berseth, Kostas
  Daniilidis, Roberto Martín-Martín, Youngwoon Lee, Percy Liang, Chelsea
  Finn, and Sergey Levine.
\newblock {RoboArena: Distributed Real-World Evaluation of Generalist Robot
  Policies}.
\newblock In \emph{Conference on Robot Learning (CoRL)}, 2025.
\newblock URL \url{https://openreview.net/forum?id=cpmwi3Xwcr}.

\bibitem[Badithela et~al.(2026{\natexlab{a}})Badithela, Snyder, Yin, Sceats,
  Pappas, Matni, Itkina, Nishimura, and Majumdar]{arxiv261014603}
Apurva Badithela, David Snyder, Tenny Yin, Jade Sceats, George Pappas, Nikolai
  Matni, Masha Itkina, Haruki Nishimura, and Anirudha Majumdar.
\newblock {AnyRank: Reliable Anytime Policy Ranking via Sequential Testing}.
\newblock In \emph{Conference on Robot Learning (CoRL)}, 2026{\natexlab{a}}.
\newblock URL \url{https://openreview.net/forum?id=iOU9zkASbg}.

\bibitem[Badithela et~al.(2026{\natexlab{b}})Badithela, Snyder, Zha, Mikhail,
  O'Kelly, Dixit, and Majumdar]{arxiv251004354}
Apurva Badithela, David Snyder, Lihan Zha, Joseph Mikhail, Matthew O'Kelly,
  Anushri Dixit, and Anirudha Majumdar.
\newblock {Reliable and Scalable Robot Policy Evaluation with Imperfect
  Simulators}.
\newblock In \emph{2026 IEEE International Conference on Robotics and
  Automation (ICRA)}, pages 21728--21735, 2026{\natexlab{b}}.
\newblock \doi{10.1109/icra57385.2026.11697329}.

\bibitem[Bai et~al.(2026)Bai, Li, Huang, and Karlsson]{arxiv260830760}
Ziyi Bai, Siqi Li, Tinglei Huang, and Börje~F. Karlsson.
\newblock {PRACTICE: From Experience to Expertise in Self-Evolving Embodied
  Agents}, 2026.
\newblock URL \url{https://arxiv.org/abs/2608.30760}.

\bibitem[Bao et~al.(2026)Bao, Huang, Lin, Feng, Li, Zhao, Li, He, Chi, and
  Zhang]{arxiv260937583}
Shifeng Bao, Fanding Huang, Yihan Lin, Youhe Feng, Guanlin Li, Chen Zhao, Yang
  Li, Jiawei He, Cheng Chi, and Jing Zhang.
\newblock {RoboHarn-Evo: Evolving Hierarchical Physical Knowledge for
  Self-Improving Robotic Manipulation}, 2026.
\newblock URL \url{https://arxiv.org/abs/2609.37583}.

\bibitem[Benaharon et~al.(2026)Benaharon, Zhang, Zheng, Zhang, Wei, and
  Sun]{arxiv261036143}
Leo Benaharon, Chi Zhang, Wei Zheng, Weifan Zhang, Yifan Wei, and Yong Sun.
\newblock {Program-Level Adversarial Co-Adaptive Self-Play for Robotic
  Security}, 2026.
\newblock URL \url{https://arxiv.org/abs/2610.36143}.

\bibitem[Bhamidipaty et~al.(2026)Bhamidipaty, Robson, Petrini, Lyu, and
  Ndousse]{arxiv261004364}
Logan~Mondal Bhamidipaty, Lauren Robson, Linda Petrini, Shengrui Lyu, and Kamal
  Ndousse.
\newblock {A Bird's-Eye View of Iterative Reward Design}, 2026.
\newblock URL \url{https://arxiv.org/abs/2610.04364}.

\bibitem[Boiko et~al.(2023)Boiko, MacKnight, Kline, and
  Gomes]{boiko2023coscientist}
Daniil~A. Boiko, Robert MacKnight, Ben Kline, and Gabe Gomes.
\newblock {Autonomous chemical research with large language models}.
\newblock \emph{Nature}, 624\penalty0 (7992):\penalty0 570--578, 2023.
\newblock \doi{10.1038/s41586-023-06792-0}.

\bibitem[Bran et~al.(2024)Bran, Cox, Schilter, Baldassari, White, and
  Schwaller]{arxiv230405376}
Andres~M Bran, Sam Cox, Oliver Schilter, Carlo Baldassari, Andrew~D White, and
  Philippe Schwaller.
\newblock {Augmenting large language models with chemistry tools}.
\newblock \emph{Nature Machine Intelligence}, 6\penalty0 (5):\penalty0
  525--535, 2024.
\newblock \doi{10.1038/s42256-024-00832-8}.

\bibitem[Burbano et~al.(2026)Burbano, Ortiz, Sun, Yang, Tu, Xie, Cao, and
  Cardenas]{arxiv251000181}
Luis Burbano, Diego Ortiz, Qi~Sun, Siwei Yang, Haoqin Tu, Cihang Xie, Yinzhi
  Cao, and Alvaro~A Cardenas.
\newblock {CHAI: Command Hijacking against embodied AI}.
\newblock In \emph{2026 IEEE Conference on Secure and Trustworthy Machine
  Learning (SaTML)}, pages 893--911, 2026.
\newblock \doi{10.1109/satml68715.2026.00055}.

\bibitem[Burns et~al.(2024)Burns, Jain, Go, Xia, Stark, Schaal, and
  Hausman]{arxiv240406645}
Kaylee Burns, Ajinkya Jain, Keegan Go, Fei Xia, Michael Stark, Stefan Schaal,
  and Karol Hausman.
\newblock {GenCHiP: Generating Robot Policy Code for High-Precision and
  Contact-Rich Manipulation Tasks}.
\newblock In \emph{2024 IEEE/RSJ International Conference on Intelligent Robots
  and Systems (IROS)}, pages 9596--9603, 2024.
\newblock \doi{10.1109/iros58592.2024.10801525}.

\bibitem[Cai et~al.(2026)Cai, Zhu, Shang, Li, Pollefeys, Wu, Vedaldi, and
  Zheng]{arxiv261013947}
Xudong Cai, Ruijie Zhu, Haoliang Shang, Ruining Li, Marc Pollefeys, Shangzhe
  Wu, Andrea Vedaldi, and Chuanxia Zheng.
\newblock {Towards Physically Grounded Articulated 3D Assets: A Multi-Agent
  Pipeline and Benchmark}, 2026.
\newblock URL \url{https://arxiv.org/abs/2610.13947}.

\bibitem[Cao et~al.(2026)Cao, Lin, Gu, Luo, and Ma]{cao-etal-2026-diffcot}
Shidong Cao, Hongzhan Lin, Yuxuan Gu, Ziyang Luo, and Jing Ma.
\newblock {D}iff{C}o{T}: Diffusion-styled chain-of-thought reasoning in {LLM}s.
\newblock In Maria Liakata, Viviane~P. Moreira, Jiajun Zhang, and David
  Jurgens, editors, \emph{Findings of the {A}ssociation for {C}omputational
  {L}inguistics: {ACL} 2026}, pages 38942--38957, San Diego, California, United
  States, July 2026. Association for Computational Linguistics.
\newblock ISBN 979-8-89176-395-1.
\newblock \doi{10.18653/v1/2026.findings-acl.1939}.
\newblock URL \url{https://aclanthology.org/2026.findings-acl.1939/}.

\bibitem[Cao et~al.(2025)Cao, Zhao, Cheng, Shu, Chen, Liu, Liang, Zhao, Yan,
  and Li]{arxiv240400282}
Yuji Cao, Huan Zhao, Yuheng Cheng, Ting Shu, Yue Chen, Guolong Liu, Gaoqi
  Liang, Junhua Zhao, Jinyue Yan, and Yun Li.
\newblock {Survey on Large Language Model-Enhanced Reinforcement Learning:
  Concept, Taxonomy, and Methods}.
\newblock \emph{IEEE Transactions on Neural Networks and Learning Systems},
  36\penalty0 (6):\penalty0 9737--9757, 2025.
\newblock \doi{10.1109/tnnls.2024.3497992}.

\bibitem[{CaP Team} et~al.(2026){CaP Team}, Wei, Qing, Ma, Liu, He, Kong, Gu,
  Liu, Zhu, Wang, Kou, Mo, Lv, Li, Wang, Li, Zhou, Liu, Liu, Jin, Yin, Wang,
  Hou, Huang, Zou, Tao, and Wang]{arxiv261116872}
{CaP Team}, Changxu Wei, Yunpeng Qing, Guozheng Ma, Bowen Liu, Zexin He, Yilun
  Kong, Haoming Gu, Jingyuan Liu, Tianyu Zhu, Jianfei Wang, Junqiang Kou,
  Huinan Mo, Hao Lv, Pengbin Li, Yuesong Wang, Bo~Li, Huan Zhou, Shunyu Liu,
  Chunxiao Liu, Bixuan Jin, Shaopeng Yin, Fei Wang, Zhi Hou, Jiaxing Huang,
  Changqing Zou, Dacheng Tao, and Xiaogang Wang.
\newblock {ACE-CaP-0: Real-World Code-as-Policies in the Era of Experience},
  2026.
\newblock URL \url{https://arxiv.org/abs/2611.16872}.

\bibitem[Cardenas et~al.(2026)Cardenas, Arnett, Yeo, Sah, and
  Kim]{arxiv260326997}
Irvin~Steve Cardenas, Marcus~Anthony Arnett, Natalie~Catherine Yeo, Lucky Sah,
  and Jong-Hoon Kim.
\newblock {ROSClaw: An OpenClaw ROS 2 Framework for Agentic Robot Control and
  Interaction}, 2026.
\newblock URL \url{https://arxiv.org/abs/2603.26997}.

\bibitem[Chen et~al.(2026{\natexlab{a}})Chen, Bagherieh, Balaji, Gonzalez,
  Heravi, Islam, Li, Regli, Ren, {\v{S}}vec, Yoon, M{\"u}ller, and
  Koltun]{arxiv261109892}
Chen Chen, Omid Bagherieh, Naveen Nikamanth~Appiah Balaji, Jesse~T. Gonzalez,
  Negin Heravi, Fahad Islam, Xueting Li, Jean-Baptiste Regli, Kejia Ren, Petr
  {\v{S}}vec, David~J. Yoon, Matthias M{\"u}ller, and Vladlen Koltun.
\newblock {On Agentic Manipulation Policies}, 2026{\natexlab{a}}.
\newblock URL \url{https://arxiv.org/abs/2611.09892}.

\bibitem[Chen et~al.(2026{\natexlab{b}})Chen, Qu, Xia, Jing, Cheng, Ma, Yang,
  Zhou, Li, Suri, Sun, Liu, Chen, Zeng, and Shi]{arxiv260923103}
Guanxiong Chen, Yiduo Qu, Qianjun Xia, Pengyu Jing, Yixian Cheng, Bole Ma,
  Pengzhi Yang, Bingyang Zhou, Ziming Li, Shashwat Suri, Gongbo Sun, Chao Liu,
  Peter~Yichen Chen, Ziqiu Zeng, and Fan Shi.
\newblock {DiagGen: Agentic Generation of Deformable Assets with Sim-based
  Diagnostics for Robotic Simulation}, 2026{\natexlab{b}}.
\newblock URL \url{https://arxiv.org/abs/2609.23103}.

\bibitem[Chen et~al.(2023)Chen, Yu, Ling, and Zhong]{arxiv231214950}
Guojun Chen, Xiaojing Yu, Neiwen Ling, and Lin Zhong.
\newblock {TypeFly: Flying Drones with Large Language Model}, 2023.
\newblock URL \url{https://arxiv.org/abs/2312.14950}.

\bibitem[Chen(2026)]{arxiv261033718}
Josef~Liyanjun Chen.
\newblock {RoboJev: An Open Vision-Language Judge for Robot Success Detection},
  2026.
\newblock URL \url{https://arxiv.org/abs/2610.33718}.

\bibitem[Chen et~al.(2024{\natexlab{a}})Chen, Mu, Yu, Wei, Wu, Yuan, Liang,
  Yang, Zhang, Shao, Qiao, Xu, Ding, and Luo]{arxiv240214623}
Junting Chen, Yao Mu, Qiaojun Yu, Tianming Wei, Silang Wu, Zhecheng Yuan,
  Zhixuan Liang, Chao Yang, Kaipeng Zhang, Wenqi Shao, Yu~Qiao, Huazhe Xu,
  Mingyu Ding, and Ping Luo.
\newblock {RoboScript: Code Generation for Free-Form Manipulation Tasks across
  Real and Simulation}, 2024{\natexlab{a}}.
\newblock URL \url{https://arxiv.org/abs/2402.14623}.

\bibitem[Chen et~al.(2026{\natexlab{c}})Chen, Zhu, Xie, Yoo, Luong, Gualtieri,
  Juette, Ren, Huang, and Goldberg]{arxiv261109030}
Kaiyuan Chen, Xinyue Zhu, Shuangyu Xie, Uksang Yoo, Thi Thu~Phuong Luong,
  Marcus Gualtieri, Christian Juette, Liu Ren, Spencer Huang, and Ken Goldberg.
\newblock {A Controlled Study of Models, Skills, Feedback, and Curriculum in
  Agentic Robotics}, 2026{\natexlab{c}}.
\newblock URL \url{https://arxiv.org/abs/2611.09030}.

\bibitem[Chen et~al.(2024{\natexlab{b}})Chen, Walsman, Memmel, Mo, Fang, Fox,
  and Gupta]{arxiv240511656}
Qiuyu Chen, Aaron Walsman, Marius Memmel, Kaichun Mo, Alex Fang, Dieter Fox,
  and Abhishek Gupta.
\newblock {URDFormer: A Pipeline for Constructing Articulated Simulation
  Environments from Real-World Images}.
\newblock In \emph{Robotics: Science and Systems XX}, 2024{\natexlab{b}}.
\newblock \doi{10.15607/rss.2024.xx.124}.

\bibitem[Chen et~al.(2026{\natexlab{d}})Chen, Zhou, Shi, Li, Feng, and
  Gu]{arxiv260937771}
Qiwei Chen, Kaijun Zhou, Nuohui Shi, Zhiyang Li, Yuxuan Feng, and Jinyu Gu.
\newblock {Faster and Better? Benchmark Bugs and Design Limitations Distort the
  Evaluation of Vision-Language-Action Acceleration}, 2026{\natexlab{d}}.
\newblock URL \url{https://arxiv.org/abs/2609.37771}.

\bibitem[Chen et~al.(2026{\natexlab{e}})Chen, Chen, Li, Tang, Su, Lu, Wan,
  Chen, Liu, Yan, Su, Dou, Wang, Zhang, Liu, Qin, Liang, Wu, Lin, Lin, Wang,
  He, Wu, Wu, Zhou, Lei, Yu, Ji, Jin, Lin, Li, Xiong, Xu, Li, Chai, Xie, Wang,
  Mu, Dong, Matusik, Ding, Ding, Luo, and Tomizuka]{arxiv260704434}
Tianxing Chen, Yue Chen, Zixuan Li, Junyuan Tang, Kailun Su, Haoran Lu, Weijie
  Wan, Baijun Chen, Songling Liu, Haowen Yan, Honghao Su, Zhiyang Dou, Kaixuan
  Wang, Dandan Zhang, Yunze Liu, Yan Qin, Qiwei Liang, Qiwei Wu, Zijian Lin,
  Wenwei Lin, Yuran Wang, Minghua He, Tianshu Wu, Ruihai Wu, Jingquan Zhou,
  Kai-Chong Lei, Haibao Yu, Yuanfeng Ji, Weiyang Jin, Guanyu Lin, Xiaofan Li,
  Qi~Xiong, Renjing Xu, Zhongyu Li, Wenhao Chai, Enze Xie, Ziwei Wang, Yao Mu,
  Hao Dong, Wojciech Matusik, Mingyu Ding, Wenbo Ding, Ping Luo, and Masayoshi
  Tomizuka.
\newblock {RoboDojo: A Unified Sim-and-Real Benchmark for Comprehensive
  Evaluation of Generalist Robot Manipulation Policies}, 2026{\natexlab{e}}.
\newblock URL \url{https://arxiv.org/abs/2607.04434}.

\bibitem[Chen et~al.(2026{\natexlab{f}})Chen, Chen, Chen, Cai, Liu, Li, Liang,
  Lin, Ge, Gu, Deng, Guo, Nian, Xie, Chen, Su, Xu, Liu, Hu, Gao, Wang, Liang,
  Qin, Yang, Luo, and Mu]{arxiv250618088}
Tianxing Chen, Zanxin Chen, Baijun Chen, Zijian Cai, Yibin Liu, Zixuan Li,
  Qiwei Liang, Xianliang Lin, Yiheng Ge, Zhenyu Gu, Weiliang Deng, Yubin Guo,
  Tian Nian, Xuanbing Xie, Qiangyu Chen, Kailun Su, Tianling Xu, Guodong Liu,
  Mengkang Hu, {Huan-ang} Gao, Kaixuan Wang, Zhixuan Liang, Yusen Qin, Xiaokang
  Yang, Ping Luo, and Yao Mu.
\newblock {RoboTwin 2.0: A Scalable Data Generator and Benchmark with Strong
  Domain Randomization for Robust Bimanual Robotic Manipulation}.
\newblock In \emph{International Conference on Machine Learning (ICML)},
  2026{\natexlab{f}}.
\newblock URL \url{https://openreview.net/forum?id=itonej9GIV}.

\bibitem[Chen et~al.(2026{\natexlab{g}})Chen, Liu, Wang, Jin, Zhu, Ruan, and
  Han]{arxiv261024464}
Yangzhi Chen, Feiran Liu, Zishuo Wang, Zijian Jin, Xin Zhu, Lecheng Ruan, and
  Wenjuan Han.
\newblock {Learning to Compose: Recursive Self-Improvement of Shared Executable
  Stacks for Zero-Shot Long-Horizon Robot Manipulation}, 2026{\natexlab{g}}.
\newblock URL \url{https://arxiv.org/abs/2610.24464}.

\bibitem[Chen et~al.(2026{\natexlab{h}})Chen, Bai, Cao, Zeng, Lin, Lin, Liang,
  Ma, Huang, and Shou]{arxiv260910522}
Yanzhe Chen, Zechen Bai, Zhijun Cao, Wenzheng Zeng, Kevin~Qinghong Lin, Yiqi
  Lin, Guoqiang Liang, Kevin~Yuchen Ma, Qiming Huang, and Mike~Zheng Shou.
\newblock {Show-Harness: Just a VLM Agent Can Play Robots}, 2026{\natexlab{h}}.
\newblock URL \url{https://arxiv.org/abs/2609.10522}.

\bibitem[Chen et~al.(2024{\natexlab{c}})Chen, Arkin, Dawson, Zhang, Roy, and
  Fan]{arxiv230606531}
Yongchao Chen, Jacob Arkin, Charles Dawson, Yang Zhang, Nicholas Roy, and
  Chuchu Fan.
\newblock {AutoTAMP: Autoregressive Task and Motion Planning with LLMs as
  Translators and Checkers}.
\newblock In \emph{2024 IEEE International Conference on Robotics and
  Automation (ICRA)}, pages 6695--6702, 2024{\natexlab{c}}.
\newblock \doi{10.1109/icra57147.2024.10611163}.

\bibitem[Chen et~al.(2025{\natexlab{a}})Chen, Hao, Zhang, and
  Fan]{arxiv250301700}
Yongchao Chen, Yilun Hao, Yang Zhang, and Chuchu Fan.
\newblock {Code-as-Symbolic-Planner: Foundation Model-Based Robot Planning via
  Symbolic Code Generation}.
\newblock In \emph{2025 IEEE/RSJ International Conference on Intelligent Robots
  and Systems (IROS)}, pages 19248--19254, 2025{\natexlab{a}}.
\newblock \doi{10.1109/iros60139.2025.11247174}.

\bibitem[Chen et~al.(2025{\natexlab{b}})Chen, Lin, Li, Luo, Deng, and
  Ma]{chen-etal-2025-memearena}
Zixin Chen, Hongzhan Lin, Kaixin Li, Ziyang Luo, Yayue Deng, and Jing Ma.
\newblock {M}eme{A}rena: Automating context-aware unbiased evaluation of
  harmfulness understanding for multimodal large language models.
\newblock In Christos Christodoulopoulos, Tanmoy Chakraborty, Carolyn Rose, and
  Violet Peng, editors, \emph{Proceedings of the 2025 Conference on Empirical
  Methods in Natural Language Processing}, pages 17637--17659, Suzhou, China,
  November 2025{\natexlab{b}}. Association for Computational Linguistics.
\newblock ISBN 979-8-89176-332-6.
\newblock \doi{10.18653/v1/2025.emnlp-main.890}.
\newblock URL \url{https://aclanthology.org/2025.emnlp-main.890/}.

\bibitem[Chen et~al.(2025{\natexlab{c}})Chen, Lin, Li, Luo, Ye, Chen, Huang,
  and Ma]{chen-etal-2025-adammeme}
Zixin Chen, Hongzhan Lin, Kaixin Li, Ziyang Luo, Zhen Ye, Guang Chen, Zhiyong
  Huang, and Jing Ma.
\newblock {A}dam{M}eme: Adaptively probe the reasoning capacity of multimodal
  large language models on harmfulness.
\newblock In Wanxiang Che, Joyce Nabende, Ekaterina Shutova, and Mohammad~Taher
  Pilehvar, editors, \emph{Proceedings of the 63rd Annual Meeting of the
  {A}ssociation for {C}omputational {L}inguistics (Volume 1: Long Papers)},
  pages 4234--4253, Vienna, Austria, July 2025{\natexlab{c}}. Association for
  Computational Linguistics.
\newblock ISBN 979-8-89176-251-0.
\newblock \doi{10.18653/v1/2025.acl-long.213}.
\newblock URL \url{https://aclanthology.org/2025.acl-long.213/}.

\bibitem[Choi et~al.(2024)Choi, Yoon, Ong, Kim, and Jang]{arxiv240208178}
Jae-Woo Choi, Youngwoo Yoon, Hyobin Ong, Jaehong Kim, and Minsu Jang.
\newblock {LoTa-Bench: Benchmarking Language-oriented Task Planners for
  Embodied Agents}.
\newblock In \emph{International Conference on Learning Representations
  (ICLR)}, 2024.
\newblock URL \url{https://openreview.net/forum?id=ADSxCpCu9s}.

\bibitem[Choi et~al.(2026{\natexlab{a}})Choi, Ye, Bui, Shrivastava, Ryu,
  Tirumala, Wulfmeier, and Mehr]{arxiv260930594}
Seoyeon Choi, Shizhao Ye, Nicholas Bui, Aayushi Shrivastava, Kanghyun Ryu,
  Dhruva Tirumala, Markus Wulfmeier, and Negar Mehr.
\newblock {HuGo: LLMs as Whole-Body Policy Code Designers for Humanoid
  Loco-Manipulation}, 2026{\natexlab{a}}.
\newblock URL \url{https://arxiv.org/abs/2609.30594}.

\bibitem[Choi et~al.(2026{\natexlab{b}})Choi, Choi, Chun, Lee, Kim, Lee, and
  Woo]{arxiv260607999}
Sera Choi, Wonje Choi, Saehun Chun, Daehee Lee, Jooyoung Kim, Chaeun Lee, and
  Honguk Woo.
\newblock {Efficient Skill Grounding via Code Refactoring with Small Language
  Models}.
\newblock In \emph{International Conference on Machine Learning (ICML)},
  2026{\natexlab{b}}.
\newblock URL \url{https://openreview.net/forum?id=GQ4zj5YiyC}.

\bibitem[Christie et~al.(2026)Christie, Dai, Bararjanianbahnamiri, Stepputtis,
  and Losey]{arxiv260304560}
Benjamin~A. Christie, Yinlong Dai, Mohammad Bararjanianbahnamiri, Simon
  Stepputtis, and Dylan~P. Losey.
\newblock {From Local Corrections to Generalized Skills: Improving
  Neuro-Symbolic Policies with MEMO}, 2026.
\newblock URL \url{https://arxiv.org/abs/2603.04560}.

\bibitem[Chu et~al.(2026)Chu, Barr, Goues, O'Hearn, Harman, Sarro, and
  Ye]{arxiv261002459}
Zhaoyang Chu, Earl~T. Barr, Claire~Le Goues, Peter O'Hearn, Mark Harman,
  Federica Sarro, and He~Ye.
\newblock {OpenRUA: Robot-Use Agents Are Zero-Shot Visuomotor Policies}, 2026.
\newblock URL \url{https://arxiv.org/abs/2610.02459}.

\bibitem[Curtis et~al.(2024)Curtis, Kumar, Cao, Lozano-Pérez, and
  Kaelbling]{arxiv240605572}
Aidan Curtis, Nishanth Kumar, Jing Cao, Tomás Lozano-Pérez, and Leslie~Pack
  Kaelbling.
\newblock {Trust the PRoC3S: Solving Long-Horizon Robotics Problems with LLMs
  and Constraint Satisfaction}.
\newblock In \emph{Conference on Robot Learning (CoRL)}, 2024.
\newblock URL \url{https://proceedings.mlr.press/v270/curtis25a.html}.

\bibitem[Dantas et~al.(2026{\natexlab{a}})Dantas, Cordeiro, and
  Junior]{arxiv260918994}
Pierre Dantas, Lucas Cordeiro, and Waldir Junior.
\newblock {A Benchmark Suite and Ground-Truth Methodology for Formal
  Verification of IEC 61131-3 Ladder Diagram Programs}, 2026{\natexlab{a}}.
\newblock URL \url{https://arxiv.org/abs/2609.18994}.

\bibitem[Dantas et~al.(2026{\natexlab{b}})Dantas, Cordeiro, Nowroozi, and
  Tihanyi]{arxiv260814590}
Pierre Dantas, Lucas Cordeiro, Ehsan Nowroozi, and Norbert Tihanyi.
\newblock {Toward Safe LLM Agents: A Survey of Specification, Verification, and
  Enforcement}, 2026{\natexlab{b}}.
\newblock URL \url{https://arxiv.org/abs/2608.14590}.

\bibitem[Deng et~al.(2026)Deng, Liu, Zhang, Yan, Cao, and Wang]{arxiv260940306}
Haoyuan Deng, Jiebin Liu, Tengxiao Zhang, Langning Yan, Hongye Cao, and Ziwei
  Wang.
\newblock {DynaHarness: A Dynamic Physical Harness for Self-Evolving Robot
  Agents}, 2026.
\newblock URL \url{https://arxiv.org/abs/2609.40306}.

\bibitem[Ding and Feng(2026)]{arxiv261034644}
Ningpei Ding and Zai~Wen Feng.
\newblock {SlowLab: Benchmarking Autonomous AI Agents on a Year of Experiments
  They Cannot Undo}, 2026.
\newblock URL \url{https://arxiv.org/abs/2610.34644}.

\bibitem[Ding et~al.(2026)Ding, Mi, Huang, Wang, Zhang, Hao, Chen, Li, Zheng,
  Guo, Wang, Li, Wu, Liu, and Cao]{arxiv260816590}
Xin Ding, Liang Mi, Mingzhe Huang, Zixuan Wang, Chao Zhang, Zixu Hao, Fu~Chen,
  Xiangyu Li, Yikai Zheng, Yaoyu Guo, Weijun Wang, Kun Li, Hao Wu, Yunxin Liu,
  and Ting Cao.
\newblock {Zetta $\zeta$: An Efficient Closed-Loop Embodied Harness for
  Self-Evolving Physical Intelligence}, 2026.
\newblock URL \url{https://arxiv.org/abs/2608.16590}.

\bibitem[dos Santos et~al.(2026)dos Santos, Khadraoui, Farhat, Yous, Teffahi,
  and Hacid]{arxiv260119510}
Vitor~Gaboardi dos Santos, Ibrahim Khadraoui, Ibrahim Farhat, Hamza Yous, Samy
  Teffahi, and Hakim Hacid.
\newblock {ALRM: Agentic LLM for Robotic Manipulation}, 2026.
\newblock URL \url{https://arxiv.org/abs/2601.19510}.

\bibitem[Du et~al.(2023)Du, Konyushkova, Denil, Raju, Landon, Hill, de~Freitas,
  and Cabi]{arxiv230307280}
Yuqing Du, Ksenia Konyushkova, Misha Denil, Akhil Raju, Jessica Landon, Felix
  Hill, Nando de~Freitas, and Serkan Cabi.
\newblock {Vision-Language Models as Success Detectors}.
\newblock In \emph{Conference on Lifelong Learning Agents (CoLLAs)}, pages
  120--136, 2023.
\newblock URL \url{https://proceedings.mlr.press/v232/du23b.html}.

\bibitem[Duan et~al.(2026)Duan, Shi, Mao, Zhang, Liang, Piao, Wu, Yao, Huang,
  Huang, Sun, Yan, and Zhang]{arxiv260821964}
Chenyuan Duan, Ge~Shi, Zineng Mao, Ge~Zhang, Hao Liang, Yinzhu Piao, Yuchen Wu,
  Zhixin Yao, Kaiyu Huang, Wenhao Huang, Linzhuang Sun, Shen Yan, and Wentao
  Zhang.
\newblock {Repo2Skill-Evo: Repository Skills Go Stale in Silence}, 2026.
\newblock URL \url{https://arxiv.org/abs/2608.21964}.

\bibitem[Duan et~al.(2025)Duan, Pumacay, Kumar, Wang, Tian, Yuan, Krishna, Fox,
  Mandlekar, and Guo]{arxiv241000371}
Jiafei Duan, Wilbert Pumacay, Nishanth Kumar, Yi~Ru Wang, Shulin Tian, Wentao
  Yuan, Ranjay Krishna, Dieter Fox, Ajay Mandlekar, and Yijie Guo.
\newblock {AHA: A Vision-Language-Model for Detecting and Reasoning Over
  Failures in Robotic Manipulation}.
\newblock In \emph{International Conference on Learning Representations
  (ICLR)}, 2025.
\newblock URL \url{https://openreview.net/forum?id=JVkdSi7Ekg}.

\bibitem[Elmaaroufi et~al.(2024)Elmaaroufi, Shanker, Cismaru,
  Vazquez-Chanlatte, Sangiovanni-Vincentelli, Zaharia, and
  Seshia]{arxiv240503709}
Karim Elmaaroufi, Devan Shanker, Ana Cismaru, Marcell Vazquez-Chanlatte,
  Alberto Sangiovanni-Vincentelli, Matei Zaharia, and Sanjit~A. Seshia.
\newblock {ScenicNL: Generating Probabilistic Scenario Programs from Natural
  Language}.
\newblock In \emph{Conference on Language Modeling (COLM)}, 2024.
\newblock URL \url{https://openreview.net/forum?id=MNLAbfZwh2}.

\bibitem[Elmaaroufi et~al.(2026)Elmaaroufi, Svegliato, Kalade, Schelle, Seshia,
  and Zaharia]{arxiv260616458}
Karim Elmaaroufi, Justin Svegliato, Sarunas Kalade, Graham Schelle, Sanjit~A.
  Seshia, and Matei Zaharia.
\newblock {RHO: Your Coding Agent is Secretly a Roboticist}, 2026.
\newblock URL \url{https://arxiv.org/abs/2606.16458}.

\bibitem[Espinola et~al.(2026)Espinola, Zhang, and Xiang]{arxiv261030051}
May Espinola, Qifan Zhang, and Yu~Xiang.
\newblock {From Reasoning to Action: Benchmarking Frontier AI Models on
  LIBERO-10 Robot Manipulation Tasks}, 2026.
\newblock URL \url{https://arxiv.org/abs/2610.30051}.

\bibitem[Fakih et~al.(2024)Fakih, Dharmaji, Moghaddas, Quiros, Ogundare, and
  Faruque]{arxiv240105443}
Mohamad Fakih, Rahul Dharmaji, Yasamin Moghaddas, Gustavo Quiros, Oluwatosin
  Ogundare, and Mohammad Abdullah~Al Faruque.
\newblock {LLM4PLC: Harnessing Large Language Models for Verifiable Programming
  of PLCs in Industrial Control Systems}.
\newblock In \emph{Proceedings of the 46th International Conference on Software
  Engineering: Software Engineering in Practice}, pages 192--203, 2024.
\newblock \doi{10.1145/3639477.3639743}.

\bibitem[Faldor et~al.(2025)Faldor, Zhang, Cully, and Clune]{arxiv240515568}
Maxence Faldor, Jenny Zhang, Antoine Cully, and Jeff Clune.
\newblock {OMNI-EPIC: Open-endedness via Models of human Notions of
  Interestingness with Environments Programmed in Code}.
\newblock In \emph{International Conference on Learning Representations
  (ICLR)}, 2025.
\newblock URL \url{https://openreview.net/forum?id=Y1XkzMJpPd}.

\bibitem[Fan et~al.(2026{\natexlab{a}})Fan, Xu, Wang, Sokolsky, Lee, and
  Kong]{arxiv260600773}
Jialiang Fan, Weizhe Xu, Zijun Wang, Oleg Sokolsky, Insup Lee, and Fanxin Kong.
\newblock {SafeVLA-Bench: A Benchmark for the Success-Safety Gap in
  Vision-Language-Action Models}, 2026{\natexlab{a}}.
\newblock URL \url{https://arxiv.org/abs/2606.00773}.

\bibitem[Fan et~al.(2026{\natexlab{b}})Fan, He, Jiang, Wu, Wang, and
  Yue]{arxiv261040799}
Yijia Fan, Xilin He, Yilei Jiang, Dongming Wu, Jiaxu Wang, and Xiangyu Yue.
\newblock {CaPtain: A Pipeline for Post-Training Code-as-Policy Robot Agents},
  2026{\natexlab{b}}.
\newblock URL \url{https://arxiv.org/abs/2610.40799}.

\bibitem[Fang et~al.(2026)Fang, Guo, Zhang, Du, Liang, and Bai]{arxiv261112885}
Heng Fang, Zehui Guo, Yuehan Zhang, Mingyang Du, Dingkang Liang, and Xiang Bai.
\newblock {World In-Context Learning for Agentic Robot Manipulation}, 2026.
\newblock URL \url{https://arxiv.org/abs/2611.12885}.

\bibitem[Fang et~al.(2025)Fang, Sun, Ma, Lu, and Yao]{arxiv250600276}
Jiawei Fang, Yuxuan Sun, Chengtian Ma, Qiuyu Lu, and Lining Yao.
\newblock {RoboMoRe: LLM-based Robot Co-design via Joint Optimization of
  Morphology and Reward}, 2025.
\newblock URL \url{https://arxiv.org/abs/2506.00276}.

\bibitem[Fares and Herbold(2026)]{arxiv261032636}
Salim Fares and Steffen Herbold.
\newblock {Can AI Agents Help Robotic Engineers? Lessons from Two Focus Days in
  Industry}, 2026.
\newblock URL \url{https://arxiv.org/abs/2610.32636}.

\bibitem[Fei et~al.(2025)Fei, Wang, Shi, Dai, Cai, Qian, Ji, He, Zhang, Fei,
  Fu, Gong, and Qiu]{arxiv251013626}
Senyu Fei, Siyin Wang, Junhao Shi, Zihao Dai, Jikun Cai, Pengfang Qian, Li~Ji,
  Xinzhe He, Shiduo Zhang, Zhaoye Fei, Jinlan Fu, Jingjing Gong, and Xipeng
  Qiu.
\newblock {LIBERO-Plus: In-depth Robustness Analysis of Vision-Language-Action
  Models}, 2025.
\newblock URL \url{https://arxiv.org/abs/2510.13626}.

\bibitem[Firoozi et~al.(2024)Firoozi, Tucker, Tian, Majumdar, Sun, Liu, Zhu,
  Song, Kapoor, Hausman, Ichter, Driess, Wu, Lu, and Schwager]{arxiv231207843}
Roya Firoozi, Johnathan Tucker, Stephen Tian, Anirudha Majumdar, Jiankai Sun,
  Weiyu Liu, Yuke Zhu, Shuran Song, Ashish Kapoor, Karol Hausman, Brian Ichter,
  Danny Driess, Jiajun Wu, Cewu Lu, and Mac Schwager.
\newblock {Foundation Models in Robotics: Applications, Challenges, and the
  Future}.
\newblock \emph{The International Journal of Robotics Research}, 44\penalty0
  (5):\penalty0 701--739, 2024.
\newblock \doi{10.1177/02783649241281508}.

\bibitem[Fu et~al.(2026)Fu, Yu, El-Refai, Kou, Xue, Huang, Xiao, Fei-Fei, Shi,
  Wu, Sastry, Zhu, Goldberg, and Fan]{arxiv260322435}
Letian Fu, Justin Yu, Karim El-Refai, Ethan Kou, Haoru Xue, Huang Huang, Wenli
  Xiao, Li~Fei-Fei, Guanya Shi, Jiajun Wu, S.~Shankar Sastry, Yuke Zhu, Ken
  Goldberg, and Linxi Fan.
\newblock {CaP-X: A Framework for Benchmarking and Improving Coding Agents for
  Robot Manipulation}.
\newblock In \emph{International Conference on Machine Learning (ICML)}, 2026.
\newblock URL \url{https://openreview.net/forum?id=4JRO9plGAI}.

\bibitem[Galanti et~al.(2026)Galanti, Shah, and Dao]{arxiv260721725}
Liane Galanti, Dhruv Shah, and Tri Dao.
\newblock {Addressing the Orchestration Gap in Generalist Robots via Physical
  Agency}, 2026.
\newblock URL \url{https://arxiv.org/abs/2607.21725}.

\bibitem[Gao et~al.(2026{\natexlab{a}})Gao, Zeng, Zhou, Ge, Shangguan, Wang,
  Wan, Zheng, Zhu, Li, Wu, Wang, He, Jia, Wu, and He]{arxiv261023479}
Hongcheng Gao, Jianshu Zeng, Alex Zhou, Shijia Ge, Xuan Shangguan, Yazhe Wang,
  Yexing Wan, Zelin Zheng, Jay Zhu, Xiping Li, Di~Wu, Yongli Wang, Lingyu He,
  Zhiqi Jia, Sihang Wu, and Xiao He.
\newblock {HexaShell: A Universal Harness for Robot Coding Agents in Simulation
  and the Real World}, 2026{\natexlab{a}}.
\newblock URL \url{https://arxiv.org/abs/2610.23479}.

\bibitem[Gao et~al.(2026{\natexlab{b}})Gao, Zhou, Zheng, Ge, Zhu, Wang, Zeng,
  Shangguan, Wu, He, Jia, Wu, and He]{arxiv260935432}
Hongcheng Gao, Jingjing Zhou, Zelin Zheng, Shijia Ge, Jay Zhu, Yazhe Wang,
  Jianshu Zeng, Xuan Shangguan, Di~Wu, Lingyu He, Zhiqi Jia, Sihang Wu, and
  Xiao He.
\newblock {Self-Evolving Coding Agents: From Digital Programs to Physical-World
  Intelligence}, 2026{\natexlab{b}}.
\newblock URL \url{https://arxiv.org/abs/2609.35432}.

\bibitem[Gao et~al.(2026{\natexlab{c}})Gao, Geng, Hua, Hu, Juan, Liu, Liu, Qiu,
  Qi, Wu, Wang, Xiao, Zhou, Zhang, Zhang, Xiang, Fang, Zhao, Liu, Ren, Qian,
  Wang, Hu, Wang, Wu, Ji, and Wang]{arxiv250721046}
{Huan-ang} Gao, Jiayi Geng, Wenyue Hua, Mengkang Hu, Xinzhe Juan, Hongzhang
  Liu, Shilong Liu, Jiahao Qiu, Xuan Qi, Yiran Wu, Hongru Wang, Han Xiao,
  Yuhang Zhou, Shaokun Zhang, Jiayi Zhang, Jinyu Xiang, Yixiong Fang, Qiwen
  Zhao, Dongrui Liu, Qihan Ren, Cheng Qian, Zhenhailong Wang, Minda Hu,
  Huazheng Wang, Qingyun Wu, Heng Ji, and Mengdi Wang.
\newblock {A Survey of Self-Evolving Agents: What, When, How, and Where to
  Evolve on the Path to Artificial Super Intelligence}.
\newblock \emph{Transactions on Machine Learning Research}, 2026{\natexlab{c}}.
\newblock URL \url{https://openreview.net/forum?id=CTr3bovS5F}.

\bibitem[Gao et~al.(2026{\natexlab{d}})Gao, Miao, Piccinini, Wang, Song, and
  Betz]{arxiv260714387}
Yuan Gao, Wenting Miao, Mattia Piccinini, Haoyu Wang, Qunying Song, and
  Johannes Betz.
\newblock {Chat2Scenic: An Iterative RAG-Based Framework for Scenario
  Generation in Autonomous Driving}, 2026{\natexlab{d}}.
\newblock URL \url{https://arxiv.org/abs/2607.14387}.

\bibitem[Garcia et~al.(2025)Garcia, Chen, and Schmid]{arxiv241001345}
Ricardo Garcia, Shizhe Chen, and Cordelia Schmid.
\newblock {Towards Generalizable Vision-Language Robotic Manipulation: A
  Benchmark and LLM-guided 3D Policy}.
\newblock In \emph{2025 IEEE International Conference on Robotics and
  Automation (ICRA)}, pages 8996--9002, 2025.
\newblock \doi{10.1109/icra55743.2025.11127315}.

\bibitem[Ge et~al.(2026)Ge, Zhou, Zeng, Wan, Wu, Zheng, Wang, Jia, Shangguan,
  Zhu, Liu, He, Wu, He, and Gao]{arxiv260939018}
Shijia Ge, Alex Zhou, Jianshu Zeng, Yexing Wan, Di~Wu, Zelin Zheng, Yazhe Wang,
  Zhiqi Jia, Xuan Shangguan, Jay Zhu, Yijun Liu, Lingyu He, Sihang Wu, Xiao He,
  and Hongcheng Gao.
\newblock {Make Code as Policy Great Again: Frontier Agents Write, Call, and
  Evolve Robot Tools}, 2026.
\newblock URL \url{https://arxiv.org/abs/2609.39018}.

\bibitem[{Gemini Robotics Team}(2025{\natexlab{a}})]{arxiv250320020}
{Gemini Robotics Team}.
\newblock {Gemini Robotics: Bringing AI into the Physical World},
  2025{\natexlab{a}}.
\newblock URL \url{https://arxiv.org/abs/2503.20020}.

\bibitem[{Gemini Robotics Team}(2025{\natexlab{b}})]{arxiv251003342}
{Gemini Robotics Team}.
\newblock {Gemini Robotics 1.5: Pushing the Frontier of Generalist Robots with
  Advanced Embodied Reasoning, Thinking, and Motion Transfer},
  2025{\natexlab{b}}.
\newblock URL \url{https://arxiv.org/abs/2510.03342}.

\bibitem[{Gemini Robotics Team} et~al.(2025){Gemini Robotics Team},
  Choromanski, Devin, Du, Dwibedi, Gao, Jindal, Kipf, Kirmani, Leal, Liu,
  Majumdar, Marmon, Parada, Rubanova, Shah, Sindhwani, Tan, Xia, Xiao, Yang,
  Yu, and Zhou]{arxiv251210675}
{Gemini Robotics Team}, Krzysztof Choromanski, Coline Devin, Yilun Du,
  Debidatta Dwibedi, Ruiqi Gao, Abhishek Jindal, Thomas Kipf, Sean Kirmani,
  Isabel Leal, Fangchen Liu, Anirudha Majumdar, Andrew Marmon, Carolina Parada,
  Yulia Rubanova, Dhruv Shah, Vikas Sindhwani, Jie Tan, Fei Xia, Ted Xiao,
  Sherry Yang, Wenhao Yu, and Allan Zhou.
\newblock {Evaluating Gemini Robotics Policies in a Veo World Simulator}, 2025.
\newblock URL \url{https://arxiv.org/abs/2512.10675}.

\bibitem[Gokhale et~al.(2025)Gokhale, Srivastava, and Bullo]{arxiv250703293}
Anand Gokhale, Vaibhav Srivastava, and Francesco Bullo.
\newblock {LogicGuard: Improving Embodied LLM agents through Temporal Logic
  based Critics}, 2025.
\newblock URL \url{https://arxiv.org/abs/2507.03293}.

\bibitem[Gong et~al.(2026)Gong, Zheng, Hou, Lee, Guo, Tan, Wen, Wang, He,
  Zhang, and Mi]{arxiv261041216}
Kehong Gong, Xiaoxu Zheng, Guanli Hou, Seunghun Lee, Mengqi Guo, Runjia Tan,
  Zhengyu Wen, Qi~Wang, Xiaoyu He, Mingyuan Zhang, and Michael~Bi Mi.
\newblock {TinyAstral: Transferring Self-Evolving Harnesses to Tiny Generalist
  Manipulation Agents}, 2026.
\newblock URL \url{https://arxiv.org/abs/2610.41216}.

\bibitem[Gong et~al.(2025)Gong, Zhang, Shang, Minniti, Patel, Pepe, Yan,
  Gundogdu, Kapelyukh, Abbas, Yan, Patel, Herlant, and
  Schmeckpeper]{arxiv251217853}
Ran Gong, Xiaohan Zhang, Jinghuan Shang, Maria~Vittoria Minniti, Jigarkumar
  Patel, Valerio Pepe, Riedana Yan, Ahmet Gundogdu, Ivan Kapelyukh, Ali Abbas,
  Xiaoqiang Yan, Harsh Patel, Laura Herlant, and Karl Schmeckpeper.
\newblock {AnyTask: an Automated Task and Data Generation Framework for
  Advancing Sim-to-Real Policy Learning}, 2025.
\newblock URL \url{https://arxiv.org/abs/2512.17853}.

\bibitem[Grover(2026)]{arxiv261027350}
Arnav Grover.
\newblock {Refuses When Asked, Complies When Acting: A Runtime Safety Check for
  Robot Control Loops}, 2026.
\newblock URL \url{https://arxiv.org/abs/2610.27350}.

\bibitem[Guan et~al.(2026)Guan, Zhang, Lyu, and Chen]{arxiv261024801}
Isabel Guan, Dingyuan Zhang, Shipeng Lyu, and I-Ming Chen.
\newblock {When Better Policies Leave Harder Failures: Coverage Shift in VLM
  Success Monitoring}, 2026.
\newblock URL \url{https://arxiv.org/abs/2610.24801}.

\bibitem[Gumbsch et~al.(2026)Gumbsch, Barcellona, Sch{\"u}nemann, Karageorgis,
  Zadaianchuk, Wang, Zakharov, Despinoy, Aljundi, and Gavves]{arxiv260600083}
Christian Gumbsch, Leonardo Barcellona, Lennard Sch{\"u}nemann, Platon
  Karageorgis, Andrii Zadaianchuk, Zehao Wang, Sergey Zakharov, Fabien
  Despinoy, Rahaf Aljundi, and Efstratios Gavves.
\newblock {From Demonstrations to Rewards: Test-Time Prompt Optimization for
  VLM Reward Models}, 2026.
\newblock URL \url{https://arxiv.org/abs/2606.00083}.

\bibitem[Ha et~al.(2023)Ha, Florence, and Song]{arxiv230714535}
Huy Ha, Pete Florence, and Shuran Song.
\newblock {Scaling Up and Distilling Down: Language-Guided Robot Skill
  Acquisition}, 2023.
\newblock URL \url{https://arxiv.org/abs/2307.14535}.

\bibitem[Han et~al.(2026)Han, Guo, Zeng, Cui, Xin, Zhou, Zhou, Zhang, and
  Li]{arxiv261016238}
Zhi Han, Zihan Guo, Chenxi Zeng, Zicai Cui, Yue Xin, Ming Zhou, Dongzhan Zhou,
  Weinan Zhang, and Yang Li.
\newblock {Embodied Skill Bench: Benchmarking How Well Agents with Embodied
  Skills Perform on Physical Tasks}, 2026.
\newblock URL \url{https://arxiv.org/abs/2610.16238}.

\bibitem[Haratian et~al.(2026)Haratian, Ansari, Malmir, and
  Knoll]{arxiv261116116}
Vahid Haratian, Vahid Ansari, Mohammadhossein Malmir, and Alois Knoll.
\newblock {Robo-SWE: An Executable Issue-Resolution Dataset Mined from Robot
  Software Repositories}, 2026.
\newblock URL \url{https://arxiv.org/abs/2611.16116}.

\bibitem[Hazra et~al.(2025)Hazra, Sygkounas, Persson, Loutfi, and
  Martires]{arxiv240601309}
Rishi Hazra, Alkis Sygkounas, Andreas Persson, Amy Loutfi, and Pedro
  Zuidberg~Dos Martires.
\newblock {REvolve: Reward Evolution with Large Language Models using Human
  Feedback}.
\newblock In \emph{International Conference on Learning Representations
  (ICLR)}, 2025.
\newblock URL \url{https://openreview.net/forum?id=cJPUpL8mOw}.

\bibitem[He et~al.(2026)He, Ma, Yu, Chen, Huang, Wu, Yang, and
  Jiang]{arxiv261002788}
Xincheng He, Siyu Ma, Chang Yu, Yunuo Chen, Yanjia Huang, Ying~Nian Wu, Yin
  Yang, and Chenfanfu Jiang.
\newblock {Skill2Real: Agentic Skill Learning for Zero-Shot Sim-to-Real Robot
  Manipulation}, 2026.
\newblock URL \url{https://arxiv.org/abs/2610.02788}.

\bibitem[Heravi et~al.(2026)Heravi, Ouyang, Xu, Kumar, Sung, and
  Stone]{arxiv260723515}
Faraz Heravi, James Ouyang, Zifan Xu, Arjun Kumar, Yoonchang Sung, and Peter
  Stone.
\newblock {LEACL: LLM-Enhanced Automatic Curriculum Learning for Reinforcement
  Learning in Long-Horizon Manipulation Tasks}, 2026.
\newblock URL \url{https://arxiv.org/abs/2607.23515}.

\bibitem[Hickok(2026)]{arxiv261036420}
Truman Hickok.
\newblock {SO-101 Bench: Accessible Spatial Manipulation}, 2026.
\newblock URL \url{https://arxiv.org/abs/2610.36420}.

\bibitem[Hsu et~al.(2026)Hsu, Setty, Butler, Lewis, Stone, Weinberg, Brettin,
  Stevens, Foster, and Ramanathan]{arxiv260105356}
Brian Hsu, Priyanka~V Setty, Rory~M Butler, Ryan Lewis, Casey Stone, Rebecca
  Weinberg, Thomas Brettin, Rick Stevens, Ian Foster, and Arvind Ramanathan.
\newblock {PRISM: Protocol Refinement through Intelligent Simulation Modeling}.
\newblock \emph{Digital Discovery}, 5\penalty0 (6):\penalty0 2613--2628, 2026.
\newblock \doi{10.1039/d6dd00004e}.

\bibitem[Hu et~al.(2026{\natexlab{a}})Hu, Martin-Martin, Stone, Duan, Jiang,
  and Shi]{arxiv260938982}
Jiaheng Hu, Roberto Martin-Martin, Peter Stone, Rocky Duan, Zhenyu Jiang, and
  Guanya Shi.
\newblock {SimEX: Simulation-Integrated Robotics AutoResearch},
  2026{\natexlab{a}}.
\newblock URL \url{https://arxiv.org/abs/2609.38982}.

\bibitem[Hu et~al.(2026{\natexlab{b}})Hu, Hu, Hong, Chen, and
  Liu]{arxiv261012369}
Kairui Hu, Siyuan Hu, Fangzhou Hong, Zhaoxi Chen, and Ziwei Liu.
\newblock {Embodied Turing Machines: Stateful Code for Robot Recursive
  Self-Improvement}, 2026{\natexlab{b}}.
\newblock URL \url{https://arxiv.org/abs/2610.12369}.

\bibitem[Hu et~al.(2023)Hu, Xie, Jain, Francis, Patrikar, Keetha, Kim, Xie,
  Zhang, Fang, Zhao, Omidshafiei, Kim, Agha-mohammadi, Sycara,
  Johnson-Roberson, Batra, Wang, Scherer, Wang, Kira, Xia, and
  Bisk]{arxiv231208782}
Yafei Hu, Quanting Xie, Vidhi Jain, Jonathan Francis, Jay Patrikar, Nikhil
  Keetha, Seungchan Kim, Yaqi Xie, Tianyi Zhang, Hao-Shu Fang, Shibo Zhao,
  Shayegan Omidshafiei, Dong-Ki Kim, {Ali-akbar} Agha-mohammadi, Katia Sycara,
  Matthew Johnson-Roberson, Dhruv Batra, Xiaolong Wang, Sebastian Scherer, Chen
  Wang, Zsolt Kira, Fei Xia, and Yonatan Bisk.
\newblock {Toward General-Purpose Robots via Foundation Models: A Survey and
  Meta-Analysis}, 2023.
\newblock URL \url{https://arxiv.org/abs/2312.08782}.

\bibitem[Hu et~al.(2024)Hu, Lucchetti, Schlesinger, Saxena, Freeman, Modak,
  Guha, and Biswas]{arxiv231111183}
Zichao Hu, Francesca Lucchetti, Claire Schlesinger, Yash Saxena, Anders
  Freeman, Sadanand Modak, Arjun Guha, and Joydeep Biswas.
\newblock {Deploying and Evaluating LLMs to Program Service Mobile Robots}.
\newblock \emph{IEEE Robotics and Automation Letters}, 9\penalty0 (3):\penalty0
  2853--2860, 2024.
\newblock \doi{10.1109/lra.2024.3360020}.

\bibitem[Hu et~al.(2025)Hu, Li, Guha, and Biswas]{arxiv240520179}
Zichao Hu, Junyi~Jessy Li, Arjun Guha, and Joydeep Biswas.
\newblock {Robo-Instruct: Simulator-Augmented Instruction Alignment For
  Finetuning Code LLMs}.
\newblock In \emph{Conference on Language Modeling (COLM)}, 2025.
\newblock URL \url{https://openreview.net/forum?id=WnZjdQOWiY}.

\bibitem[Hua et~al.(2024)Hua, Liu, Macaluso, Lin, Zhang, Xu, and
  Wang]{arxiv241003645}
Pu~Hua, Minghuan Liu, Annabella Macaluso, Yunfeng Lin, Weinan Zhang, Huazhe Xu,
  and Lirui Wang.
\newblock {GenSim2: Scaling Robot Data Generation with Multi-modal and
  Reasoning LLMs}.
\newblock In \emph{Conference on Robot Learning (CoRL)}, 2024.
\newblock URL \url{https://openreview.net/forum?id=5u9l6U61S7}.

\bibitem[Huang et~al.(2024{\natexlab{a}})Huang, Lin, Liu, Luo, Chen, and
  Ma]{huang-etal-2024-towards}
Jianzhao Huang, Hongzhan Lin, Ziyan Liu, Ziyang Luo, Guang Chen, and Jing Ma.
\newblock Towards low-resource harmful meme detection with {LMM} agents.
\newblock In Yaser Al-Onaizan, Mohit Bansal, and Yun-Nung Chen, editors,
  \emph{Proceedings of the 2024 Conference on Empirical Methods in Natural
  Language Processing}, pages 2269--2293, Miami, Florida, USA, November
  2024{\natexlab{a}}. Association for Computational Linguistics.
\newblock \doi{10.18653/v1/2024.emnlp-main.136}.
\newblock URL \url{https://aclanthology.org/2024.emnlp-main.136/}.

\bibitem[Huang et~al.(2026{\natexlab{a}})Huang, Zhang, Chen, Li, Wang, Lyu, Ye,
  Hong, Wang, Su, Xiong, Wu, Xie, Manderson, Ermon, Abbeel, Torralba, Song, Ye,
  and Du]{arxiv261022496}
Runhan Huang, Xiangcheng Zhang, Haoyu Chen, Yitong Li, Yen-Jen Wang, Bohan Lyu,
  Haotian Ye, Chuye Hong, Yixuan Wang, Chloe Su, Haoyu Xiong, Yangzhen Wu, Enze
  Xie, Travis Manderson, Stefano Ermon, Pieter Abbeel, Antonio Torralba, Dawn
  Song, Weirui Ye, and Yilun Du.
\newblock {Scaling Online Exploration for Robot Agents}, 2026{\natexlab{a}}.
\newblock URL \url{https://arxiv.org/abs/2610.22496}.

\bibitem[Huang et~al.(2023{\natexlab{a}})Huang, Jiang, Dong, Qiao, Gao, and
  Li]{arxiv230511176}
Siyuan Huang, Zhengkai Jiang, Hao Dong, Yu~Qiao, Peng Gao, and Hongsheng Li.
\newblock {Instruct2Act: Mapping Multi-modality Instructions to Robotic Actions
  with Large Language Model}, 2023{\natexlab{a}}.
\newblock URL \url{https://arxiv.org/abs/2305.11176}.

\bibitem[Huang et~al.(2026{\natexlab{b}})Huang, Bi, Yang, Si, Ni, Wang, Jiang,
  Zhang, Ruan, Zhang, Huang, Gong, and Zhou]{arxiv261108318}
Wenbo Huang, Jianxin Bi, Shunyu Yang, Ruiyang Si, Rui Ni, Qiang Wang, Shulong
  Jiang, Yintianrun Zhang, Feixiang Ruan, Kewei Zhang, Ye~Huang, Shukai Gong,
  and Daquan Zhou.
\newblock {Learning the Task or Learning the Verifier? Rethinking Criteria
  Exposure in Embodied Self-Improvement}, 2026{\natexlab{b}}.
\newblock URL \url{https://arxiv.org/abs/2611.08318}.

\bibitem[Huang et~al.(2022{\natexlab{a}})Huang, Abbeel, Pathak, and
  Mordatch]{arxiv220107207}
Wenlong Huang, Pieter Abbeel, Deepak Pathak, and Igor Mordatch.
\newblock {Language Models as Zero-Shot Planners: Extracting Actionable
  Knowledge for Embodied Agents}.
\newblock In \emph{International Conference on Machine Learning (ICML)}, pages
  9118--9147, 2022{\natexlab{a}}.
\newblock URL \url{https://proceedings.mlr.press/v162/huang22a.html}.

\bibitem[Huang et~al.(2022{\natexlab{b}})Huang, Xia, Xiao, Chan, Liang,
  Florence, Zeng, Tompson, Mordatch, Chebotar, Sermanet, Jackson, Brown, Luu,
  Levine, Hausman, and Ichter]{arxiv220705608}
Wenlong Huang, Fei Xia, Ted Xiao, Harris Chan, Jacky Liang, Pete Florence, Andy
  Zeng, Jonathan Tompson, Igor Mordatch, Yevgen Chebotar, Pierre Sermanet,
  Tomas Jackson, Noah Brown, Linda Luu, Sergey Levine, Karol Hausman, and Brian
  Ichter.
\newblock {Inner Monologue: Embodied Reasoning through Planning with Language
  Models}.
\newblock In \emph{Conference on Robot Learning (CoRL)}, pages 1769--1782,
  2022{\natexlab{b}}.
\newblock URL \url{https://proceedings.mlr.press/v205/huang23c.html}.

\bibitem[Huang et~al.(2023{\natexlab{b}})Huang, Wang, Zhang, Li, Wu, and
  Fei-Fei]{arxiv230705973}
Wenlong Huang, Chen Wang, Ruohan Zhang, Yunzhu Li, Jiajun Wu, and Li~Fei-Fei.
\newblock {VoxPoser: Composable 3D Value Maps for Robotic Manipulation with
  Language Models}.
\newblock In \emph{Conference on Robot Learning (CoRL)}, 2023{\natexlab{b}}.
\newblock URL \url{https://proceedings.mlr.press/v229/huang23b.html}.

\bibitem[Huang et~al.(2024{\natexlab{b}})Huang, Wang, Li, Zhang, and
  Fei-Fei]{arxiv240901652}
Wenlong Huang, Chen Wang, Yunzhu Li, Ruohan Zhang, and Li~Fei-Fei.
\newblock {ReKep: Spatio-Temporal Reasoning of Relational Keypoint Constraints
  for Robotic Manipulation}.
\newblock In \emph{Conference on Robot Learning (CoRL)}, 2024{\natexlab{b}}.
\newblock URL \url{https://proceedings.mlr.press/v270/huang25g.html}.

\bibitem[Huang et~al.(2026{\natexlab{c}})Huang, Yang, Shen, Ma, and
  Zheng]{arxiv260301414}
Xinyu Huang, Qiang Yang, Leming Shen, Zijing Ma, and Yuanqing Zheng.
\newblock {Jailbreaking Embodied LLMs via Action-level Manipulation}.
\newblock In \emph{Proceedings of the 2026 ACM/IEEE International Conference on
  Embedded Artificial Intelligence and Sensing Systems}, pages 1057--1071,
  2026{\natexlab{c}}.
\newblock \doi{10.1145/3774906.3802758}.

\bibitem[Hundt et~al.(2025)Hundt, Azeem, Mansouri, and
  Brandão]{arxiv240608824}
Andrew Hundt, Rumaisa Azeem, Masoumeh Mansouri, and Martim Brandão.
\newblock {LLM-Driven Robots Risk Enacting Discrimination, Violence, and
  Unlawful Actions}.
\newblock \emph{International Journal of Social Robotics}, 17\penalty0
  (11):\penalty0 2663--2711, 2025.
\newblock \doi{10.1007/s12369-025-01301-x}.

\bibitem[Iacob et~al.(2026)Iacob, Jovanović, Shen, Burkhardt, Kurmanji,
  Tastan, Sani, Venanzi, Nie, Odonnat, Cao, Marino, Qiu, Antonova, and
  Lane]{arxiv260626294}
Alex Iacob, Andrej Jovanović, William~F. Shen, Daniel Burkhardt, Meghdad
  Kurmanji, Nurbek Tastan, Lorenzo Sani, Niccolò Alberto~Elia Venanzi, Jiayi
  Nie, Ambroise Odonnat, Zeyu Cao, Bill Marino, Xinchi Qiu, Rika Antonova, and
  Nicholas~D. Lane.
\newblock {The Red Queen G\"odel Machine: Co-Evolving Agents and Their
  Evaluators}, 2026.
\newblock URL \url{https://arxiv.org/abs/2606.26294}.

\bibitem[Ichter et~al.(2022)Ichter, Brohan, Chebotar, Finn, Hausman, Herzog,
  Ho, Ibarz, Irpan, Jang, Julian, Kalashnikov, Levine, Lu, Parada, Rao,
  Sermanet, Toshev, Vanhoucke, Xia, Xiao, Xu, Yan, Brown, Ahn, Cortes, Sievers,
  Tan, Xu, Reyes, Rettinghouse, Quiambao, Pastor, Luu, Lee, Kuang, Jesmonth,
  Joshi, Jeffrey, Ruano, Hsu, Gopalakrishnan, David, Zeng, and
  Fu]{arxiv220401691}
Brian Ichter, Anthony Brohan, Yevgen Chebotar, Chelsea Finn, Karol Hausman,
  Alexander Herzog, Daniel Ho, Julian Ibarz, Alex Irpan, Eric Jang, Ryan
  Julian, Dmitry Kalashnikov, Sergey Levine, Yao Lu, Carolina Parada, Kanishka
  Rao, Pierre Sermanet, Alexander~T. Toshev, Vincent Vanhoucke, Fei Xia, Ted
  Xiao, Peng Xu, Mengyuan Yan, Noah Brown, Michael Ahn, Omar Cortes, Nicolas
  Sievers, Clayton Tan, Sichun Xu, Diego Reyes, Jarek Rettinghouse, Jornell
  Quiambao, Peter Pastor, Linda Luu, Kuang-Huei Lee, Yuheng Kuang, Sally
  Jesmonth, Nikhil~J. Joshi, Kyle Jeffrey, Rosario~Jauregui Ruano, Jasmine Hsu,
  Keerthana Gopalakrishnan, Byron David, Andy Zeng, and Chuyuan~Kelly Fu.
\newblock {Do As I Can, Not As I Say: Grounding Language in Robotic
  Affordances}.
\newblock In \emph{Conference on Robot Learning (CoRL)}, pages 287--318, 2022.
\newblock URL \url{https://proceedings.mlr.press/v205/ichter23a.html}.

\bibitem[Inagaki et~al.(2023)Inagaki, Kato, Takahashi, Ozaki, and
  Kanda]{arxiv230410267}
Takashi Inagaki, Akari Kato, Koichi Takahashi, Haruka Ozaki, and Genki~N.
  Kanda.
\newblock {LLMs can generate robotic scripts from goal-oriented instructions in
  biological laboratory automation}, 2023.
\newblock URL \url{https://arxiv.org/abs/2304.10267}.

\bibitem[Iscan(2026)]{arxiv260930219}
Mehmet Iscan.
\newblock {Requirement-Bound Verified Commissioning: A Frozen
  Four-Billion-Parameter Local Model as a Candidate Generator under an External
  Acceptance Layer with Verification and Release Authority}, 2026.
\newblock URL \url{https://arxiv.org/abs/2609.30219}.

\bibitem[Islam(2026)]{arxiv261100896}
Rafiqul Islam.
\newblock {Verifying Test-State Identity in Paired Evaluation of
  Vision-Language-Action Models}, 2026.
\newblock URL \url{https://arxiv.org/abs/2611.00896}.

\bibitem[Izzo et~al.(2026)Izzo, Bardaro, and Matteucci]{arxiv260201870}
Riccardo~Andrea Izzo, Gianluca Bardaro, and Matteo Matteucci.
\newblock {BTGenBot-2: Efficient Behavior Tree Generation with Small Language
  Models}.
\newblock In \emph{2026 International Joint Conference on Neural Networks
  (IJCNN)}, pages 1--8, 2026.
\newblock \doi{10.1109/ijcnn69854.2026.11728453}.

\bibitem[Jain et~al.(2026)Jain, Zhang, Arora, Chen, Torne, Irshad, Zakharov,
  Wang, Levine, Finn, Ma, Shah, Gupta, and Pertsch]{arxiv251216881}
Arhan Jain, Mingtong Zhang, Kanav Arora, William Chen, Marcel Torne,
  Muhammad~Zubair Irshad, Sergey Zakharov, Yue Wang, Sergey Levine, Chelsea
  Finn, Wei-Chiu Ma, Dhruv Shah, Abhishek Gupta, and Karl Pertsch.
\newblock {PolaRiS: Scalable Real-to-Sim Evaluations for Generalist Robot
  Policies}.
\newblock In \emph{Robotics: Science and Systems XXII}, 2026.
\newblock \doi{10.15607/rss.2026.xxii.062}.

\bibitem[Jangir et~al.(2026)Jangir, Zhang, Lo, Yamazaki, Zhang, Tu, Ke, Ke,
  Bisk, and Fragkiadaki]{arxiv251023571}
Yash Jangir, Yidi Zhang, Pang-Chi Lo, Kashu Yamazaki, Chenyu Zhang, Kuan-Hsun
  Tu, Tsung-Wei Ke, Lei Ke, Yonatan Bisk, and Katerina Fragkiadaki.
\newblock {RobotArena $\infty$: Scalable Robot Benchmarking via Real-to-Sim
  Translation}.
\newblock In \emph{International Conference on Learning Representations
  (ICLR)}, 2026.
\newblock URL \url{https://openreview.net/forum?id=OutljIofvS}.

\bibitem[Je-Gal et~al.(2026)Je-Gal, Kim, Yi, and Lee]{arxiv261104974}
Hong Je-Gal, Jinwoo Kim, Chanbin Yi, and Hyun-Suk Lee.
\newblock {LLMs as MDP Designers: A Dependency-Aware Agentic Framework for
  Automated Robot Policy Generation}, 2026.
\newblock URL \url{https://arxiv.org/abs/2611.04974}.

\bibitem[Jedlicki et~al.(2026)Jedlicki, Sethi, and Yang]{arxiv261017350}
Jack~B. Jedlicki, Amish Sethi, and Heng Yang.
\newblock {Can a Frontier VLM Serve as a Robot Policy for Memory-Dependent
  Manipulation?}, 2026.
\newblock URL \url{https://arxiv.org/abs/2610.17350}.

\bibitem[Jena et~al.(2026)Jena, Wanaskar, Jain, Chadha, Sharma, and
  Das]{arxiv260801851}
Gaytri Jena, Kapil Wanaskar, Vinija Jain, Aman Chadha, Vasu Sharma, and Amitava
  Das.
\newblock {Weights or Skills? A Survey of Robot-Learning Techniques: from
  Action-Predicting Weights to Robots that Write their Own Skills}, 2026.
\newblock URL \url{https://arxiv.org/abs/2608.01851}.

\bibitem[Jeon et~al.(2026)Jeon, Ye, Doo, Kim, Seo, Son, and
  Lee]{arxiv260701060}
Byeongguk Jeon, Seonghyeon Ye, JaeHyeok Doo, Sungdong Kim, Minjoon Seo,
  Hyungmok Son, and Kimin Lee.
\newblock {RoboWorld: Fast and Reliable Neural Simulators for Generalist Robot
  Policy Evaluation}, 2026.
\newblock URL \url{https://arxiv.org/abs/2607.01060}.

\bibitem[Jeung et~al.(2026)Jeung, Yoon, Hong, Cho, Jeon, Kim, Oh, Yu, and
  No]{arxiv260905401}
Wonje Jeung, Sangyeon Yoon, Hyesoo Hong, Yoonjun Cho, Dongjae Jeon, Bumjun Kim,
  Jean Oh, Youngjae Yu, and Albert No.
\newblock {Same Trajectory, Contradictory Rewards (ROBORMBENCH): Paraphrase
  Fragility in Vision Language Reward Models}.
\newblock In \emph{Proceedings of the 2026 Conference on Empirical Methods in
  Natural Language Processing}, pages 38376--38395, 2026.
\newblock \doi{10.18653/2026.emnlp-main.1687}.

\bibitem[Ji et~al.(2026)Ji, Chen, Chen, Zhu, Xu, Groß, Zhou, Cao, and
  Zhao]{arxiv250323875}
Wenkang Ji, Huaben Chen, Mingyang Chen, Guobin Zhu, Lufeng Xu, Roderich Groß,
  Rui Zhou, Ming Cao, and Shiyu Zhao.
\newblock {GenSwarm: Scalable Multi-Robot Code-Policy Generation and Deployment
  via Language Models}.
\newblock \emph{npj Robotics}, 4\penalty0 (1), 2026.
\newblock \doi{10.1038/s44182-025-00065-w}.

\bibitem[Jia et~al.(2026)Jia, Lin, Zhang, Zhang, Liu, and
  Jiang]{arxiv260912541}
Mengzhao Jia, Yang Lin, Xixin Zhang, Zhihan Zhang, Xiaobai Liu, and Meng Jiang.
\newblock {Agent as Policy for Robotic Manipulation}, 2026.
\newblock URL \url{https://arxiv.org/abs/2609.12541}.

\bibitem[Jiang et~al.(2026{\natexlab{a}})Jiang, Yang, Jiang, Bertasius, Szafir,
  and Lioutikov]{arxiv260919413}
Jing Jiang, Yue Yang, Xinkai Jiang, Gedas Bertasius, Daniel~J. Szafir, and
  Rudolf Lioutikov.
\newblock {From Rollout to Reset: A Graph-Based Harness for Autonomous
  Long-Horizon Manipulation Evaluation}, 2026{\natexlab{a}}.
\newblock URL \url{https://arxiv.org/abs/2609.19413}.

\bibitem[Jiang et~al.(2026{\natexlab{b}})Jiang, Wang, Shen, Kim, and
  Kim]{arxiv240600515}
Juyong Jiang, Fan Wang, Jiasi Shen, Sungju Kim, and Sunghun Kim.
\newblock {A Survey on Large Language Models for Code Generation}.
\newblock \emph{ACM Transactions on Software Engineering and Methodology},
  35\penalty0 (2):\penalty0 1--72, 2026{\natexlab{b}}.
\newblock \doi{10.1145/3747588}.

\bibitem[Jimenez et~al.(2024)Jimenez, Yang, Wettig, Yao, Pei, Press, and
  Narasimhan]{arxiv231006770}
Carlos~E. Jimenez, John Yang, Alexander Wettig, Shunyu Yao, Kexin Pei, Ofir
  Press, and Karthik Narasimhan.
\newblock {SWE-bench: Can Language Models Resolve Real-World GitHub Issues?}
\newblock In \emph{International Conference on Learning Representations
  (ICLR)}, 2024.
\newblock URL \url{https://openreview.net/forum?id=VTF8yNQM66}.

\bibitem[Jin et~al.(2024)Jin, Li, A, Shi, Hao, Sun, Zhang, and
  Fang]{arxiv231201421}
Yixiang Jin, Dingzhe Li, Yong A, Jun Shi, Peng Hao, Fuchun Sun, Jianwei Zhang,
  and Bin Fang.
\newblock {RobotGPT: Robot Manipulation Learning from ChatGPT}.
\newblock \emph{IEEE Robotics and Automation Letters}, 9\penalty0 (3):\penalty0
  2543--2550, 2024.
\newblock \doi{10.1109/lra.2024.3357432}.

\bibitem[Kagaya et~al.(2025{\natexlab{a}})Kagaya, Lakshmi, Lou, Yuan, Karlekar,
  Pranata, Murakami, Kinose, and You]{arxiv250924160}
Tomoyuki Kagaya, Subramanian Lakshmi, Yuxuan Lou, Thong~Jing Yuan, Jayashree
  Karlekar, Sugiri Pranata, Natsuki Murakami, Akira Kinose, and Yang You.
\newblock {Memory Transfer Planning: LLM-driven Context-Aware Code Adaptation
  for Robot Manipulation}, 2025{\natexlab{a}}.
\newblock URL \url{https://arxiv.org/abs/2509.24160}.

\bibitem[Kagaya et~al.(2025{\natexlab{b}})Kagaya, Lakshmi, Ye, Yuan, Karlekar,
  Pranata, Murakami, Kinose, and You]{arxiv250924219}
Tomoyuki Kagaya, Subramanian Lakshmi, Anbang Ye, Thong~Jing Yuan, Jayashree
  Karlekar, Sugiri Pranata, Natsuki Murakami, Akira Kinose, and Yang You.
\newblock {ViReSkill: Vision-Grounded Replanning with Skill Memory for
  LLM-Based Planning in Lifelong Robot Learning}, 2025{\natexlab{b}}.
\newblock URL \url{https://arxiv.org/abs/2509.24219}.

\bibitem[Kang et~al.(2026{\natexlab{a}})Kang, Ye, Liu, Mantri, Mao, Fleming,
  Regmi, and Qin]{arxiv260509423}
Haoqiang Kang, Xiaokang Ye, Yuhan Liu, Siddhant~Hitesh Mantri, Lingjun Mao,
  James Fleming, Drishti Regmi, and Lianhui Qin.
\newblock {SimWorld Studio: Automatic Environment Generation with Evolving
  Coding Agent for Embodied Agent Learning}, 2026{\natexlab{a}}.
\newblock URL \url{https://arxiv.org/abs/2605.09423}.

\bibitem[Kang et~al.(2026{\natexlab{b}})Kang, Wang, Fan, and
  Liu]{arxiv260937359}
Yifan Kang, Zihan Wang, Zhiwen Fan, and Bangya Liu.
\newblock {Encore: Few-Shot Agentic Discovery of Manipulation Strategies},
  2026{\natexlab{b}}.
\newblock URL \url{https://arxiv.org/abs/2609.37359}.

\bibitem[Katara et~al.(2024)Katara, Xian, and Fragkiadaki]{arxiv231018308}
Pushkal Katara, Zhou Xian, and Katerina Fragkiadaki.
\newblock {Gen2Sim: Scaling up Robot Learning in Simulation with Generative
  Models}.
\newblock In \emph{2024 IEEE International Conference on Robotics and
  Automation (ICRA)}, pages 6672--6679, 2024.
\newblock \doi{10.1109/icra57147.2024.10610566}.

\bibitem[Kawaharazuka et~al.(2024)Kawaharazuka, Matsushima, Gambardella, Guo,
  Paxton, and Zeng]{arxiv240205741}
Kento Kawaharazuka, Tatsuya Matsushima, Andrew Gambardella, Jiaxian Guo, Chris
  Paxton, and Andy Zeng.
\newblock {Real-World Robot Applications of Foundation Models: A Review}.
\newblock \emph{Advanced Robotics}, 38\penalty0 (18):\penalty0 1232--1254,
  2024.
\newblock \doi{10.1080/01691864.2024.2408593}.

\bibitem[Kim et~al.(2026{\natexlab{a}})Kim, White, He, Sala, and
  Mutlu]{arxiv260306842}
Callie~Y. Kim, Nathan~Thomas White, Evan He, Frederic Sala, and Bilge Mutlu.
\newblock {RoboCritics: Enabling Reliable End-to-End LLM Robot Programming
  through Expert-Informed Critics}.
\newblock In \emph{Proceedings of the 21st ACM/IEEE International Conference on
  Human-Robot Interaction}, pages 914--923, 2026{\natexlab{a}}.
\newblock \doi{10.1145/3757279.3785550}.

\bibitem[Kim et~al.(2026{\natexlab{b}})Kim, Lee, Seo, Choi, and
  Lee]{arxiv261002982}
Yunji Kim, Yunseok Lee, Hyunwoo Seo, Jaerim Choi, and Woojin Lee.
\newblock {PLCWorld: Benchmarking LLM-Generated PLC Programs in Closed-Loop
  Plant Simulation}, 2026{\natexlab{b}}.
\newblock URL \url{https://arxiv.org/abs/2610.02982}.

\bibitem[Kong et~al.(2025)Kong, Luo, Lin, Fan, Fan, Sun, and
  Ma]{kong-etal-2025-sharp}
Chuyi Kong, Ziyang Luo, Hongzhan Lin, Zhiyuan Fan, Yaxin Fan, Yuxi Sun, and
  Jing Ma.
\newblock {SHARP}: Unlocking interactive hallucination via stance transfer in
  role-playing {LLM}s.
\newblock In Wanxiang Che, Joyce Nabende, Ekaterina Shutova, and Mohammad~Taher
  Pilehvar, editors, \emph{Findings of the {A}ssociation for {C}omputational
  {L}inguistics: {ACL} 2025}, pages 839--866, Vienna, Austria, July 2025.
  Association for Computational Linguistics.
\newblock ISBN 979-8-89176-256-5.
\newblock \doi{10.18653/v1/2025.findings-acl.47}.
\newblock URL \url{https://aclanthology.org/2025.findings-acl.47/}.

\bibitem[Koziolek et~al.(2023)Koziolek, Gruener, and Ashiwal]{arxiv230515809}
Heiko Koziolek, Sten Gruener, and Virendra Ashiwal.
\newblock {ChatGPT for PLC/DCS Control Logic Generation}.
\newblock In \emph{2023 IEEE 28th International Conference on Emerging
  Technologies and Factory Automation (ETFA)}, pages 1--8, 2023.
\newblock \doi{10.1109/etfa54631.2023.10275411}.

\bibitem[Koziolek et~al.(2024)Koziolek, Ashiwal, Bandyopadhyay, and
  R]{arxiv240501874}
Heiko Koziolek, Virendra Ashiwal, Soumyadip Bandyopadhyay, and Chandrika~K R.
\newblock {Automated Control Logic Test Case Generation using Large Language
  Models}.
\newblock In \emph{2024 IEEE 29th International Conference on Emerging
  Technologies and Factory Automation (ETFA)}, pages 1--8, 2024.
\newblock \doi{10.1109/etfa61755.2024.10711016}.

\bibitem[Kress-Gazit et~al.(2024)Kress-Gazit, Hashimoto, Kuppuswamy, Shah,
  Horgan, Richardson, Feng, and Burchfiel]{arxiv240909491}
Hadas Kress-Gazit, Kunimatsu Hashimoto, Naveen Kuppuswamy, Paarth Shah, Phoebe
  Horgan, Gordon Richardson, Siyuan Feng, and Benjamin Burchfiel.
\newblock {Robot Learning as an Empirical Science: Best Practices for Policy
  Evaluation}, 2024.
\newblock URL \url{https://arxiv.org/abs/2409.09491}.

\bibitem[Kuroki et~al.(2026)Kuroki, Kawaharazuka, and Tang]{arxiv260919906}
So~Kuroki, Kento Kawaharazuka, and Yujin Tang.
\newblock {Learning and Transferring Closed-Loop Robot Software}, 2026.
\newblock URL \url{https://arxiv.org/abs/2609.19906}.

\bibitem[Larchenko et~al.(2025)Larchenko, Zarin, and Karnatak]{arxiv251206951}
Ilia Larchenko, Gleb Zarin, and Akash Karnatak.
\newblock {Task adaptation of Vision-Language-Action model: 1st Place Solution
  for the 2025 BEHAVIOR Challenge}, 2025.
\newblock URL \url{https://arxiv.org/abs/2512.06951}.

\bibitem[Le et~al.(2025)Le, Xie, Liang, Wang, Yang, Ma, Vedder, Krishna,
  Jayaraman, and Eaton]{arxiv241013882}
Long Le, Jason Xie, William Liang, Hung-Ju Wang, Yue Yang, Yecheng~Jason Ma,
  Kyle Vedder, Arjun Krishna, Dinesh Jayaraman, and Eric Eaton.
\newblock {Articulate-Anything: Automatic Modeling of Articulated Objects via a
  Vision-Language Foundation Model}.
\newblock In \emph{International Conference on Learning Representations
  (ICLR)}, 2025.
\newblock URL \url{https://openreview.net/forum?id=s3FTX4Ay55}.

\bibitem[Lee et~al.(2026{\natexlab{a}})Lee, Subramanian, Abbatematteo,
  Veerabadran, Matias, Ridgeway, and Kamra]{arxiv260601672}
Hojoon Lee, Ajay Subramanian, Ben Abbatematteo, Vijay Veerabadran, Pedro
  Matias, Karl Ridgeway, and Nitin Kamra.
\newblock {RDA: Reward Design Agent for Reinforcement Learning},
  2026{\natexlab{a}}.
\newblock URL \url{https://arxiv.org/abs/2606.01672}.

\bibitem[Lee et~al.(2026{\natexlab{b}})Lee, Seo, Cho, Choe, Wang, Saravanan,
  Huang, Huang, Scherer, Shi, Kim, Lee, and Lee]{arxiv261012549}
Jaewoo Lee, Jeongyeon Seo, Sihyun Cho, Gyeongrak Choe, Yutong Wang, Bavin
  Saravanan, Jia-Bin Huang, Furong Huang, Sebastian Scherer, Guanya Shi, H.~Jin
  Kim, Seungjae Lee, and Dongjae Lee.
\newblock {Fly-by-Code: Embodied Coding Agents for Aerial Manipulation with
  Active Visual and Physical Feedback}, 2026{\natexlab{b}}.
\newblock URL \url{https://arxiv.org/abs/2610.12549}.

\bibitem[Lee et~al.(2026{\natexlab{c}})Lee, Wagenmaker, Pertsch, Liang, Levine,
  and Finn]{arxiv260100675}
Tony Lee, Andrew Wagenmaker, Karl Pertsch, Percy Liang, Sergey Levine, and
  Chelsea Finn.
\newblock {RoboReward: General-Purpose Vision-Language Reward Models for
  Robotics}, 2026{\natexlab{c}}.
\newblock URL \url{https://arxiv.org/abs/2601.00675}.

\bibitem[Lehman et~al.(2022)Lehman, Gordon, Jain, Ndousse, Yeh, and
  Stanley]{arxiv220608896}
Joel Lehman, Jonathan Gordon, Shawn Jain, Kamal Ndousse, Cathy Yeh, and
  Kenneth~O. Stanley.
\newblock {Evolution through Large Models}, 2022.
\newblock URL \url{https://arxiv.org/abs/2206.08896}.

\bibitem[Lei et~al.(2026)Lei, Liu, Zhang, Liu, Cai, Chen, Wu, Wang, Li, Wen,
  Zhao, Zhang, Lian, and Chen]{arxiv260121570}
Zixing Lei, Genjia Liu, Yuanshuo Zhang, Qipeng Liu, Yuzhu Cai, Sixiang Chen,
  Jixian Wu, Yunhong Wang, Weixin Li, Chuan Wen, Bo~Zhao, Shanghang Zhang,
  Wenzhao Lian, and Siheng Chen.
\newblock {From Digital to Physical: Digital Agents as Autonomous Coaches for
  Physical Intelligence}, 2026.
\newblock URL \url{https://arxiv.org/abs/2601.21570}.

\bibitem[Li et~al.(2024{\natexlab{a}})Li, Liang, Zeng, Chen, Hausman, Sadigh,
  Levine, Fei-Fei, Xia, and Ichter]{arxiv231204474}
Chengshu Li, Jacky Liang, Andy Zeng, Xinyun Chen, Karol Hausman, Dorsa Sadigh,
  Sergey Levine, Li~Fei-Fei, Fei Xia, and Brian Ichter.
\newblock {Chain of Code: Reasoning with a Language Model-Augmented Code
  Emulator}.
\newblock In \emph{International Conference on Machine Learning (ICML)},
  2024{\natexlab{a}}.
\newblock URL \url{https://openreview.net/forum?id=vKtomqlSxm}.

\bibitem[Li et~al.(2026{\natexlab{a}})Li, Zhao, Li, Gao, Liu, Li, Wang, Si,
  Pan, Ouyang, and Zhou]{arxiv261002717}
Chenxi Li, Zhangrui Zhao, Rui Li, Yuan Gao, Kehui Liu, Jiarui Li, Dong Wang,
  Tong Si, Minting Pan, Wanli Ouyang, and Dongzhan Zhou.
\newblock {RoboBridge: A Self-Evolving Embodied Agent Framework for Sim-to-Real
  Transfer}, 2026{\natexlab{a}}.
\newblock URL \url{https://arxiv.org/abs/2610.02717}.

\bibitem[Li et~al.(2024{\natexlab{b}})Li, Chen, Wu, Zheng, Xu, and
  Jia]{arxiv240603757}
Jingyao Li, Pengguang Chen, Sitong Wu, Chuanyang Zheng, Hong Xu, and Jiaya Jia.
\newblock {RoboCoder: Robotic Learning from Basic Skills to General Tasks with
  Large Language Models}, 2024{\natexlab{b}}.
\newblock URL \url{https://arxiv.org/abs/2406.03757}.

\bibitem[Li et~al.(2025)Li, Meng, Lin, Luo, Tian, Ma, Huang, and
  Chua]{arxiv250407981}
Kaixin Li, Ziyang Meng, Hongzhan Lin, Ziyang Luo, Yuchen Tian, Jing Ma, Zhiyong
  Huang, and Tat-Seng Chua.
\newblock {ScreenSpot-Pro: GUI Grounding for Professional High-Resolution
  Computer Use}.
\newblock In \emph{Proceedings of the 33rd ACM International Conference on
  Multimedia}, pages 8778--8786, 2025.
\newblock \doi{10.1145/3746027.3755688}.

\bibitem[Li et~al.(2024{\natexlab{c}})Li, Zhao, Wang, Wang, Zhou, Srivastava,
  Gokmen, Lee, Li, Zhang, Liu, Liang, Fei-Fei, Mao, and Wu]{arxiv241007166}
Manling Li, Shiyu Zhao, Qineng Wang, Kangrui Wang, Yu~Zhou, Sanjana Srivastava,
  Cem Gokmen, Tony Lee, Li~Erran Li, Ruohan Zhang, Weiyu Liu, Percy Liang,
  Li~Fei-Fei, Jiayuan Mao, and Jiajun Wu.
\newblock {Embodied Agent Interface: Benchmarking LLMs for Embodied Decision
  Making}.
\newblock In \emph{Advances in Neural Information Processing Systems 37}, pages
  100428--100534, 2024{\natexlab{c}}.
\newblock \doi{10.52202/079017-3188}.

\bibitem[Li et~al.(2026{\natexlab{b}})Li, Chi, Zheng, Liu, Wang, Gao, Jiang,
  Hong, Xiao, Yang, and Na]{arxiv261018867}
Wenyi Li, Yizhe Chi, Youjie Zheng, Chunyu Liu, Xiaoqiu Wang, Mingju Gao, Zheng
  Jiang, Deyao Hong, Calvin Xiao, Kaisen Yang, and Qinhuai Na.
\newblock {EmbodiedEvo: Self-Evolving Coding Agent for Physical Intelligence},
  2026{\natexlab{b}}.
\newblock URL \url{https://arxiv.org/abs/2610.18867}.

\bibitem[Li et~al.(2026{\natexlab{c}})Li, Xu, Kong, Zhou, Liu, Yang, Huang, and
  Shan]{arxiv260923133}
Xiaoshi Li, Yule Xu, Chunghiu Kong, Yizhou Zhou, Yang Liu, Hao Yang, Zihao
  Huang, and Yunxiao Shan.
\newblock {AquaCap: A Training-Free Underwater Embodied Agent with
  Code-as-Policy}, 2026{\natexlab{c}}.
\newblock URL \url{https://arxiv.org/abs/2609.23133}.

\bibitem[Li et~al.(2024{\natexlab{d}})Li, Hsu, Gu, Mees, Pertsch, Walke, Fu,
  Lunawat, Sieh, Kirmani, Levine, Wu, Finn, Su, Vuong, and
  Xiao]{arxiv240505941}
Xuanlin Li, Kyle Hsu, Jiayuan Gu, Oier Mees, Karl Pertsch, Homer~Rich Walke,
  Chuyuan Fu, Ishikaa Lunawat, Isabel Sieh, Sean Kirmani, Sergey Levine, Jiajun
  Wu, Chelsea Finn, Hao Su, Quan Vuong, and Ted Xiao.
\newblock {Evaluating Real-World Robot Manipulation Policies in Simulation}.
\newblock In \emph{Conference on Robot Learning (CoRL)}, 2024{\natexlab{d}}.
\newblock URL \url{https://proceedings.mlr.press/v270/li25c.html}.

\bibitem[Li et~al.(2026{\natexlab{d}})Li, Hao, Wu, Fei, Lee, and
  Hsu]{arxiv260931847}
Yanlin Li, Mingyang Hao, Shengqiong Wu, Hao Fei, Mong-Li Lee, and Wynne Hsu.
\newblock {Omni-IO Skills: Harnessing Your Agent Omni-Native},
  2026{\natexlab{d}}.
\newblock URL \url{https://arxiv.org/abs/2609.31847}.

\bibitem[Li et~al.(2026{\natexlab{e}})Li, Feng, Sun, Chen, Ren, Yang, Xu, Wu,
  Cheng, and Xu]{arxiv261010479}
Yihan Li, Yating Feng, Shengjiu Sun, Jianing Chen, Hao Ren, Bowen Yang,
  Weisheng Xu, Qiwei Wu, Hui Cheng, and Renjing Xu.
\newblock {Agentic RSR: Real-to-Sim-to-Real through Scene Reconstruction and
  Execution-Grounded Robot Policies}, 2026{\natexlab{e}}.
\newblock URL \url{https://arxiv.org/abs/2610.10479}.

\bibitem[Li et~al.(2026{\natexlab{f}})Li, Cheng, Ma, Zou, Yang, Cheng, Li,
  Chen, and Chen]{arxiv260319583}
Yiming Li, Yuhan Cheng, Mingchen Ma, Yihang Zou, Ningyuan Yang, Wei Cheng,
  Hai~"Helen" Li, Yiran Chen, and Tingjun Chen.
\newblock {Skilled AI Agents for Embedded and IoT Systems Development}.
\newblock In \emph{Proceedings of the ACM Conference on AI and Agentic
  Systems}, pages 1248--1252, 2026{\natexlab{f}}.
\newblock \doi{10.1145/3786335.3813205}.

\bibitem[Li et~al.(2026{\natexlab{g}})Li, Jin, Liu, Liu, Prasad, D'Eramo, and
  Chalvatzaki]{arxiv260608610}
Zechu Li, Yufeng Jin, Xiaoyang Liu, Puze Liu, Vignesh Prasad, Carlo D'Eramo,
  and Georgia Chalvatzaki.
\newblock {HARBOR: A Harness Framework for Agentic Robot Reinforcement
  Learning}.
\newblock In \emph{Conference on Robot Learning (CoRL)}, 2026{\natexlab{g}}.
\newblock URL \url{https://openreview.net/forum?id=JqZEEaNDLJ}.

\bibitem[Li et~al.(2026{\natexlab{h}})Li, Zhang, Huang, Zhou, Li, Wang, Chang,
  Wei, Zhang, Chen, Zhou, Li, and Cheng]{arxiv260929394}
Zexi Li, Yehang Zhang, Haojian Huang, Bohan Zhou, Wenqian Li, Chenxu Wang,
  Yifan Chang, Yangkai Wei, Tianyi Zhang, Ying-Cong Chen, Kaiwen Zhou, Yinchuan
  Li, and James Cheng.
\newblock {RACaP: Agentic Reasoning, Acting, and Coding as Policies for
  Evolvable Robot Learning}, 2026{\natexlab{h}}.
\newblock URL \url{https://arxiv.org/abs/2609.29394}.

\bibitem[Liang et~al.(2023)Liang, Huang, Xia, Xu, Hausman, Ichter, Florence,
  and Zeng]{arxiv220907753}
Jacky Liang, Wenlong Huang, Fei Xia, Peng Xu, Karol Hausman, Brian Ichter, Pete
  Florence, and Andy Zeng.
\newblock {Code as Policies: Language Model Programs for Embodied Control}.
\newblock In \emph{2023 IEEE International Conference on Robotics and
  Automation (ICRA)}, pages 9493--9500, 2023.
\newblock \doi{10.1109/icra48891.2023.10160591}.

\bibitem[Liang et~al.(2024{\natexlab{a}})Liang, Xia, Yu, Zeng, Attarian,
  Villalonga, Bennice, Bewley, Dostmohamed, Fu, Gileadi, Giustina,
  Gopalakrishnan, Hasenclever, Humplik, Hsu, Joshi, Jyenis, Kew, Kirmani, Lee,
  Lee, Michaely, Moore, Oslund, Rao, Ren, Tabanpour, Vuong, Wahid, Xiao, Xu,
  Zhuang, Xu, Frey, Caluwaerts, Zhang, Ichter, Tompson, Takayama, Vanhoucke,
  Shafran, Mataric, Sadigh, Heess, Rao, Stewart, Tan, and
  Parada]{arxiv240211450}
Jacky Liang, Fei Xia, Wenhao Yu, Andy Zeng, Maria Attarian, Maria~Bauzá
  Villalonga, Matthew Bennice, Alex Bewley, Adil Dostmohamed, Chuyuan Fu,
  Nimrod Gileadi, Marissa Giustina, Keerthana Gopalakrishnan, Leonard
  Hasenclever, Jan Humplik, Jasmine Hsu, Nikhil~J. Joshi, Ben Jyenis, J.~Chase
  Kew, Sean Kirmani, Tsang-Wei~Edward Lee, Kuang-Huei Lee, Assaf~Hurwitz
  Michaely, Joss Moore, Kenneth Oslund, Dushyant Rao, Allen~Z. Ren, Baruch
  Tabanpour, Quan Vuong, Ayzaan Wahid, Ted Xiao, Ying Xu, Vincent Zhuang, Peng
  Xu, Erik Frey, Ken Caluwaerts, Tingnan Zhang, Brian Ichter, Jonathan Tompson,
  Leila Takayama, Vincent Vanhoucke, Izhak Shafran, Maja~J. Mataric, Dorsa
  Sadigh, Nicolas Heess, Kanishka Rao, Nik Stewart, Jie Tan, and Carolina
  Parada.
\newblock {Learning to Learn Faster from Human Feedback with Language Model
  Predictive Control}.
\newblock In \emph{Robotics: Science and Systems (RSS)}, 2024{\natexlab{a}}.
\newblock URL \url{https://doi.org/10.15607/RSS.2024.XX.125}.

\bibitem[Liang et~al.(2024{\natexlab{b}})Liang, Wang, Wang, Bastani, Jayaraman,
  and Ma]{arxiv241101775}
William Liang, Sam Wang, Hung-Ju Wang, Osbert Bastani, Dinesh Jayaraman, and
  Yecheng~Jason Ma.
\newblock {Environment Curriculum Generation via Large Language Models}.
\newblock In \emph{Conference on Robot Learning (CoRL)}, 2024{\natexlab{b}}.
\newblock URL \url{https://openreview.net/forum?id=F0rWEID2gb}.

\bibitem[Lin et~al.(2024)Lin, Luo, Gao, Ma, Wang, and Yang]{lin2024explainhm}
Hongzhan Lin, Ziyang Luo, Wei Gao, Jing Ma, Bo~Wang, and Ruichao Yang.
\newblock {Towards Explainable Harmful Meme Detection through Multimodal Debate
  between Large Language Models}.
\newblock In \emph{Proceedings of the ACM Web Conference 2024}, pages
  2359--2370, 2024.
\newblock \doi{10.1145/3589334.3645381}.
\newblock URL \url{https://doi.org/10.1145/3589334.3645381}.

\bibitem[Lin et~al.(2025)Lin, Deng, Gu, Zhang, Ma, Ng, and
  Chua]{lin-etal-2025-fact}
Hongzhan Lin, Yang Deng, Yuxuan Gu, Wenxuan Zhang, Jing Ma, See-Kiong Ng, and
  Tat-Seng Chua.
\newblock {FACT}-{AUDIT}: An adaptive multi-agent framework for dynamic
  fact-checking evaluation of large language models.
\newblock In Wanxiang Che, Joyce Nabende, Ekaterina Shutova, and Mohammad~Taher
  Pilehvar, editors, \emph{Proceedings of the 63rd Annual Meeting of the
  {A}ssociation for {C}omputational {L}inguistics (Volume 1: Long Papers)},
  pages 360--381, Vienna, Austria, July 2025. Association for Computational
  Linguistics.
\newblock ISBN 979-8-89176-251-0.
\newblock \doi{10.18653/v1/2025.acl-long.17}.
\newblock URL \url{https://aclanthology.org/2025.acl-long.17/}.

\bibitem[Lin et~al.(2026{\natexlab{a}})Lin, Cao, Luo, Chai, Lee, and
  Hsu]{arxiv261007753}
Hongzhan Lin, Shidong Cao, Ziyang Luo, Wenhao Chai, Mong-Li Lee, and Wynne Hsu.
\newblock {From Evidence to Action: How Tool-Using Agents Fail},
  2026{\natexlab{a}}.
\newblock URL \url{https://arxiv.org/abs/2610.07753}.

\bibitem[Lin et~al.(2026{\natexlab{b}})Lin, Chen, Shen, Luo, Ye, Ma, Chua, and
  Xu]{arxiv260102669}
Hongzhan Lin, Zixin Chen, Zhiqi Shen, Ziyang Luo, Zhen Ye, Jing Ma, Tat-Seng
  Chua, and Guandong Xu.
\newblock {Towards Comprehensive Stage-wise Benchmarking of Large Language
  Models in Fact-Checking}, 2026{\natexlab{b}}.
\newblock URL \url{https://arxiv.org/abs/2601.02669}.

\bibitem[Lin et~al.(2026{\natexlab{c}})Lin, Gao, Ma, Deng, Luo, Wang, Yang, and
  Chua]{lin2026explainhmpp}
Hongzhan Lin, Wei Gao, Jing Ma, Yang Deng, Ziyang Luo, Bo~Wang, Ruichao Yang,
  and Tat-Seng Chua.
\newblock {ExplainHM++: Explainable Harmful Meme Detection With
  Retrieval-Augmented Debate Between Large Multimodal Models}.
\newblock \emph{IEEE Transactions on Knowledge and Data Engineering},
  38\penalty0 (2):\penalty0 1040--1053, 2026{\natexlab{c}}.
\newblock \doi{10.1109/TKDE.2025.3637552}.
\newblock URL \url{https://doi.org/10.1109/TKDE.2025.3637552}.

\bibitem[Lin et~al.(2026{\natexlab{d}})Lin, Xu, Bian, Wang, and
  Gui]{arxiv261002196}
Zhuo Lin, Sirui Xu, Liuyu Bian, Yu-Xiong Wang, and Liang-Yan Gui.
\newblock {InterEvolve: Test-Time Evolution of Reward Programs for Humanoid
  Loco-Manipulation}, 2026{\natexlab{d}}.
\newblock URL \url{https://arxiv.org/abs/2610.02196}.

\bibitem[Liu et~al.(2026{\natexlab{a}})Liu, Nedungadi, Cai, Kleinman, Bhasin,
  and Donoughe]{arxiv260611150}
Andrew~Bo Liu, Samira Nedungadi, Bryce Cai, Alex Kleinman, Harmon Bhasin, and
  Seth Donoughe.
\newblock {ABC-Bench: An Agentic Bio-Capabilities Benchmark for Biosecurity}.
\newblock In \emph{International Conference on Machine Learning (ICML)},
  2026{\natexlab{a}}.
\newblock URL \url{https://openreview.net/forum?id=yiaf7VlPpH}.

\bibitem[Liu et~al.(2023{\natexlab{a}})Liu, Jiang, Zhang, Liu, Zhang, Biswas,
  and Stone]{arxiv230411477}
Bo~Liu, Yuqian Jiang, Xiaohan Zhang, Qiang Liu, Shiqi Zhang, Joydeep Biswas,
  and Peter Stone.
\newblock {LLM+P: Empowering Large Language Models with Optimal Planning
  Proficiency}, 2023{\natexlab{a}}.
\newblock URL \url{https://arxiv.org/abs/2304.11477}.

\bibitem[Liu et~al.(2023{\natexlab{b}})Liu, Zhu, Gao, Feng, Liu, Zhu, and
  Stone]{arxiv230603310}
Bo~Liu, Yifeng Zhu, Chongkai Gao, Yihao Feng, Qiang Liu, Yuke Zhu, and Peter
  Stone.
\newblock {LIBERO: Benchmarking Knowledge Transfer for Lifelong Robot
  Learning}.
\newblock In \emph{Advances in Neural Information Processing Systems 36}, pages
  44776--44791, 2023{\natexlab{b}}.
\newblock \doi{10.52202/075280-1939}.

\bibitem[Liu et~al.(2026{\natexlab{b}})Liu, Ye, Yuan, Feng, Huang, Dinh,
  Vorobeychik, Zhang, and Zhang]{arxiv261036533}
Buzhao Liu, Jianing Ye, Xiu Yuan, Yangtong Feng, Chengsong Huang, Tyler Dinh,
  Yevgeniy Vorobeychik, Ning Zhang, and Chongjie Zhang.
\newblock {RoboAuto: Autonomous Robot Program Search through Multi-Level
  Physical Understanding}, 2026{\natexlab{b}}.
\newblock URL \url{https://arxiv.org/abs/2610.36533}.

\bibitem[Liu et~al.(2026{\natexlab{c}})Liu, Wang, Chen, Li, and
  Huang]{arxiv260818827}
Chenglin Liu, Xun Wang, Ruishuo Chen, Zhuoran Li, and Longbo Huang.
\newblock {MLREF: Efficient Module Reuse for Reward Design in Reinforcement
  Learning via Large Language Models}, 2026{\natexlab{c}}.
\newblock URL \url{https://arxiv.org/abs/2608.18827}.

\bibitem[Liu et~al.(2023{\natexlab{c}})Liu, Yang, Idrees, Liang, Schornstein,
  Tellex, and Shah]{arxiv230211649}
Jason~Xinyu Liu, Ziyi Yang, Ifrah Idrees, Sam Liang, Benjamin Schornstein,
  Stefanie Tellex, and Ankit Shah.
\newblock {Grounding Complex Natural Language Commands for Temporal Tasks in
  Unseen Environments}.
\newblock In \emph{Conference on Robot Learning (CoRL)}, pages 1084--1110,
  2023{\natexlab{c}}.
\newblock URL \url{https://proceedings.mlr.press/v229/liu23d.html}.

\bibitem[Liu et~al.(2026{\natexlab{d}})Liu, Wang, Chen, Peng, Chen, Zhang, and
  Lou]{arxiv240902977}
Junwei Liu, Kaixin Wang, Yixuan Chen, Xin Peng, Zhenpeng Chen, Lingming Zhang,
  and Yiling Lou.
\newblock {Large Language Model-Based Agents for Software Engineering: A
  Survey}.
\newblock \emph{ACM Transactions on Software Engineering and Methodology},
  2026{\natexlab{d}}.
\newblock \doi{10.1145/3796507}.

\bibitem[Liu et~al.(2026{\natexlab{e}})Liu, Corcodel, and Zhao]{arxiv261038388}
Shiqi Liu, Radu Corcodel, and Ding Zhao.
\newblock {ELEVATE: Agentic Controller Acquisition from Task Failure for
  Quadrupedal Manipulation}, 2026{\natexlab{e}}.
\newblock URL \url{https://arxiv.org/abs/2610.38388}.

\bibitem[Liu et~al.(2025)Liu, Li, Zhang, Cui, Fang, Zheng, Zheng, and
  Song]{arxiv240715325}
Shunyu Liu, Yaoru Li, Kongcheng Zhang, Zhenyu Cui, Wenkai Fang, Yuxuan Zheng,
  Tongya Zheng, and Mingli Song.
\newblock {Odyssey: Empowering Minecraft Agents with Open-World Skills}.
\newblock In \emph{Proceedings of the Thirty-Fourth International Joint
  Conference on Artificial Intelligence}, pages 187--195, 2025.
\newblock \doi{10.24963/ijcai.2025/22}.

\bibitem[Liu et~al.(2026{\natexlab{f}})Liu, Xu, Bao, Li, Bao, Chen, and
  Shang]{arxiv261028589}
Xinxin Liu, Bingxin Xu, Wenrui Bao, Ming Li, Wentao Bao, Chen Chen, and Yuzhang
  Shang.
\newblock {AgenticRoboRL: Bringing Coding-Agent Post-Training to Agentic
  Robotics}, 2026{\natexlab{f}}.
\newblock URL \url{https://arxiv.org/abs/2610.28589}.

\bibitem[Liu et~al.(2026{\natexlab{g}})Liu, Chen, Song, Pu, Mo, Bai, Chen, Sun,
  Zhong, Shen, and Lin]{arxiv260716636}
Yang Liu, Weixing Chen, Xinshuai Song, Tao Pu, Siwen Mo, Yongjie Bai, Zihao
  Chen, Qianran Sun, Liruo Zhong, Ying Shen, and Liang Lin.
\newblock {PhyAgentOS: A Self-Evolving Operating System for Embodied Agents
  with Decoupled Cognitive Planning and Physical Execution},
  2026{\natexlab{g}}.
\newblock URL \url{https://arxiv.org/abs/2607.16636}.

\bibitem[Liu et~al.(2026{\natexlab{h}})Liu, Wang, and Ye]{arxiv260115164}
Yaru Liu, {Ao-bo} Wang, and Nanyang Ye.
\newblock {V-CAGE: Context-Aware Generation and Verification for Scalable
  Long-Horizon Embodied Tasks}, 2026{\natexlab{h}}.
\newblock URL \url{https://arxiv.org/abs/2601.15164}.

\bibitem[Liu et~al.(2026{\natexlab{i}})Liu, Mao, Hsu, Kaelbling, and
  Lozano-Pérez]{arxiv260930249}
Yuyao Liu, Jiayuan Mao, David Hsu, Leslie~Pack Kaelbling, and Tomás
  Lozano-Pérez.
\newblock {RAPID: Robot Agentic Programming from Demonstrations},
  2026{\natexlab{i}}.
\newblock URL \url{https://arxiv.org/abs/2609.30249}.

\bibitem[Liu et~al.(2023{\natexlab{d}})Liu, Bahety, and Song]{arxiv230615724}
Zeyi Liu, Arpit Bahety, and Shuran Song.
\newblock {REFLECT: Summarizing Robot Experiences for Failure Explanation and
  Correction}.
\newblock In \emph{Conference on Robot Learning (CoRL)}, 2023{\natexlab{d}}.
\newblock URL \url{https://proceedings.mlr.press/v229/liu23g.html}.

\bibitem[Liu et~al.(2026{\natexlab{j}})Liu, Zeng, Wang, Peng, Liu, Liu, Liu,
  Wang, and Wang]{arxiv241014209}
Zihan Liu, Ruinan Zeng, Dongxia Wang, Gengyun Peng, Xiaoxia Liu, Qiang Liu,
  Peiyu Liu, Wenhai Wang, and Jingyi Wang.
\newblock {Agents4PLC: Automating Closed-loop PLC Code Generation and
  Verification in Industrial Control Systems using LLM-based Agents}.
\newblock \emph{IEEE Transactions on Software Engineering}, 52\penalty0
  (5):\penalty0 1672--1687, 2026{\natexlab{j}}.
\newblock \doi{10.1109/tse.2026.3667895}.

\bibitem[Lu et~al.(2026)Lu, Wu, Kou, Lin, Fu, Xiao, Mandlekar, Xu, Shi,
  Goldberg, Chen, Chowdhury, Kautz, Zhu, Fan, and Wang]{arxiv260700272}
Runyu Lu, Yubo Wu, Ethan Kou, Haotian Lin, Letian Fu, Wenli Xiao, Ajay
  Mandlekar, Yinzhen Xu, Guanya Shi, Ken Goldberg, Ang Chen, Mosharaf
  Chowdhury, Jan Kautz, Yuke Zhu, Linxi~"Jim" Fan, and Guanzhi Wang.
\newblock {ASPIRE: Agentic /Skills Discovery for Robotics}, 2026.
\newblock URL \url{https://arxiv.org/abs/2607.00272}.

\bibitem[Lu et~al.(2024)Lu, Huang, Li, Zhang, Ji, and Xu]{arxiv241216633}
Xuancun Lu, Zhengxian Huang, Xinfeng Li, Chi Zhang, Xiaoyu Ji, and Wenyuan Xu.
\newblock {Easier Said Than Done: Unpacking Intent-Behavior Gap in Jailbreaking
  LLM-Based Robots}, 2024.
\newblock URL \url{https://arxiv.org/abs/2412.16633}.

\bibitem[Luo et~al.(2026{\natexlab{a}})Luo, Xu, Wang, Zhou, and
  Zhang]{arxiv261002527}
Jiaxuan Luo, Xingguo Xu, Shanshan Wang, Yuhan Zhou, and Zhen Zhang.
\newblock {CriticHack: Evaluating Visual Rewards Under Robot Policy
  Optimization}, 2026{\natexlab{a}}.
\newblock URL \url{https://arxiv.org/abs/2610.02527}.

\bibitem[Luo et~al.(2026{\natexlab{b}})Luo, Tian, Cao, Luo, Lin, Li, Kong,
  Yang, and Ma]{luo-etal-2026-storage}
Jinghao Luo, Yuchen Tian, Chuxue Cao, Ziyang Luo, Hongzhan Lin, Kaixin Li,
  Chuyi Kong, Ruichao Yang, and Jing Ma.
\newblock From storage to experience: A survey on the evolution of {LLM} agent
  memory mechanisms.
\newblock In Maria Liakata, Viviane~P. Moreira, Jiajun Zhang, and David
  Jurgens, editors, \emph{Findings of the {A}ssociation for {C}omputational
  {L}inguistics: {ACL} 2026}, pages 41622--41652, San Diego, California, United
  States, July 2026{\natexlab{b}}. Association for Computational Linguistics.
\newblock ISBN 979-8-89176-395-1.
\newblock \doi{10.18653/v1/2026.findings-acl.2069}.
\newblock URL \url{https://aclanthology.org/2026.findings-acl.2069/}.

\bibitem[Luo et~al.(2026{\natexlab{c}})Luo, Bai, Xu, Li, Li, Qian, Lu, Zhang,
  Zou, Luo, Chen, and Lin]{arxiv261029677}
Meng Luo, Zechen Bai, Shanqing Xu, Yanlin Li, Kaixin Li, Mengchen Qian, Weifeng
  Lu, Ran Zhang, Yingtian Zou, Ziyang Luo, Wei Chen, and Hongzhan Lin.
\newblock {Toward Recursive Self-Improvement in Physical AI},
  2026{\natexlab{c}}.
\newblock URL \url{https://arxiv.org/abs/2610.29677}.

\bibitem[Luo et~al.(2026{\natexlab{d}})Luo, Li, Li, Lin, Zhou, Ju, Zhang, Jin,
  Lee, and Hsu]{arxiv260916679}
Meng Luo, Yanlin Li, Hao Li, Hongzhan Lin, Pengfei Zhou, Tianjie Ju, Ran Zhang,
  Yeying Jin, Mong-Li Lee, and Wynne Hsu.
\newblock {AI for Games in the Foundation Model Era}, 2026{\natexlab{d}}.
\newblock URL \url{https://arxiv.org/abs/2609.16679}.

\bibitem[Luo et~al.(2026{\natexlab{e}})Luo, Liu, Wang, Zhang, Tao, Wan, Gai,
  and Fei]{arxiv260911242}
Meng Luo, Yicheng Liu, Jiahao Wang, Yuanxing Zhang, Xin Tao, Pengfei Wan, Kun
  Gai, and Hao Fei.
\newblock {From Evaluation to Enhancement: Benchmarking and Improving
  Think-with-Video Reasoning for Video Generative Models}.
\newblock In \emph{European Conference on Computer Vision (ECCV)},
  2026{\natexlab{e}}.

\bibitem[Lyu et~al.(2026)Lyu, Jiang, Li, Lu, and Zhang]{arxiv260933807}
Yiheng Lyu, Xueying Jiang, Wenhao Li, Shijian Lu, and Gongjie Zhang.
\newblock {CodeActionBench: Evaluating Agentic Code-as-Policy for Embodied
  Manipulation}, 2026.
\newblock URL \url{https://arxiv.org/abs/2609.33807}.

\bibitem[Ma et~al.(2026{\natexlab{a}})Ma, Gao, Qiang, Dai, and
  Li]{arxiv260934210}
Haitong Ma, Chenxiao Gao, Rushi Qiang, Bo~Dai, and Na~Li.
\newblock {RLE-Bench: A Qualifying Exam for Coding Agents as Robot Learning
  Engineers}, 2026{\natexlab{a}}.
\newblock URL \url{https://arxiv.org/abs/2609.34210}.

\bibitem[Ma and Wu(2026)]{arxiv260910873}
Qinzhen Ma and Ruihai Wu.
\newblock {When Validation Stops Learning: Auditing Update Admission for
  Continual Embodied Agents}, 2026.
\newblock URL \url{https://arxiv.org/abs/2609.10873}.

\bibitem[Ma et~al.(2024{\natexlab{a}})Ma, Liang, Wang, Zhu, Fan, Bastani, and
  Jayaraman]{arxiv240601967}
Yecheng Ma, William Liang, Hung-Ju Wang, Yuke Zhu, Linxi Fan, Osbert Bastani,
  and Dinesh Jayaraman.
\newblock {DrEureka: Language Model Guided Sim-To-Real Transfer}.
\newblock In \emph{Robotics: Science and Systems XX}, 2024{\natexlab{a}}.
\newblock \doi{10.15607/rss.2024.xx.094}.

\bibitem[Ma et~al.(2024{\natexlab{b}})Ma, Liang, Wang, Huang, Bastani,
  Jayaraman, Zhu, Fan, and Anandkumar]{arxiv231012931}
Yecheng~Jason Ma, William Liang, Guanzhi Wang, De-An Huang, Osbert Bastani,
  Dinesh Jayaraman, Yuke Zhu, Linxi Fan, and Anima Anandkumar.
\newblock {Eureka: Human-Level Reward Design via Coding Large Language Models}.
\newblock In \emph{International Conference on Learning Representations
  (ICLR)}, 2024{\natexlab{b}}.
\newblock URL \url{https://openreview.net/forum?id=IEduRUO55F}.

\bibitem[Ma et~al.(2025)Ma, Hejna, Fu, Shah, Liang, Xu, Kirmani, Xu, Driess,
  Xiao, Bastani, Jayaraman, Yu, Zhang, Sadigh, and Xia]{arxiv241104549}
Yecheng~Jason Ma, Joey Hejna, Chuyuan Fu, Dhruv Shah, Jacky Liang, Zhuo Xu,
  Sean Kirmani, Peng Xu, Danny Driess, Ted Xiao, Osbert Bastani, Dinesh
  Jayaraman, Wenhao Yu, Tingnan Zhang, Dorsa Sadigh, and Fei Xia.
\newblock {Vision Language Models are In-Context Value Learners}.
\newblock In \emph{International Conference on Learning Representations
  (ICLR)}, 2025.
\newblock URL \url{https://openreview.net/forum?id=friHAl5ofG}.

\bibitem[Ma et~al.(2026{\natexlab{b}})Ma, Song, Zhuang, Hao, and
  King]{arxiv240514093}
Yueen Ma, Zixing Song, Yuzheng Zhuang, Jianye Hao, and Irwin King.
\newblock {A Survey on Vision-Language-Action Models for Embodied AI}.
\newblock \emph{IEEE Transactions on Neural Networks and Learning Systems},
  37\penalty0 (7):\penalty0 3031--3051, 2026{\natexlab{b}}.
\newblock \doi{10.1109/tnnls.2025.3650584}.

\bibitem[Ma et~al.(2024{\natexlab{c}})Ma, Cui, Cao, Ye, Liu, Lu, Abdelraouf,
  Gupta, Han, Bera, Rehg, and Wang]{arxiv231204372}
Yunsheng Ma, Can Cui, Xu~Cao, Wenqian Ye, Peiran Liu, Juanwu Lu, Amr
  Abdelraouf, Rohit Gupta, Kyungtae Han, Aniket Bera, James~M. Rehg, and Ziran
  Wang.
\newblock {LaMPilot: An Open Benchmark Dataset for Autonomous Driving with
  Language Model Programs}.
\newblock In \emph{2024 IEEE/CVF Conference on Computer Vision and Pattern
  Recognition (CVPR)}, pages 15141--15151, 2024{\natexlab{c}}.
\newblock \doi{10.1109/cvpr52733.2024.01434}.

\bibitem[Mandi et~al.(2025)Mandi, Weng, Bauer, and Song]{arxiv240608474}
Zhao Mandi, Yijia Weng, Dominik Bauer, and Shuran Song.
\newblock {Real2Code: Reconstruct Articulated Objects via Code Generation}.
\newblock In \emph{International Conference on Learning Representations
  (ICLR)}, 2025.
\newblock URL \url{https://openreview.net/forum?id=CAssIgPN4I}.

\bibitem[Mendez-Mendez(2025)]{arxiv251000182}
Jorge Mendez-Mendez.
\newblock {A Systematic Study of Large Language Models for Task and Motion
  Planning With PDDLStream}, 2025.
\newblock URL \url{https://arxiv.org/abs/2510.00182}.

\bibitem[Meng et~al.(2026)Meng, Sun, Fest, Yao, Li, Bing, and
  Knoll]{arxiv250918597}
Yuan Meng, Zhenguo Sun, Max Fest, Xiangtong Yao, Xukun Li, Zhenshan Bing, and
  Alois Knoll.
\newblock {Robotic Ultra-Long-Horizon Manipulation Skills via Human-guided
  Lifelong Code Generation}.
\newblock \emph{IEEE Robotics and Automation Letters}, 11\penalty0
  (12):\penalty0 13845--13852, 2026.
\newblock \doi{10.1109/lra.2026.3744999}.

\bibitem[Merler et~al.(2026)Merler, Li, Roy, Liang, Wang, Huang, and
  Silver]{arxiv260930233}
Matteo Merler, Bowen Li, Josh Roy, Yichao Liang, Qianwei Wang, Yixuan Huang,
  and Tom Silver.
\newblock {Coding Agents for Generalized Task and Motion Planning Problems},
  2026.
\newblock URL \url{https://arxiv.org/abs/2609.30233}.

\bibitem[Mower et~al.(2026)Mower, Wan, Yu, Grosnit, Gonzalez-Billandon, Zimmer,
  Liu, Palenicek, Tateo, Peters, Qu, Zhang, Lan, Cramariuc, Cadena, Hutter,
  Tian, Zhuang, Shao, Quan, Hao, Wang, and Bou-Ammar]{arxiv240619741}
Christopher~E. Mower, Yuhui Wan, Hongzhan Yu, Antoine Grosnit, Jonas
  Gonzalez-Billandon, Matthieu Zimmer, Puze Liu, Daniel Palenicek, Davide
  Tateo, Jan Peters, Kaixian Qu, Mike Zhang, Guowei Lan, Andrei Cramariuc,
  Cesar Cadena, Marco Hutter, Guangjian Tian, Yuzhen Zhuang, Kun Shao, Xingyue
  Quan, Jianye Hao, Jun Wang, and Haitham Bou-Ammar.
\newblock {A robot operating system framework for using large language models
  in embodied AI}.
\newblock \emph{Nature Machine Intelligence}, 8\penalty0 (3):\penalty0
  313--325, 2026.
\newblock \doi{10.1038/s42256-026-01186-z}.

\bibitem[Mu et~al.(2024)Mu, Chen, Zhang, Chen, Yu, Ge, Chen, Liang, Hu, Tao,
  Sun, Yu, Yang, Shao, Wang, Dai, Qiao, Ding, and Luo]{arxiv240216117}
Yao Mu, Junting Chen, Qinglong Zhang, Shoufa Chen, Qiaojun Yu, Chongjian Ge,
  Runjian Chen, Zhixuan Liang, Mengkang Hu, Chaofan Tao, Peize Sun, Haibao Yu,
  Chao Yang, Wenqi Shao, Wenhai Wang, Jifeng Dai, Yu~Qiao, Mingyu Ding, and
  Ping Luo.
\newblock {RoboCodeX: Multimodal Code Generation for Robotic Behavior
  Synthesis}.
\newblock In \emph{International Conference on Machine Learning (ICML)}, 2024.
\newblock URL \url{https://openreview.net/forum?id=xnQ1qoly7Q}.

\bibitem[Nagaraja et~al.(2026)Nagaraja, Bagari, and Bahsi]{arxiv260800747}
Neha Nagaraja, Amisha Bagari, and Hayretdin Bahsi.
\newblock {When Prompts Control Robots: Prompt Injection Attacks in Multi-Agent
  Robotic Systems}, 2026.
\newblock URL \url{https://arxiv.org/abs/2608.00747}.

\bibitem[Naouali et~al.(2026)Naouali, Wu, Wong, Puthran, and
  Younis]{arxiv260816978}
Dhia Naouali, Minghan Wu, Claudia Wong, Abhinav Puthran, and Omar~G. Younis.
\newblock {VLCP: Vision Language Control Policy Closed-Loop Code Replanning for
  Robot Manipulation}, 2026.
\newblock URL \url{https://arxiv.org/abs/2608.16978}.

\bibitem[Navasardyan et~al.(2026)Navasardyan, Danielyan, and
  Davtyan]{arxiv260903611}
Zaruhi Navasardyan, Tatul Danielyan, and Hrant Davtyan.
\newblock {FailBench: How Reliable are VLMs at Judging Robot Task Success?},
  2026.
\newblock URL \url{https://arxiv.org/abs/2609.03611}.

\bibitem[Nayak et~al.(2026)Nayak, Parameswaran, Peri, and
  Ramanan]{arxiv260924145}
Lucky~Kant Nayak, Narayanan~Palghat Parameswaran, Neehar Peri, and Deva
  Ramanan.
\newblock {MimicAgent: Quadruped Skills via Prompt-to-Trajectory Generation},
  2026.
\newblock URL \url{https://arxiv.org/abs/2609.24145}.

\bibitem[Ng et~al.(2026)Ng, Zhang, Wu, Lui, Zhao, Gao, and Ren]{arxiv261013376}
Chi~Kit Ng, Yidong Zhang, Tianchun Wu, Siu~Hing Lui, Hanlin Zhao, Huxin Gao,
  and Hongliang Ren.
\newblock {EndoRSI: Physical-Evidence-Driven Recursive Improvement of
  Endoscopic Robot Policies}, 2026.
\newblock URL \url{https://arxiv.org/abs/2610.13376}.

\bibitem[Nie et~al.(2026)Nie, Liu, and Wang]{arxiv260817209}
Chang Nie, Zhe Liu, and Hesheng Wang.
\newblock {Teach and Grow: An Agent-Centered Architecture for General Robot
  Learning}, 2026.
\newblock URL \url{https://arxiv.org/abs/2608.17209}.

\bibitem[Ning et~al.(2026)Ning, Tieu, Fu, Wei, Li, Bei, Zou, Ai, Liu, Li, Chen,
  Zhao, Yang, Li, Qian, Li, Lin, Zeng, Qiu, Chen, Sun, Yang, Wang, Pan, Yang,
  Zhang, Fang, Cui, Cao, Chen, Sun, Chen, Srinivasan, Mathur, Xia, Li, Yan, Lu,
  Zhang, Zhang, Tong, and He]{arxiv260518747}
Xuying Ning, Katherine Tieu, Dongqi Fu, Tianxin Wei, Zihao Li, Yuanchen Bei,
  Jiaru Zou, Mengting Ai, Zhining Liu, Ting-Wei Li, Lingjie Chen, Yanjun Zhao,
  Ke~Yang, Bingxuan Li, Cheng Qian, Gaotang Li, Xiao Lin, Zhichen Zeng,
  Ruizhong Qiu, Sirui Chen, Yifan Sun, Xiyuan Yang, Ruida Wang, Rui Pan,
  Chenyuan Yang, Dylan Zhang, Liri Fang, Zikun Cui, Yang Cao, Pan Chen, Dorothy
  Sun, Ren Chen, Mahesh Srinivasan, Nipun Mathur, Yinglong Xia, Hong Li, Hong
  Yan, Pan Lu, Lingming Zhang, Tong Zhang, Hanghang Tong, and Jingrui He.
\newblock {Code as Agent Harness}, 2026.
\newblock URL \url{https://arxiv.org/abs/2605.18747}.

\bibitem[Obi et~al.(2026)Obi, Venkatesh, Wang, Wang, Suh, Amosa, Jo, and
  Min]{arxiv260405427}
Ike Obi, Vishnunandan L.~N. Venkatesh, Weizheng Wang, Ruiqi Wang, Dayoon Suh,
  Temitope~I. Amosa, Wonse Jo, and Byung-Cheol Min.
\newblock {Pre-Execution Safety Gate \& Task Safety Contracts for
  LLM-Controlled Robot Systems}, 2026.
\newblock URL \url{https://arxiv.org/abs/2604.05427}.

\bibitem[Ouyang et~al.(2025)Ouyang, Li, Li, Li, Yu, Sreenath, and
  Wu]{arxiv240405291}
Yutao Ouyang, Jinhan Li, Yunfei Li, Zhongyu Li, Chao Yu, Koushil Sreenath, and
  Yi~Wu.
\newblock {Long-horizon Locomotion and Manipulation on a Quadrupedal Robot with
  Large Language Models}.
\newblock In \emph{2025 IEEE/RSJ International Conference on Intelligent Robots
  and Systems (IROS)}, pages 11157--11164, 2025.
\newblock \doi{10.1109/iros60139.2025.11246632}.

\bibitem[Qi et~al.(2025)Qi, Zhang, Ding, Dong, Yu, Li, Xu, Li, He, Fan, Zhang,
  He, Gu, Jin, Ma, Zhang, Wang, and Yi]{arxiv250213143}
Zekun Qi, Wenyao Zhang, Yufei Ding, Runpei Dong, Xinqiang Yu, Jingwen Li,
  Lingyun Xu, Baoyu Li, Xialin He, Guofan Fan, Jiazhao Zhang, Jiawei He,
  Jiayuan Gu, Xin Jin, Kaisheng Ma, Zhizheng Zhang, He~Wang, and Li~Yi.
\newblock {SoFar: Language-Grounded Orientation Bridges Spatial Reasoning and
  Object Manipulation}.
\newblock In \emph{Advances in Neural Information Processing Systems 38}, pages
  85008--85053, 2025.
\newblock \doi{10.52202/085713-2565}.

\bibitem[Qin et~al.(2026)Qin, Wu, Schommer, Amini-Naieni, and
  Gongora]{arxiv261111690}
Tony Qin, Ziheng Wu, Maximilian Schommer, Roya Amini-Naieni, and Aldair
  Gongora.
\newblock {A Physical Testbed for Self-Driving Laboratories: Comparing Bayesian
  and Agentic Optimization in a Minimal Closed-Loop Color Formulation Task},
  2026.
\newblock URL \url{https://arxiv.org/abs/2611.11690}.

\bibitem[Qiu and Cygan(2026)]{arxiv251025850}
Kevin Qiu and Marek Cygan.
\newblock {Debate2Create: Robot Co-design via Multi-Agent LLM Debate}.
\newblock In \emph{International Conference on Machine Learning (ICML)}, 2026.
\newblock URL \url{https://openreview.net/forum?id=1ufVo73uzD}.

\bibitem[Qiu et~al.(2026)Qiu, Pałucki, Ciebiera, Fijałkowski, Cygan, and
  Łukasz Kuciński]{arxiv240708626}
Kevin Qiu, Władysław Pałucki, Krzysztof Ciebiera, Paweł Fijałkowski, Marek
  Cygan, and Łukasz Kuciński.
\newblock {RoboMorph: Evolving Robot Morphology using Large Language Models}.
\newblock In \emph{2026 IEEE International Conference on Robotics and
  Automation (ICRA)}, pages 17611--17618, 2026.
\newblock \doi{10.1109/icra57385.2026.11696020}.

\bibitem[Qu et~al.(2026)Qu, Chen, and Zhang]{arxiv261014018}
Bince Qu, Wei Chen, and Bo~Zhang.
\newblock {RoboHarness: A Simple Harness Outperforms VLA and World Action
  Models}, 2026.
\newblock URL \url{https://arxiv.org/abs/2610.14018}.

\bibitem[Ramani et~al.(2025)Ramani, Tawosi, Alamir, and
  Borrajo]{arxiv251003469}
Keshav Ramani, Vali Tawosi, Salwa Alamir, and Daniel Borrajo.
\newblock {Bridging LLM Planning Agents and Formal Methods: A Case Study in
  Plan Verification}, 2025.
\newblock URL \url{https://arxiv.org/abs/2510.03469}.

\bibitem[Ravichandran et~al.(2026)Ravichandran, Robey, Kumar, Pappas, and
  Hassani]{arxiv250307885}
Zachary Ravichandran, Alexander Robey, Vijay Kumar, George~J. Pappas, and Hamed
  Hassani.
\newblock {Safety Guardrails for LLM-Enabled Robots}.
\newblock \emph{IEEE Robotics and Automation Letters}, 11\penalty0
  (4):\penalty0 4649--4656, 2026.
\newblock \doi{10.1109/lra.2026.3667488}.

\bibitem[Robey et~al.(2025)Robey, Ravichandran, Kumar, Hassani, and
  Pappas]{arxiv241013691}
Alexander Robey, Zachary Ravichandran, Vijay Kumar, Hamed Hassani, and
  George~J. Pappas.
\newblock {Jailbreaking LLM-Controlled Robots}.
\newblock In \emph{2025 IEEE International Conference on Robotics and
  Automation (ICRA)}, pages 11948--11956, 2025.
\newblock \doi{10.1109/icra55743.2025.11128119}.

\bibitem[Rui et~al.(2026)Rui, Zuo, Wang, Jia, Ma, Zhou, and Li]{arxiv260706256}
Ke~Rui, Yushen Zuo, Jiawei Wang, Haoran Jia, Jinming Ma, Weitao Zhou, and
  Minglei Li.
\newblock {Diagnosing Semantic Handoff Failures in Agent-Orchestrated
  Vision-Language-Action Skill Composition}, 2026.
\newblock URL \url{https://arxiv.org/abs/2607.06256}.

\bibitem[Ryu et~al.(2025)Ryu, Liao, Li, Delgosha, Sreenath, and
  Mehr]{arxiv240918382}
Kanghyun Ryu, Qiayuan Liao, Zhongyu Li, Payam Delgosha, Koushil Sreenath, and
  Negar Mehr.
\newblock {CurricuLLM: Automatic Task Curricula Design for Learning Complex
  Robot Skills using Large Language Models}.
\newblock In \emph{2025 IEEE International Conference on Robotics and
  Automation (ICRA)}, pages 4470--4477, 2025.
\newblock \doi{10.1109/icra55743.2025.11128783}.

\bibitem[Salfity et~al.(2026)Salfity, Anderson, and Pryor]{arxiv260712220}
Jonathan Salfity, Robert~Blake Anderson, and Mitch Pryor.
\newblock {Contract-Grounded Behavior Tree Synthesis via Coding Agents}, 2026.
\newblock URL \url{https://arxiv.org/abs/2607.12220}.

\bibitem[Samarakoon et~al.(2026)Samarakoon, Muthugala, Sachinthana, and
  Elara]{arxiv260805715}
S.~M . Bhagya~P. Samarakoon, M.~A. Viraj~J. Muthugala, W.~K.~R. Sachinthana,
  and Mohan~Rajesh Elara.
\newblock {Hijacking Robots with a Piece of Paper: A Systematic Study of
  Physical Prompt Injection in VLM-Controlled Robots}, 2026.
\newblock URL \url{https://arxiv.org/abs/2608.05715}.

\bibitem[Sarch et~al.(2024)Sarch, Jang, Tarr, Cohen, Marino, and
  Fragkiadaki]{arxiv240614596}
Gabriel Sarch, Lawrence Jang, Michael~J. Tarr, William~W. Cohen, Kenneth
  Marino, and Katerina Fragkiadaki.
\newblock {VLM Agents Generate Their Own Memories: Distilling Experience into
  Embodied Programs of Thought}.
\newblock In \emph{Advances in Neural Information Processing Systems 37}, pages
  75942--75985, 2024.
\newblock \doi{10.52202/079017-2418}.

\bibitem[Schlesinger et~al.(2024)Schlesinger, Guha, and Biswas]{arxiv241018893}
Claire Schlesinger, Arjun Guha, and Joydeep Biswas.
\newblock {Creating and Repairing Robot Programs in Open-World Domains}, 2024.
\newblock URL \url{https://arxiv.org/abs/2410.18893}.

\bibitem[Schroder et~al.(2026)Schroder, Hwang, Kim, Cheng, Liu, Zheng, He, and
  Mutlu]{arxiv260808884}
Mya Schroder, Yuna Hwang, Callie~Y. Kim, Leqian Cheng, Jeffrey~{Li-cheng} Liu,
  Chenchen Zheng, Xinning He, and Bilge Mutlu.
\newblock {SHRIMP: Iterative Refinement of Robot Task Plans}.
\newblock In \emph{Proceedings of the 39th Annual ACM Symposium on User
  Interface Software and Technology}, pages 1--18, 2026.
\newblock \doi{10.1145/3830398.3830644}.

\bibitem[Setty et~al.(2026)Setty, Ramanathan, Foster, and
  Stevens]{arxiv260715620}
Priyanka~V. Setty, Arvind Ramanathan, Ian Foster, and Rick Stevens.
\newblock {AEGIS: Assay-Aware Protocol Validation and Runtime Monitoring for
  Open-Source Liquid Handling Robots}, 2026.
\newblock URL \url{https://arxiv.org/abs/2607.15620}.

\bibitem[Shang et~al.(2026)Shang, Li, Ma, Su, Jin, Wang, Jin, Zhang, Tang, Su,
  Gao, Wu, Liu, Shah, Zhang, Chen, Zhu, Tian, Chua, Wang, Zhu, and
  Li]{arxiv260208971}
Yu~Shang, Zhuohang Li, Yiding Ma, Weikang Su, Xin Jin, Ziyou Wang, Lei Jin, Xin
  Zhang, Yinzhou Tang, Haisheng Su, Chen Gao, Wei Wu, Xihui Liu, Dhruv Shah,
  Zhaoxiang Zhang, Zhibo Chen, Jun Zhu, Yonghong Tian, Tat-Seng Chua, Xin Wang,
  Wenwu Zhu, and Yong Li.
\newblock {WorldArena: A Unified Benchmark for Evaluating Perception and
  Functional Utility of Embodied World Models}.
\newblock \emph{AI Open}, 7:\penalty0 208--226, 2026.
\newblock \doi{10.1016/j.aiopen.2026.07.002}.

\bibitem[Sharma et~al.(2026)Sharma, Hu, Kamat, Sadigh, and
  Bohg]{arxiv261108209}
Satvik Sharma, Hengyuan Hu, Mihika Kamat, Dorsa Sadigh, and Jeannette Bohg.
\newblock {RoboARCade: Benchmarking Agentic Robot Control}, 2026.
\newblock URL \url{https://arxiv.org/abs/2611.08209}.

\bibitem[Shen et~al.(2026)Shen, You, Liu, Zheng, Zha, Yamazaki, Zhang, Huang,
  Sun, Chen, He, Liu, Chang, Fragkiadaki, Shah, Schwager, Henderson, Abraham,
  and Xu]{arxiv260927308}
Zeyu Shen, Haoxiang You, Yilang Liu, Zhicheng Zheng, Lihan Zha, Kashu Yamazaki,
  Mingtong Zhang, Suning Huang, Jiankai Sun, Qianzhong Chen, Lucy He, Kaiyuan
  Liu, Haoran Chang, Katerina Fragkiadaki, Dhruv Shah, Mac Schwager, Peter
  Henderson, Ian Abraham, and Canwen Xu.
\newblock {EmbodiedSWE: Coding Agents for Long Horizon Dexterous Robotics},
  2026.
\newblock URL \url{https://arxiv.org/abs/2609.27308}.

\bibitem[Shirai et~al.(2024)Shirai, Beltran-Hernandez, Hamaya, Hashimoto,
  Tanaka, Kawaharazuka, Tanaka, Ushiku, and Mori]{arxiv231100967}
Keisuke Shirai, Cristian~C. Beltran-Hernandez, Masashi Hamaya, Atsushi
  Hashimoto, Shohei Tanaka, Kento Kawaharazuka, Kazutoshi Tanaka, Yoshitaka
  Ushiku, and Shinsuke Mori.
\newblock {Vision-Language Interpreter for Robot Task Planning}.
\newblock In \emph{2024 IEEE International Conference on Robotics and
  Automation (ICRA)}, pages 2051--2058, 2024.
\newblock \doi{10.1109/icra57147.2024.10611112}.

\bibitem[Si et~al.(2026)Si, Bi, Yang, Ni, Huang, Wang, Jiang, Zhang, Ruan,
  Zhang, Wang, Li, Feng, Dong, and Zhou]{arxiv261001939}
Ruiyang Si, Jianxin Bi, Shunyu Yang, Rui Ni, Wenbo Huang, Qiang Wang, Shulong
  Jiang, Yintianrun Zhang, Feixiang Ruan, Kewei Zhang, Duomin Wang, Xiuyu Li,
  Haiwen Feng, Zhen Dong, and Daquan Zhou.
\newblock {Fewer Tokens, Better Action: GPT-6 Astra Robot Agents with 14\%
  Higher Success Rate but 65\% Fewer Tokens}, 2026.
\newblock URL \url{https://arxiv.org/abs/2610.01939}.

\bibitem[Silva and Burke(2026)]{arxiv260115486}
Javier Noé~Ramos Silva and Peter~J. Burke.
\newblock {An LLM-Agnostic, MAVLink-Based Drone Command and Control Interface
  and Agentic Harness Using the Model Context Protocol}, 2026.
\newblock URL \url{https://arxiv.org/abs/2601.15486}.

\bibitem[Singh et~al.(2023)Singh, Blukis, Mousavian, Goyal, Xu, Tremblay, Fox,
  Thomason, and Garg]{arxiv220911302}
Ishika Singh, Valts Blukis, Arsalan Mousavian, Ankit Goyal, Danfei Xu, Jonathan
  Tremblay, Dieter Fox, Jesse Thomason, and Animesh Garg.
\newblock {ProgPrompt: Generating Situated Robot Task Plans using Large
  Language Models}.
\newblock In \emph{2023 IEEE International Conference on Robotics and
  Automation (ICRA)}, pages 11523--11530, 2023.
\newblock \doi{10.1109/icra48891.2023.10161317}.

\bibitem[Skreta et~al.(2023)Skreta, Yoshikawa, Arellano-Rubach, Ji, Kristensen,
  Darvish, Aspuru-Guzik, Shkurti, and Garg]{arxiv230314100}
Marta Skreta, Naruki Yoshikawa, Sebastian Arellano-Rubach, Zhi Ji, Lasse~Bjørn
  Kristensen, Kourosh Darvish, Alán Aspuru-Guzik, Florian Shkurti, and Animesh
  Garg.
\newblock {Errors are Useful Prompts: Instruction Guided Task Programming with
  Verifier-Assisted Iterative Prompting}, 2023.
\newblock URL \url{https://arxiv.org/abs/2303.14100}.

\bibitem[Snyder et~al.(2025)Snyder, Hancock, Badithela, Dixon, Miller, Ambrus,
  Majumdar, Itkina, and Nishimura]{arxiv250310966}
David Snyder, Asher~James Hancock, Apurva Badithela, Emma Dixon, Patrick
  Miller, Rares~Andrei Ambrus, Anirudha Majumdar, Masha Itkina, and Haruki
  Nishimura.
\newblock {Is Your Imitation Learning Policy Better than Mine? Policy
  Comparison with Near-Optimal Stopping}.
\newblock In \emph{Robotics: Science and Systems XXI}, 2025.
\newblock \doi{10.15607/rss.2025.xxi.077}.

\bibitem[Snyder et~al.(2026)Snyder, Badithela, Matni, Pappas, Majumdar, Itkina,
  and Nishimura]{arxiv260313616}
David Snyder, Apurva Badithela, Nikolai Matni, George Pappas, Anirudha
  Majumdar, Masha Itkina, and Haruki Nishimura.
\newblock {Beyond Binary Success: Sample-Efficient and Statistically Rigorous
  Robot Policy Comparison}.
\newblock In \emph{Robotics: Science and Systems XXII}, 2026.
\newblock \doi{10.15607/rss.2026.xxii.076}.

\bibitem[Song et~al.(2023)Song, Zhou, Liu, Fang, Shu, and Ma]{arxiv230906687}
Jiayang Song, Zhehua Zhou, Jiawei Liu, Chunrong Fang, Zhan Shu, and Lei Ma.
\newblock {Self-Refined Large Language Model as Automated Reward Function
  Designer for Deep Reinforcement Learning in Robotics}, 2023.
\newblock URL \url{https://arxiv.org/abs/2309.06687}.

\bibitem[Song et~al.(2026)Song, Liang, Fu, Wang, Yang, and Liu]{arxiv261010498}
Python Song, Zhixuan Liang, Kelsey Fu, Mengdi Wang, Junfeng Yang, and Shilong
  Liu.
\newblock {EmbodiedRSI: Active Continual Robot Learning Through
  Hypothesis-Guided Co-Evolution}, 2026.
\newblock URL \url{https://arxiv.org/abs/2610.10498}.

\bibitem[Sun et~al.(2026{\natexlab{a}})Sun, Leet, Machcha, Menon, Chan, Cui,
  and Chooi]{arxiv261014043}
Edward Sun, Christopher Leet, Sravanthi Machcha, Achyutha Menon, Tzu~Kit Chan,
  Yuchen Cui, and Jay Chooi.
\newblock {Refuses in Chat, Complies with Arms: LLM Refusal Does Not Survive
  Robot Control}, 2026{\natexlab{a}}.
\newblock URL \url{https://arxiv.org/abs/2610.14043}.

\bibitem[Sun et~al.(2025)Sun, Liu, Lyu, Yang, Zhang, and Li]{arxiv241014660}
Shengjie Sun, Runze Liu, Jiafei Lyu, Jing-Wen Yang, Liangpeng Zhang, and Xiu
  Li.
\newblock {A Large Language Model-Driven Reward Design Framework via Dynamic
  Feedback for Reinforcement Learning}.
\newblock \emph{Knowledge-Based Systems}, 326:\penalty0 114065, 2025.
\newblock \doi{10.1016/j.knosys.2025.114065}.

\bibitem[Sun and Zhang(2026)]{arxiv261111438}
Yiming Sun and Mi~Zhang.
\newblock {When Coding Agents Get Arms: Safety Risks Across the
  Digital-Physical Boundary}, 2026.
\newblock URL \url{https://arxiv.org/abs/2611.11438}.

\bibitem[Sun et~al.(2026{\natexlab{b}})Sun, Li, Li, Liu, Wang, and
  Chen]{arxiv261110510}
Zheng Sun, Zhihao Li, Zhuo Li, Wenrui Liu, Xin Wang, and Fei Chen.
\newblock {ReSkill: Selective Self-Improvement for Robot Manipulation},
  2026{\natexlab{b}}.
\newblock URL \url{https://arxiv.org/abs/2611.10510}.

\bibitem[Sygkounas et~al.(2026)Sygkounas, Aregbede, Loutfi, and
  Persson]{arxiv260722832}
Alkis Sygkounas, Victor Aregbede, Amy Loutfi, and Andreas Persson.
\newblock {MEMENTO: Memory-Guided Memetic Code-as-Policy Evolution}, 2026.
\newblock URL \url{https://arxiv.org/abs/2607.22832}.

\bibitem[Talwar et~al.(2026)Talwar, Nijs, Verleysen, and
  wyffels]{arxiv260926499}
Raman Talwar, Elias Nijs, Andreas Verleysen, and Francis wyffels.
\newblock {Generalizing Manipulation Skills with a Local Coding Agent}, 2026.
\newblock URL \url{https://arxiv.org/abs/2609.26499}.

\bibitem[Tan et~al.(2026)Tan, Wang, Wang, and Yuan]{arxiv261031841}
Xiao Tan, Yixuan Wang, Zhexi Wang, and Wenzhen Yuan.
\newblock {FORCA: A Force-Aware Coding Agent for Contact-Rich Robot
  Manipulation}, 2026.
\newblock URL \url{https://arxiv.org/abs/2610.31841}.

\bibitem[Tang et~al.(2026{\natexlab{a}})Tang, Zhang, Yang, Yan, Zhou, Mao, Han,
  Cao, Xu, He, Fu, Feng, and Huang]{arxiv261004432}
Jinzhou Tang, Zijun Zhang, Jing Yang, Yuchen Yan, Kun Zhou, Lingjun Mao,
  Ruobing Han, Jinglin Cao, Wenpeng Xu, Lukun He, Minghao Fu, Fan Feng, and
  Biwei Huang.
\newblock {Video2World: Benchmarking Coding Agents for Interactive World
  Modeling from Embodied Videos}, 2026{\natexlab{a}}.
\newblock URL \url{https://arxiv.org/abs/2610.04432}.

\bibitem[Tang et~al.(2026{\natexlab{b}})Tang, Wang, Zhang, Wang, Chen, Zhang,
  Ma, Jiang, Guo, Li, Jiao, Cui, Yin, Liang, Wan, Lin, Zhao, Zhang, Zhou,
  Zhang, and You]{arxiv261036077}
Weidong Tang, Ruiheng Wang, Jingxuan Zhang, Hexin Wang, Haoyuan Chen,
  Zhengfeiyang Zhang, Yixing Ma, Zichen Jiang, Ziyu Guo, Haodong Li, Zhengbo
  Jiao, Zhiqing Cui, Xinxiang Yin, Shuo Liang, Zhenglin Wan, Hongzhan Lin,
  Wangbo Zhao, Kaipeng Zhang, Pengfei Zhou, Odin Zhang, and Yang You.
\newblock {The Trinity: Capability-Driven Recursive Self-Improvement for
  Multimodal Agents}, 2026{\natexlab{b}}.
\newblock URL \url{https://arxiv.org/abs/2610.36077}.

\bibitem[Tang et~al.(2026{\natexlab{c}})Tang, Zhang, Chan, B{\i}y{\i}k, Koenig,
  and Chen]{arxiv261102391}
Yimin Tang, Han Zhang, Shao-Hung Chan, Erdem B{\i}y{\i}k, Sven Koenig, and
  Jingkai Chen.
\newblock {Late Autumn: LLM Agents Specialize a General Multi-Robot Planner for
  One Warehouse}, 2026{\natexlab{c}}.
\newblock URL \url{https://arxiv.org/abs/2611.02391}.

\bibitem[Tang et~al.(2026{\natexlab{d}})Tang, Chen, and Ho]{arxiv241102317}
Yung-Chen Tang, Pin-Yu Chen, and Tsung-Yi Ho.
\newblock {Defining and Evaluating Physical Safety for Large Language Models}.
\newblock \emph{Communications of the ACM}, 69\penalty0 (8):\penalty0 88--95,
  2026{\natexlab{d}}.
\newblock \doi{10.1145/3816260}.

\bibitem[Timperley et~al.(2024)Timperley, van~der Hoorn, Santos, Deshpande, and
  Wąsowski]{arxiv240403629}
Christopher~S. Timperley, Gijs van~der Hoorn, André Santos, Harshavardhan
  Deshpande, and Andrzej Wąsowski.
\newblock {ROBUST: 221 Bugs in the Robot Operating System}.
\newblock \emph{Empirical Software Engineering}, 29\penalty0 (3), 2024.
\newblock \doi{10.1007/s10664-024-10440-0}.

\bibitem[{TRI LBM Team}(2026)]{arxiv250705331}
{TRI LBM Team}.
\newblock {A Careful Examination of Large Behavior Models for Multitask
  Dexterous Manipulation}.
\newblock \emph{Science Robotics}, 11\penalty0 (113):\penalty0 eaea6201, 2026.
\newblock \doi{10.1126/scirobotics.aea6201}.

\bibitem[Tripathi et~al.(2026)Tripathi, Bafghi, and Raissi]{arxiv260300455}
Shivanshu Tripathi, Reza~Akbarian Bafghi, and Maziar Raissi.
\newblock {Test-Driven Agentic Framework for Reliable Robot Controller}, 2026.
\newblock URL \url{https://arxiv.org/abs/2603.00455}.

\bibitem[Tsui et~al.(2026)Tsui, Fang, and Hwu]{arxiv260120334}
Brian~Y. Tsui, Alan~Y. Fang, and Tiffany~J. Hwu.
\newblock {Demonstration-Free Robotic Control via LLM Agents}.
\newblock In \emph{IEEE/RSJ International Conference on Intelligent Robots and
  Systems (IROS)}, 2026.
\newblock URL \url{https://openreview.net/forum?id=nh704PUegV}.

\bibitem[Tsutsumi et~al.(2026)Tsutsumi, Itsuka, Kasahara, Kouno, Akinari,
  Yamauchi, Endo, Abe, Hashimoto, Nagatani, and Kurazume]{arxiv260201041}
Akinosuke Tsutsumi, Tomoya Itsuka, Yuichiro Kasahara, Tomoya Kouno, Kota
  Akinari, Genki Yamauchi, Daisuke Endo, Taro Abe, Takeshi Hashimoto, Keiji
  Nagatani, and Ryo Kurazume.
\newblock {Coordinated Control of Multiple Construction Machines Using
  LLM-Generated Behavior Trees with Flag-Based Synchronization}, 2026.
\newblock URL \url{https://arxiv.org/abs/2602.01041}.

\bibitem[Tu et~al.(2026)Tu, Wang, Xie, Zhou, Zhou, Zhu, Chen, Chen, Zhou, Zhou,
  Yang, Mou, Zhang, and Xu]{arxiv260818565}
Yanlun Tu, Huacan Wang, Jin Xie, Ziyue Zhou, Jie Zhou, Ningyan Zhu, Ge~Chen,
  Wangyi Chen, Tengfei Zhou, Yifan Zhou, Dasheng Yang, Xiaofeng Mou, Hui Zhang,
  and Yi~Xu.
\newblock {SemaPLC: A Project-Grounded, Verification-Gated Agent Harness for
  PLC Code Generation}, 2026.
\newblock URL \url{https://arxiv.org/abs/2608.18565}.

\bibitem[Turcato et~al.(2025)Turcato, Iovino, Synodinos, Libera, Carli, and
  Falco]{arxiv250304280}
Niccolò Turcato, Matteo Iovino, Aris Synodinos, Alberto~Dalla Libera, Ruggero
  Carli, and Pietro Falco.
\newblock {Towards Autonomous Reinforcement Learning for Real-World Robotic
  Manipulation with Large Language Models}.
\newblock \emph{IEEE Robotics and Automation Letters}, 10\penalty0
  (9):\penalty0 8850--8857, 2025.
\newblock \doi{10.1109/lra.2025.3589162}.

\bibitem[Tuzikov(2026)]{arxiv261016369}
Alexey Tuzikov.
\newblock {RoboBugBench: Benchmarking Automated Repair of Real Robot Software
  Under an Audited Behavioral Oracle}, 2026.
\newblock URL \url{https://arxiv.org/abs/2610.16369}.

\bibitem[Tziafas and Kasaei(2024)]{arxiv240618746}
Georgios Tziafas and Hamidreza Kasaei.
\newblock {Lifelong Robot Library Learning: Bootstrapping Composable and
  Generalizable Skills for Embodied Control with Language Models}.
\newblock In \emph{2024 IEEE International Conference on Robotics and
  Automation (ICRA)}, pages 515--522, 2024.
\newblock \doi{10.1109/icra57147.2024.10611448}.

\bibitem[van~der Vleuten et~al.(2025)van~der Vleuten, Flores, Mathur, Rakitin,
  Hopkins, Yager, and Tsai]{arxiv251109964}
Noah van~der Vleuten, Anthony Flores, Shray Mathur, Max Rakitin, Thomas
  Hopkins, Kevin~G. Yager, and Esther H.~R. Tsai.
\newblock {EnvTrace: Simulation-Based Semantic Evaluation of LLM Code via
  Execution Trace Alignment -- Demonstrated at Synchrotron Beamlines}, 2025.
\newblock URL \url{https://arxiv.org/abs/2511.09964}.

\bibitem[Vemprala et~al.(2024)Vemprala, Bonatti, Bucker, and
  Kapoor]{arxiv230617582}
Sai Vemprala, Rogerio Bonatti, Arthur Bucker, and Ashish Kapoor.
\newblock {ChatGPT for Robotics: Design Principles and Model Abilities}.
\newblock \emph{IEEE Access}, 12:\penalty0 55682--55696, 2024.
\newblock \doi{10.1109/access.2024.3387941}.

\bibitem[Verma et~al.(2026)Verma, Cilio, Christodoulou, Fermelis, Ng, Chung,
  and Liang]{arxiv261000731}
Sanya Verma, Luca Cilio, Velissarios Christodoulou, Tyler Fermelis, Qi~Ting Ng,
  Rachel Chung, and Sofia Liang.
\newblock {Measuring Asset and Scene Reconstruction Effects in Real-to-Sim
  Robot Evaluation}, 2026.
\newblock URL \url{https://arxiv.org/abs/2610.00731}.

\bibitem[Vo et~al.(2026)Vo, Tran, Hanyu, Ikebe, Nguyen, Bui, Vu, Gunderman,
  Rainwater, Nguyen, and Le]{arxiv260422238}
Khoa Vo, Sieu Tran, Taisei Hanyu, Yuki Ikebe, Duy Nguyen, Nghi D.~Q. Bui, Minh
  Vu, Anthony Gunderman, Chase Rainwater, Anh Nguyen, and Ngan Le.
\newblock {CodeGraphVLP: Code-as-Planner Meets Semantic-Graph State for
  Non-Markovian Vision-Language-Action Models}, 2026.
\newblock URL \url{https://arxiv.org/abs/2604.22238}.

\bibitem[Wan et~al.(2026)Wan, Yue, Liu, Chu, Wang, Chen, Jiang, Zhu, Dong, and
  Zhu]{arxiv260909250}
Yang Wan, Xihang Yue, Zhirui Liu, Ziyuan Chu, Shuxun Wang, Yuhan Chen, Xiaonan
  Jiang, Xukun Zhu, Yubo Dong, and Linchao Zhu.
\newblock {No Free Checker: A Survey of Verifiers for Robot Policies}, 2026.
\newblock URL \url{https://arxiv.org/abs/2609.09250}.

\bibitem[Wang et~al.(2026{\natexlab{a}})Wang, Shi, Xu, Wang, Zhao, Xia, Liao,
  Peng, Bi, Ma, Liu, Wang, Wang, Wu, Dong, Su, Huang, Guo, Du, Wang, and
  Xu]{arxiv261101112}
Chengcheng Wang, Yuheng Shi, Siyu Xu, Yutong Wang, Jinjing Zhao, Chenghao Xia,
  Jinqi Liao, Jiankun Peng, Ligong Bi, Yuechuan Ma, Jiaqi Liu, Wenbo Wang,
  Zijian Wang, Yuemin Wu, Minjing Dong, Xiu Su, Tao Huang, Jianyuan Guo,
  Chengbin Du, Yunke Wang, and Chang Xu.
\newblock {Auriga Technical Report: A Task-Agnostic Agent Harness for Zero-Shot
  Embodied Manipulation}, 2026{\natexlab{a}}.
\newblock URL \url{https://arxiv.org/abs/2611.01112}.

\bibitem[Wang et~al.(2026{\natexlab{b}})Wang, Qi, Xie, Linke, and
  Brown]{arxiv260627231}
Duanyang Wang, Lu~Qi, Yuanheng Xie, Norbert~M. Linke, and Kenneth~R. Brown.
\newblock {A hardware-safety-gated system for LLM-written native ARTIQ control
  code on a trapped-ion platform}, 2026{\natexlab{b}}.
\newblock URL \url{https://arxiv.org/abs/2606.27231}.

\bibitem[Wang et~al.(2024{\natexlab{a}})Wang, Xie, Jiang, Mandlekar, Xiao, Zhu,
  Fan, and Anandkumar]{arxiv230516291}
Guanzhi Wang, Yuqi Xie, Yunfan Jiang, Ajay Mandlekar, Chaowei Xiao, Yuke Zhu,
  Linxi Fan, and Anima Anandkumar.
\newblock {Voyager: An Open-Ended Embodied Agent with Large Language Models}.
\newblock \emph{Transactions on Machine Learning Research}, 2024{\natexlab{a}}.
\newblock URL \url{https://openreview.net/forum?id=ehfRiF0R3a}.

\bibitem[Wang et~al.(2026{\natexlab{c}})Wang, Zhang, Liu, Zhang, Cai, Liu, and
  Liu]{arxiv260206556}
Guodong Wang, Chenkai Zhang, Qingjie Liu, Jinjin Zhang, Jiancheng Cai, Junjie
  Liu, and Xinmin Liu.
\newblock {LIBERO-X: Robustness Litmus for Vision-Language-Action Models}.
\newblock In \emph{Robotics: Science and Systems XXII}, 2026{\natexlab{c}}.
\newblock \doi{10.15607/rss.2026.xxii.097}.

\bibitem[Wang et~al.(2026{\natexlab{d}})Wang, Cai, Chen, Chi, Sun, Dai, Hung,
  Guo, Ren, Yao, Liu, Long, Duan, Gao, Lyu, Liu, and Wu]{arxiv260827549}
Hanyang Wang, Yimo Cai, Weiliang Chen, Jiawei Chi, Haowen Sun, Qiyu Dai,
  Yi-Hsin Hung, Xingzhuo Guo, Jinshan Ren, Runmao Yao, Ziwei Liu, Mingsheng
  Long, Yueqi Duan, Jun Gao, Jiangran Lyu, Fangfu Liu, and Jialong Wu.
\newblock {Code as Worlds: Agentic Discovery of Executable World
  Representations for Physical Reasoning}, 2026{\natexlab{d}}.
\newblock URL \url{https://arxiv.org/abs/2608.27549}.

\bibitem[Wang et~al.(2023)Wang, Gonzalez-Pumariega, Sharma, and
  Choudhury]{arxiv230516744}
Huaxiaoyue Wang, Gonzalo Gonzalez-Pumariega, Yash Sharma, and Sanjiban
  Choudhury.
\newblock {Demo2Code: From Summarizing Demonstrations to Synthesizing Code via
  Extended Chain-of-Thought}.
\newblock In \emph{Advances in Neural Information Processing Systems 36}, pages
  14848--14956, 2023.
\newblock \doi{10.52202/075280-0655}.

\bibitem[Wang(2026)]{arxiv260931760}
Jiaming Wang.
\newblock {What Stops Recursive Self-Improvement in Robotics? Lessons from 123
  Rounds of Agentic Skill Discovery}, 2026.
\newblock URL \url{https://arxiv.org/abs/2609.31760}.

\bibitem[Wang et~al.(2026{\natexlab{e}})Wang, Zhang, Lu, and
  Chen]{arxiv261034108}
Jie Wang, Bo~Zhang, Renzhi Lu, and Jianyu Chen.
\newblock {RoboDuo: Attempt, Check and Atomic Recovery through Code for
  Vision-Language-Action Models}, 2026{\natexlab{e}}.
\newblock URL \url{https://arxiv.org/abs/2610.34108}.

\bibitem[Wang et~al.(2026{\natexlab{f}})Wang, Jie, Yan, Heng, and
  Li]{arxiv260927612}
Kailin Wang, Haoxiang Jie, Yaoyuan Yan, Zhiyou Heng, and Zhaosong Li.
\newblock {RegenHarness: A Robot Agent Harness with Evidence-Gated Recursive
  Self-Improvement}, 2026{\natexlab{f}}.
\newblock URL \url{https://arxiv.org/abs/2609.27612}.

\bibitem[Wang et~al.(2026{\natexlab{g}})Wang, Wang, Chen, Zhang, Gui, Wang,
  Chen, Xiao, Huang, Zhang, Li, Wang, Kim, Sun, Deng, Hou, Zhu, Liu, Xing, Yao,
  Fan, Liu, Song, Abbeel, Fei-Fei, Wu, Zhu, Shi, Zhang, and Li]{arxiv261112177}
Kangrui Wang, Qineng Wang, Changhe Chen, Pingyue Zhang, Chi Gui, Feishi Wang,
  Canyu Chen, Wenli Xiao, Xiaoyu Huang, Tonghe Zhang, Linjie Li, Zihan Wang,
  Minyeong Kim, Yiyou Sun, Mingkai Deng, Jinyu Hou, Qi~Zhu, Han Liu, Eric~P.
  Xing, Huaxiu Yao, Zhiwen Fan, Yuejiang Liu, Dawn Song, Pieter Abbeel,
  Li~Fei-Fei, Jiajun Wu, Yuke Zhu, Guanya Shi, Ruohan Zhang, and Manling Li.
\newblock {Embodied First Exam}, 2026{\natexlab{g}}.
\newblock URL \url{https://arxiv.org/abs/2611.12177}.

\bibitem[Wang et~al.(2024{\natexlab{b}})Wang, Ling, Yuan, Shridhar, Bao, Qin,
  Wang, Xu, and Wang]{arxiv231001361}
Lirui Wang, Yiyang Ling, Zhecheng Yuan, Mohit Shridhar, Chen Bao, Yuzhe Qin,
  Bailin Wang, Huazhe Xu, and Xiaolong Wang.
\newblock {GenSim: Generating Robotic Simulation Tasks via Large Language
  Models}.
\newblock In \emph{International Conference on Learning Representations
  (ICLR)}, 2024{\natexlab{b}}.
\newblock URL \url{https://openreview.net/forum?id=OI3RoHoWAN}.

\bibitem[Wang et~al.(2026{\natexlab{h}})Wang, Ma, Chang, Luo, Zhang, Wang,
  Yang, Feng, Yang, and Li]{arxiv260811350}
Peidong Wang, Zhiming Ma, Ying Chang, Xufang Luo, Yiqun Zhang, Zihan Wang,
  Xiaocui Yang, Shi Feng, Yuqing Yang, and Dongsheng Li.
\newblock {Self-Evolving Embodied Agents via Skill-Harness Evolution},
  2026{\natexlab{h}}.
\newblock URL \url{https://arxiv.org/abs/2608.11350}.

\bibitem[Wang et~al.(2026{\natexlab{i}})Wang, Wang, Li, Ban, Chen, Ge, Qin, Li,
  and Zhu]{arxiv260811246}
Qi~Wang, Tianyi Wang, Chengyang Li, Shikun Ban, Yurun Chen, Yizhong Ge, Jason
  Qin, Chengtai Li, and Wentao Zhu.
\newblock {Towards the Harness of Embodied Agents}, 2026{\natexlab{i}}.
\newblock URL \url{https://arxiv.org/abs/2608.11246}.

\bibitem[Wang et~al.(2026{\natexlab{j}})Wang, Bi, Jiang, Si, Huang, Chen, Li,
  Li, Li, and Zhou]{arxiv261106768}
Qiang Wang, Jianxin Bi, Shulong Jiang, Ruiyang Si, Wenbo Huang, Kaiqi Chen,
  Jiaxin Li, Liyun Li, Xuanyi Li, and Daquan Zhou.
\newblock {PyRUA-REAL: Robot Exploration and Adaptive Learning through RSI in
  the Real World}, 2026{\natexlab{j}}.
\newblock URL \url{https://arxiv.org/abs/2611.06768}.

\bibitem[Wang et~al.(2026{\natexlab{k}})Wang, Li, Jiao, and
  Yuan]{arxiv250701930}
Wenhao Wang, Yanyan Li, Long Jiao, and Jiawei Yuan.
\newblock {Large Language Model-Driven Closed-Loop UAV Operation with Semantic
  Observations}.
\newblock \emph{IEEE Internet of Things Journal}, 13\penalty0 (7):\penalty0
  14465--14476, 2026{\natexlab{k}}.
\newblock \doi{10.1109/jiot.2025.3649376}.

\bibitem[Wang et~al.(2026{\natexlab{l}})Wang, Rong, Li, Jiao, and
  Yuan]{arxiv251202002}
Wenhao Wang, Yi~Rong, Yanyan Li, Long Jiao, and Jiawei Yuan.
\newblock {LLM-Driven Corrective Robot Operation Code Generation with Static
  Text-Based Simulation}.
\newblock In \emph{2026 IEEE International Conference on Robotics and
  Automation (ICRA)}, pages 13716--13723, 2026{\natexlab{l}}.
\newblock \doi{10.1109/icra57385.2026.11696332}.

\bibitem[Wang et~al.(2026{\natexlab{m}})Wang, Wu, Zhang, Wang, Shao, Luan, Li,
  Li, Gu, Zhou, Shi, Ni, Lu, Meng, Li, Wu, Zhao, Guo, Li, Tang, Zhang, Xie, and
  Qu]{arxiv260929166}
Xin Wang, Wenhao Wu, Menghao Zhang, Zhi Wang, Kun Shao, Jian Luan, Yang Li,
  Qing Li, Shangding Gu, Huichi Zhou, Shuqing Shi, Fei Ni, Shuo Lu, Weicheng
  Meng, Kang Li, Jin Wu, Kang Zhao, Shangmin Guo, Gen Li, Yongqiang Tang,
  Zhizhong Zhang, Yuan Xie, and Heng Qu.
\newblock {HarnessPAI: An Evolving Harness for Physical AI},
  2026{\natexlab{m}}.
\newblock URL \url{https://arxiv.org/abs/2609.29166}.

\bibitem[Wang et~al.(2024{\natexlab{c}})Wang, Chen, Yuan, Zhang, Li, Peng, and
  Ji]{arxiv240201030}
Xingyao Wang, Yangyi Chen, Lifan Yuan, Yizhe Zhang, Yunzhu Li, Hao Peng, and
  Heng Ji.
\newblock {Executable Code Actions Elicit Better LLM Agents}.
\newblock In \emph{International Conference on Machine Learning (ICML)},
  2024{\natexlab{c}}.
\newblock URL \url{https://openreview.net/forum?id=jJ9BoXAfFa}.

\bibitem[Wang et~al.(2025)Wang, Li, Song, Xu, Tang, Zhuge, Pan, Song, Li,
  Singh, Tran, Li, Ma, Zheng, Qian, Shao, Muennighoff, Zhang, Hui, Lin,
  Brennan, Peng, Ji, and Neubig]{arxiv240716741}
Xingyao Wang, Boxuan Li, Yufan Song, Frank~F. Xu, Xiangru Tang, Mingchen Zhuge,
  Jiayi Pan, Yueqi Song, Bowen Li, Jaskirat Singh, Hoang~H. Tran, Fuqiang Li,
  Ren Ma, Mingzhang Zheng, Bill Qian, Yanjun Shao, Niklas Muennighoff, Yizhe
  Zhang, Binyuan Hui, Junyang Lin, Robert Brennan, Hao Peng, Heng Ji, and
  Graham Neubig.
\newblock {OpenHands: An Open Platform for AI Software Developers as Generalist
  Agents}.
\newblock In \emph{International Conference on Learning Representations
  (ICLR)}, 2025.
\newblock URL \url{https://openreview.net/forum?id=OJd3ayDDoF}.

\bibitem[Wang et~al.(2026{\natexlab{n}})Wang, Li, Wang, Zhang, Cui, Tang, Buch,
  and Xu]{arxiv261018792}
Xinyi Wang, Bangzheng Li, Hanyu Wang, Lin Zhang, Andrea~Yaoyun Cui, Zhenggang
  Tang, Shyamal Buch, and Dejia Xu.
\newblock {SEAR: Self-Evolving Agentic Robotics}, 2026{\natexlab{n}}.
\newblock URL \url{https://arxiv.org/abs/2610.18792}.

\bibitem[Wang et~al.(2026{\natexlab{o}})Wang, Jiang, Deng, Xue, Ye, Duan,
  Haghtalab, Sastry, Abbeel, and Qi]{arxiv261002204}
Yen-Jen Wang, Haozhe Jiang, Shuying Deng, Haoru Xue, Weirui Ye, Rocky Duan,
  Nika Haghtalab, S.~Shankar Sastry, Pieter Abbeel, and Haozhi Qi.
\newblock {Reconstruct, Practice, Go Real: Guided Self-Improvement for Embodied
  Agents}, 2026{\natexlab{o}}.
\newblock URL \url{https://arxiv.org/abs/2610.02204}.

\bibitem[Wang et~al.(2024{\natexlab{d}})Wang, Xian, Chen, Wang, Wang,
  Fragkiadaki, Erickson, Held, and Gan]{arxiv231101455}
Yufei Wang, Zhou Xian, Feng Chen, Tsun-Hsuan Wang, Yian Wang, Katerina
  Fragkiadaki, Zackory Erickson, David Held, and Chuang Gan.
\newblock {RoboGen: Towards Unleashing Infinite Data for Automated Robot
  Learning via Generative Simulation}.
\newblock In \emph{International Conference on Machine Learning (ICML)},
  2024{\natexlab{d}}.
\newblock URL \url{https://openreview.net/forum?id=SQIDlJd3hN}.

\bibitem[Wang et~al.(2026{\natexlab{p}})Wang, Xu, Li, and
  Huang]{arxiv260525832}
Yunfei Wang, Xiaohao Xu, Yang Li, and Xiaonan Huang.
\newblock {When Search Becomes Memory: Accelerating Robot Design Discovery with
  Self-Evolving Skills}.
\newblock In \emph{Proceedings of the 2026 Conference on Empirical Methods in
  Natural Language Processing}, pages 8417--8436, 2026{\natexlab{p}}.
\newblock \doi{10.18653/2026.emnlp-main.368}.

\bibitem[Wang et~al.(2026{\natexlab{q}})Wang, Dang, Shi, Gao, Yin, Tam, Liu,
  Wang, Lu, Zhang, Cong, and Yao]{arxiv261116009}
Zefan Wang, Yufan Dang, Xuanwei Shi, Mingqi Gao, Cheng Yin, SikYuen Tam, Hanyu
  Liu, Yasheng Wang, Yaxi Lu, Songyang Zhang, Xin Cong, and Yuan Yao.
\newblock {JustRUA: A Controlled Study of GPT-6 Robot-Use Agents},
  2026{\natexlab{q}}.
\newblock URL \url{https://arxiv.org/abs/2611.16009}.

\bibitem[Wei et~al.(2026{\natexlab{a}})Wei, Huang, Ma, zhang, and
  Cui]{arxiv260301505}
Bingchuan Wei, Bingqi Huang, Jingheng Ma, Zeyu zhang, and Sen Cui.
\newblock {FATE: Closed-Loop Feasibility-Aware Task Generation with Active
  Repair for Physically Grounded Robotic Curricula}, 2026{\natexlab{a}}.
\newblock URL \url{https://arxiv.org/abs/2603.01505}.

\bibitem[Wei et~al.(2026{\natexlab{b}})Wei, Tan, Wang, and
  Zhou]{arxiv260939304}
Zhijie Wei, Ferris Tan, Jinghui Wang, and Hao Zhou.
\newblock {Scale and Selection: What Makes Automatic Harness Evolution Work for
  Visual-Interface Robot Agents}, 2026{\natexlab{b}}.
\newblock URL \url{https://arxiv.org/abs/2609.39304}.

\bibitem[Woods and Seifi(2026)]{arxiv260814944}
John Woods and Hasti Seifi.
\newblock {SkillComposer: Learning Reusable Skills for Natural-Language Robot
  Programming}, 2026.
\newblock URL \url{https://arxiv.org/abs/2608.14944}.

\bibitem[Wu et~al.(2026{\natexlab{a}})Wu, Zheng, Wang, ming Dai, and
  Li]{arxiv260428056}
Feiyu Wu, Xu~Zheng, Zhuocheng Wang, Yi~ming Dai, and Hui Li.
\newblock {RHyVE: Competence-Aware Verification and Phase-Aware Deployment for
  LLM-Generated Reward Hypotheses}, 2026{\natexlab{a}}.
\newblock URL \url{https://arxiv.org/abs/2604.28056}.

\bibitem[Wu et~al.(2026{\natexlab{b}})Wu, Huang, Ho, Doshi, Zhu, and
  Yeung-Levy]{arxiv261042141}
Shengguang Wu, Suning Huang, Cherie Ho, Ria Doshi, Hao Zhu, and Serena
  Yeung-Levy.
\newblock {RoboArchitect: Automating the Design of Robotic Agents with
  Persistent Memory}, 2026{\natexlab{b}}.
\newblock URL \url{https://arxiv.org/abs/2610.42141}.

\bibitem[Wu et~al.(2025)Wu, Xiong, Hu, Iyengar, Jiang, Bera, Tan, and
  Jagannathan]{arxiv240919471}
Yi~Wu, Zikang Xiong, Yiran Hu, Shreyash~S. Iyengar, Nan Jiang, Aniket Bera, Lin
  Tan, and Suresh Jagannathan.
\newblock {SELP: Generating Safe and Efficient Task Plans for Robot Agents with
  Large Language Models}.
\newblock In \emph{2025 IEEE International Conference on Robotics and
  Automation (ICRA)}, pages 2599--2605, 2025.
\newblock \doi{10.1109/icra55743.2025.11128420}.

\bibitem[Xiao et~al.(2026)Xiao, Xie, Zhang, Lin, Fu, Xue, Lu, Yang, Dai, Wang,
  Wu, Wang, Sastry, Goldberg, Fan, Zhu, and Shi]{arxiv260619980}
Wenli Xiao, Jia Xie, Tonghe Zhang, Haotian Lin, Letian~"Max" Fu, Haoru Xue,
  Jalen Lu, Yi~Yang, Cunxi Dai, Zi~Wang, Jimmy Wu, Guanzhi Wang, S.~Shankar
  Sastry, Ken Goldberg, Linxi~"Jim" Fan, Yuke Zhu, and Guanya Shi.
\newblock {ENPIRE: Agentic Robot Policy Self-Improvement in the Real World}.
\newblock In \emph{Conference on Robot Learning (CoRL)}, 2026.
\newblock URL \url{https://openreview.net/forum?id=0cK9Wf9Eeq}.

\bibitem[Xie et~al.(2025)Xie, Wang, Xiao, Wang, and Chen]{arxiv250104268}
Senwei Xie, Hongyu Wang, Zhanqi Xiao, Ruiping Wang, and Xilin Chen.
\newblock {Robotic Programmer: Video Instructed Policy Code Generation for
  Robotic Manipulation}.
\newblock In \emph{2025 IEEE/RSJ International Conference on Intelligent Robots
  and Systems (IROS)}, pages 14923--14930, 2025.
\newblock \doi{10.1109/iros60139.2025.11247043}.

\bibitem[Xie et~al.(2026{\natexlab{a}})Xie, Chen, and Goldberg]{arxiv260818227}
Shuangyu Xie, Kaiyuan Chen, and Ken Goldberg.
\newblock {Revisiting the "Push-T" Robot Manipulation Task with Agentic
  Robotics}, 2026{\natexlab{a}}.
\newblock URL \url{https://arxiv.org/abs/2608.18227}.

\bibitem[Xie et~al.(2026{\natexlab{b}})Xie, Chen, Yoo, Zhu, Shen, Drake,
  Solowjow, Gulich, Millane, Cathomen, Stuyck, Subramanian, Pan, McGowan,
  Benedict, Fan, Zhu, Huang, and Goldberg]{arxiv261106881}
Shuangyu Xie, Kaiyuan Chen, Uksang Yoo, Xinyue Zhu, Meadow Shen, Jaimyn Drake,
  Eugen Solowjow, Lionel Gulich, Alexander Millane, Rafael Cathomen, Tuur
  Stuyck, Sangeeta Subramanian, Yichao Pan, Bruce McGowan, Michael Benedict,
  Linxi Fan, Yuke Zhu, Spencer Huang, and Ken Goldberg.
\newblock {VABAR: A Variational Automation Benchmark for Agentic Robotics},
  2026{\natexlab{b}}.
\newblock URL \url{https://arxiv.org/abs/2611.06881}.

\bibitem[Xie et~al.(2026{\natexlab{c}})Xie, Chen, Cao, Lu, Wu, and
  Jiang]{arxiv260937810}
Sicheng Xie, Yitong Chen, Haidong Cao, Shunlin Lu, Zuxuan Wu, and Yu-Gang
  Jiang.
\newblock {Explore, Execute, Evolve: A Skill Acquisition and Reuse Loop for
  Embodied Agents}, 2026{\natexlab{c}}.
\newblock URL \url{https://arxiv.org/abs/2609.37810}.

\bibitem[Xie et~al.(2024)Xie, Zhao, Wu, Liu, Luo, Zhong, Yang, and
  Yu]{arxiv230911489}
Tianbao Xie, Siheng Zhao, Chen~Henry Wu, Yitao Liu, Qian Luo, Victor Zhong,
  Yanchao Yang, and Tao Yu.
\newblock {Text2Reward: Reward Shaping with Language Models for Reinforcement
  Learning}.
\newblock In \emph{International Conference on Learning Representations
  (ICLR)}, 2024.
\newblock URL \url{https://openreview.net/forum?id=tUM39YTRxH}.

\bibitem[Xu et~al.(2026{\natexlab{a}})Xu, Shang, Dong, and
  Ferrara]{arxiv260920822}
Bingxin Xu, Yuzhang Shang, Zhen Dong, and Emilio Ferrara.
\newblock {Coding Agents with Harness for Safe Robot Control},
  2026{\natexlab{a}}.
\newblock URL \url{https://arxiv.org/abs/2609.20822}.

\bibitem[Xu et~al.(2026{\natexlab{b}})Xu, Shang, and Ferrara]{arxiv260816889}
Bingxin Xu, Yuzhang Shang, and Emilio Ferrara.
\newblock {Don't Drop the BATON: Long-Horizon Robot Manipulation via Agentic
  Subtask Exploration and Transition-aware Memory}, 2026{\natexlab{b}}.
\newblock URL \url{https://arxiv.org/abs/2608.16889}.

\bibitem[Xu et~al.(2025)Xu, Zhang, Guo, Wen, Yang, Lin, Huang, Li, Zhang, Wang,
  Kuang, Cao, Zheng, and Liang]{arxiv250412636}
Rongtao Xu, Jian Zhang, Minghao Guo, Youpeng Wen, Haoting Yang, Min Lin,
  Jianzheng Huang, Zhe Li, Kaidong Zhang, Liqiong Wang, Yuxuan Kuang, Meng Cao,
  Feng Zheng, and Xiaodan Liang.
\newblock {A0: An Affordance-Aware Hierarchical Model for General Robotic
  Manipulation}.
\newblock In \emph{2025 IEEE/CVF International Conference on Computer Vision
  (ICCV)}, pages 13491--13501, 2025.
\newblock \doi{10.1109/iccv51701.2025.01253}.

\bibitem[Xu et~al.(2026{\natexlab{c}})Xu, Cao, Wu, Zhong, Lu, He, Han, Cheung,
  and Sun]{arxiv250611003}
Ruiyang Xu, Jialun Cao, Mingyuan Wu, Wenliang Zhong, Yaojie Lu, Ben He, Xianpei
  Han, Shing-Chi Cheung, and Le~Sun.
\newblock {EmbedAgent: Benchmarking Large Language Models in Embedded System
  Development}.
\newblock In \emph{Proceedings of the 2026 IEEE/ACM 48th International
  Conference on Software Engineering}, pages 630--641, 2026{\natexlab{c}}.
\newblock \doi{10.1145/3744916.3787788}.

\bibitem[Yang et~al.(2025{\natexlab{a}})Yang, Wu, Zhang, Zhang, Lian, Liu, Ren,
  Tian, and Che]{arxiv241202410}
Donghao Yang, Aolang Wu, Tianyi Zhang, Li~Zhang, Xiaoli Lian, Fang Liu, Yuming
  Ren, Jiaji Tian, and Xiaoyin Che.
\newblock {AutoPLC: Generating Vendor-Aware Structured Text for Programmable
  Logic Controllers}.
\newblock In \emph{2025 40th IEEE/ACM International Conference on Automated
  Software Engineering (ASE)}, pages 3593--3604, 2025{\natexlab{a}}.
\newblock \doi{10.1109/ase63991.2025.00297}.

\bibitem[Yang et~al.(2026{\natexlab{a}})Yang, Zhang, Wu, Cheng, Zheng, Xu,
  Grant, Chai, Du, Li, LI, Xing, Hao, Tao, Gong, Liu, Li, Tang, Lin, Chen,
  Zhao, Liu, Zhou, Dai, and Lv]{arxiv260403144}
Jian Yang, Wei Zhang, Jiajun Wu, Junhang Cheng, Tuney Zheng, Fanglin Xu,
  William Grant, Linzheng Chai, Yaxin Du, Haau-Sing Li, Yizhi LI, Yan Xing,
  Chuan Hao, Ran Tao, Ruihao Gong, Aishan Liu, Zhoujun Li, Mingjie Tang,
  Chenghua Lin, Siheng Chen, Xin Zhao, Xianglong Liu, Ming Zhou, Bryan Dai, and
  Weifeng Lv.
\newblock {InCoder-32B-Thinking: Industrial Code World Model for Thinking}.
\newblock In \emph{Proceedings of the 2026 Conference on Empirical Methods in
  Natural Language Processing}, pages 41611--41628, 2026{\natexlab{a}}.
\newblock \doi{10.18653/2026.emnlp-main.1827}.

\bibitem[Yang et~al.(2024{\natexlab{a}})Yang, Dong, Liu, Li, Wang, Tan, Jiang,
  Kang, Zhang, Zhou, and Liu]{arxiv231008588}
Jingkang Yang, Yuhao Dong, Shuai Liu, Bo~Li, Ziyue Wang, Haoran Tan, Chencheng
  Jiang, Jiamu Kang, Yuanhan Zhang, Kaiyang Zhou, and Ziwei Liu.
\newblock {Octopus: Embodied Vision-Language Programmer from Environmental
  Feedback}.
\newblock In \emph{European Conference on Computer Vision (ECCV)}, pages
  20--38, 2024{\natexlab{a}}.
\newblock \doi{10.1007/978-3-031-73232-4_2}.

\bibitem[Yang et~al.(2024{\natexlab{b}})Yang, Jimenez, Wettig, Lieret, Yao,
  Narasimhan, and Press]{arxiv240515793}
John Yang, Carlos~E. Jimenez, Alexander Wettig, Kilian Lieret, Shunyu Yao,
  Karthik Narasimhan, and Ofir Press.
\newblock {SWE-agent: Agent-Computer Interfaces Enable Automated Software
  Engineering}.
\newblock In \emph{Advances in Neural Information Processing Systems 37}, pages
  50528--50652, 2024{\natexlab{b}}.
\newblock \doi{10.52202/079017-1601}.

\bibitem[Yang et~al.(2024{\natexlab{c}})Yang, Liu, Wu, Yang, Fung, Li, Huang,
  Cao, Wang, Wang, Ji, and Zhai]{arxiv240100812}
Ke~Yang, Jiateng Liu, John Wu, Chaoqi Yang, Yi~R. Fung, Sha Li, Zixuan Huang,
  Xu~Cao, Xingyao Wang, Yiquan Wang, Heng Ji, and Chengxiang Zhai.
\newblock {If LLM Is the Wizard, Then Code Is the Wand: A Survey on How Code
  Empowers Large Language Models to Serve as Intelligent Agents},
  2024{\natexlab{c}}.
\newblock URL \url{https://arxiv.org/abs/2401.00812}.

\bibitem[Yang et~al.(2025{\natexlab{b}})Yang, Chen, Zhang, Zhao, Qian, Wang,
  Wang, Koripella, Movahedi, Li, Ji, Zhang, and Zhang]{arxiv250209560}
Rui Yang, Hanyang Chen, Junyu Zhang, Mark Zhao, Cheng Qian, Kangrui Wang,
  Qineng Wang, Teja~Venkat Koripella, Marziyeh Movahedi, Manling Li, Heng Ji,
  Huan Zhang, and Tong Zhang.
\newblock {EmbodiedBench: Comprehensive Benchmarking Multi-modal Large Language
  Models for Vision-Driven Embodied Agents}.
\newblock In \emph{International Conference on Machine Learning (ICML)},
  2025{\natexlab{b}}.
\newblock URL \url{https://proceedings.mlr.press/v267/yang25f.html}.

\bibitem[Yang et~al.(2026{\natexlab{b}})Yang, Bian, Gao, Zhang, Ma, Lin, Luo,
  Zhu, and Yin]{yang-etal-2026-dialectical}
Ruichao Yang, Yufan Bian, Wei Gao, Bo-Wen Zhang, Jing Ma, Hongzhan Lin, Ziyang
  Luo, Xiaobin Zhu, and Xu-Cheng Yin.
\newblock Dialectical structured reasoning for explainable multimodal fake news
  detection.
\newblock In Maria Liakata, Viviane~P. Moreira, Jiajun Zhang, and David
  Jurgens, editors, \emph{Findings of the {A}ssociation for {C}omputational
  {L}inguistics: {ACL} 2026}, pages 31998--32013, San Diego, California, United
  States, July 2026{\natexlab{b}}. Association for Computational Linguistics.
\newblock ISBN 979-8-89176-395-1.
\newblock \doi{10.18653/v1/2026.findings-acl.1601}.
\newblock URL \url{https://aclanthology.org/2026.findings-acl.1601/}.

\bibitem[Yang et~al.(2025{\natexlab{c}})Yang, Zhou, Ding, and
  Xin]{arxiv250702256}
Yang Yang, Xiaolu Zhou, Bosong Ding, and Miao Xin.
\newblock {Uncertainty-aware Reward Design Process}.
\newblock \emph{Transactions on Machine Learning Research}, 2025{\natexlab{c}}.
\newblock URL \url{https://openreview.net/forum?id=CId5tW1HxR}.

\bibitem[Yang et~al.(2026{\natexlab{c}})Yang, Tong, Zhang, Ding, Yang, Zhang,
  Li, Yang, and Xin]{arxiv260623280}
Yang Yang, Yuchuang Tong, Zhengtao Zhang, Xu~Ding, Ning Yang, Yifan Zhang,
  Haipeng Li, Kehu Yang, and Miao Xin.
\newblock {Causal Reward World Models: Zero-shot Reward Design for Automated
  Skill Generation}, 2026{\natexlab{c}}.
\newblock URL \url{https://arxiv.org/abs/2606.23280}.

\bibitem[Yang et~al.(2025{\natexlab{d}})Yang, Gu, Wen, Li, Zhao, Wang, and
  Liu]{arxiv251011660}
Yi~Yang, Kefan Gu, Yuqing Wen, Hebei Li, Yucheng Zhao, Tiancai Wang, and Xudong
  Liu.
\newblock {ManiAgent: An Agentic Framework for General Robotic Manipulation},
  2025{\natexlab{d}}.
\newblock URL \url{https://arxiv.org/abs/2510.11660}.

\bibitem[Yang et~al.(2026{\natexlab{d}})Yang, Chen, Shao, Dong, Li, Wang, Fang,
  Wu, Wang, Man, Zhang, Cui, Liu, Lu, Ma, Ma, Sun, Yao, Dong, Zhao, Mu, Jiang,
  Deng, Zhou, Chen, Zhang, Yan, Yuan, You, Wang, Li, Xu, Cao, and
  Su]{arxiv261106654}
Yixuan Yang, Qiang Chen, Run Shao, Shuwei Dong, Wenhao Li, Yunpeng Wang, Wenhan
  Fang, Haomiao Wu, Xingyu Wang, Lichen Man, Rong Zhang, Xuezhi Cui, Yuan Liu,
  Bingyan Lu, Jiawei Ma, Xiaoyang Ma, Zhemin Sun, Zhuorui Yao, Yutong Dong,
  Yuxuan Zhao, Yichong Mu, Jiabao Jiang, Wenxuan Deng, Junling Zhou, Yi~Chen,
  Deyu Zhang, Yiming Yan, Yige Yuan, Shan You, Fei Wang, Haifeng Li, Hongyan
  Xu, Yichao Cao, and Xiu Su.
\newblock {Compounding Physical RSI: Turning Execution Failure into Inherited
  Capability}, 2026{\natexlab{d}}.
\newblock URL \url{https://arxiv.org/abs/2611.06654}.

\bibitem[Yang et~al.(2024{\natexlab{d}})Yang, Sun, Weihs, VanderBilt, Herrasti,
  Han, Wu, Haber, Krishna, Liu, Callison-Burch, Yatskar, Kembhavi, and
  Clark]{arxiv231209067}
Yue Yang, Fan-Yun Sun, Luca Weihs, Eli VanderBilt, Alvaro Herrasti, Winson Han,
  Jiajun Wu, Nick Haber, Ranjay Krishna, Lingjie Liu, Chris Callison-Burch,
  Mark Yatskar, Aniruddha Kembhavi, and Christopher Clark.
\newblock {Holodeck: Language Guided Generation of 3D Embodied AI
  Environments}.
\newblock In \emph{2024 IEEE/CVF Conference on Computer Vision and Pattern
  Recognition (CVPR)}, pages 16277--16287, 2024{\natexlab{d}}.
\newblock \doi{10.1109/cvpr52733.2024.01536}.

\bibitem[Yang et~al.(2024{\natexlab{e}})Yang, Bhatt, Ward, Hu, Biswas, and
  Topcu]{arxiv241014865}
Yunhao Yang, Neel~P. Bhatt, William Ward, Zichao Hu, Joydeep Biswas, and Ufuk
  Topcu.
\newblock {Joint Verification and Refinement of Language Models for
  Safety-Constrained Planning}, 2024{\natexlab{e}}.
\newblock URL \url{https://arxiv.org/abs/2410.14865}.

\bibitem[Yang et~al.(2026{\natexlab{e}})Yang, Bhatt, Wang, Tetteh, Wang, and
  Topcu]{arxiv260605395}
Yunhao Yang, Neel~P. Bhatt, Kevin Wang, Samuel Tetteh, Zhangyang Wang, and Ufuk
  Topcu.
\newblock {VASO: Formally Verifiable Self-Evolving Skills for Physical AI
  Agents}, 2026{\natexlab{e}}.
\newblock URL \url{https://arxiv.org/abs/2606.05395}.

\bibitem[Yang et~al.(2026{\natexlab{f}})Yang, Li, Hu, Liu, Huang, Sun, Li, Wu,
  Huang, Huang, Sun, Liu, Tong, Huang, Hu, Huang, Bai, Wang, Lin, Tang, Zhang,
  Chen, Song, Dai, Chen, Ji, Zhan, Hu, Xue, Zhang, Hu, Ho, and
  Guo]{arxiv261010409}
Zhiqin Yang, Chenxin Li, Xiaomeng Hu, Yibin Liu, Weidong Huang, Jiankai Sun,
  Haitao Li, Zijian Wu, Yuzhi Huang, Fanding Huang, Hanwen Sun, Jiashun Liu,
  Jingqi Tong, Mingxin Huang, Shaoli Hu, Shijue Huang, Tianyi Bai, Xinyuan
  Wang, Yunlong Lin, Zhengyang Tang, Zhexin Zhang, Zhuo Chen, Xierui Song,
  Juntao Dai, Boyuan Chen, Jiaming Ji, Fangneng Zhan, Mengkang Hu, Wei Xue,
  Yonggang Zhang, Han Hu, Tsung-Yi Ho, and Yike Guo.
\newblock {RobotWorld: Benchmarking Multimodal Agents for Robot Use Across
  Diverse Tasks and Embodiments}, 2026{\natexlab{f}}.
\newblock URL \url{https://arxiv.org/abs/2610.10409}.

\bibitem[Yang et~al.(2024{\natexlab{f}})Yang, Raman, Shah, and
  Tellex]{arxiv230909919}
Ziyi Yang, Shreyas~S. Raman, Ankit Shah, and Stefanie Tellex.
\newblock {Plug in the Safety Chip: Enforcing Constraints for LLM-driven Robot
  Agents}.
\newblock In \emph{2024 IEEE International Conference on Robotics and
  Automation (ICRA)}, pages 14435--14442, 2024{\natexlab{f}}.
\newblock \doi{10.1109/icra57147.2024.10611447}.

\bibitem[Yao et~al.(2026{\natexlab{a}})Yao, Zhou, Mees, Meng, Xiao, Bisk, Oh,
  Johns, Shridhar, Shah, Thomason, Huang, Chai, Bing, and
  Knoll]{arxiv231210807}
Xiangtong Yao, Hongkuan Zhou, Oier Mees, Yuan Meng, Ted Xiao, Yonatan Bisk,
  Jean Oh, Edward Johns, Mohit Shridhar, Dhruv Shah, Jesse Thomason, Kai Huang,
  Joyce Chai, Zhenshan Bing, and Alois Knoll.
\newblock {Bridging Language and Action: A Survey of Language-Conditioned Robot
  Manipulation}.
\newblock \emph{The International Journal of Robotics Research},
  2026{\natexlab{a}}.
\newblock \doi{10.1177/02783649261468360}.

\bibitem[Yao et~al.(2026{\natexlab{b}})Yao, Zou, Jiao, Gao, Gao, Wang, Chen,
  Zhu, Li, Yan, Wang, Ma, Guo, Zhang, and Shen]{arxiv261035832}
Yifei Yao, Yude Zou, Guanbo Jiao, Xing Gao, Ning Gao, Yukai Wang, Jiantong
  Chen, Zihou Zhu, Xinyu Li, Ningyu Yan, Shuai Wang, Jianming Ma, Haoyu Guo,
  Weinan Zhang, and Chunhua Shen.
\newblock {Profiling a Frontier Agent with EBench: From Direct Control to
  Harness with Policy}, 2026{\natexlab{b}}.
\newblock URL \url{https://arxiv.org/abs/2610.35832}.

\bibitem[Yeke et~al.(2026)Yeke, Zhou, Lin, Cai, Bianchi, and
  Celik]{arxiv260519328}
Doguhuan Yeke, Yanming Zhou, Leo~Y. Lin, Hongyu Cai, Antonio Bianchi, and
  Z.~Berkay Celik.
\newblock {RoboJailBench: Benchmarking Adversarial Attacks and Defenses in
  Embodied Robotic Agents}, 2026.
\newblock URL \url{https://arxiv.org/abs/2605.19328}.

\bibitem[Yin et~al.(2025)Yin, Li, Wang, McConachie, Shah, Hashimoto, Zhang,
  Liu, and Li]{arxiv250616652}
Guang Yin, Yitong Li, Yixuan Wang, Dale McConachie, Paarth Shah, Kunimatsu
  Hashimoto, Huan Zhang, Katherine Liu, and Yunzhu Li.
\newblock {CodeDiffuser: Attention-Enhanced Diffusion Policy via VLM-Generated
  Code for Instruction Ambiguity}.
\newblock In \emph{Robotics: Science and Systems XXI}, 2025.
\newblock \doi{10.15607/rss.2025.xxi.072}.

\bibitem[Yin et~al.(2024)Yin, Pang, Ding, Chen, Bi, Xiong, Huang, Xiang, Shao,
  and Chen]{arxiv241213178}
Sheng Yin, Xianghe Pang, Yuanzhuo Ding, Menglan Chen, Yutong Bi, Yichen Xiong,
  Wenhao Huang, Zhen Xiang, Jing Shao, and Siheng Chen.
\newblock {SafeAgentBench: A Benchmark for Safe Task Planning of Embodied LLM
  Agents}, 2024.
\newblock URL \url{https://arxiv.org/abs/2412.13178}.

\bibitem[Yoon et~al.(2025)Yoon, Lee, and Sim]{arxiv260603047}
Gina Yoon, Sumin Lee, and Joo~Yong Sim.
\newblock {ModuLoop: Low-Level Code Generation using Modular Synthesizer and
  Closed-Loop Debugger for Robotic Control}.
\newblock \emph{IEEE Robotics and Automation Letters}, 10\penalty0
  (12):\penalty0 12812--12819, 2025.
\newblock \doi{10.1109/lra.2025.3623437}.

\bibitem[Yu(2026)]{arxiv260807555}
Hang Yu.
\newblock {You Don't Need To Stay in The Loop: An Agentic Robotics Loop for
  Robot-Policy Improvement}, 2026.
\newblock URL \url{https://arxiv.org/abs/2608.07555}.

\bibitem[Yu et~al.(2026)Yu, Li, Li, Yan, Peng, Zhong, Xin, Yao, Tian, and
  Li]{arxiv261035604}
Heyang Yu, Haomin Li, Sihang Li, Jixin Yan, Shiyun Peng, Wanqi Zhong, Tianyu
  Xin, Yuan Yao, Xiaoyu Tian, and Yiming Li.
\newblock {RoboDoctor: Training-Free Recursive Self-Improvement for Robot
  Policies}, 2026.
\newblock URL \url{https://arxiv.org/abs/2610.35604}.

\bibitem[Yu et~al.(2023)Yu, Gileadi, Fu, Kirmani, Lee, Arenas, Chiang, Erez,
  Hasenclever, Humplik, Ichter, Xiao, Xu, Zeng, Zhang, Heess, Sadigh, Tan,
  Tassa, and Xia]{arxiv230608647}
Wenhao Yu, Nimrod Gileadi, Chuyuan Fu, Sean Kirmani, Kuang-Huei Lee,
  Montse~Gonzalez Arenas, Hao-Tien~Lewis Chiang, Tom Erez, Leonard Hasenclever,
  Jan Humplik, Brian Ichter, Ted Xiao, Peng Xu, Andy Zeng, Tingnan Zhang,
  Nicolas Heess, Dorsa Sadigh, Jie Tan, Yuval Tassa, and Fei Xia.
\newblock {Language to Rewards for Robotic Skill Synthesis}.
\newblock In \emph{Conference on Robot Learning (CoRL)}, 2023.
\newblock URL \url{https://proceedings.mlr.press/v229/yu23a.html}.

\bibitem[Yuan et~al.(2026)Yuan, Zuo, Huang, Zhang, Park, Yang, Zhang, Chen,
  Ouyang, He, Lv, Wu, Liu, Zhao, Luo, Li, Qian, Ge, ang Gao, Yang, Deng, Xu,
  Huang, Zhou, Xu, Liu, Abbeel, and Ding]{arxiv261101560}
Jiarui Yuan, Yuxin Zuo, Xiaoyu Huang, Wenbo Zhang, Kwanyoung Park, Junlin Yang,
  Tianxiang Zhang, Ruoqu Chen, Yutao Ouyang, Bingxiang He, Xingtai Lv, Hongjin
  Wu, Jiajun Liu, Jiahao Zhao, Tianwei Luo, Haozhan Li, Cheng Qian, Jiaxin Ge,
  Huan ang Gao, Kaisen Yang, Jiaqi Deng, Qixin Xu, Siqiao Huang, Bowen Zhou,
  Mengdi Xu, Zhiyuan Liu, Pieter Abbeel, and Ning Ding.
\newblock {Democratizing Robot-Use Agents: A Unified Harness with Open Models},
  2026.
\newblock URL \url{https://arxiv.org/abs/2611.01560}.

\bibitem[Zeng et~al.(2023)Zeng, Gan, Huai, Sun, Chen, Wang, Liu, and
  Yu]{arxiv231107226}
Fanlong Zeng, Wensheng Gan, Zezheng Huai, Lichao Sun, Hechang Chen, Yongheng
  Wang, Ning Liu, and Philip~S. Yu.
\newblock {Large Language Models for Robotics: A Survey}, 2023.
\newblock URL \url{https://arxiv.org/abs/2311.07226}.

\bibitem[Zeng et~al.(2025)Zeng, Luo, Lin, Tian, Li, Gong, Guo, and
  Ma]{zeng-etal-2025-mm}
Gailun Zeng, Ziyang Luo, Hongzhan Lin, Yuchen Tian, Kaixin Li, Ziyang Gong,
  Jianxiong Guo, and Jing Ma.
\newblock {MM}-{CRITIC}: A holistic evaluation of large multimodal models as
  multimodal critique.
\newblock In Christos Christodoulopoulos, Tanmoy Chakraborty, Carolyn Rose, and
  Violet Peng, editors, \emph{Findings of the {A}ssociation for {C}omputational
  {L}inguistics: {EMNLP} 2025}, pages 13603--13630, Suzhou, China, November
  2025. Association for Computational Linguistics.
\newblock ISBN 979-8-89176-335-7.
\newblock \doi{10.18653/v1/2025.findings-emnlp.733}.
\newblock URL \url{https://aclanthology.org/2025.findings-emnlp.733/}.

\bibitem[Zha et~al.(2024)Zha, Cui, Lin, Kwon, Arenas, Zeng, Xia, and
  Sadigh]{arxiv231110678}
Lihan Zha, Yuchen Cui, Li-Heng Lin, Minae Kwon, Montserrat~Gonzalez Arenas,
  Andy Zeng, Fei Xia, and Dorsa Sadigh.
\newblock {Distilling and Retrieving Generalizable Knowledge for Robot
  Manipulation via Language Corrections}.
\newblock In \emph{2024 IEEE International Conference on Robotics and
  Automation (ICRA)}, pages 15172--15179, 2024.
\newblock \doi{10.1109/icra57147.2024.10610455}.

\bibitem[Zhang et~al.(2025{\natexlab{a}})Zhang, Hong, Pacchiano, and
  Agrawal]{arxiv241013837}
Chen Bo~Calvin Zhang, Zhang-Wei Hong, Aldo Pacchiano, and Pulkit Agrawal.
\newblock {ORSO: Accelerating Reward Design via Online Reward Selection and
  Policy Optimization}.
\newblock In \emph{International Conference on Learning Representations
  (ICLR)}, 2025{\natexlab{a}}.
\newblock URL \url{https://openreview.net/forum?id=0uRc3CfJIQ}.

\bibitem[Zhang et~al.(2025{\natexlab{b}})Zhang, Zhu, Wang, Zhou, Yin, Li, Xue,
  Wang, Hu, Liu, Guo, and Zhang]{arxiv240720242}
Hangtao Zhang, Chenyu Zhu, Xianlong Wang, Ziqi Zhou, Changgan Yin, Minghui Li,
  Lulu Xue, Yichen Wang, Shengshan Hu, Aishan Liu, Peijin Guo, and Leo~Yu
  Zhang.
\newblock {BadRobot: Jailbreaking Embodied LLM Agents in the Physical World}.
\newblock In \emph{International Conference on Learning Representations
  (ICLR)}, 2025{\natexlab{b}}.
\newblock URL \url{https://openreview.net/forum?id=ei3qCntB66}.

\bibitem[Zhang et~al.(2024)Zhang, Xu, and Li]{arxiv240514062}
Jiawei Zhang, Chejian Xu, and Bo~Li.
\newblock {ChatScene: Knowledge-Enabled Safety-Critical Scenario Generation for
  Autonomous Vehicles}.
\newblock In \emph{2024 IEEE/CVF Conference on Computer Vision and Pattern
  Recognition (CVPR)}, pages 15459--15469, 2024.
\newblock \doi{10.1109/cvpr52733.2024.01464}.

\bibitem[Zhang et~al.(2026{\natexlab{a}})Zhang, Zhang, Song, Zhou, Song, and
  Wang]{arxiv260932416}
Jiawei Zhang, Xiangrong Zhang, Rui Song, Huanbin Zhou, Chengye Song, and
  Hongzhou Wang.
\newblock {RE-0: Verified Recursive Improvement of Embodied Code-as-Policy
  Agents through Local On-Policy Distillation}, 2026{\natexlab{a}}.
\newblock URL \url{https://arxiv.org/abs/2609.32416}.

\bibitem[Zhang et~al.(2026{\natexlab{b}})Zhang, Wang, Yu, Cai, Chow, Choy, Dou,
  Hua, Li, Liu, Wang, Wei, Yang, Yuan, Zhan, Zhang, Zheng, Li, Liu, Matusik,
  Tu, and Wang]{arxiv261108946}
Jingdong Zhang, Jionghao Wang, Zhengming Yu, Yancheng Cai, Wei Chow, Chris
  Choy, Zhiyang Dou, Hang Hua, Jiachen Li, Zhixuan Liu, Chen Wang, Songlin Wei,
  Siyuan Yang, Yu~Yuan, Xiaohang Zhan, Yucheng Zhang, Shuhong Zheng, Xin Li,
  Lingjie Liu, Wojciech Matusik, Zhengzhong Tu, and Wenping Wang.
\newblock {Flex-RSI: Benchmarking Recursive Self-Improvement of Frontier
  Agents}, 2026{\natexlab{b}}.
\newblock URL \url{https://arxiv.org/abs/2611.08946}.

\bibitem[Zhang et~al.(2026{\natexlab{c}})Zhang, Ge, Yoo, Fu, Yang, Liu,
  Saravanan, Yin, Yu, Niu, Wang, Herzig, Goldberg, Bai, Chan, Stoica, Kanazawa,
  Lei, Feng, and Darrell]{arxiv260619419}
Junyi Zhang, Jiaxin Ge, Hanjun Yoo, Letian Fu, Zihan Yang, Yaowei Liu, Raj
  Saravanan, Shaofeng Yin, Justin Yu, Dantong Niu, Zirui Wang, Roei Herzig, Ken
  Goldberg, Yutong Bai, David~M. Chan, Ion Stoica, Angjoo Kanazawa, Jiahui Lei,
  Haiwen Feng, and Trevor Darrell.
\newblock {Playful Agentic Robot Learning}.
\newblock In \emph{Conference on Robot Learning (CoRL)}, 2026{\natexlab{c}}.
\newblock URL \url{https://openreview.net/forum?id=GeL3GdwSYo}.

\bibitem[Zhang et~al.(2026{\natexlab{d}})Zhang, Ma, Lv, Wang, Ruan, Si, Zhang,
  Chen, Chen, Liu, Zhang, Deng, Xie, and Zhou]{arxiv261105320}
Kewei Zhang, Juncheng Ma, Guodong Lv, Duomin Wang, Feixiang Ruan, Ruiyang Si,
  Chuanrui Zhang, Yuxuan Chen, Hao Chen, Yichen Liu, Yintianrun Zhang, Yufan
  Deng, Enze Xie, and Daquan Zhou.
\newblock {PyRUA-DataFarm: LLM Agent as a Cost-Efficient Teleoperator for
  Real-Ready Robot Data}, 2026{\natexlab{d}}.
\newblock URL \url{https://arxiv.org/abs/2611.05320}.

\bibitem[Zhang et~al.(2025{\natexlab{c}})Zhang, Dong, Bai, Ni, Fu, Marton,
  Chen, and Zhang]{arxiv251204308}
Lei Zhang, Ju~Dong, Kaixin Bai, Minheng Ni, Yu~Fu, Zoltan-Csaba Marton,
  Zhaopeng Chen, and Jianwei Zhang.
\newblock {ResponsibleRobotBench: Benchmarking Risk-Aware Task-Level Decision
  Making for Robotic Manipulation with Large Multimodal Models},
  2025{\natexlab{c}}.
\newblock URL \url{https://arxiv.org/abs/2512.04308}.

\bibitem[Zhang et~al.(2025{\natexlab{d}})Zhang, Xu, Liu, Yu, Li, Gao, Fei, Yin,
  Wu, Jiang, and Qiu]{arxiv241218194}
Shiduo Zhang, Zhe Xu, Peiju Liu, Xiaopeng Yu, Yuan Li, Qinghui Gao, Zhaoye Fei,
  Zhangyue Yin, Zuxuan Wu, Yu-Gang Jiang, and Xipeng Qiu.
\newblock {VLABench: A Large-Scale Benchmark for Language-Conditioned Robotics
  Manipulation with Long-Horizon Reasoning Tasks}.
\newblock In \emph{2025 IEEE/CVF International Conference on Computer Vision
  (ICCV)}, pages 11142--11152, 2025{\natexlab{d}}.
\newblock \doi{10.1109/iccv51701.2025.01037}.

\bibitem[Zhang et~al.(2026{\natexlab{e}})Zhang, Wang, Ouyang, Huang, Li, Su,
  Jin, Chai, Liang, Dou, Chen, and Chen]{arxiv260924170}
Wenbo Zhang, Kaixuan Wang, Yutao Ouyang, Xiaoyu Huang, Liyang Li, Kailun Su,
  Weiyang Jin, Wenhao Chai, Haotian Liang, Zhiyang Dou, Yue Chen, and Tianxing
  Chen.
\newblock {An Unexpected Robot Policy: Early Evaluations of GPT-6 Astra on
  RoboDojo and Beyond}, 2026{\natexlab{e}}.
\newblock URL \url{https://arxiv.org/abs/2609.24170}.

\bibitem[Zhang et~al.(2026{\natexlab{f}})Zhang, Wang, Bauer, Yu, Huang, Wu, and
  Pavone]{arxiv261009228}
Xilun Zhang, Maggie Wang, Erik Bauer, Hong-Xing Yu, Huang Huang, Jiajun Wu, and
  Marco Pavone.
\newblock {Co-Evolving Robot Orchestrators and Policies through Deployment},
  2026{\natexlab{f}}.
\newblock URL \url{https://arxiv.org/abs/2610.09228}.

\bibitem[Zhang et~al.(2026{\natexlab{g}})Zhang, Cui, Wang, Li, Qiu, Zhu, and
  He]{arxiv260522148}
Xing Zhang, Yanwei Cui, Guanghui Wang, Ziyuan Li, Wei Qiu, Bing Zhu, and
  Peiyang He.
\newblock {Ratchet: How Reliable Must an LLM Judge Be to Retire a Skill?},
  2026{\natexlab{g}}.
\newblock URL \url{https://arxiv.org/abs/2605.22148}.

\bibitem[Zhang et~al.(2026{\natexlab{h}})Zhang, Huang, Chang, Su, Zhou, Xu,
  Chen, Zhou, Wang, Zhang, Wei, Li, Deng, Li, Chen, and Li]{arxiv260929964}
Yehang Zhang, Haojian Huang, Yifan Chang, Jianchong Su, Bohan Zhou, Yingjie Xu,
  Wosong Chen, Tianhao Zhou, Chenxu Wang, Tianyi Zhang, Yangkai Wei, Wenqian
  Li, Shiyuan Deng, Yinchuan Li, Ying-Cong Chen, and Zexi Li.
\newblock {World Action Agent: Harnessing VLMs for Robot Manipulation via World
  Action Rehearsal}, 2026{\natexlab{h}}.
\newblock URL \url{https://arxiv.org/abs/2609.29964}.

\bibitem[Zhang et~al.(2026{\natexlab{i}})Zhang, Zhang, Gao, Li, Liu, Nie, Zhu,
  Qiu, Yan, Liu, Tang, Rao, Fang, Wei, Wang, Ding, and Yu]{arxiv260708448}
Yixian Zhang, Huanming Zhang, Feng Gao, Xiao Li, Zhihao Liu, Yi~Nie, Chunyang
  Zhu, Jiaxing Qiu, Yuchen Yan, Jiyuan Liu, Wenhao Tang, Jiaji Rao, Zhengru
  Fang, Changxu Wei, Yu~Wang, Wenbo Ding, and Chao Yu.
\newblock {Harness VLA: Steering Frozen VLAs into Reliable Manipulation
  Primitives via Memory-Guided Agents}, 2026{\natexlab{i}}.
\newblock URL \url{https://arxiv.org/abs/2607.08448}.

\bibitem[Zhang et~al.(2026{\natexlab{j}})Zhang, Zhu, Chen, Tuya, and
  James]{arxiv261007594}
Zexi Zhang, Zecheng Zhu, Zidong Chen, Zulkhuu Tuya, and Stephen James.
\newblock {BiGym 2.0: Benchmarking Learned and Agent-Developed Policies for
  Humanoid Household Manipulation}, 2026{\natexlab{j}}.
\newblock URL \url{https://arxiv.org/abs/2610.07594}.

\bibitem[Zhao et~al.(2026{\natexlab{a}})Zhao, Yue, He, and
  Wang]{arxiv261113034}
Delin Zhao, Zhengrong Yue, Junlin He, and Yali Wang.
\newblock {RoboACE: Adversarial Co-Evolution for Robot Self-Improvement},
  2026{\natexlab{a}}.
\newblock URL \url{https://arxiv.org/abs/2611.13034}.

\bibitem[Zhao and Yang(2026)]{arxiv261034151}
Weiyan Zhao and Sunny Yang.
\newblock {Execution Logs Sway VLM Judges of Robot Progress}, 2026.
\newblock URL \url{https://arxiv.org/abs/2610.34151}.

\bibitem[Zhao et~al.(2026{\natexlab{b}})Zhao, Weber, and
  Wermter]{arxiv240515019}
Xufeng Zhao, Cornelius Weber, and Stefan Wermter.
\newblock {Agentic Skill Discovery}.
\newblock \emph{Robotics and Autonomous Systems}, 196:\penalty0 105248,
  2026{\natexlab{b}}.
\newblock \doi{10.1016/j.robot.2025.105248}.

\bibitem[Zhao et~al.(2026{\natexlab{c}})Zhao, Shao, Yu, Shen, Chen, Tam, and
  Shafique]{arxiv261041168}
Yichen Zhao, Minghao Shao, Zhuoyan Yu, Yijun Shen, Boyuan Chen, Yik-Cheung Tam,
  and Muhammad Shafique.
\newblock {ROSTrans: Empirically Benchmarking LLMs for ROS1-to-ROS2 Migration},
  2026{\natexlab{c}}.
\newblock URL \url{https://arxiv.org/abs/2610.41168}.

\bibitem[Zhao et~al.(2026{\natexlab{d}})Zhao, Ni, Chen, Zhang, and
  Zhu]{arxiv260809410}
Yunhao Zhao, Zhenyang Ni, Haoyang Chen, Ruohan Zhang, and Qi~Zhu.
\newblock {Skills in Weights, Memory in Code: Hybrid Learning for
  Memory-Dependent Robot Manipulation}, 2026{\natexlab{d}}.
\newblock URL \url{https://arxiv.org/abs/2608.09410}.

\bibitem[Zhao et~al.(2025)Zhao, Luo, Tian, Lin, Yan, Li, and
  Ma]{zhao-etal-2025-codejudge}
Yuwei Zhao, Ziyang Luo, Yuchen Tian, Hongzhan Lin, Weixiang Yan, Annan Li, and
  Jing Ma.
\newblock {C}ode{J}udge-eval: Can large language models be good judges in code
  understanding?
\newblock In Owen Rambow, Leo Wanner, Marianna Apidianaki, Hend Al-Khalifa,
  Barbara~Di Eugenio, and Steven Schockaert, editors, \emph{Proceedings of the
  31st International Conference on Computational Linguistics}, pages 73--95,
  Abu Dhabi, UAE, January 2025. Association for Computational Linguistics.
\newblock URL \url{https://aclanthology.org/2025.coling-main.7/}.

\bibitem[Zhao et~al.(2027)Zhao, Mao, Zhang, Li, and Xiao]{zhao2026rosdeveval}
Yuxin Zhao, Xinjun Mao, Tanghaoran Zhang, Tun Li, and Zhiqun Xiao.
\newblock {Evaluating LLMs in ROS Robotic Software Code Generation}.
\newblock \emph{Empirical Software Engineering}, 32:\penalty0 19, 2027.
\newblock \doi{10.1007/s10664-026-10910-7}.

\bibitem[Zhao et~al.(2026{\natexlab{e}})Zhao, Hua, Berto, Ma, Lee, Li, and
  Park]{arxiv260511859}
Zhikai Zhao, Chuanbo Hua, Federico Berto, Zihan Ma, Kanghoon Lee, Jiachen Li,
  and Jinkyoo Park.
\newblock {EvoNav: Evolutionary Reward Function Design for Robot Navigation
  with Large Language Models}, 2026{\natexlab{e}}.
\newblock URL \url{https://arxiv.org/abs/2605.11859}.

\bibitem[Zhong et~al.(2025)Zhong, Bai, Cai, Huang, Chen, Zhang, Wang, Guo,
  Guan, Lui, Qi, Liang, Chen, and Yang]{arxiv250701925}
Yifan Zhong, Fengshuo Bai, Shaofei Cai, Xuchuan Huang, Zhang Chen, Xiaowei
  Zhang, Yuanfei Wang, Shaoyang Guo, Tianrui Guan, Ka~Nam Lui, Zhiquan Qi,
  Yitao Liang, Yuanpei Chen, and Yaodong Yang.
\newblock {A Survey on Vision-Language-Action Models: An Action Tokenization
  Perspective}, 2025.
\newblock URL \url{https://arxiv.org/abs/2507.01925}.

\bibitem[Zhou et~al.(2025{\natexlab{a}})Zhou, Su, Chi, Zhang, Wang, Huang,
  Sheng, and Wang]{arxiv241204455}
Enshen Zhou, Qi~Su, Cheng Chi, Zhizheng Zhang, Zhongyuan Wang, Tiejun Huang,
  Lu~Sheng, and He~Wang.
\newblock {Code-as-Monitor: Constraint-aware Visual Programming for Reactive
  and Proactive Robotic Failure Detection}.
\newblock In \emph{2025 IEEE/CVF Conference on Computer Vision and Pattern
  Recognition (CVPR)}, pages 6919--6929, 2025{\natexlab{a}}.
\newblock \doi{10.1109/cvpr52734.2025.00649}.

\bibitem[Zhou et~al.(2026{\natexlab{a}})Zhou, Yang, Chen, Zhang, Fu, Li, Zhu,
  Liu, Yang, Meng, Wang, Han, Sun, Ye, Wang, Li, Zhou, Li, Jin, Shao, Chen,
  Chen, Ma, Zhou, Spanos, Hu, Schwager, Zhang, Feng, and Gu]{arxiv261018842}
Hai Zhou, Fan Yang, Zeyi Chen, Lejun Zhang, Chenyang Fu, Jianping Li, Jiayi
  Zhu, Xianyang Liu, Yingxuan Yang, Lili Meng, Lu~Wang, Xinyang Han, Jiankai
  Sun, Weirui Ye, Guanchu Wang, Wanhua Li, Ming Zhou, Boyi Li, Ming Jin, Kun
  Shao, Muhao Chen, Guang Chen, Lei Ma, Joey~Tianyi Zhou, Costas Spanos, Xia
  Hu, Mac Schwager, Weinan Zhang, Qian Feng, and Shangding Gu.
\newblock {Who Improves the Robot? Toward Physical Recursive Self-Improvement},
  2026{\natexlab{a}}.
\newblock URL \url{https://arxiv.org/abs/2610.18842}.

\bibitem[Zhou et~al.(2026{\natexlab{b}})Zhou, Li, Lyu, Song, Huang, Zheng,
  Rupprecht, Vedaldi, and Wu]{arxiv260515187}
Matt Zhou, Ruining Li, Xiaoyang Lyu, Zhaomou Song, Zhening Huang, Chuanxia
  Zheng, Christian Rupprecht, Andrea Vedaldi, and Shangzhe Wu.
\newblock {Articraft: An Agentic System for Scalable Articulated 3D Asset
  Generation}, 2026{\natexlab{b}}.
\newblock URL \url{https://arxiv.org/abs/2605.15187}.

\bibitem[Zhou et~al.(2025{\natexlab{b}})Zhou, Xu, Tie, Chen, Zhang, Chu, Zhou,
  and Sun]{arxiv251003827}
Xueyang Zhou, Yangming Xu, Guiyao Tie, Yongchao Chen, Guowen Zhang, Duanfeng
  Chu, Pan Zhou, and Lichao Sun.
\newblock {LIBERO-PRO: Towards Robust and Fair Evaluation of
  Vision-Language-Action Models Beyond Memorization}, 2025{\natexlab{b}}.
\newblock URL \url{https://arxiv.org/abs/2510.03827}.

\bibitem[Zhou et~al.(2026{\natexlab{c}})Zhou, Chen, Salunkhe, Bekar, Stahre,
  and Skoogh]{arxiv260821417}
Zhichao Zhou, Siyuan Chen, Omkar Salunkhe, Ebru~Turanoglu Bekar, Johan Stahre,
  and Anders Skoogh.
\newblock {Retrieval-grounded robot program generation and simulation-based
  correction via Model Context Protocol}, 2026{\natexlab{c}}.
\newblock URL \url{https://arxiv.org/abs/2608.21417}.

\bibitem[Zhou et~al.(2025{\natexlab{c}})Zhou, Atreya, Tan, Pertsch, and
  Levine]{arxiv250324278}
Zhiyuan Zhou, Pranav Atreya, You~Liang Tan, Karl Pertsch, and Sergey Levine.
\newblock {AutoEval: Autonomous Evaluation of Generalist Robot Manipulation
  Policies in the Real World}.
\newblock In \emph{Conference on Robot Learning (CoRL)}, 2025{\natexlab{c}}.
\newblock URL \url{https://openreview.net/forum?id=isrcFrgwZp}.

\bibitem[Zhu et~al.(2026{\natexlab{a}})Zhu, Tanaka, Goldberg, and
  Hong]{arxiv250602507}
Alvin Zhu, Yusuke Tanaka, Andrew Goldberg, and Dennis Hong.
\newblock {AURA: Autonomous Upskilling with Retrieval-Augmented Agents}.
\newblock In \emph{2026 IEEE International Conference on Robotics and
  Automation (ICRA)}, pages 6216--6223, 2026{\natexlab{a}}.
\newblock \doi{10.1109/icra57385.2026.11696789}.

\bibitem[Zhu et~al.(2026{\natexlab{b}})Zhu, Zhao, Sun, Guo, Jiang, Zhang, Fan,
  and Huang]{arxiv261114115}
Qinfeng Zhu, Weiguang Zhao, Ke~Sun, Xihao Guo, Yunxi Jiang, Rui Zhang, Lei Fan,
  and Kaizhu Huang.
\newblock {LEASH: Stay Safe in Agentic Robotics}, 2026{\natexlab{b}}.
\newblock URL \url{https://arxiv.org/abs/2611.14115}.

\bibitem[Zook et~al.(2025)Zook, Sun, Spjut, Blukis, Birchfield, and
  Tremblay]{arxiv241015536}
Alex Zook, Fan-Yun Sun, Josef Spjut, Valts Blukis, Stan Birchfield, and
  Jonathan Tremblay.
\newblock {GRS: Generating Robotic Simulation Tasks from Real-World Images}.
\newblock In \emph{2025 IEEE/CVF Conference on Computer Vision and Pattern
  Recognition Workshops (CVPRW)}, pages 594--603, 2025.
\newblock \doi{10.1109/cvprw67362.2025.00064}.

\bibitem[Zuo et~al.(2025)Zuo, Velez, Li, Littman, and Bach]{arxiv240703321}
Max Zuo, Francisco~Piedrahita Velez, Xiaochen Li, Michael~L. Littman, and
  Stephen~H. Bach.
\newblock {Planetarium: A Rigorous Benchmark for Translating Text to Structured
  Planning Languages}.
\newblock In \emph{Proceedings of the 2025 Conference of the Nations of the
  Americas Chapter of the Association for Computational Linguistics: Human
  Language Technologies (Volume 1: Long Papers)}, pages 11223--11240, 2025.
\newblock \doi{10.18653/v1/2025.naacl-long.560}.

\end{thebibliography}
